%% file: preprint.tex
\documentclass[runningheads]{llncs}

\usepackage[preprint,year=2026,ID=1669]{eccv}

\usepackage{eccvabbrv}
\usepackage[export]{adjustbox}
\usepackage{microtype}
\usepackage{graphicx}
\usepackage{subcaption}
\usepackage{booktabs} 
\usepackage{xcolor}
\usepackage{caption}
\usepackage{wrapfig}
\usepackage[table]{xcolor}

\usepackage[utf8]{inputenc}
\usepackage[T1]{fontenc}
\usepackage{lmodern}

\usepackage[accsupp]{axessibility}  

\definecolor{best}{RGB}{220,50,47}      
\definecolor{second}{RGB}{255,159,10}   
\definecolor{third}{RGB}{255,214,10}    

\definecolor{colorA}{RGB}{247, 166, 166} 
\definecolor{colorB}{RGB}{250, 195, 184} 
\definecolor{colorC}{RGB}{253, 225, 197} 
\definecolor{colorD}{RGB}{255, 253, 205} 

\newcommand{\ca}{\cellcolor{colorA}}
\newcommand{\cb}{\cellcolor{colorB}}
\newcommand{\cc}{\cellcolor{colorC}}
\newcommand{\cd}{\cellcolor{colorD}}

\usepackage{lipsum}

\def\our{IRIS}

\usepackage{hyperref}

\usepackage{orcidlink}

\begin{document}

\title{\our{}: Intersection-aware Ray-based Implicit Editable Scenes} 

\titlerunning{\our{}: Intersection-aware Ray-based Implicit Editable Scenes}

\author{Grzegorz Wilczy\'nski\inst{1, 3}\orcidlink{0009-0002-6053-4410} \and
Miko\l{}aj Zieli\'nski\inst{2}\orcidlink{0000-0003-1107-1149} \and
Krzysztof Byrski\inst{1}\orcidlink{0000-0003-0307-4200} \and
Joanna Waczy\'nska\inst{1, 3}\orcidlink{0009-0003-8593-9307} \and
Dominik Belter\inst{2}\orcidlink{0000-0003-3002-9747} \and
Przemys\l{}aw Spurek\inst{1, 3}\orcidlink{0000-0003-0097-5521}}

\authorrunning{G. Wilczyński et al.}

\institute{Jagiellonian University, Faculty of Mathematics and Computer Science \and
Poznań University of Technology, Institute of Robotics and Machine Intelligence \and
IDEAS Research Institute
}

\maketitle
\vspace{-0.25cm}
\begin{figure}
\begin{center}
\vspace{-0.4cm}
\includegraphics[width=0.95\linewidth]{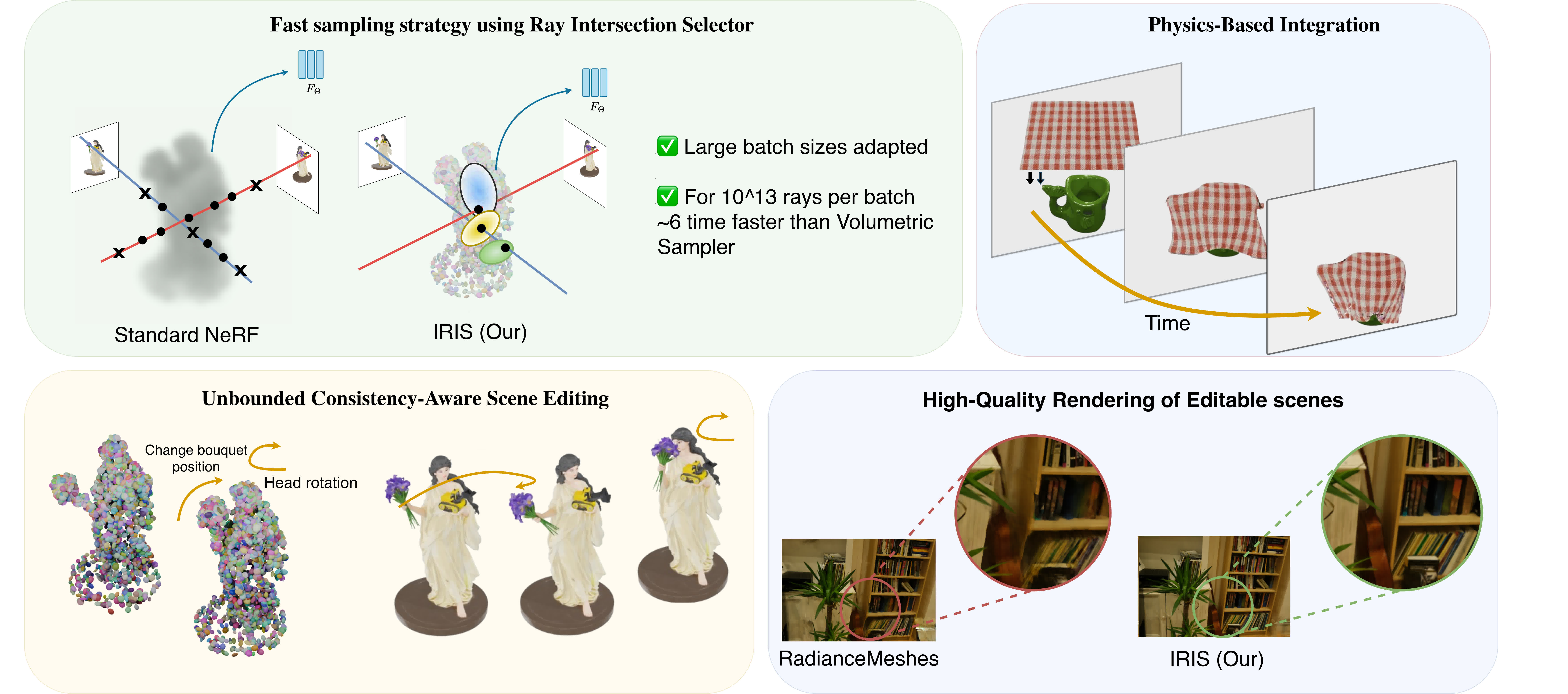}
\vspace{-0.4cm}
\end{center}
    \caption{The framework introduces the Ray Intersection Selector (RIS), which eliminates redundant sampling to achieve high-throughput inference compared to volumetric baselines. This explicit anchor-based representation facilitates seamless integration with physics engines and robust part-level scene editing. Consequently, high-fidelity reconstruction is maintained.}
    \label{fig:teaser}
    \vspace{-1.4cm}
\end{figure}

\begin{abstract}
Neural Radiance Fields achieve high-fidelity scene representation but suffer from costly training and rendering, while 3D Gaussian splatting offers real-time performance with strong empirical results. Recently, solutions that harness the best of both worlds by using Gaussians as proxies to guide neural field evaluations, still suffer from significant computational inefficiencies. They typically rely on stochastic volumetric sampling to aggregate features, which severely limits rendering performance.
To address this issue, a novel framework named \textbf{\our{}} (\textbf{I}ntersection-aware \textbf{R}ay-based \textbf{I}mplicit Editable \textbf{S}cenes) is introduced as a method designed for efficient and interactive scene editing.
To overcome the limitations of standard ray marching, an analytical sampling strategy is employed that precisely identifies interaction points between rays and scene primitives, effectively eliminating empty space processing. Furthermore, to address the computational bottleneck of spatial neighbor lookups, a continuous feature aggregation mechanism is introduced that operates directly along the ray. By interpolating latent attributes from sorted intersections, costly 3D searches are bypassed, ensuring geometric consistency, enabling high-fidelity, real-time rendering, and flexible shape editing. Code can be found at \url{https://github.com/gwilczynski95/iris}.
\end{abstract}

\section{Introduction}
\label{sec:intro}

\begin{figure}
    \centering
    \includegraphics[width=\linewidth]{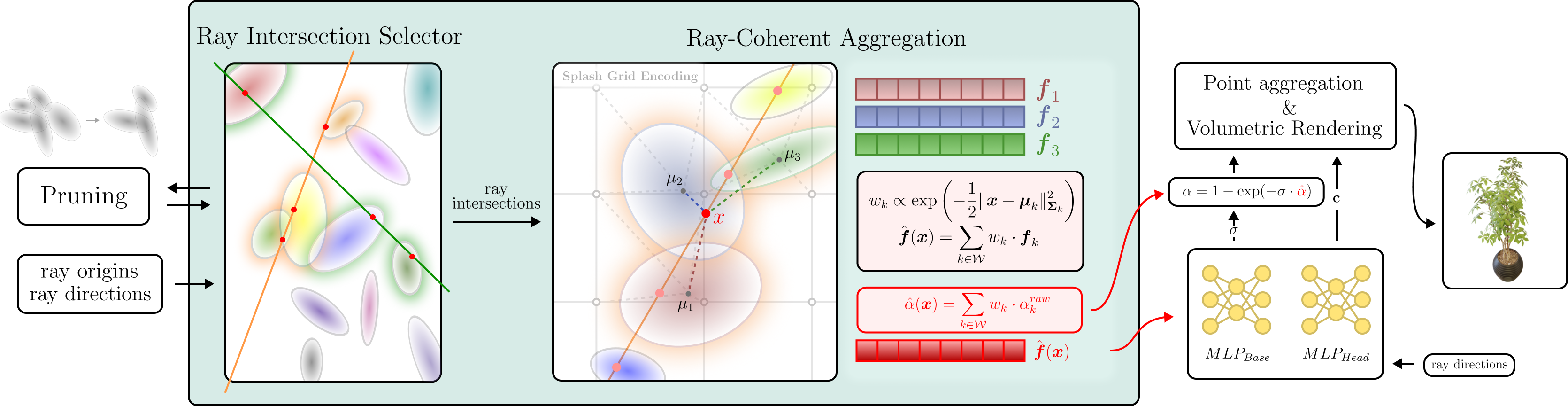}
    \caption{Overview of the IRIS pipeline. The rendering process begins with the \textbf{Ray Intersection Selector (RIS)}, which analytically determines the precise intersection points between camera rays and 3D Gaussians, thereby significantly reducing the number of required samples. In the subsequent \textbf{Ray-Coherent Aggregation (RCA)} stage, local neural features are aggregated from these intersected primitives utilizing Mahalanobis distance-based weighting to synthesize a continuous latent vector. Shallow MLPs decode this representation into view-dependent color and material density, $\sigma$. Finally, the neural density is spatially modulated by the explicit Gaussian opacity and integrated via point-based volumetric rendering to produce the final image.}
    \label{fig:pipeline}
    \vspace{-0.4cm}
\end{figure}

The field of three-dimensional scene reconstruction and rendering has been transformed by two competing paradigms: Neural Radiance Fields (NeRF)~\cite{mildenhall2020nerf} and 3D Gaussian Splatting (3DGS)~\cite{kerbl20233d}. NeRF represents scenes as continuous implicit functions, excelling at capturing view-dependent effects and complex geometries through neural synthesis. Conversely, 3DGS utilizes an explicit collection of anisotropic Gaussians, enabling unprecedented rendering speeds and providing a natural pathway for scene manipulation. As demand for interactive, physically plausible 3D content grows, researchers have increasingly turned to hybrid representations that aim to combine NeRF's high-fidelity continuity with the explicit control of 3DGS.

Despite their potential, current hybrid frameworks like EKS \cite{zielinski2025genie}, particularly those based on Gaussian-encoded neural fields, face a severe performance ceiling. The primary bottleneck stems from the adoption of traditional volumetric rendering pipelines. In these systems, rendering a single pixel requires dense sampling of numerous points along a camera ray. For each sampled point, the model must perform a K-Nearest Neighbors (KNN) search to identify and aggregate features from surrounding Gaussians, then pass them through a multilayer perceptron (MLP). This dual burden of redundant ray sampling and computationally intensive KNN lookups scales poorly with scene complexity, often reducing frame rates to non-interactive levels and hindering the integration of real-time physics engines.

\begin{figure*}[t]
\centering
{\fontsize{6.8pt}{11pt}\selectfont
\begin{tabular}{cccccc}
 & Counter&\hspace{4mm}& Bonsai &\hspace{4mm}& Room\\
\rotatebox{90}{\tiny\hspace{0pt}{RadianceMeshes}} &
\includegraphics[width=0.15\textwidth]{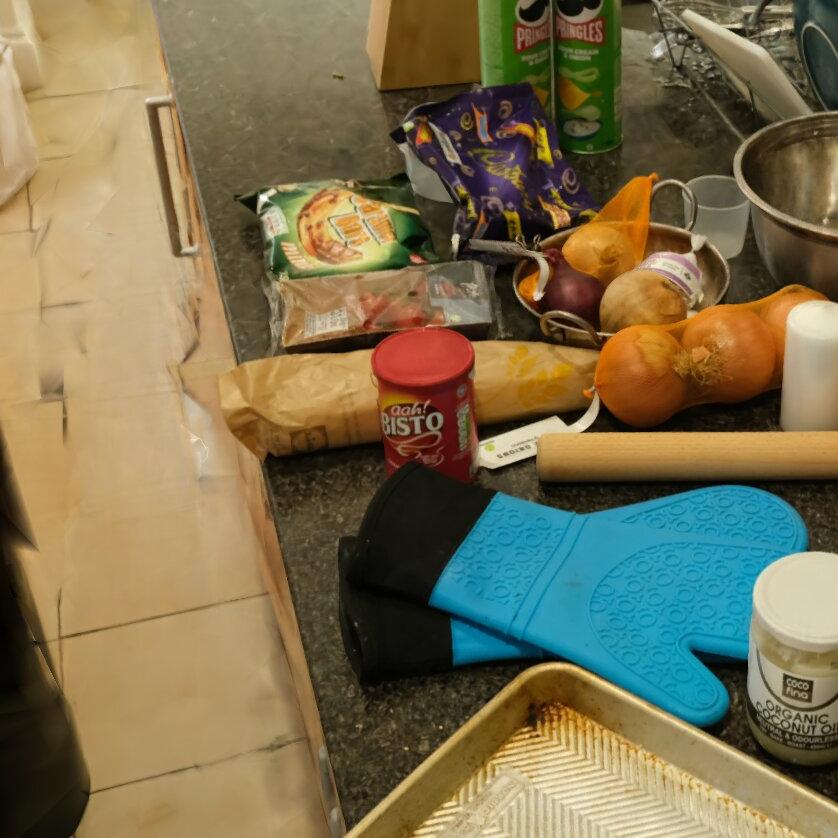} 
\includegraphics[width=0.15\textwidth]{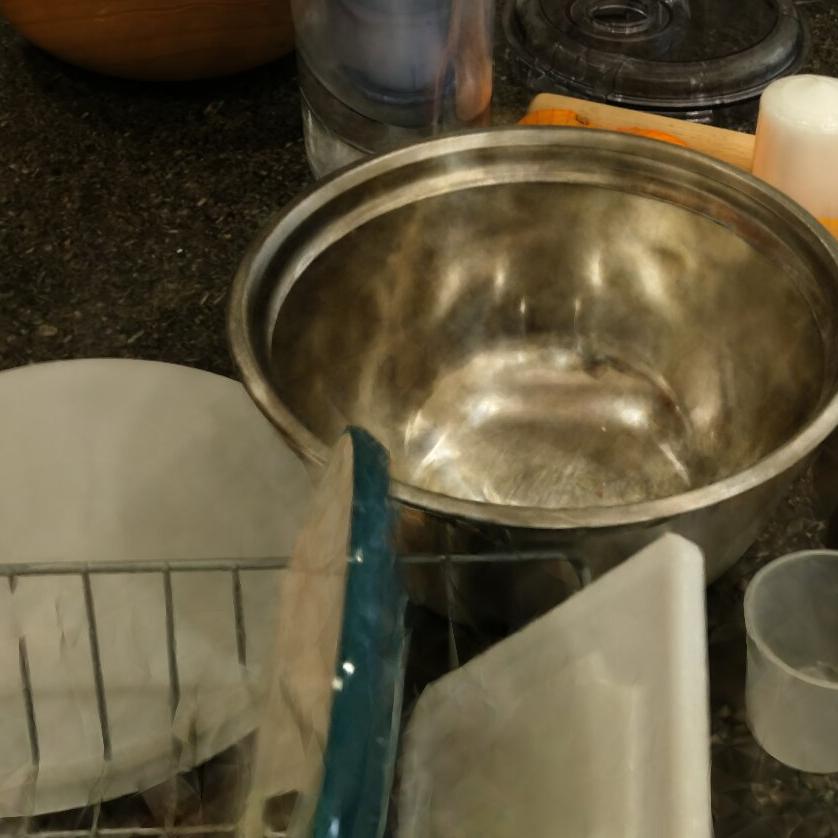}  
&&
\includegraphics[width=0.15\textwidth]{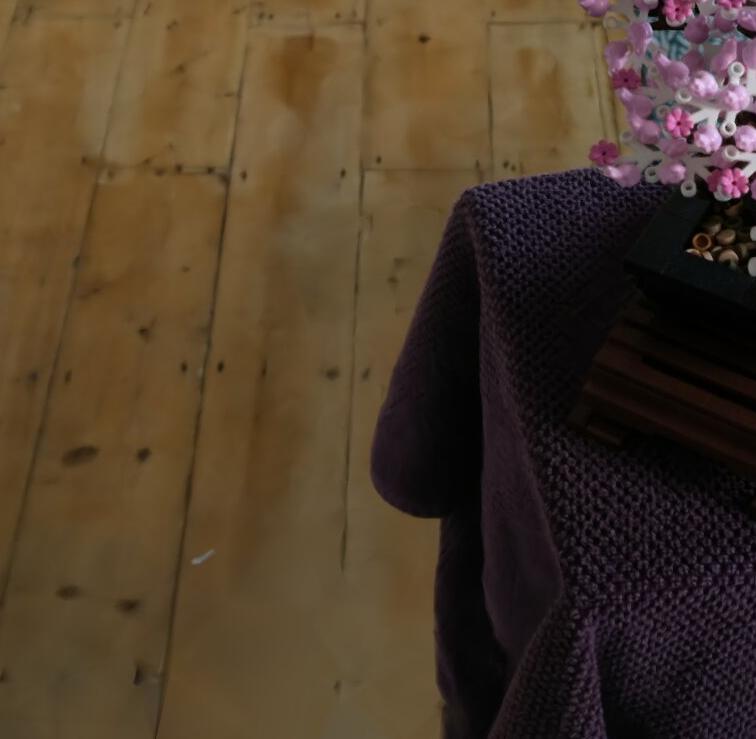} 
\includegraphics[width=0.15\textwidth]{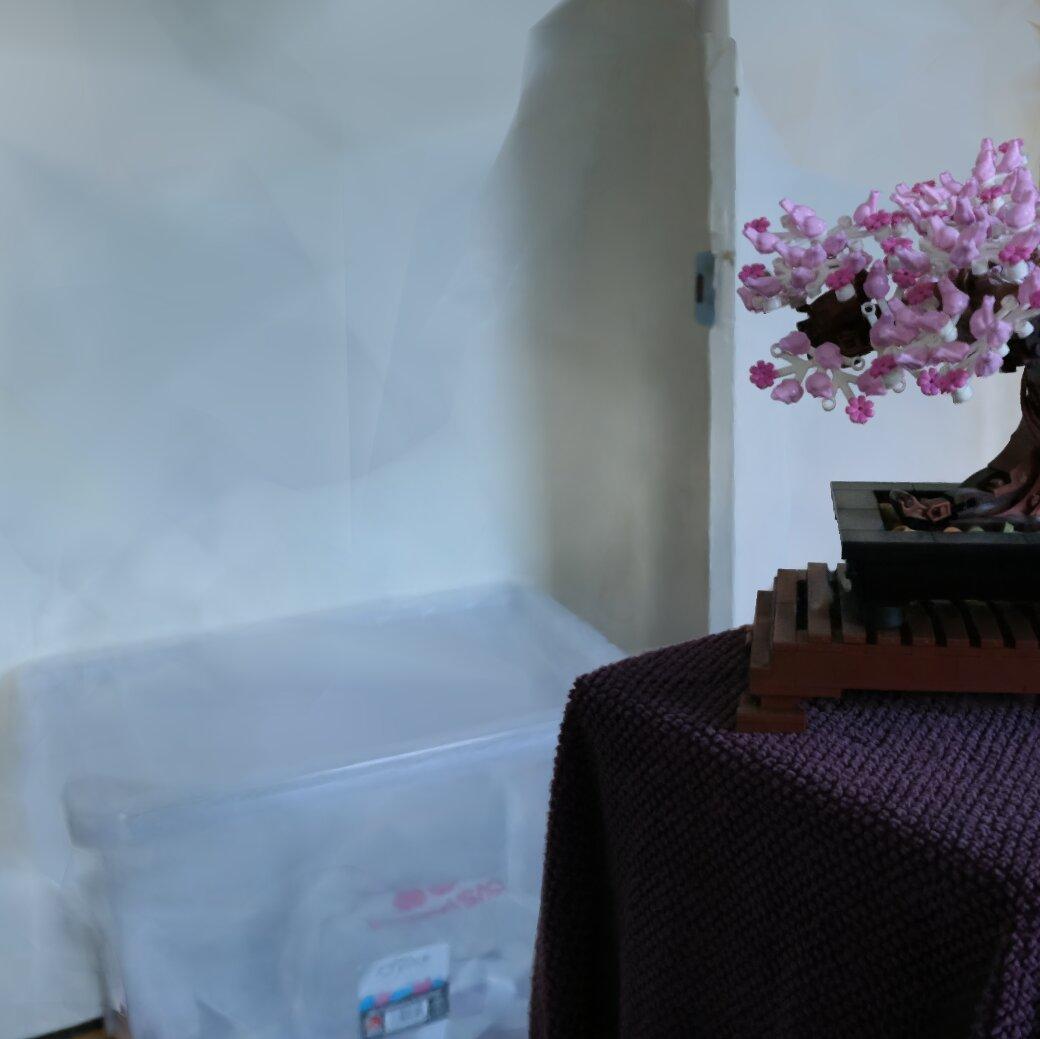}  
&& 
\includegraphics[width=0.15\textwidth]{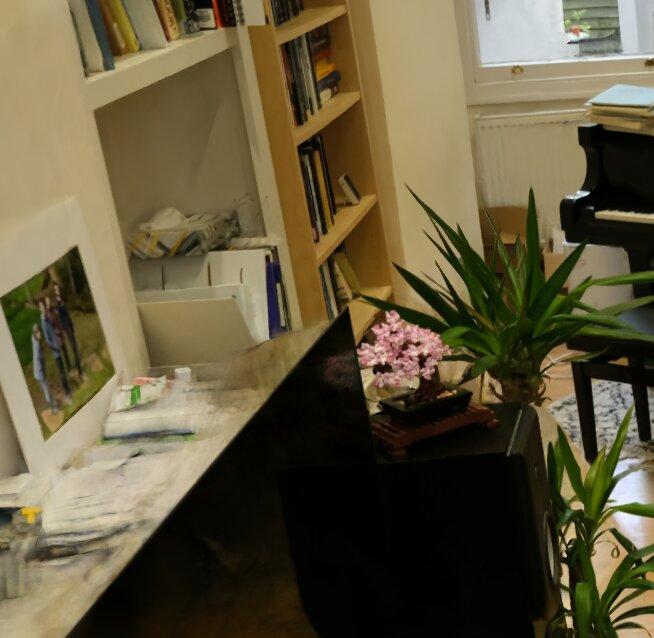} 
\includegraphics[width=0.15\textwidth]{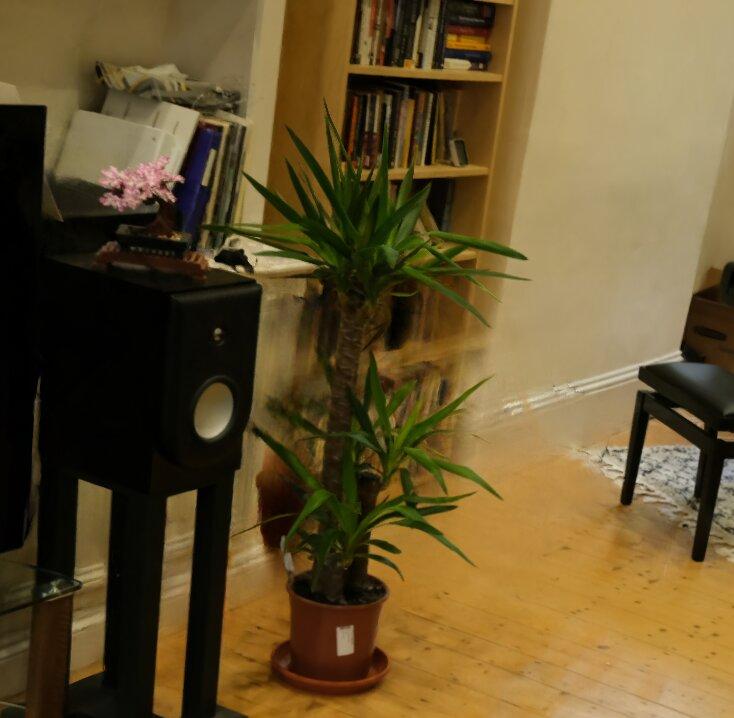}  \\
\rotatebox{90}{\tiny\hspace{12pt}{EKS}} &
\includegraphics[width=0.15\textwidth]{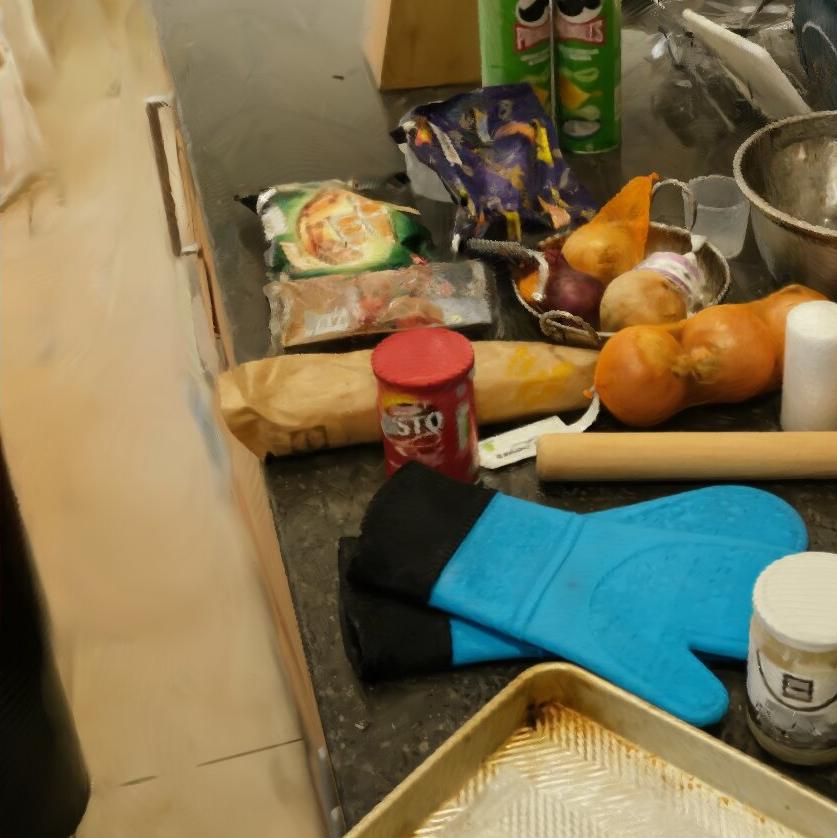}
\includegraphics[width=0.15\textwidth]{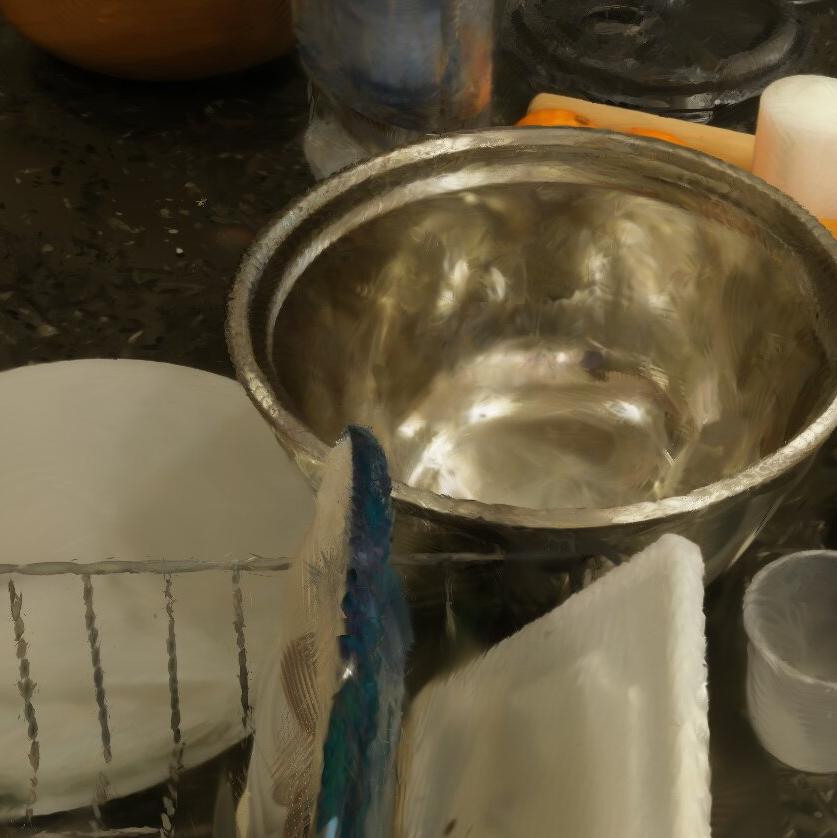}  &&
\includegraphics[width=0.15\textwidth]{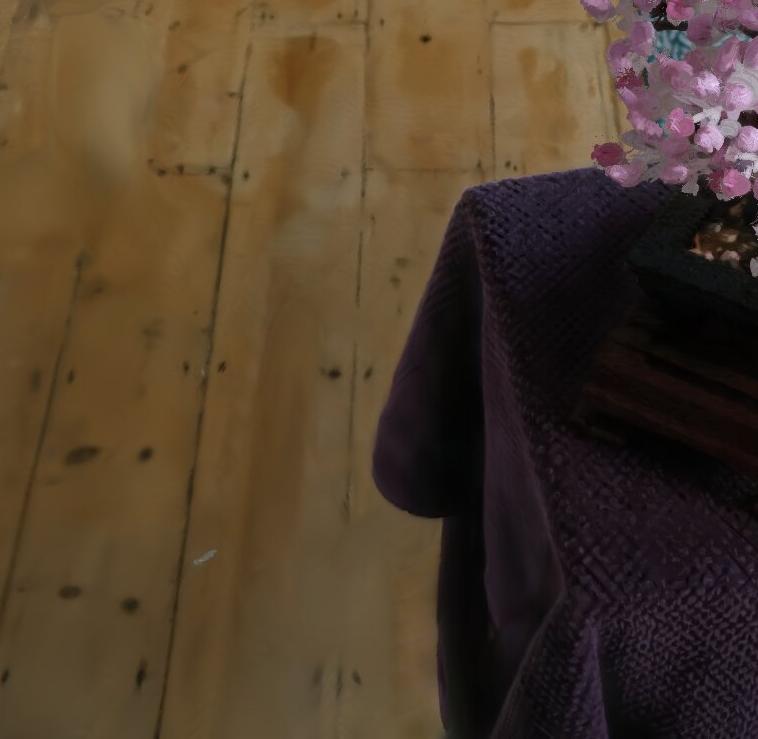}
\includegraphics[width=0.15\textwidth]{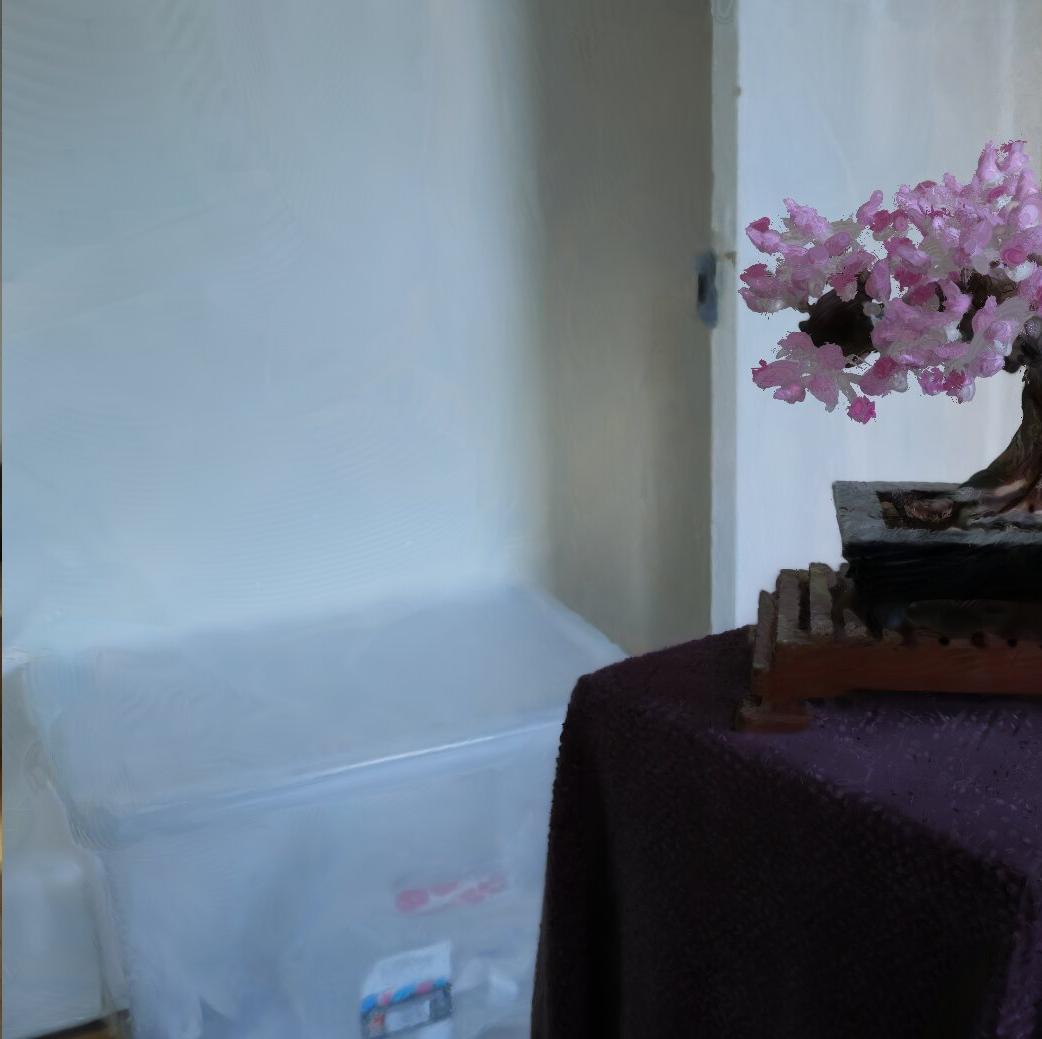}
& &
\includegraphics[width=0.15\textwidth]{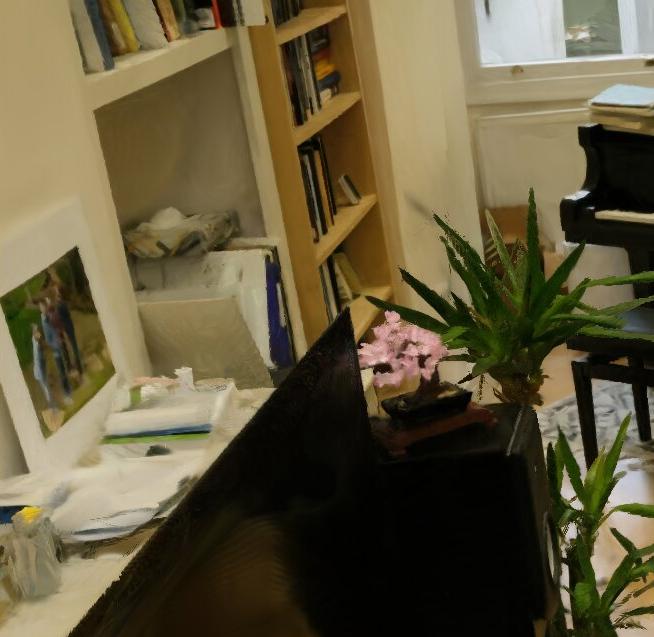}
\includegraphics[width=0.15\textwidth]{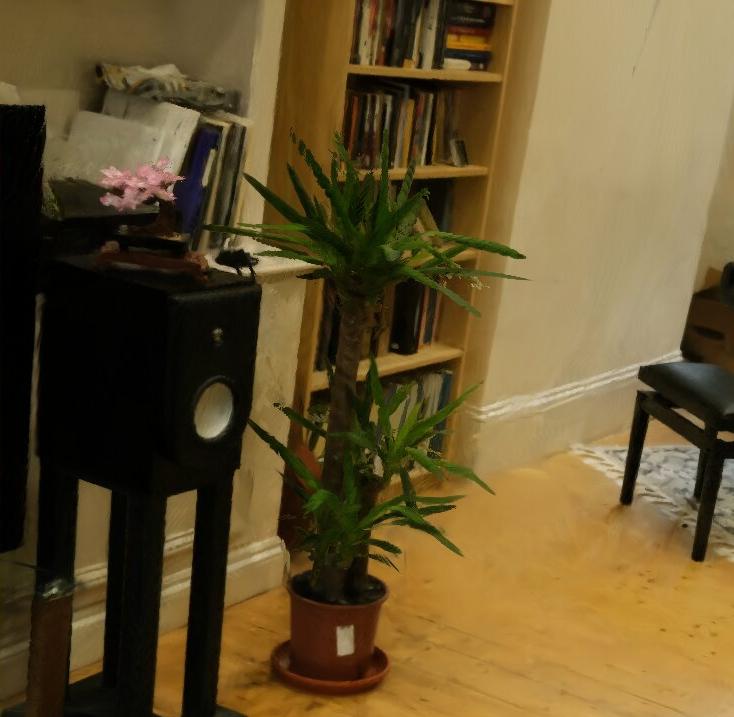} \\
\rotatebox{90}{\tiny\hspace{12pt}{\our{} (our)}} &
\includegraphics[width=0.15\textwidth]{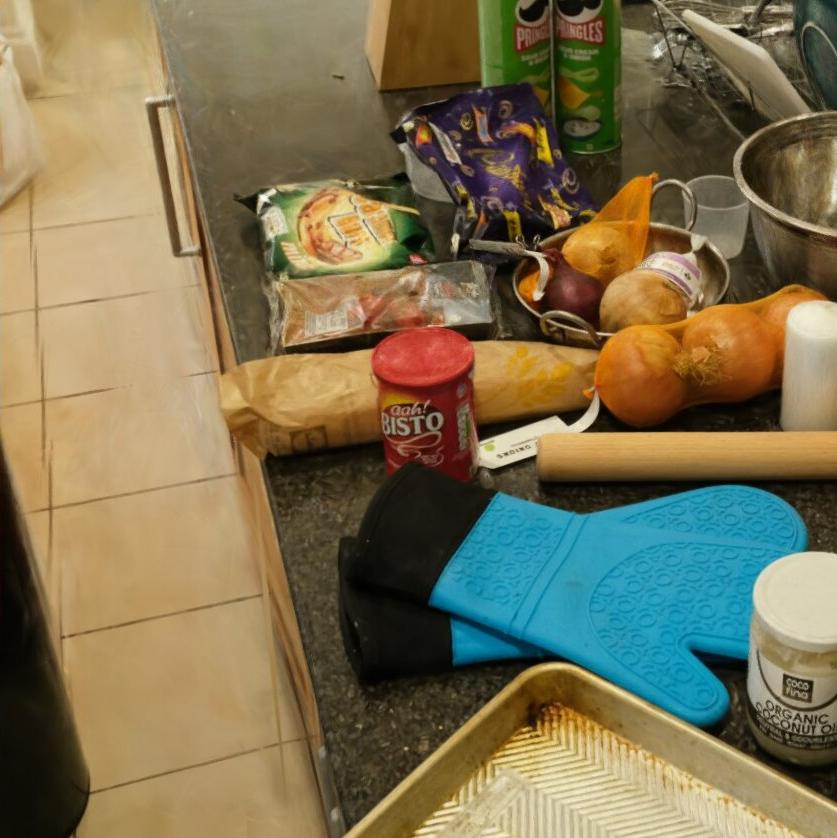}
\includegraphics[width=0.15\textwidth]{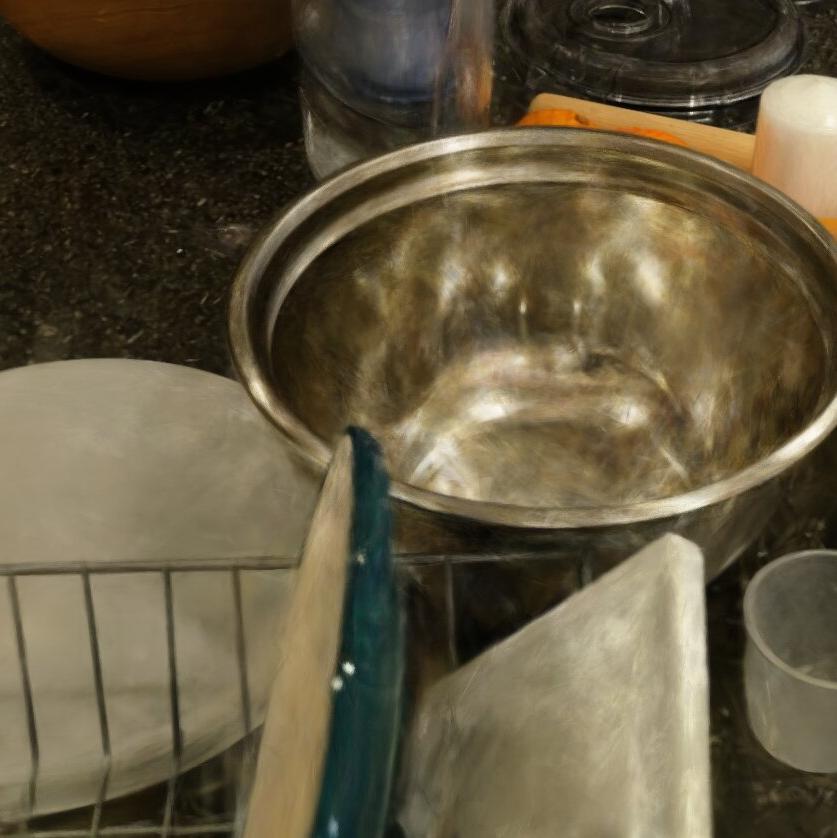}&&
\includegraphics[width=0.15\textwidth]{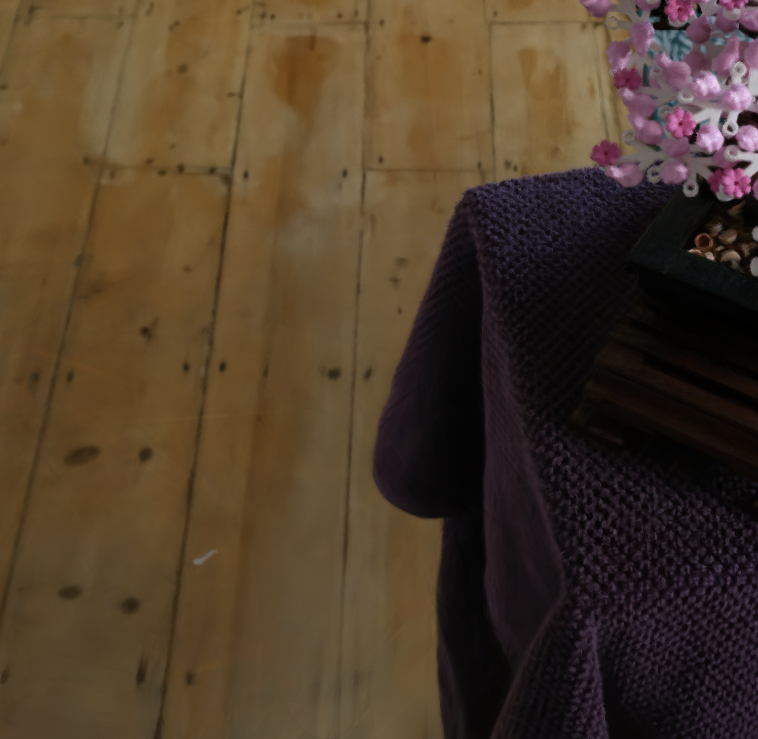}
\includegraphics[width=0.15\textwidth]{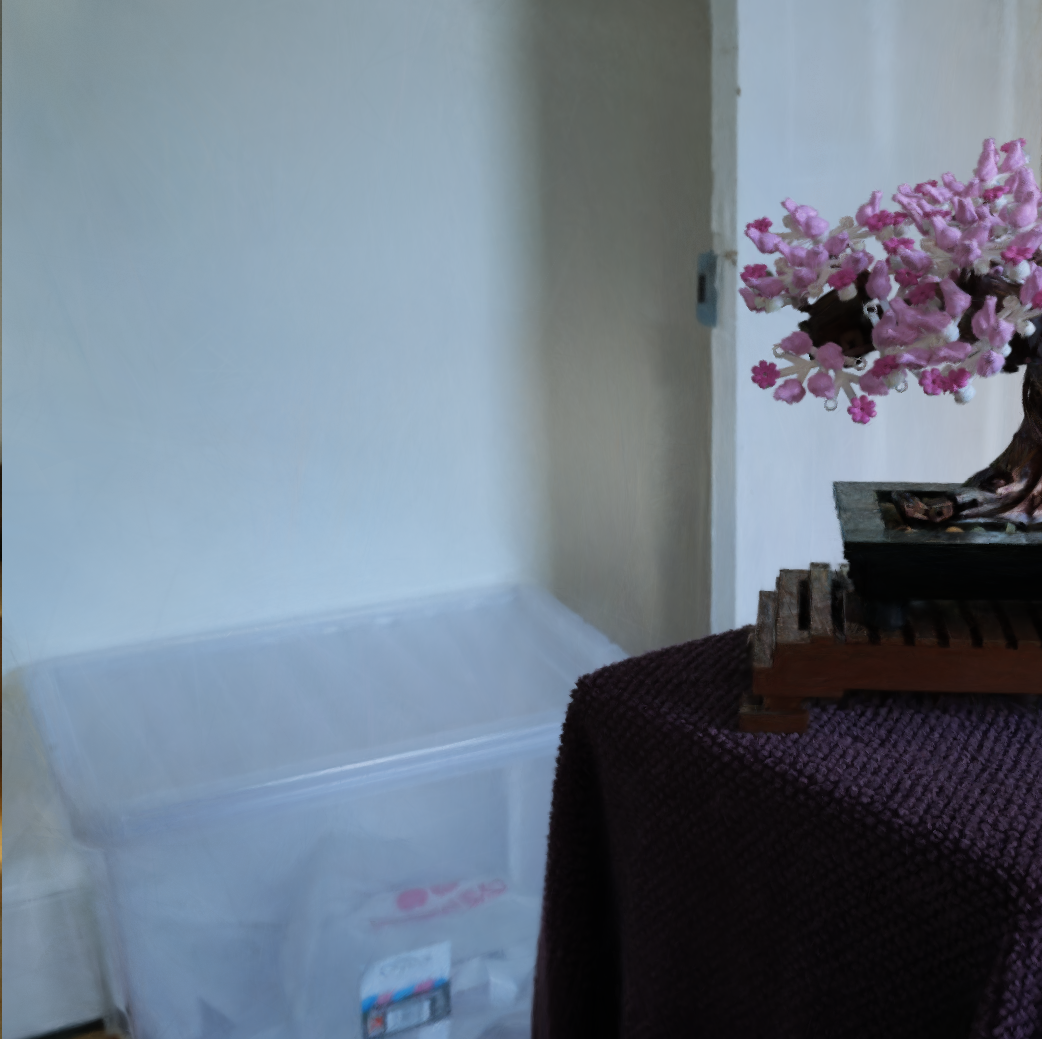} &&
\includegraphics[width=0.15\textwidth]{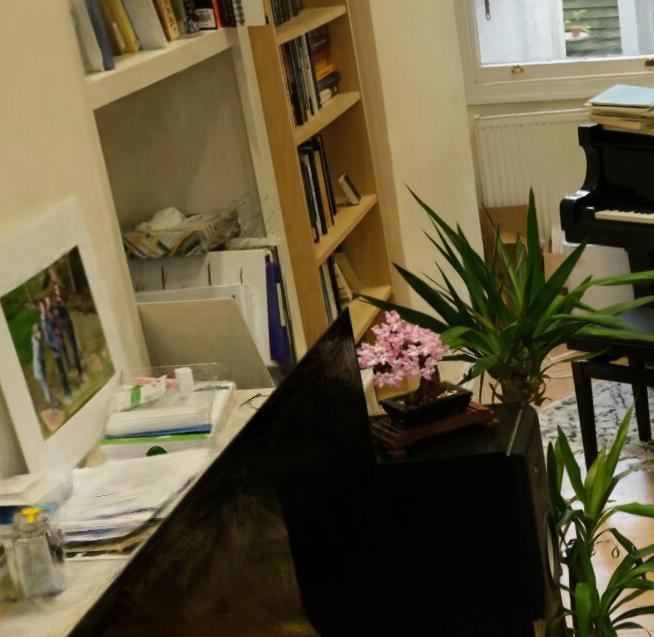}
\includegraphics[width=0.15\textwidth]{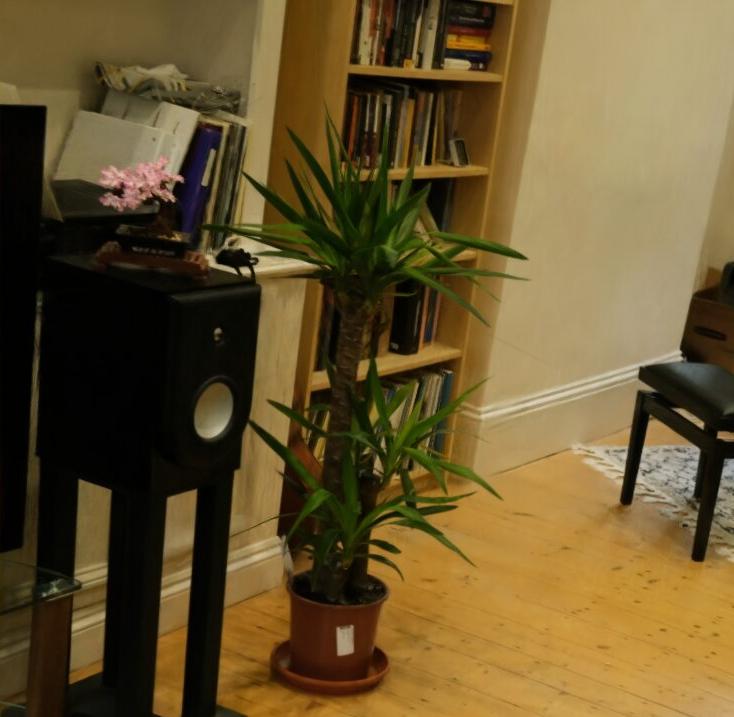} 
\end{tabular}
}

\caption{Qualitative comparison on indoor scenes from the Mip-NeRF 360 dataset. Reconstruction fidelity is evaluated against editable baselines: Radiance Meshes and EKS. It is observed that \our{} consistently achieves superior visual quality, effectively recovering high-frequency details that are blurred or lost in competing methods. 
}
\vspace{-0.5cm}
\label{fig:comp}
\end{figure*}

In this paper, \our{} (Intersection-aware Ray-based Implicit Editable Scenes) is presented, a novel approach that fundamentally rethinks the interaction between rays and neural-encoded Gaussians (see Fig.~\ref{fig:teaser}). It is observed that dense volumetric sampling is inherently inefficient when the underlying geometry is already explicitly defined by Gaussian primitives. Consequently, instead of reliance on blind stochastic sampling, an analytical ray-intersection strategy is employed. By solving the intersection equations for a ray traversing an oriented ellipsoid, the precise locations that require neural queries are identified, thereby aligning the sampling process with the scene's explicit structure (see Fig.~\ref{fig:comp}).

The proposed framework offers three distinct advantages. First, the computational bottleneck of explicit 3D spatial searches is eliminated: local features are aggregated directly along the ray using a sliding-window mechanism, bypassing costly 3D KNN lookups. Second, the efficiency of the neural rendering pipeline is maximized. Rather than relying on stochastic sampling that often queries empty space, the proposed method focuses exclusively on structurally meaningful intersection points. This ensures that every MLP evaluation contributes directly to the surface reconstruction, eliminating the computational redundancy inherent in standard volumetric integration. Finally, the explicit nature of the Gaussian anchors is preserved, ensuring full compatibility with modern physics solvers while simultaneously maintaining the high visual fidelity characteristic of neural fields.
Main contributions are as follows:
\vspace{-0.3cm}
\begin{itemize}
    \item \our{} is introduced, utilizing the Ray Intersection Selector (RIS) to replace heuristic probabilistic sampling with precise geometric intersections. This approach eliminates empty space processing by evaluating only structurally meaningful samples, significantly accelerating rendering throughput.
    \item A robust pipeline for topology-agnostic editing and physics simulation is established. By leveraging explicit Neural Anchors with RIS, rendering integrity is maintained even when objects traverse beyond the initial training volume, overcoming the spatial constraints of traditional grid-based methods.
    \item High-fidelity rendering is achieved without the typical trade-off between editability and quality. The proposed framework outperforms all editable baselines in challenging indoor environments, delivering reconstruction quality competitive with leading static representations.
\end{itemize}

\section{Related Works}

\begin{figure*}[t]
    \centering
    \includegraphics[width=\linewidth]{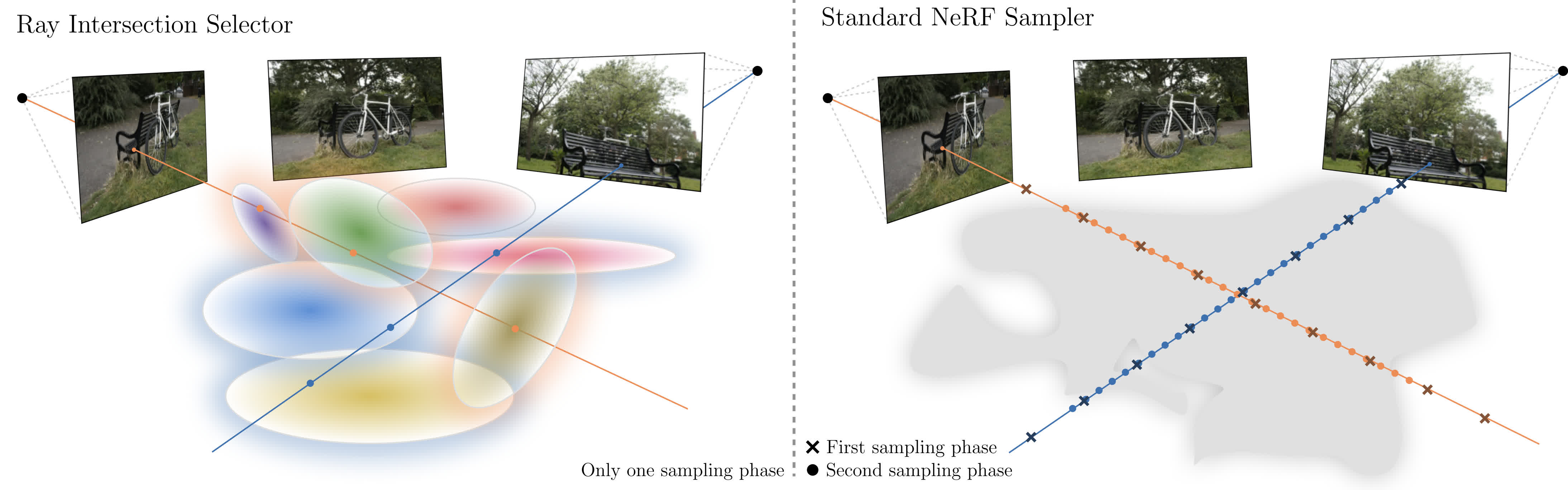}
    \caption{Sampling strategy comparison. The proposed Ray Intersection Selector (RIS) (left) analytically determines precise interaction points with scene primitives, generating meaningful samples in a single pass without processing empty space. In contrast, standard volumetric sampling (right) relies on a heuristic, multi-stage process (coarse and fine) that inherently allocates the computational budget to unoccupied regions.}
    \label{fig:ray_intersection}
    \vspace{-0.5cm}
\end{figure*}

Recent technological advances in 3D reconstruction based on photographs have led to the development of various models, such as NeRF \cite{mildenhall2020nerf} and Gaussian splatting (GS)\cite{kerbl20233d}. Currently, solutions that combine the best features of both methods, using Gaussian distributions as a proxy for guiding neural field evaluations, still suffer from significant computational inefficiency. These methods employ stochastic volumetric sampling for feature aggregation, a process that imposes a severe penalty on rendering speed \cite{zielinski2025genie}. Editing 3D scenes represented by NeRFs has consequently become an active research area. A first line of work focuses on semantic-level editing, including object insertion and removal within learned radiance fields \cite{kobayashi2022decomposing,lazova2023control,weder2023removing}. A complementary direction addresses geometric editing directly within the NeRF framework Conerf \cite{kania2022conerf}, \cite{yuan2023interactive}, Editablenerf \cite{zheng2023editablenerf}. 
RIP-NeRF~\cite{wang2023rip} introduces a novel point-based radiance field representation as the network input.  RIP-NeRF~\cite{wang2023rip} introduces a novel point-based radiance field representation as the network input. NeuralEditor~\cite{chen2023neuraleditor} rendering scheme based on deterministic integration within K-D tree-guided density-adaptive voxels.  NeRF-Editing~\cite{yuan2022nerf} and NeuMesh \cite{yang2022neumesh}, which found correspondence between the extracted explicit mesh representation and the implicit neural representation of the target scene. This enables the use of mesh editing tools for NeRF model editability. ~\cite{hofherr2023neural} propose to combine neural implicit representations for appearance modeling with neural ordinary differential equations (ODEs) for modelling physical phenomena to obtain a dynamic scene representation that can be identified directly from visual observations. Neuphysics \cite{qiao2022neuphysics} used a differentiable physics simulator to connect the neural field and its corresponding hexahedral mesh.

In the context of 3DGS-based models such as SuGaR \cite{guedon2023sugar}, similar to them we also can use an estimated mesh that allows for the editing of Gaussian representations. GASP \cite{2024gasp} uses flat Gaussians as the base model, which are then parameterised using a soup of triangles. Editing resembling a triangle-based method rather than a point-based method. In contrast, PhysGaussian \cite{xie2024physgaussian} operates directly on Gaussian distributions, whereas DreamPhysics \cite{huang2024dreamphysics} infers the physical attributes of 3DGS based on video diffusion priors. Finally, Radiance Meshes~\cite{radiance_meshes} represent radiance fields using constantly density tetrahedral cells, enabling exact and fast volume rendering via hardware-native rasterization. Their use of a Zip-NeRF-style~\cite{barron2023zipnerf} backbone allows for robust optimization, supporting real-time view synthesis as well as editing and physics-based simulations.

\begin{figure*}[t]
\centering
\setlength{\tabcolsep}{1pt}
{\fontsize{6.8pt}{11pt}\selectfont
\begin{tabular}{ccccccc}
 & GT & TS & 3DGS & INGP   & MipNeRF-360 & \our{} (our) \\
\rotatebox{90}{\footnotesize\hspace{8pt}{Bonsai}} &
\includegraphics[width=0.15\textwidth]{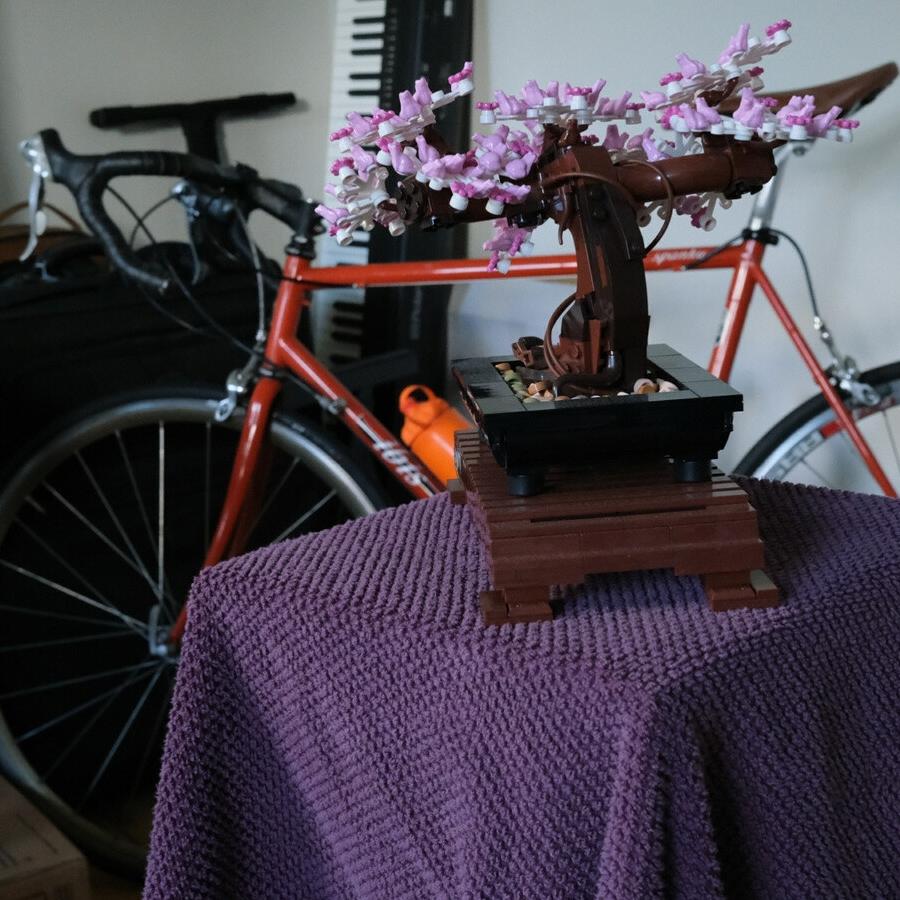}   & \includegraphics[width=0.15\textwidth]{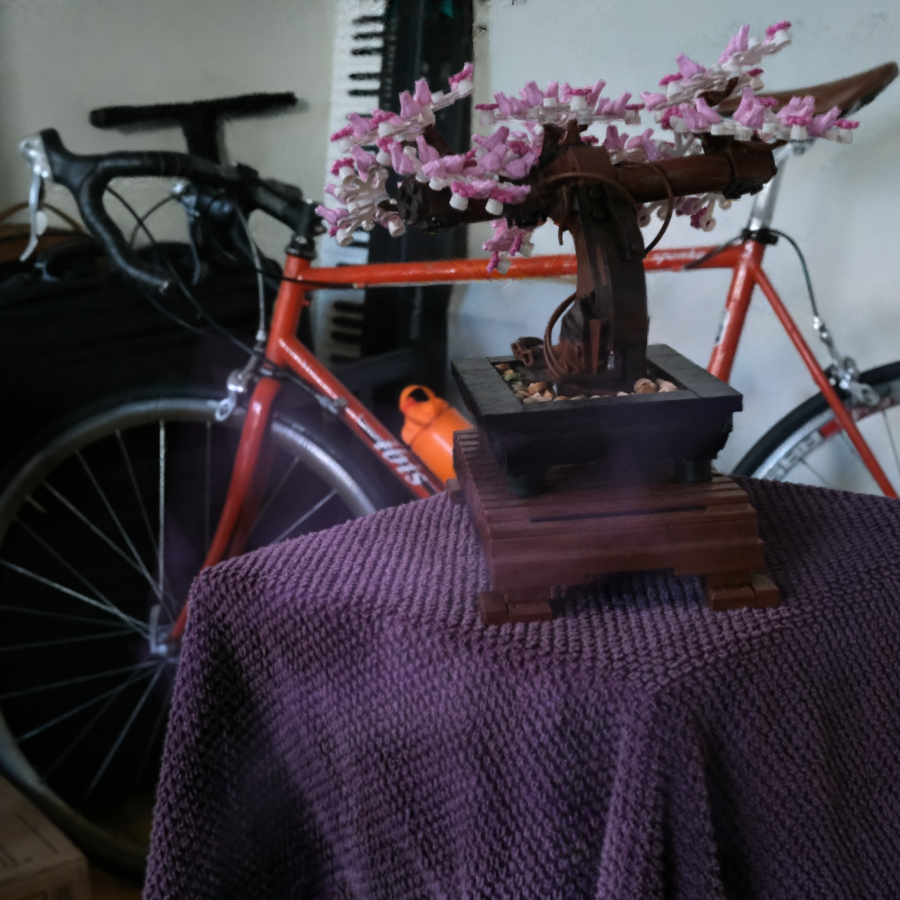}  & 
\includegraphics[width=0.15\textwidth]{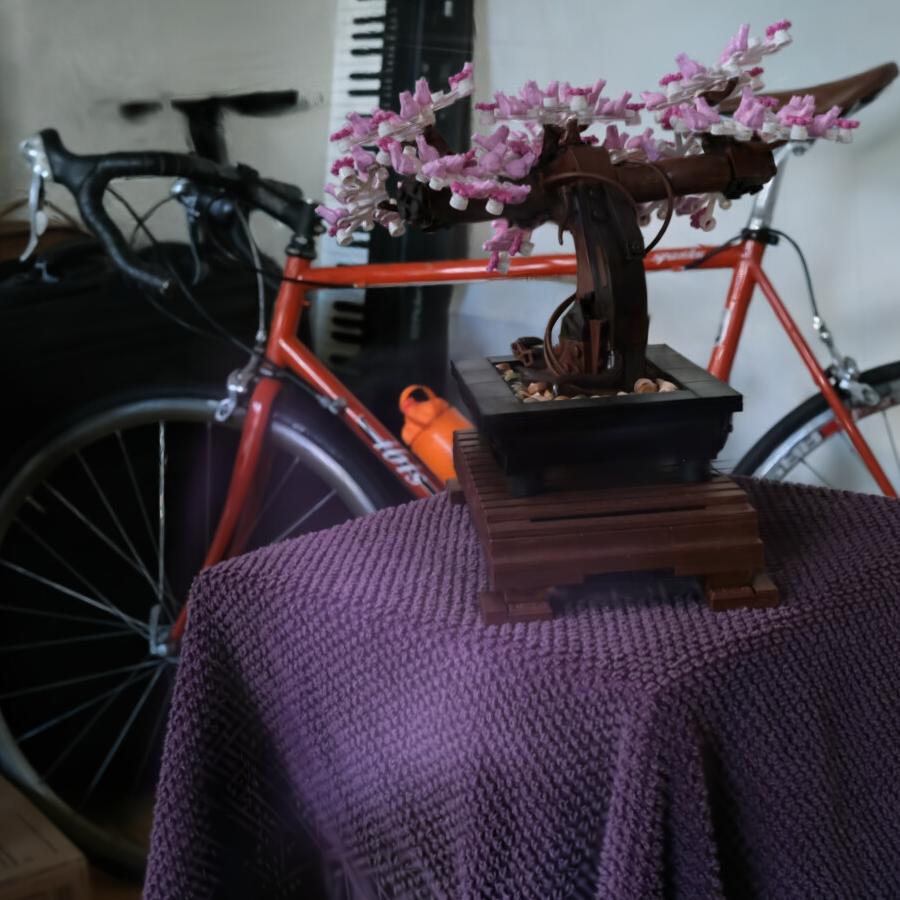}    &  \includegraphics[width=0.15\textwidth]{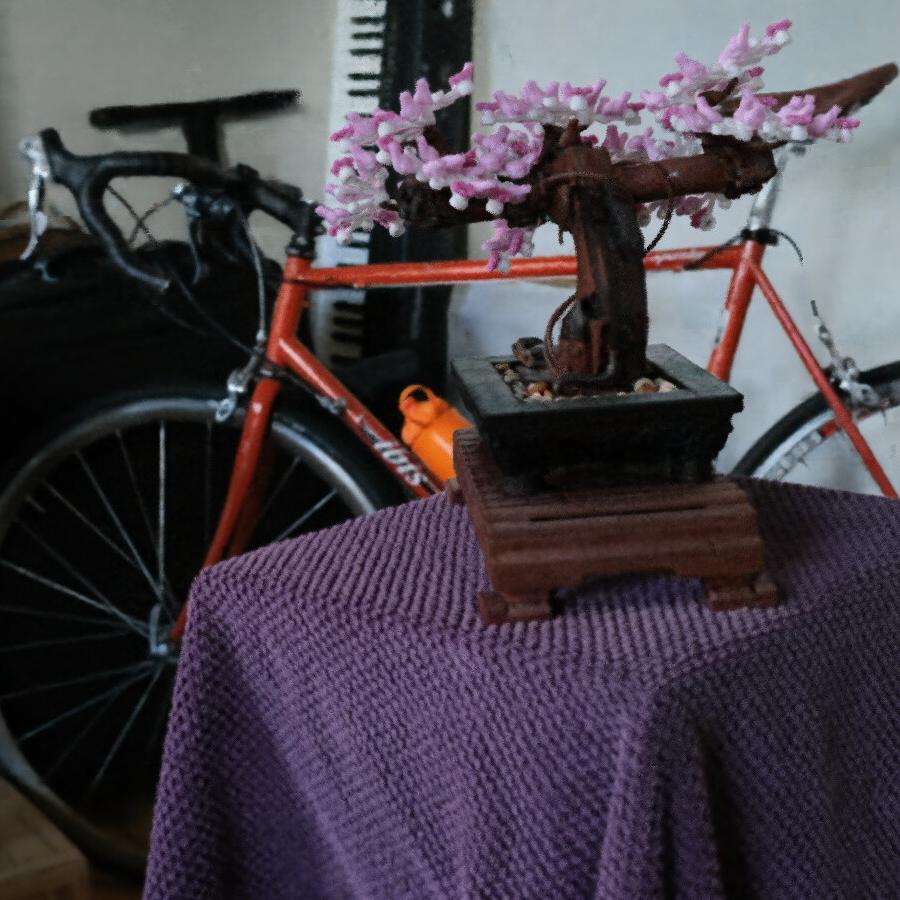}  &
\includegraphics[width=0.15\textwidth]{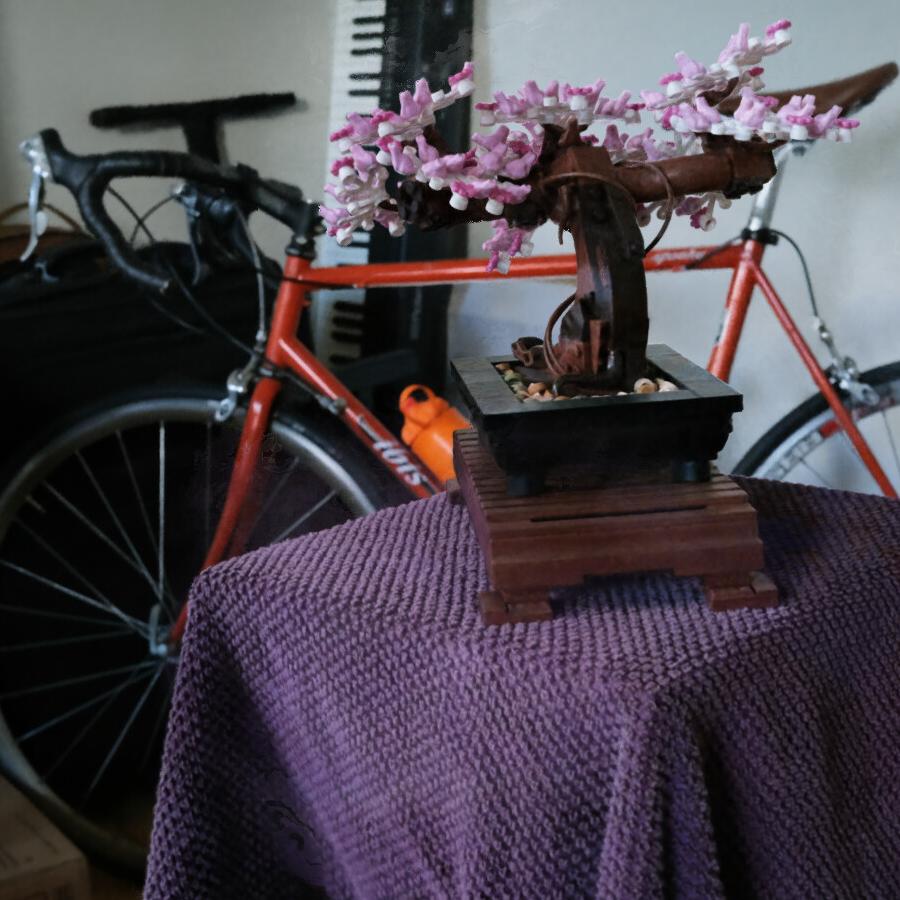}    & 
\includegraphics[width=0.15\textwidth]{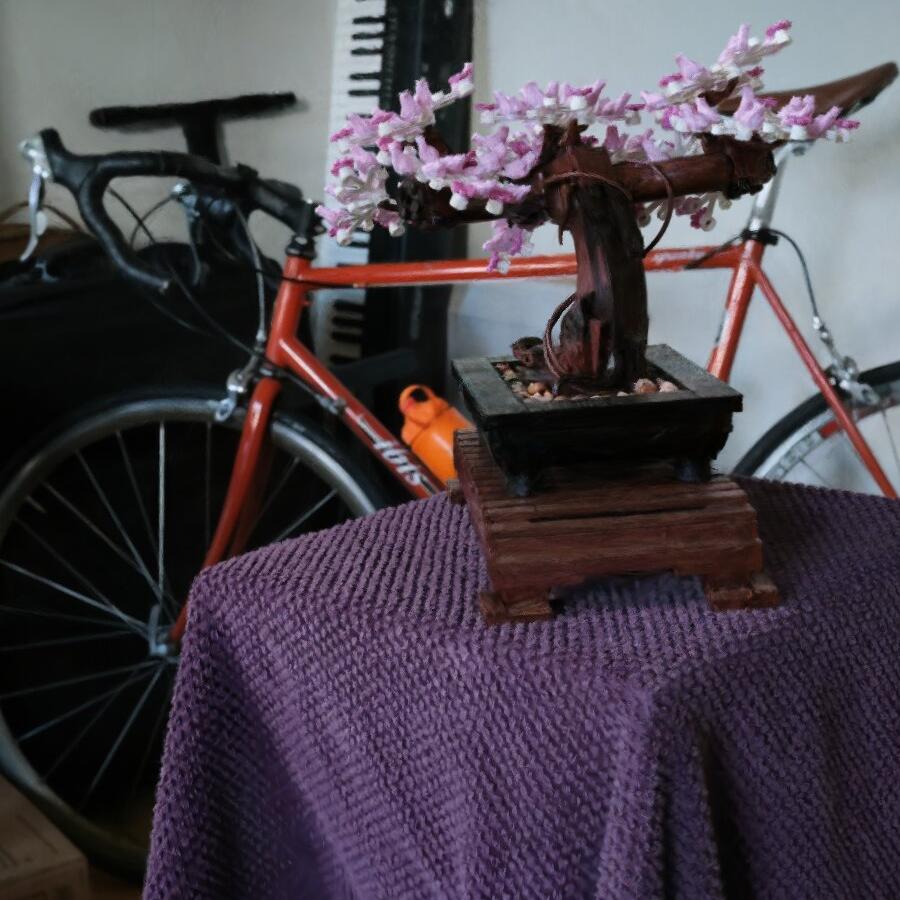}   \\
\rotatebox{90}{\footnotesize\hspace{8pt}{Bicycle}} &\includegraphics[width=0.15\textwidth]{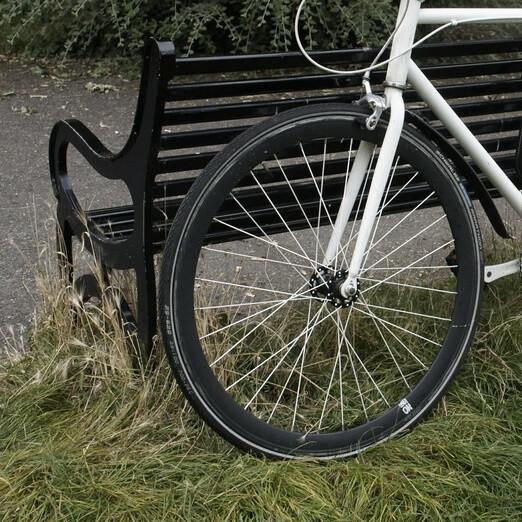}   & \includegraphics[width=0.15\textwidth]{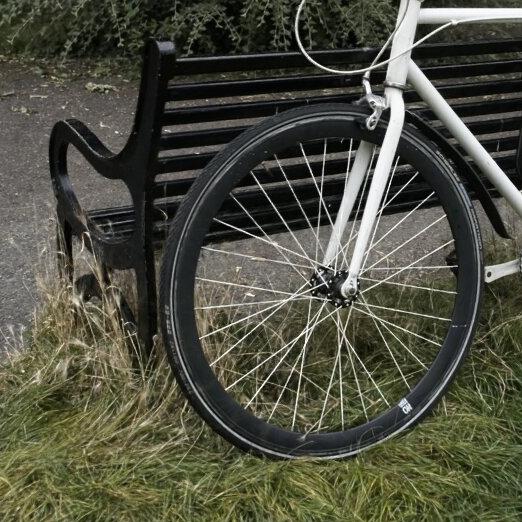}  & 
\includegraphics[width=0.15\textwidth]{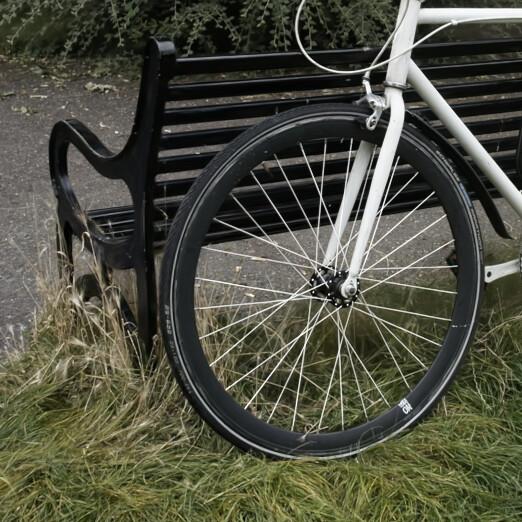}    &  \includegraphics[width=0.15\textwidth]{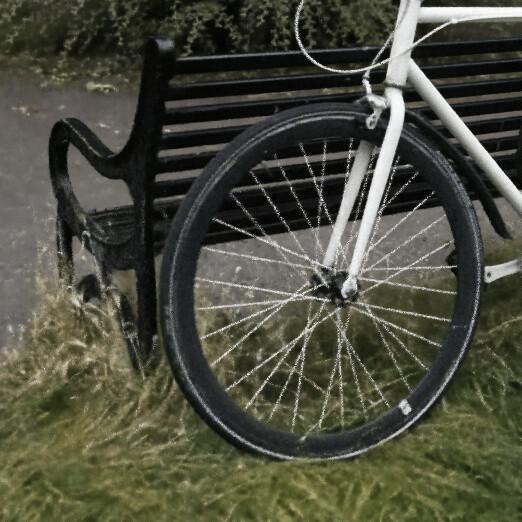}  &
\includegraphics[width=0.15\textwidth]{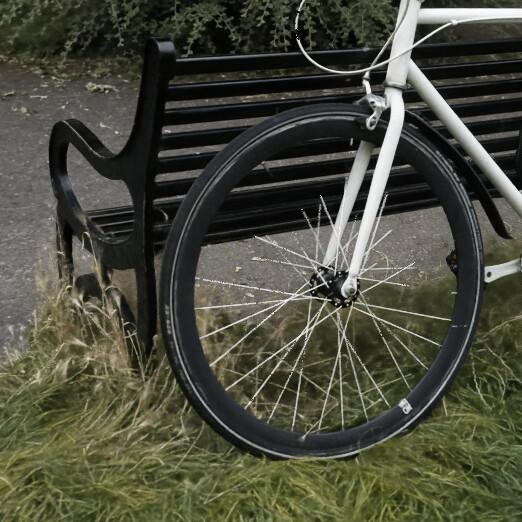}    & 
\includegraphics[width=0.15\textwidth]{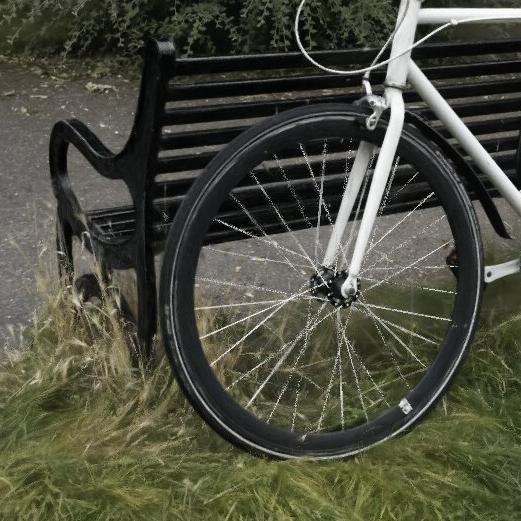}   \\
\rotatebox{90}{\footnotesize\hspace{12pt}{Truck}} & \includegraphics[width=0.15\textwidth]{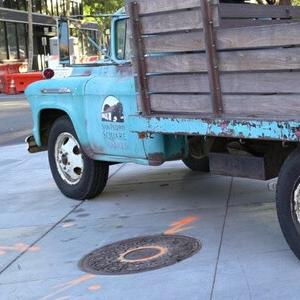}   & \includegraphics[width=0.15\textwidth]{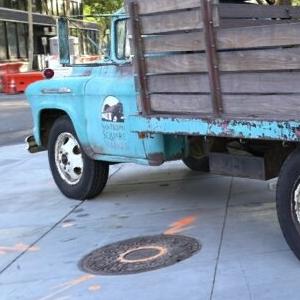}  & 
\includegraphics[width=0.15\textwidth]{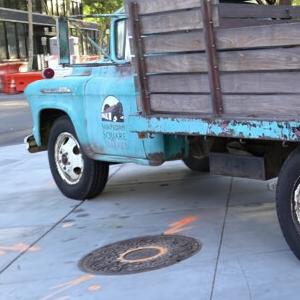}    &  \includegraphics[width=0.15\textwidth]{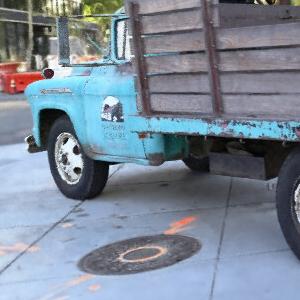}  &
\includegraphics[width=0.15\textwidth]{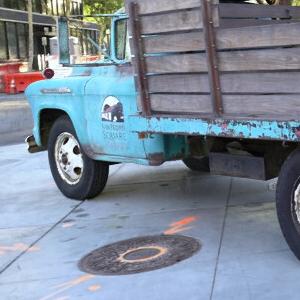}    & 
\includegraphics[width=0.15\textwidth]{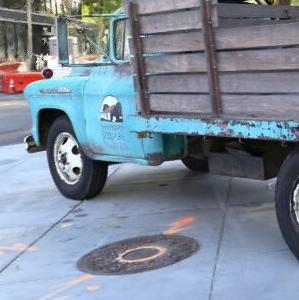}   \\
\rotatebox{90}{\footnotesize\hspace{12pt}{Train}} &\includegraphics[width=0.15\textwidth]{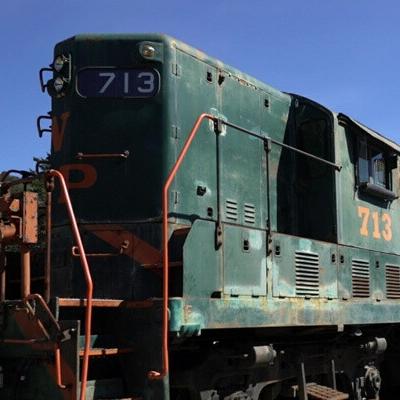}   & \includegraphics[width=0.15\textwidth]{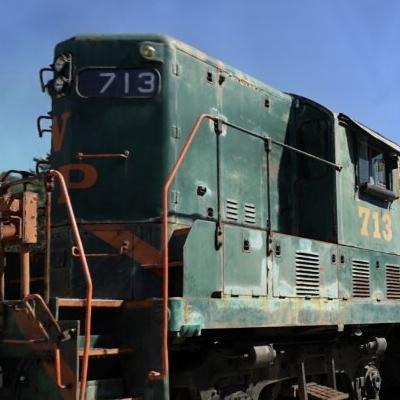}  & 
\includegraphics[width=0.15\textwidth]{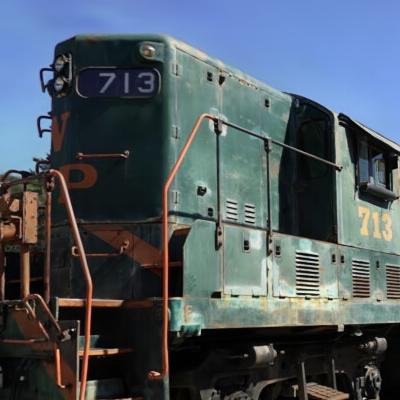}    &  \includegraphics[width=0.15\textwidth]{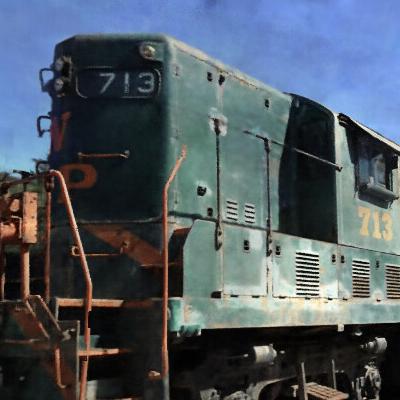}  &
\includegraphics[width=0.15\textwidth]{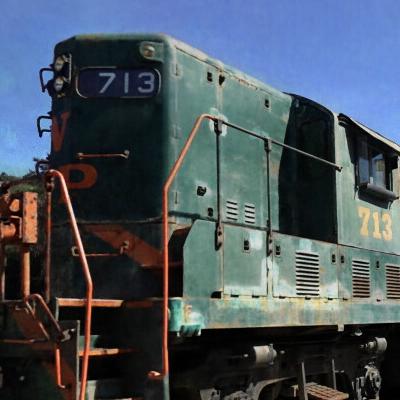}    & 
\includegraphics[width=0.15\textwidth]{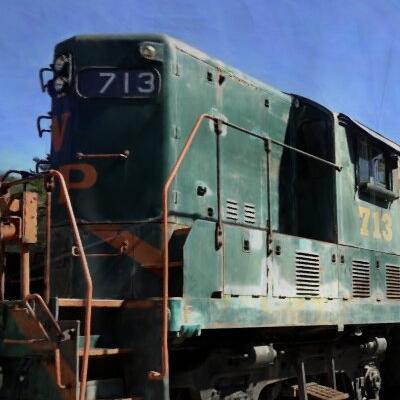}   
\end{tabular}}
\caption{Qualitative comparison against static state-of-the-art baselines on complex real-world scenes. TS denotes Triangle Splatting~\cite{triangle_splatting}. It is observed that \our{} achieves reconstruction fidelity highly competitive with leading static representations such as 3DGS or Mip-NeRF 360. Despite the additional architectural constraints required for editability, fine geometric details-such as the thin structures in the \textit{Bicycle} scene and the surface textures of the \textit{Truck}-are preserved with precision. 
}
\vspace{-0.7cm}
\label{fig:comparisionmain}
\end{figure*}

\vspace{-0.2cm}
\section{\our{} - model}
The proposed framework is designed as a hybrid representation that unifies the efficiency of explicit geometry with the continuity of implicit neural fields. An overview of the pipeline is illustrated in Fig.~\ref{fig:pipeline}. The method operates by parameterizing the scene as a set of Neural Anchors-explicit 3D Gaussian primitives serving as spatial carriers for learnable latent features. To render the scene, the Ray Intersection Selector efficiently identifies relevant interaction points, bypassing dense volumetric sampling. Subsequently, a continuous latent field is synthesized using the Ray-Coherent Aggregation mechanism and decoded by a neural network into volumetric properties, which are then accumulated to produce the final image.

\textbf{Hybrid Gaussian-Neural Representation}
The scene is parametrized as a set of unstructured Neural Anchors, formally defined as a collection of $N$ primitives 
$$
\boldsymbol{\mathcal{P}}=\{ ( \boldsymbol{\mathcal{N}}(\boldsymbol{\mu}_i, \boldsymbol{\Sigma}_i), \omega_i,\boldsymbol{f}_i ) \}_{i=1}^{N}.
$$
This architecture utilizes explicit primitives as spatial anchors for the implicit scene representation, thereby binding neural features to movable geometry to facilitate editing.

The geometric structure of each anchor is represented by 3D normal distribution. Specifically, its spatial extent is parametrized by a mean position $\boldsymbol{\mu}_i\in\mathbb{R}^3$ and a 3D covariance matrix $\boldsymbol{\Sigma}_i$. 
The geometric influence of the \textit{i}-th anchor at any spatial point $\boldsymbol{x}$ is determined by the following Gaussian density $\boldsymbol{\mathcal{N}}(\boldsymbol{x}; \boldsymbol{\mu}_i, \boldsymbol{\Sigma}_i )$.
To allow for valid optimization, the covariance matrix $\boldsymbol{\Sigma}_i$ is factorized into a rotation matrix $\boldsymbol{R}_i$ stored as a quaternion and a scaling vector $\boldsymbol{S}_i$. Additionally, a logit parameter $\omega_i\in\mathbb{R}$ is maintained to control the opacity and existence probability of each primitive.

Beyond geometry, each anchor carries a high-dimensional latent feature vector $\boldsymbol{f}_i\in\mathbb{R}^D$. This vector serves as a localized neural embedding that encodes compressed scene properties, such as texture or material density. These embeddings are designed to be decoded by a Multi-Layer Perceptron (MLP), effectively allowing the discrete anchors to support a continuous neural field representation.

\begin{figure*}[t]
    \centering
    \includegraphics[width=1\linewidth]{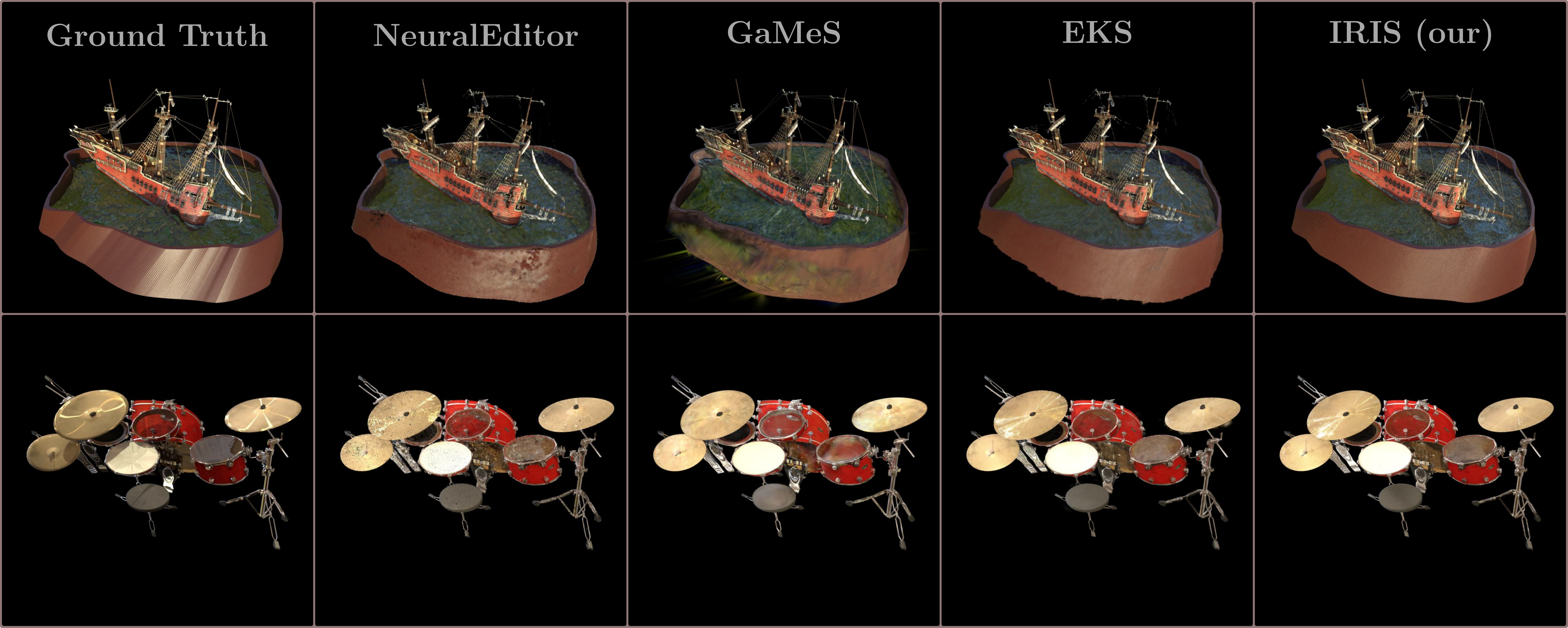}
    \caption{Qualitative comparison on the NeuralEditor deformation benchmark. The rendered views demonstrate the capability of \our{} to maintain high visual fidelity under significant geometric deformation. It is observed that while baselines such as GaMeS and NeuralEditor exhibit artifacts of texture degradation, \our{} produces crisp, photorealistic renderings with preserved view-dependent effects (e.g., specular highlights on the cymbals).}
    \vspace{-0.5cm}
    \label{fig:edit_big_comparison}
\end{figure*}

To ensure high-fidelity reconstruction, the feature vectors are initialized using a multi-resolution $\mathcal{H}_{enc}$ Hash Grid strategy. During the training phase, $\boldsymbol{f}_i$ is queried from the grid based on the anchor's position $\boldsymbol{\mu}_i$, enforcing spatial consistency. For inference, this dependency is severed, and the optimized features are explicitly stored within the primitives. This ensures that the neural features remain strictly attached to their respective anchors during deformation, enabling topology-agnostic editing without grid warping.

\begin{wraptable}{r}{0.5\textwidth}
\vspace{-1.2cm}
    \centering
    \caption{Quantitative analysis. TS: Triangle Splatting, RM: Radiance Meshes ($^\star$re-evaluated). \our{} achieves state-of-the-art fidelity on Deep Blending and competitive performance on Tanks\&Temples, confirming its overall robustness against all baselines.}
    \label{tab:db_tandt}
    \setlength{\tabcolsep}{2.0pt} 
    \resizebox{\linewidth}{!}{%
        {\fontsize{6.8pt}{11pt}\selectfont
        \begin{tabular}{lcc|cc|c}
         & \multicolumn{2}{c|}{Deep Blending} & \multicolumn{2}{c|}{Tanks\&Temples} & \\
         & drjohnson & playroom & \makebox[0.9cm]{train} & truck & Avg.\\
        \hline
        \multicolumn{6}{c}{Static} \\
        \hline
        INGP & \cd 28.26 & \cd 21.67 & \cc 20.46 & \cd 23.38 & \cd 23.44 \\
        MNeRF360 & \ca 29.14 & \cb 29.66 & \cd 19.52 & \cc 24.91 & \cc 25.81 \\
        3DGS & \cc 28.77 & \ca 30.04 & \cb 21.10 & \ca 25.19 & \ca 26.27 \\
        TS & \cb 28.91 & \cc 29.35 & \ca 21.33 & \cb 24.94 & \cb 26.13 \\
        \hline
        \multicolumn{6}{c}{Editable} \\
        \hline
        RM$^\star$ & \cd 28.96 & \cb 30.19 & \ca 21.72 & \ca 24.70 & \ca 26.39 \\
        EKS & \cb 29.22 & \cd 29.67 & \cd 20.06 & \cd 22.49 & \cd 25.36 \\
        \our{} & \ca 29.33 & \ca 30.32 & \cb 21.46 & \cb 24.01 & \cb 26.28\\
        \hline
        \end{tabular}
        }
    }
\vspace{-0.5cm}
\end{wraptable}

\textbf{Ray Intersection Selector (RIS)}
Efficient volumetric rendering relies heavily on identifying high-density regions to allocate samples effectively. To achieve this, we introduce the Ray Intersection Selector (RIS), which reduces the sampling process to a precise geometric intersection problem rather than a probabilistic estimation. By leveraging the explicit nature of Neural Anchors, RIS completely eliminates the need for heuristic geometry searching or auxiliary networks.

This analytical approach stands in stark contrast to existing methods, which typically rely on heuristic or stochastic means, often introducing significant computational overhead. For instance, Nerfacto \cite{nerfacto} employs a Proposal Sampler that requires querying an auxiliary neural network to estimate density distributions prior to the main rendering pass. Similarly, Instant-NGP \cite{instantngp} uses an Occupancy Grid to skip empty regions, but this requires ray marching through a voxel grid and periodic queries of the density field to update occupancy bits. Furthermore, classical approaches such as Mip-NeRF \cite{mipnerf} rely on hierarchical sampling (coarse-to-fine), thereby requiring two passes of volume integration per pixel. Crucially, by decoupling the sampling domain from static global structures, our method inherently enables rendering objects that have been manipulated and moved entirely outside the established training bounding box.

To determine sample positions efficiently, an analytical intersection strategy is employed rather than probabilistic sampling. By leveraging the OptiX API, a proxy geometry approach inspired by 3DGRT~\cite{moenne20243d} is adopted, where Gaussians are conservatively bounded by affine-transformed icosahedrons. Consequently, the heavy intersection logic is offloaded to dedicated RT cores, allowing for precise identification of maximum density points. Detailed shader logic and buffer management strategies are provided in the Appendix.

\begin{figure}[t]
    \centering
    
    \begin{minipage}{0.48\linewidth}
        \centering
        \includegraphics[width=0.94\linewidth]{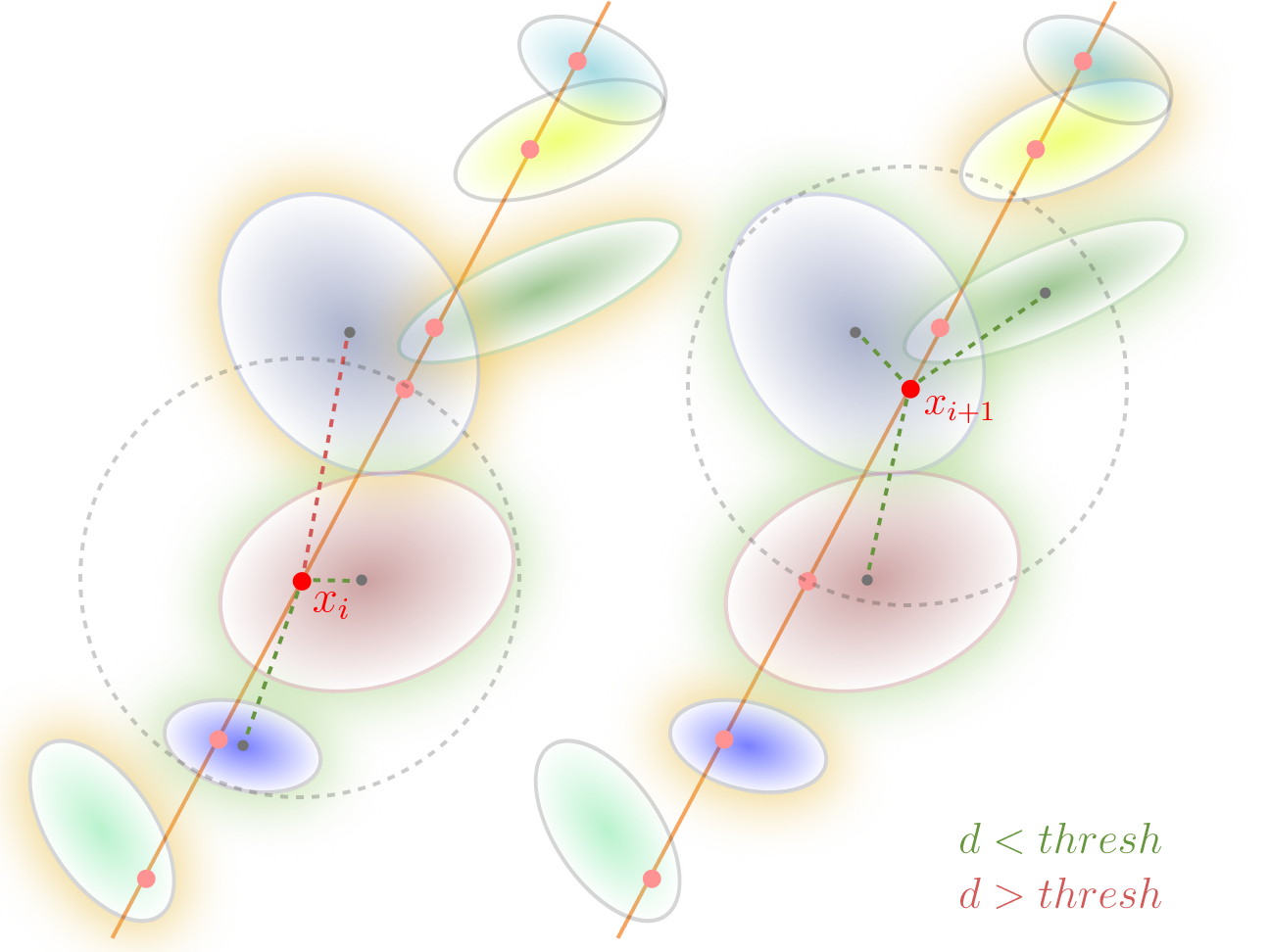}
        \vspace{1mm}
        \captionof{figure}{KNN mechanism. Illuminated primitives represent the sliding window. \textbf{Left}: At spatial discontinuity $\boldsymbol{x}_i$, distant candidates are masked; only spatially consistent neighbors (green) contribute to aggregation. \textbf{Right}: On the continuous surface $\boldsymbol{x}_{i+1}$, all candidates within the window satisfy proximity constraints and are utilized.}
        \label{fig:knn}
    \end{minipage}
    \hfill
    \begin{minipage}{0.48\linewidth}
        \centering
        \includegraphics[width=0.9\linewidth]{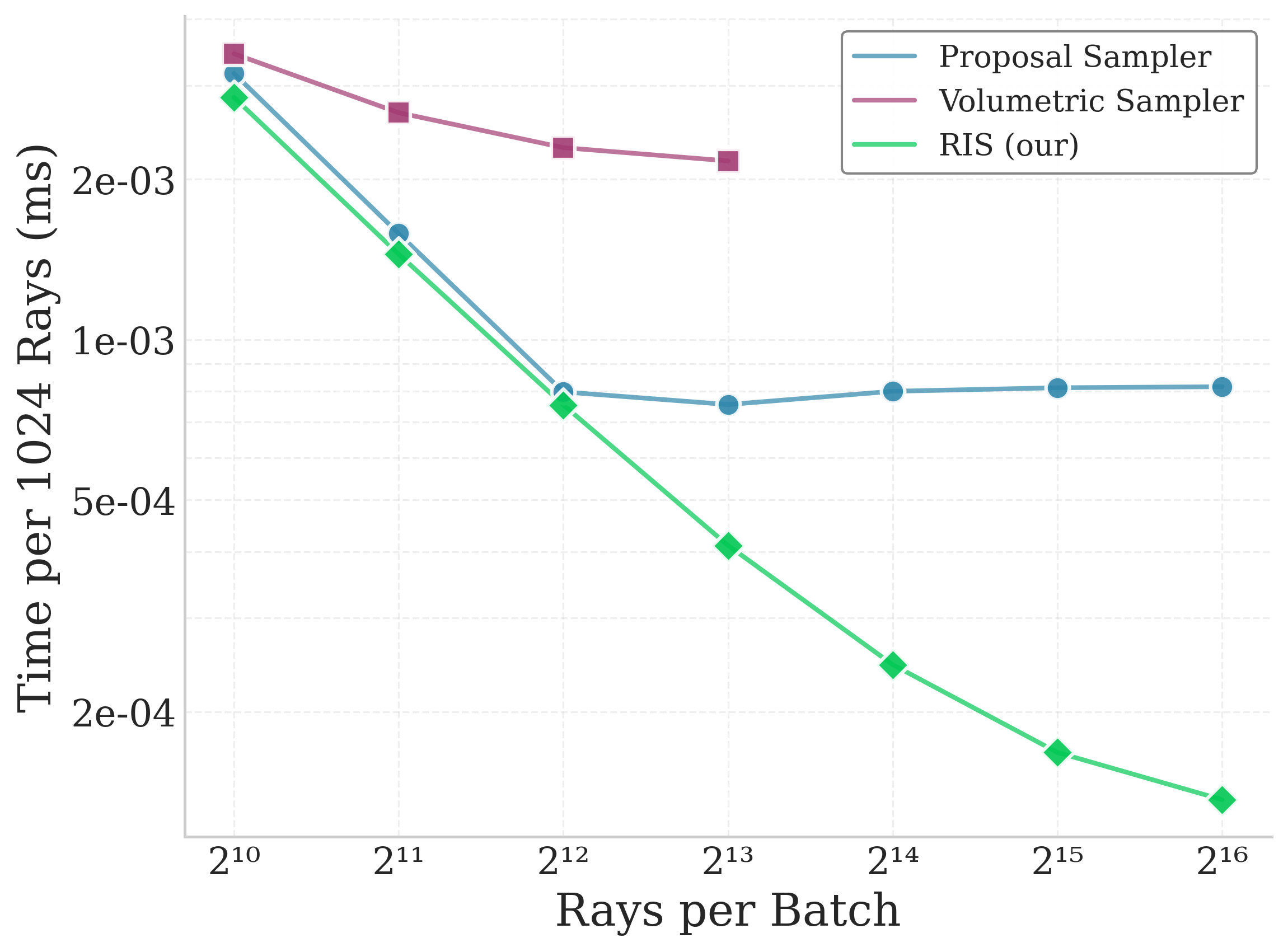}
        \captionof{figure}{Comparison of sampling performance. While the Proposal Sampler plateaus due to overhead and the Volumetric Sampler terminates early due to memory exhaustion, RIS demonstrates linear scalability, efficiently utilizing GPU parallelism to minimize per-ray costs at large batch sizes.}
        \label{fig:sampler}
    \end{minipage}
\vspace{-0.3cm}
\end{figure}

A defining feature of RIS is its ability to operate in a single pass, as visualized in Fig.\ref{fig:ray_intersection}. While standard methods must iterate through empty space, either via a proposal network or a coarse sampling phase, to locate the surface, RIS directly identifies the relevant interaction points. Since the anchors explicitly define the spatial bounds of the neural field, the sampler generates a sorted sequence of samples only where the density contribution is guaranteed to be significant. This ,,structure-aware'' sampling inherently skips empty space without maintaining a voxel grid or evaluating a proposal MLP. As demonstrated empirically in Section~\ref{sec:experiments}, RIS exhibits superior scalability compared to proposal-based and voxel-based samplers, maintaining high throughput even at large batch sizes.

\begin{table}[t]
\centering
\caption{Averaged quantitative comparisons on the NeRF Synthetic dataset. While reconstruction fidelity remains comparable to the state-of-the-art editable baseline (EKS), a decisive advantage is demonstrated in efficiency, with significantly reduced training times and rendering speeds approximately \textbf{30$\times$ higher} compared to the EKS method.}
\vspace{-0.3cm}
\setlength{\tabcolsep}{4pt}
{\fontsize{6.8pt}{11pt}\selectfont
\begin{tabular}{lccccc}
Model              & PSNR $\uparrow$ &SSIM $\uparrow$ & LPIPS $\downarrow$ & Training time $\downarrow$ & FPS $\uparrow$ \\ 
\hline
\multicolumn{6}{c}{Static} \\
\hline
InstantNGP        & \ca 29.31  & \ca 0.939 & \ca 0.053 & \ca 7min 27s &  \ca 1.894  \\
Nerfacto        & \cd 29.19  & \cd 0.941 & \cd 0.095 & \cd 9m 38s & \cd 0.860 \\
\hline
\multicolumn{6}{c}{Editable} \\
\hline
NeuralEditor    & \cd 31.41  & \cd 0.955 & \cd 0.036 & \cd 103h 45min & \cd 0.083 \\
EKS              & \ca 33.12 & \ca 0.962 & \ca 0.031 & \cb 26h 2min & \cb 0.183   \\
\our{} (our)    & \cb 32.65 & \cb 0.957 & \cb 0.034 & \ca 51min 18s & \ca 6.038   \\ 
\end{tabular}
}
\label{tab:exptime}
\vspace{-0.5cm}
\end{table}

\textbf{Ray-Coherent Feature Aggregation (RCA)}
To transition from a discrete particle representation to a continuous neural field, individual sample queries are not treated in isolation. Instead, a Ray-Coherent Feature Aggregation (RCA) mechanism is employed, which acts as a vectorized, 1D approximated K-Nearest Neighbors search along the ray. This module is responsible for synthesizing a locally smooth feature vector $\hat{\boldsymbol{f}_k}$ and opacity $\hat{\boldsymbol{\alpha}_k}$ for each sample point $\boldsymbol{x}_k$.

Given the sorted sequence of intersection points generated by the RIS, the neighborhood $\boldsymbol{\mathcal{W}}_k$ for the \textit{k}-th sample is defined as a symmetric window of size $W=2N+1$ indices: $\{k-N,...,k,...k+N\}$. To avoid the computational overhead of iterative queries, this operation is implemented using vectorized strided tensor views (unfolding), enabling the simultaneous retrieval of parameters for all neighbors in a single pass.

Naive index-based aggregation carries the risk of blending geometrically disjoint objects (e.g. a foreground object and a distant background wall) that happen to be adjacent in the sorted ray list. To enforce strict locality, a binary validity mask $\boldsymbol{\mathcal{M}}_{kj}$ is applied between the current sample $\boldsymbol{x}_k$ and a neighbor $j\in\boldsymbol{\mathcal{W}}_k$ as illustrated in Fig.\ref{fig:knn}. The neighbor is considered valid only if it belongs to the same ray index and the Euclidean distance $\lVert\boldsymbol{x}_k-\boldsymbol{\mu}_j\rVert_{2}$ is below a fixed threshold $\tau_{dist}$.

For every valid neighbor, the unnormalized logit $l_{kj}$ is computed based on the Mahalanobis distance $\Delta^2(\boldsymbol{x}_k,\boldsymbol{\mu}_j,\boldsymbol{\Sigma}_j)$. Aggregation weights $w_{kj}$ are obtained via a Softmax operation, where invalid neighbors are masked with $-\infty$. Simultaneously, the raw alpha $\alpha_{kj}^{raw}$ is derived from the Gaussian falloff $exp(l_{kj})$ modulated by the anchor's sigmoid-activated opacity parameter $\omega_{j}$. Finally, the continuous neural features and volumetric opacity are synthesized through a weighted summation:

\begin{equation}
\label{eq:rca_combined}
w_{kj} = \frac{\exp(l_{kj})}{\sum_{m \in \mathcal{W}_k} \exp(l_{km})}, \quad
\hat{\boldsymbol{f}}(\boldsymbol{x}_k) = \sum_{j \in \mathcal{W}_k} w_{kj}\boldsymbol{f}_j, \quad
\hat{\alpha}(\boldsymbol{x}_k) = \sum_{j \in \mathcal{W}_k} w_{kj}\alpha_{kj}^{raw}
\end{equation}

This formulation ensures that the opacity is not merely an interpolated parameter but is physically grounded in the Gaussian geometry of the neighbors, modulated by their relative influence weights.

\textbf{Neural-Explicit Geometry Decoding}
\label{sec:decoding}
Once the locally consistent feature vector $\hat{\boldsymbol{f}}(\boldsymbol{x}_k)$ and the aggregated geometric weight $\hat{\alpha}(\boldsymbol{x}_k)$ are synthesized via RCA, the final volumetric properties of the sample are resolved. This stage employs a shallow MLP $\mathcal{F}_\Theta$ to decode the implicit scene attributes from the latent embedding.

\begin{figure*}[t]
\centering
\setlength{\tabcolsep}{1pt}
{\fontsize{6.8pt}{11pt}\selectfont
\begin{tabular}{cc ccc c c ccc}
& GT & RM & EKS & \our{} (our) & \hspace{2mm} & GT & RM & EKS & \our{} (our)   \\
 \rotatebox{90}{\footnotesize\hspace{8pt}{Sculpture 1}} &
 \includegraphics[width=0.11\textwidth]{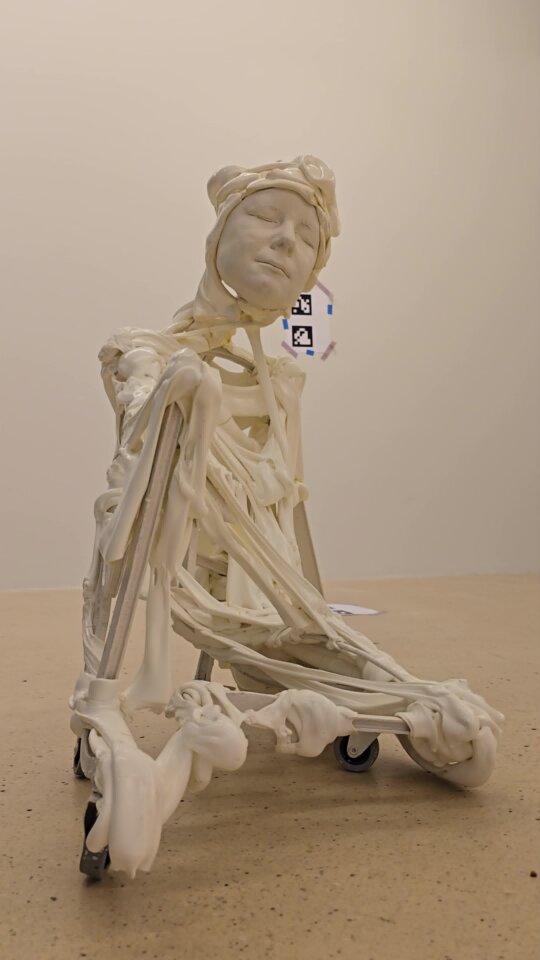} &
 \includegraphics[height=2.4cm, width=0.11\textwidth]{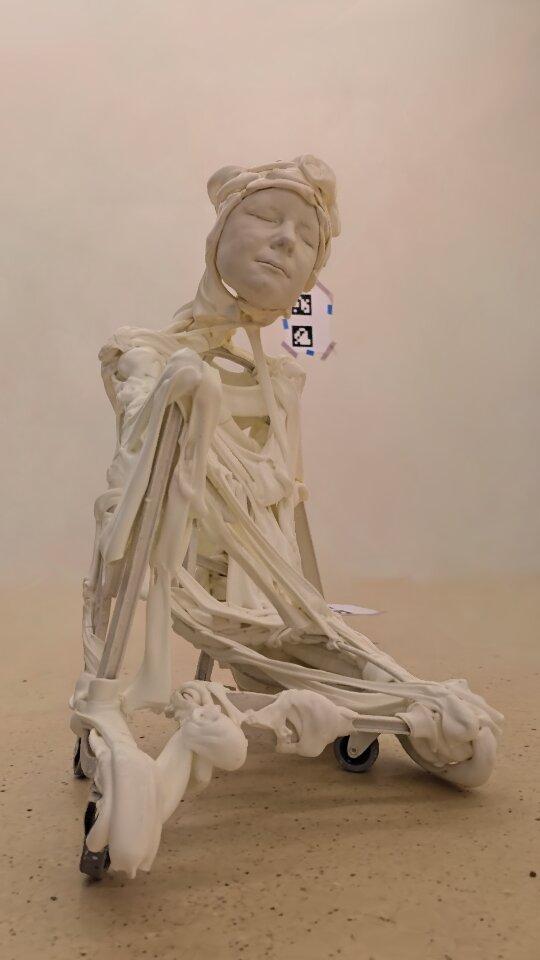} 
 &
 \includegraphics[width=0.11\textwidth]{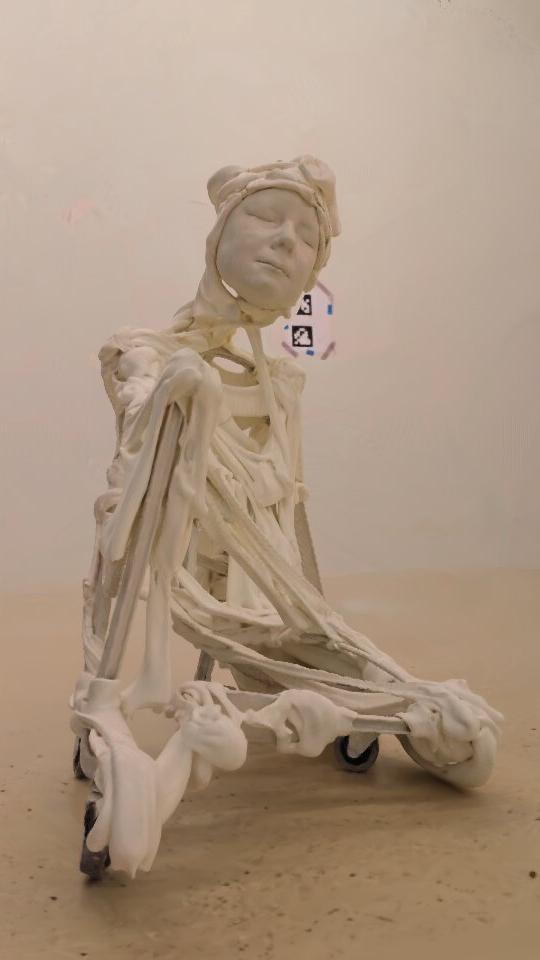} 
 &
 \includegraphics[width=0.11\textwidth]{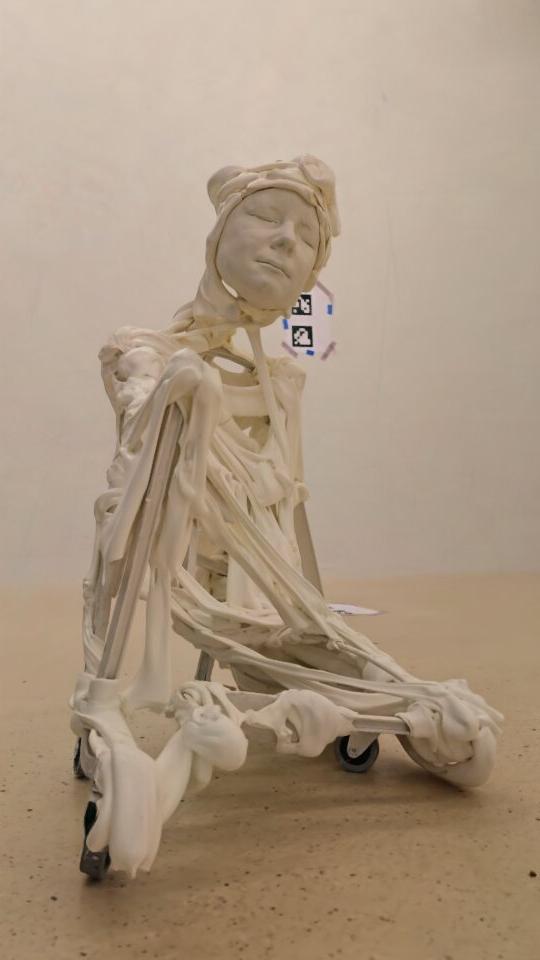} 
  &&
\includegraphics[width=0.11\textwidth]{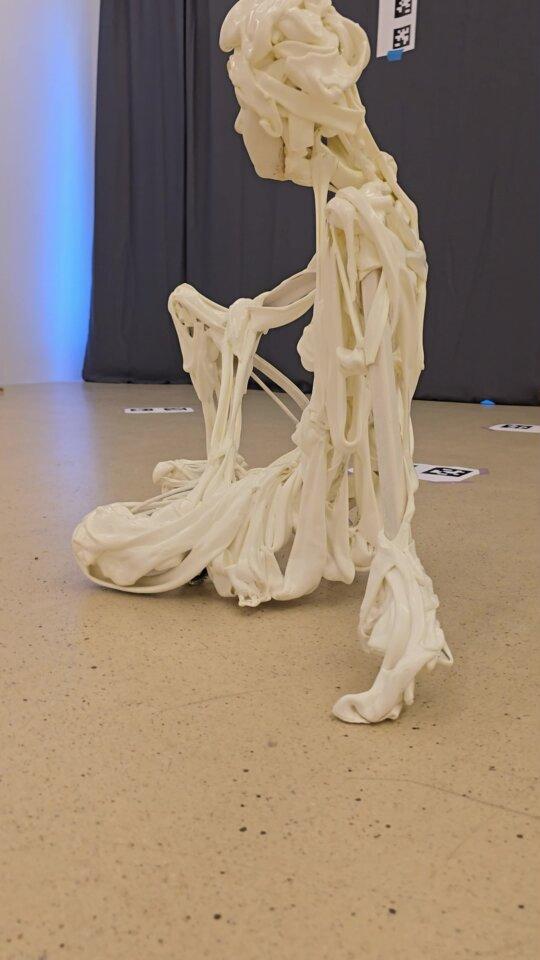} &
\includegraphics[height=2.4cm, width=0.11\textwidth]{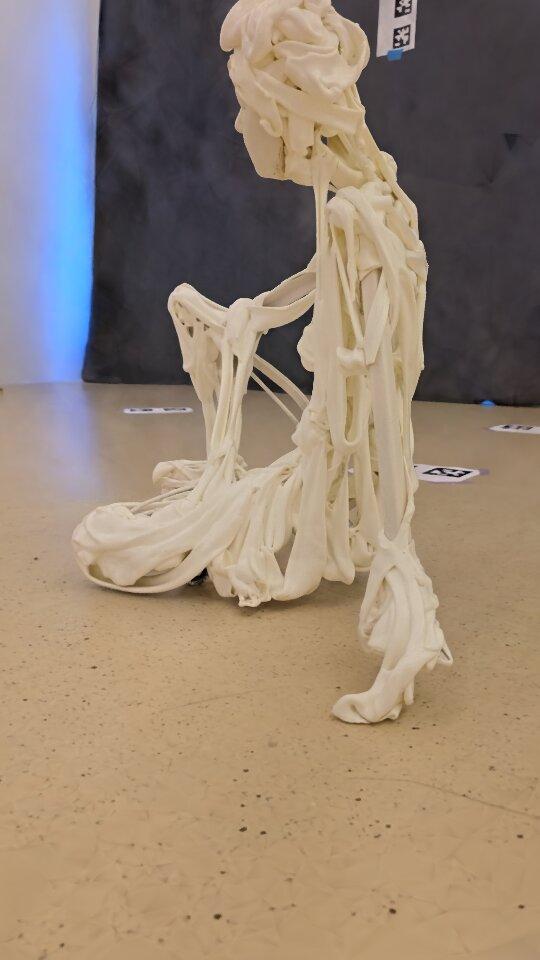} & 
\includegraphics[width=0.11\textwidth]{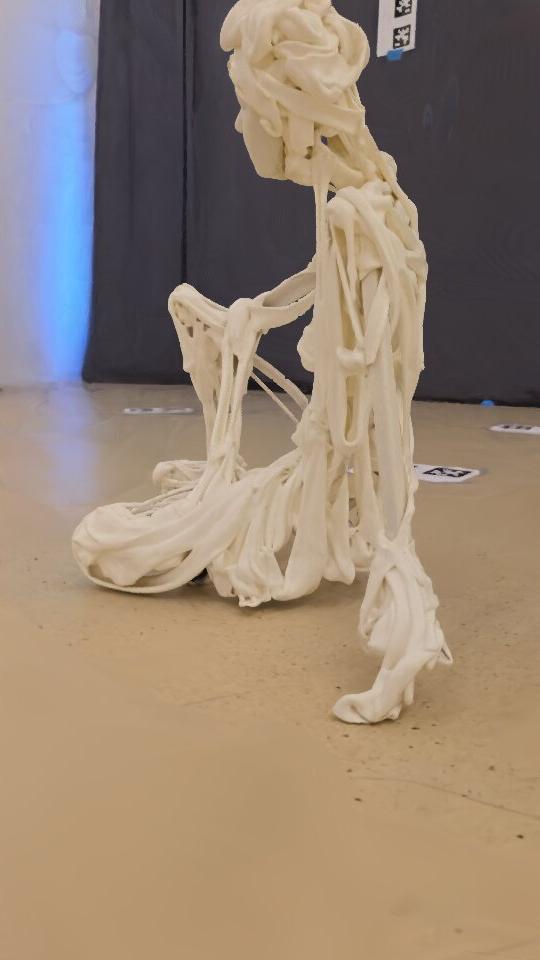} &
\includegraphics[width=0.11\textwidth]{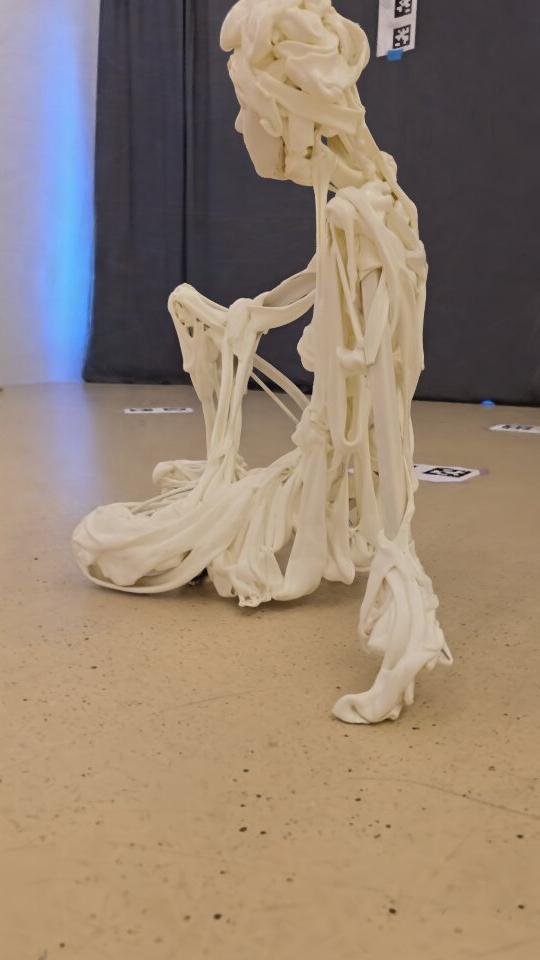}
  \\
   \rotatebox{90}{\footnotesize\hspace{8pt}{Sculpture 2}} &
 \includegraphics[width=0.11\textwidth]{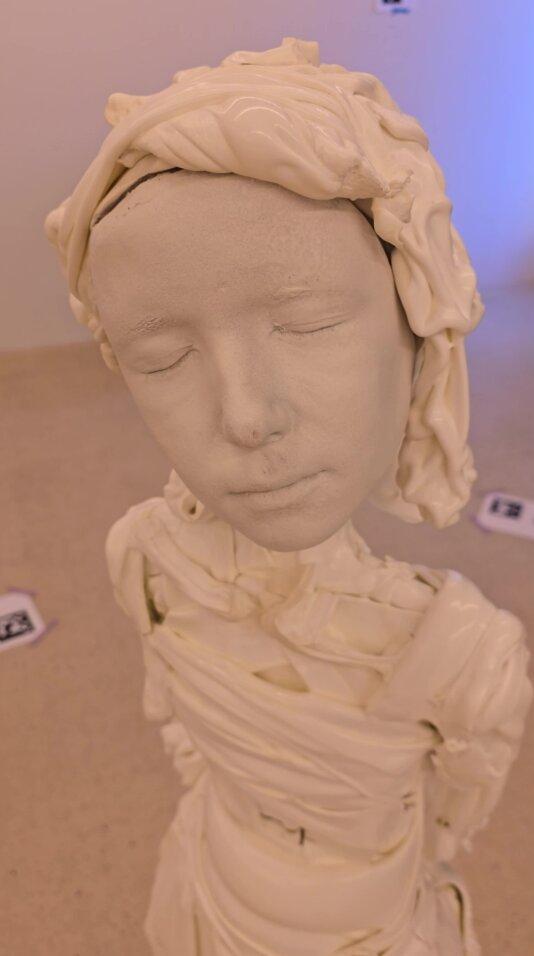} 
&
 \includegraphics[height=2.4cm, width=0.11\textwidth]{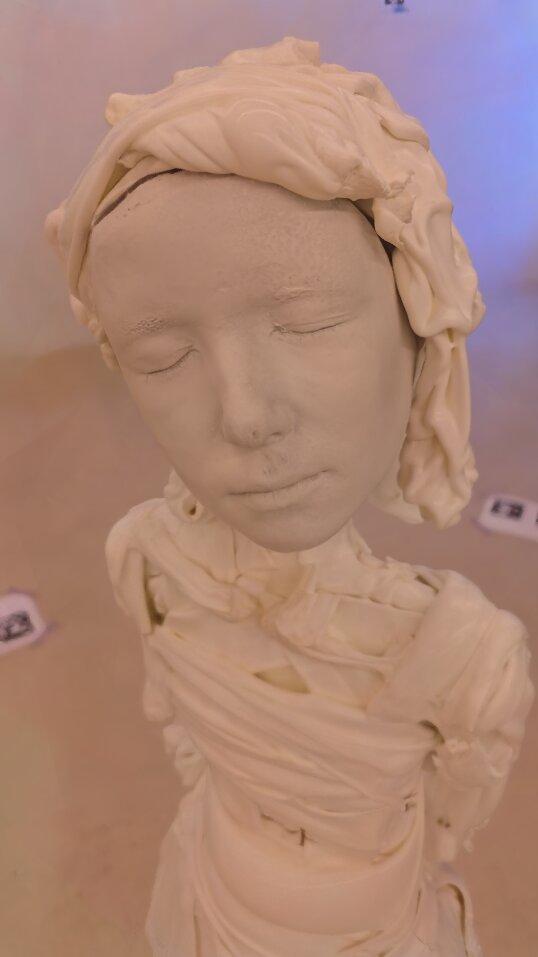} 
 &
 \includegraphics[width=0.11\textwidth]{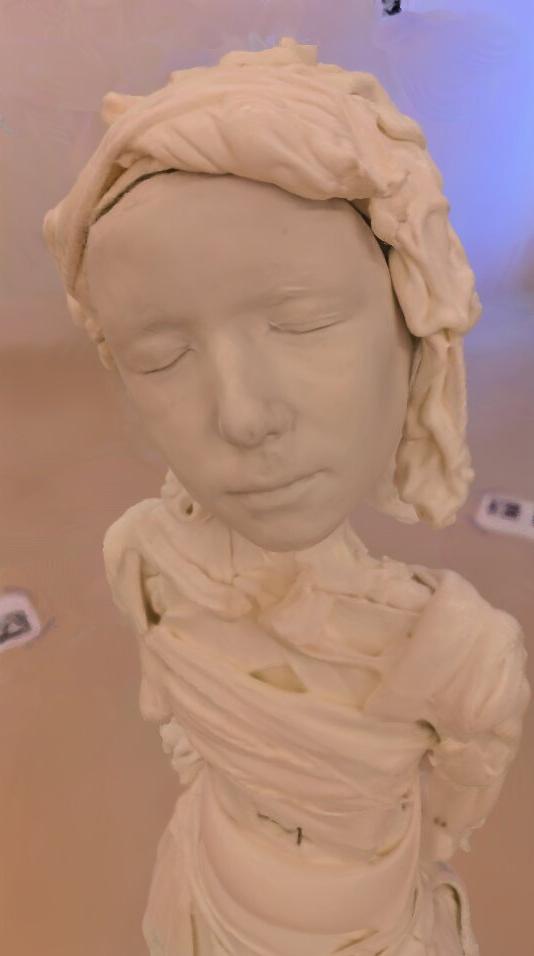}
 &
 \includegraphics[width=0.11\textwidth]{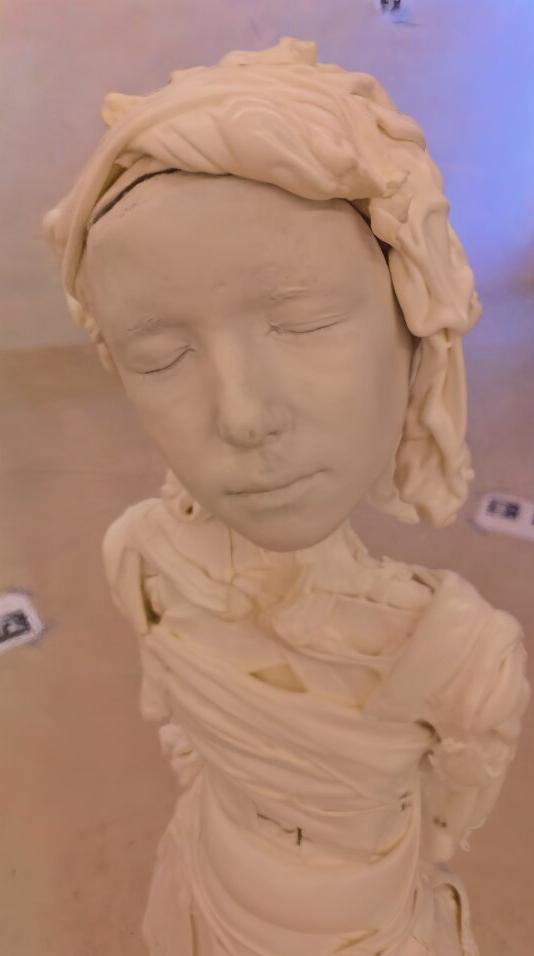} &&
    \includegraphics[width=0.11\textwidth]{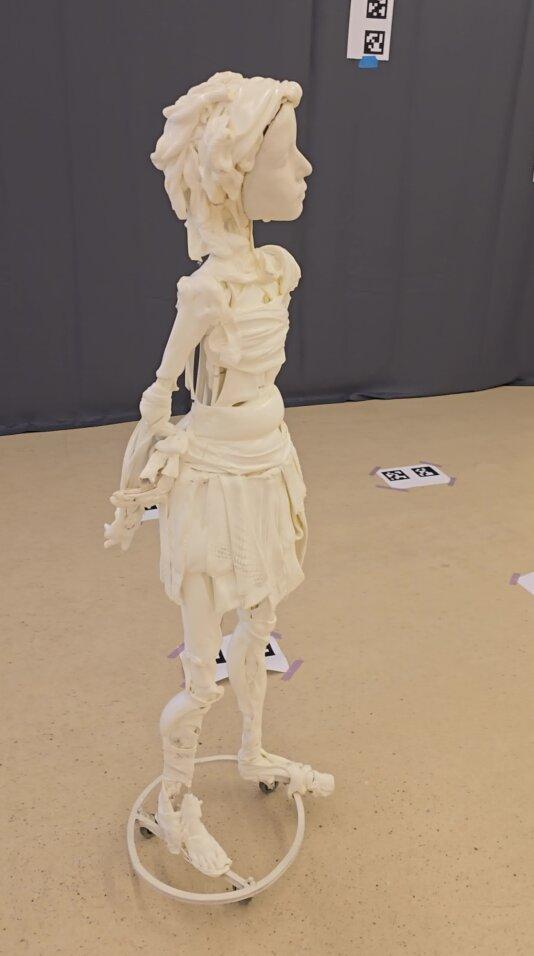}
    &
     \includegraphics[height=2.4cm, width=0.11\textwidth]{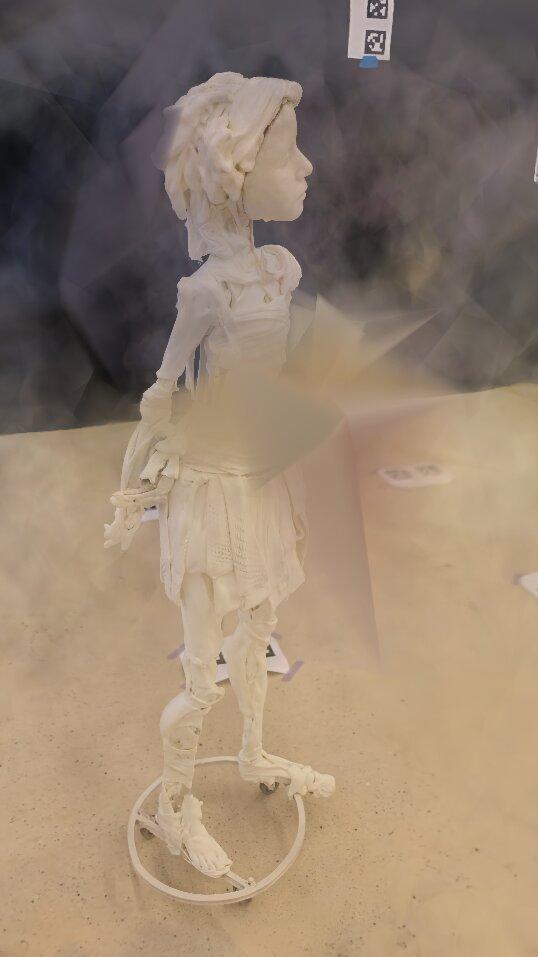}  & 
      \includegraphics[width=0.11\textwidth]{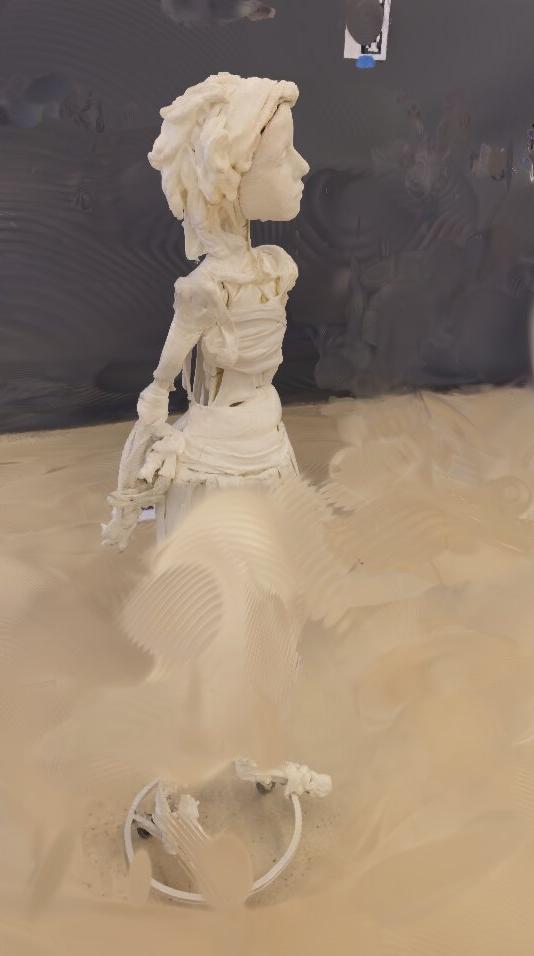} & 
      \includegraphics[width=0.11\textwidth]{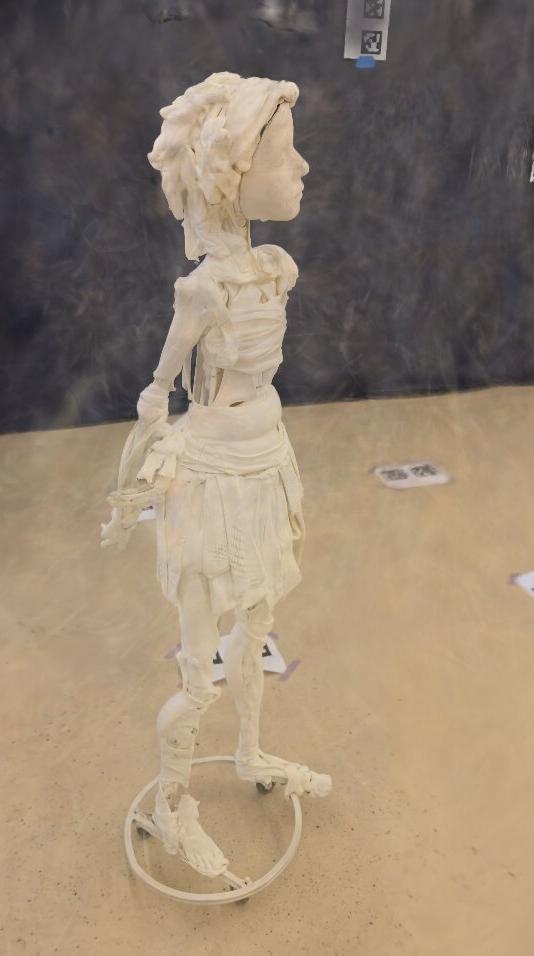}
\end{tabular}}
\caption{Qualitative comparison on the challenging HuSc3D \cite{husc3d} dataset. Compared to editable baselines, \our{} effectively suppresses floating artifacts prevalent in Radiance Meshes (RM) and preserves fine textures and sharp geometry that EKS over-smooths.}
\vspace{-0.5cm}
\label{fig:human}
\end{figure*}

The network takes the aggregated feature vector and the view direction $\boldsymbol{d}$ as inputs. It outputs a view-dependent color $\boldsymbol{c}_k$ and a raw scalar density value $\sigma_{mlp}$:

\begin{equation}
    (\sigma_{mlp},\boldsymbol{c}_k)=\mathcal{F}_\Theta\Bigl(\hat{\boldsymbol{f}}(\boldsymbol{x}_k),\gamma(\boldsymbol{d})\Bigl)
\end{equation}

Crucially, $\sigma_{mlp}$ represents the material density inferred from the neural field. To ensure physical plausibility and stable initialization, a truncated exponential activation with a negative bias is applied: $\sigma_{mlp}' = trunc\_exp(\sigma_{mlp} - 1)$.

To prevent artifacts typical of unbounded fields, the neural density is spatially constrained by the explicit Gaussian structure. The final effective density is computed as $\sigma_{eff}(\boldsymbol{x}_k)=\sigma_{mlp}'\hat{\alpha}(\boldsymbol{x}_k)$, effectively bounding the neural field within the probability mass of the anchors. This effective density is then converted into opacity $\alpha_k=1-exp(-\sigma_{eff}(\boldsymbol{x}_k))$, which serves as the input for point-based volumetric accumulation. The final pixel color $\boldsymbol{C}(\boldsymbol{r})$ is computed by compositing the sorted samples along the ray:

\begin{equation}
    \boldsymbol{C}(\boldsymbol{r})=\sum_{k=1}^KT_k\alpha_k\boldsymbol{c}_k, \:\: where \; T_k=\prod_{j=1}^{k-1}(1-\alpha_j)
\end{equation}

\begin{wrapfigure}{r}{0.5\textwidth}
\vspace{-0.6cm}
 \centering
 \resizebox{\linewidth}{!}{
 \setlength{\tabcolsep}{2pt} 
 {\fontsize{6.8pt}{11pt}\selectfont
 \begin{tabular}{c c c c}
  & $t_1$ & $t_2$ & $t_3$ \\ 
  
  \raisebox{0.7cm}{\rotatebox{90}{\tiny \our{}}} & 
  \includegraphics[width=2cm]{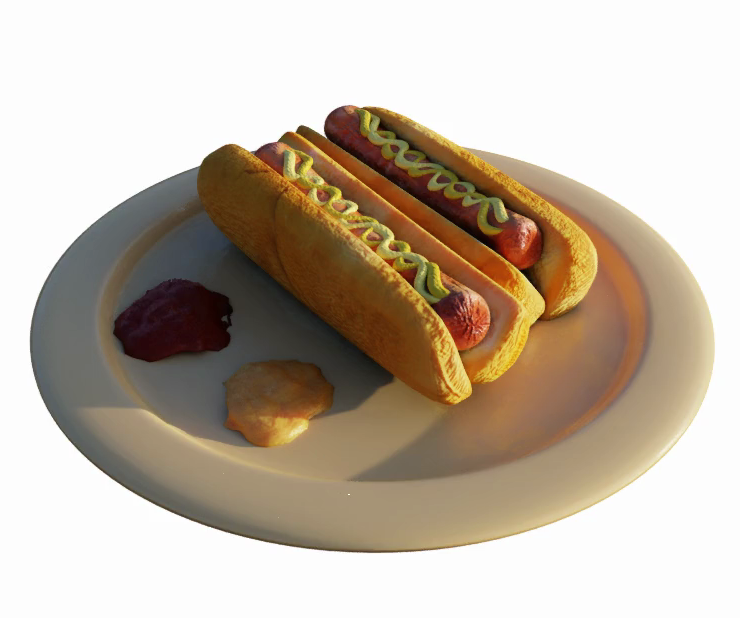} &
  \includegraphics[width=2cm]{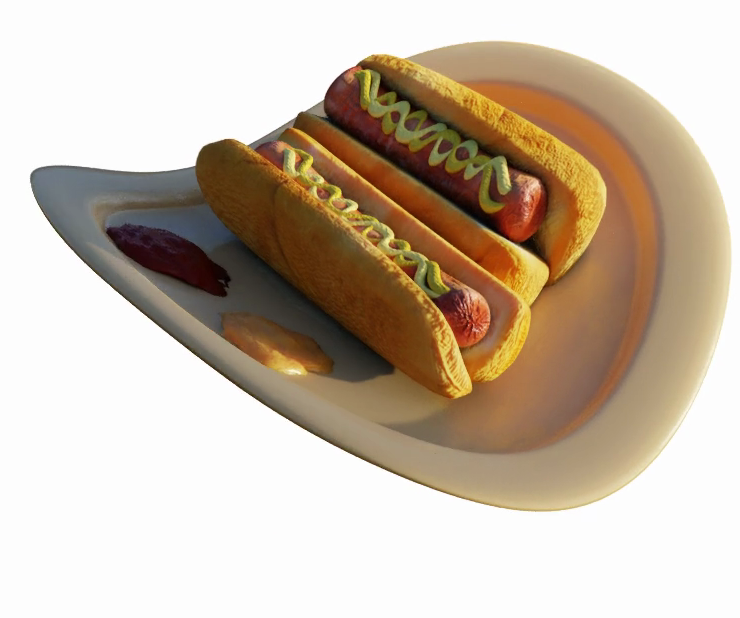} &
  \includegraphics[width=2cm]{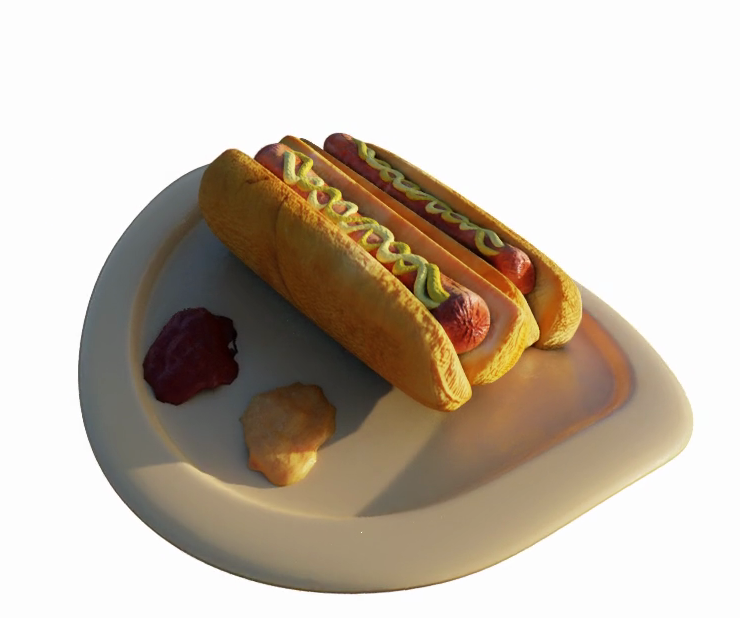} \\
  
  \raisebox{0.7cm}{\rotatebox{90}{\tiny EKS}} & 
  \includegraphics[width=2cm]{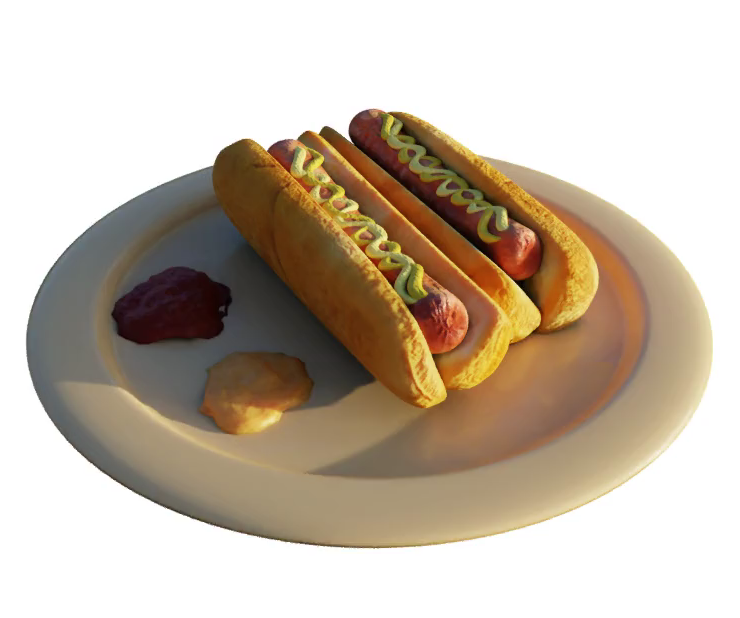} &
  \includegraphics[width=2cm]{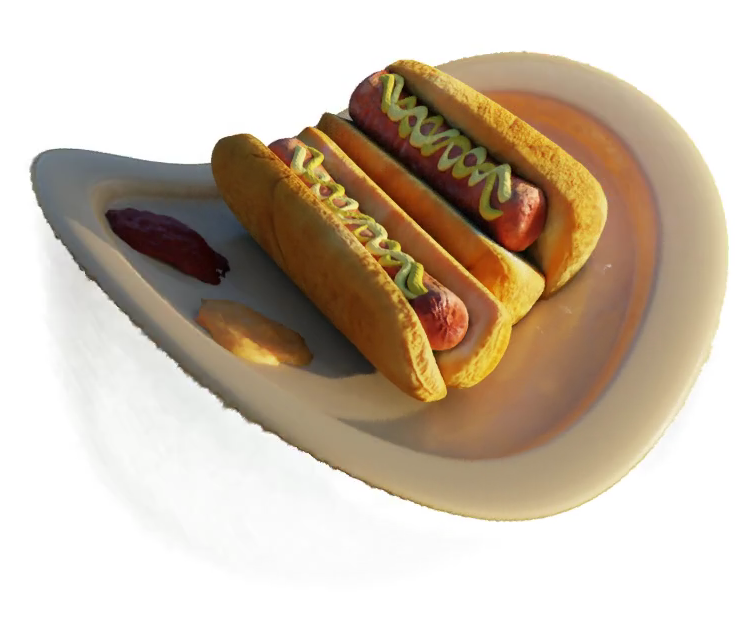} &
  \includegraphics[width=2cm]{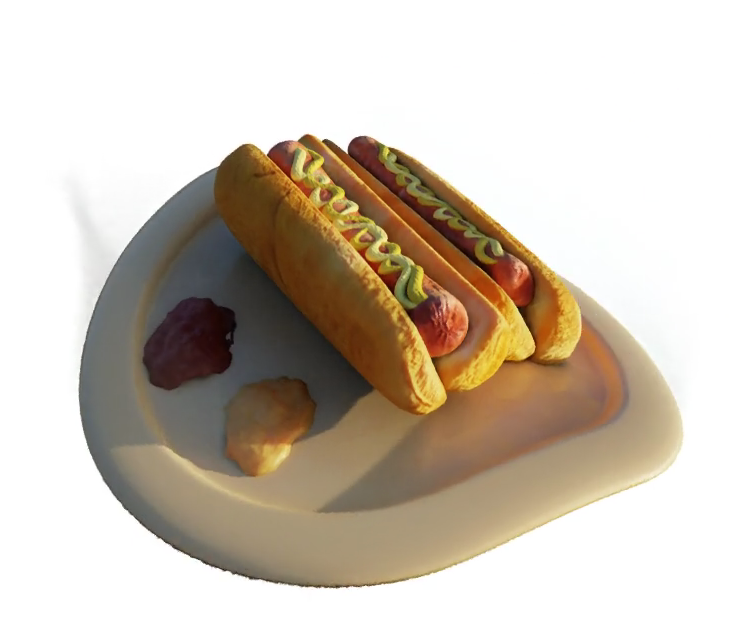} \\
  
  \raisebox{0.7cm}{\rotatebox{90}{\tiny \our{}}} & 
  \includegraphics[width=2cm, height=2cm]{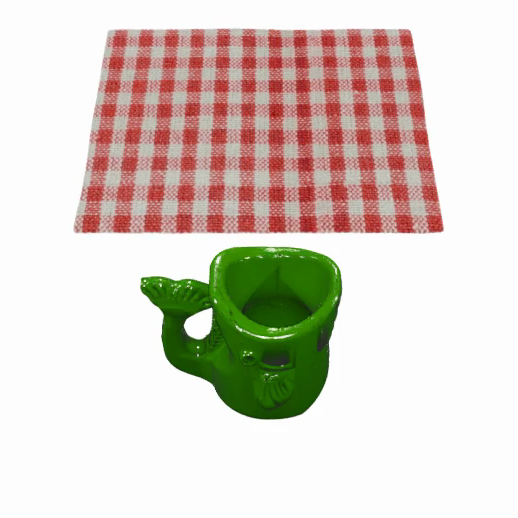} &
  \includegraphics[width=2cm, height=2cm]{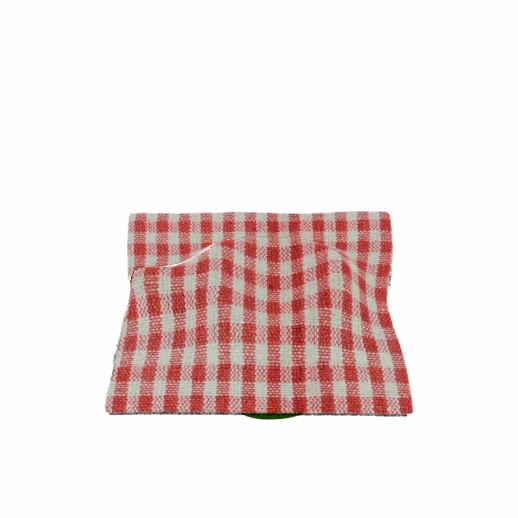} &
  \includegraphics[width=2cm, height=2cm]{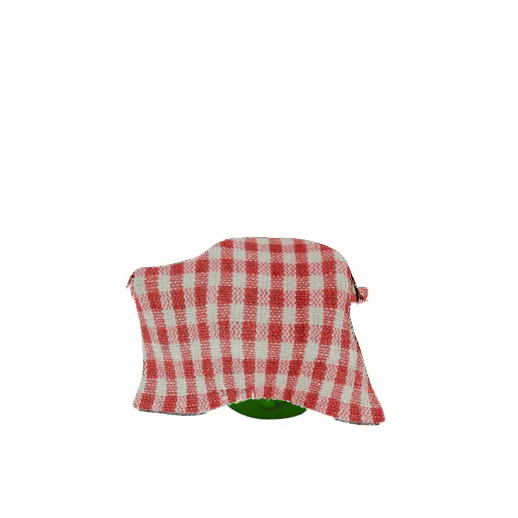} \\
  
  \raisebox{0.7cm}{\rotatebox{90}{\tiny EKS}} & 
  \includegraphics[width=2cm, height=2cm]{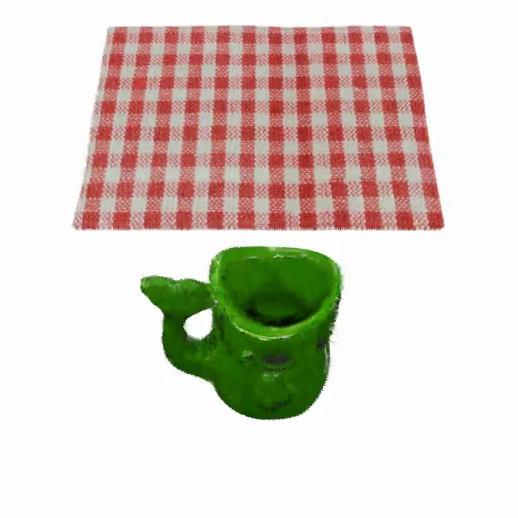} &
  \includegraphics[width=2cm, height=2cm]{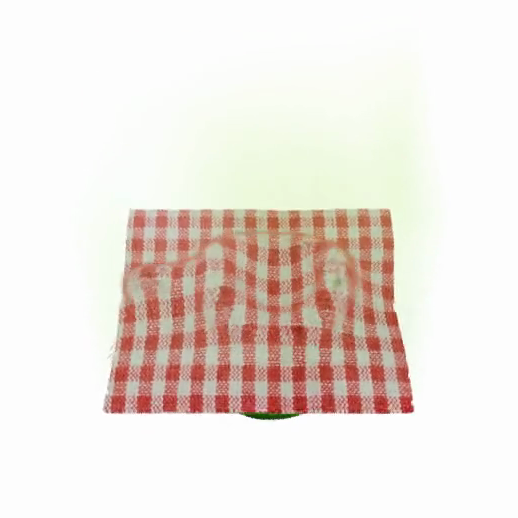} &
  \includegraphics[width=2cm, height=2cm]{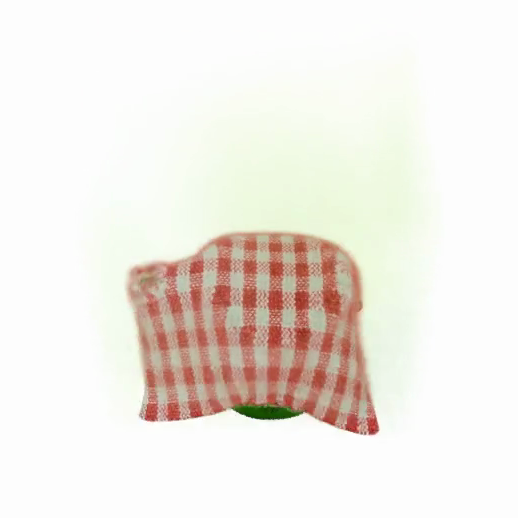} \\
 \end{tabular}
 }
 }
 \caption{Qualitative comparison on manual (top) and physics-based (bottom) editing. It is observed that EKS exhibits visible ghosting artifacts, whereas \our{} maintains superior fidelity and clean geometry, effectively eliminating trailing noise during motion.}
 \label{fig:editables}
 \vspace{-0.6cm}
\end{wrapfigure}

This formulation leverages the discrete nature of RIS samples to perform efficient single-pass integration while maintaining the differentiability of the volumetric model.

\textbf{Visibility-Aware Pruning}
To eliminate redundant geometry and artifacts, a visibility-driven pruning strategy inspired by \cite{zielinski2025genie} is implemented. The existence of each Neural Anchor is tracked via a confidence score $c_i$. Throughout optimization, a global decay factor is continuously applied to all anchors, gradually reducing their confidence. In contrast, anchors that are actively selected by the RIS as candidate samples receive a positive confidence update. This creates a competitive mechanism where only geometry that significantly contributes to the rendering process survives. Anchors that remain unvisited by the sampler, with confidence score below $\tau_{prune}$, are permanently removed from the scene.

\textbf{Consistency-Aware Scene Editing}
The hybrid architecture of \our{} funudamentally distinguishes it from purely implicit editing methods by enabling direct manipulation of the scene's neural representation. The inference stage operates on ,,baked'' primitives, where the dependency between the feature vector $\boldsymbol{f}_i$ and the global hash grid position is severed. Consequently, each Neural Anchor becomes a self-contained entity carrying both its explicit geometry $(\boldsymbol{\mu}_i,\boldsymbol{\Sigma}_i)$ and its implicit appearance code $\boldsymbol{f}_i$. This decoupling ensures that neural features remain strictly attached to their respective anchors, regardless of spatial location, enabling topology-agnostic editing. Operations such as rigid transformation, or non-rigid deformation are performed by simply updating the explicit parameters, with no requirement for retraining or complex volumetric warping.

This explicit structure facilitates seamless integration with standard graphic pipelines. Neural Anchors can be bound to external physics engines (e.g., via tetrahedral cages or mesh skinning) to drive realistic dynamic simulations, including gravity, collisions, and soft-body deformations. However, rigid rotation of the latent features can lead to inconsistencies in view-dependent effects, where specular reflections appear static relative to the surface. To address this, the Infinitesimal Surface Transformation (IST)~\cite{chen2023neuraleditor} is incorporated. During rendering, the query view direction $\boldsymbol{d}$ is dynamically adjusted into the local frame of the deformed surface via the deformation gradient $\boldsymbol{R}_{def}$, such that $\boldsymbol{d}'=\boldsymbol{R}_{def}^T\boldsymbol{d}$. This ensures that high-frequency optical effects, decoded by the MLP from the invariant features react naturally to changes in surface orientation, preserving photorealism under dynamic motion.


\begin{table*}[t]
{
\caption{Quantitative comparisons of reconstruction capability (PSNR) on the Mip-NeRF 360 dataset. It is observed that in indoor environments, \our{} outperforms all editable baselines and explicit static methods, being surpassed only by the implicit Mip-NeRF 360. In outdoor scenarios, while a performance gap is observed relative to Radiance Meshes, competitive fidelity is maintained against other baselines, with results significantly exceeding those of EKS. Radiance Meshes results ($^\star$) were re-evaluated using the official implementation.}
\vspace{-0.3cm}
\centering
\resizebox{\columnwidth}{!}{%
\setlength{\tabcolsep}{4.3pt}
{\fontsize{6.8pt}{11pt}\selectfont
\begin{tabular}{lcccccc|ccccc}
 & \multicolumn{6}{c}{Outdoor scenes} & \multicolumn{5}{c}{Indoor scenes} \\
 & {bicycle} & {flowers} & {garden} & {stump} & {treehill} & {Avg.}
 & {room} & {counter} & {kitchen} & {bonsai} & {Avg.} \\
 \hline
 \multicolumn{12}{c}{Static} \\
\hline
INGP
 & \cd 22.17 & \cd 20.65 & \cd 25.07 & \cd 23.47 & \cc 22.37 & \cd 22.75
 & \cd 29.69 & \cd 26.69 & \cd 29.48 & \cd 30.69 & \cd 29.14 \\
Mip-NeRF 360
 & \cb 24.37 & \ca 21.73 & \cc 26.98 & \cb 26.40 & \ca 22.87 & \cb 24.47
 & \ca 31.63 & \ca 29.55 & \ca 32.23 & \ca 33.46 & \ca 31.72 \\
3DGS
 & \ca 25.25 & \cb 21.52 & \ca 27.41 & \ca 26.55 & \cb 22.49 & \ca 24.64
 & \cc 30.63 & \cc 28.70 & \cc 30.32 & \cb 31.98 & \cc 30.41 \\
 Triangle Splatting & \cb 24.90 & \cc 20.85 & \cb 27.20 & \cc 26.29 & \cd 21.94 & \cc 24.27 & \cb 31.05 & \cb 28.90 & \cb 31.32 & \cc 31.95 & \cb 30.80 \\
\hline
\multicolumn{12}{c}{Editable} \\
\hline
Radiance Meshes$^\star$ & \ca 25.24 & \ca 21.70 & \ca 26.86 & \ca 26.71 & \ca 23.16 & \ca 24.73 & \cd 28.71 & \cb 28.32 & \cb 29.44 & \cb 30.77 & \cb 29.29\\
EKS
 & \cd 21.07 & \cd 19.62 & \cd 24.62 & \cd 20.54 & \cd 17.64 & \cd 20.70
 & \cb 30.54 & \cd 26.34 & \cd 28.42 & \cd 28.63 & \cd 28.48 \\
\our{} (ours) & \cb 23.89 & \cb 21.38 & \cb 26.19 & \cb 26.28 & \cb 22.80 & \cb 24.11 & \ca 31.14 & \ca 29.14 & \ca 31.03 & \ca 32.08 & \ca 30.85 \\
\hline
\end{tabular}
    }
    }
\label{tab:mipnerf360psnr}
}
\vspace{-0.5cm}
\end{table*}
\vspace{-0.2cm}
\section{Experiments}
\label{sec:experiments}

An extensive evaluation of \our{} demonstrates its capabilities in high-fidelity reconstruction, real-time rendering, and consistent scene editing. The experiments are designed to validate the efficiency of the proposed method.

\textbf{Experimental Setup}
The evaluation protocol encompasses standard benchmarks and large-scale environments. Object-centric reconstruction is assessed on the NeRF Synthetic dataset~\cite{mildenhall2020nerf}. For complex, unbounded environments, the Mip-NeRF 360~\cite{barron2022mipnerf360}, Tanks\&Temples~\cite{tandt}, and Deep Blending~\cite{deepblending} datasets are utilized. Uniquely, the method is also stress-tested on the HuSc3D~\cite{husc3d} dataset, characterized by challenging indoor scenes with significant lighting variations, to validate robustness. For editing tasks, the deformation benchmark from NeuralEditor~\cite{chen2023neuraleditor} is adopted. Standard image quality metrics (PSNR, SSIM, LPIPS) are reported, alongside inference speed (FPS) and training time.

\textbf{Evaluation on Large-Scale Scenes}
Quantitative results for unbounded and large-scale environments are presented in Table \ref{tab:mipnerf360psnr} and Table \ref{tab:db_tandt}. A distinction in performance characteristics is observed between outdoor and indoor scenarios. In outdoor environments (outdoor split of Mip-NeRF 360 and Tanks\&Temples), \our{} remains competitive with static methods and Radiance Meshes, while significantly outperforming EKS.

In indoor scenarios, a dominant performance is established. On the Indoor split of Mip-NeRF 360, \our{} surpasses all editable baselines and explicit static methods. Within the domain of neural radiance fields, performance is exceeded only by the purely implicit Mip-NeRF360. Furthermore, on the Deep Blending dataset, state-of-the-art fidelity is achieved across all scenes.
Consequently, these combined results validate the effectiveness of the proposed NeRF architecture, demonstrating its ability to deliver high-fidelity reconstructions in challenging environments without compromising structural requirements for editing.

\begin{wrapfigure}{r}{0.5\textwidth}
\vspace{-0.8cm}
 \centering
 \resizebox{\linewidth}{!}{
 \setlength{\tabcolsep}{2pt} 
 {\fontsize{6.8pt}{11pt}\selectfont
 \begin{tabular}{c c c}
  & $t_1$ & $t_2$ \\ 
  
  \raisebox{0.6cm}{\rotatebox{90}{\tiny Bicycle}} & 
  \includegraphics[width=3cm]{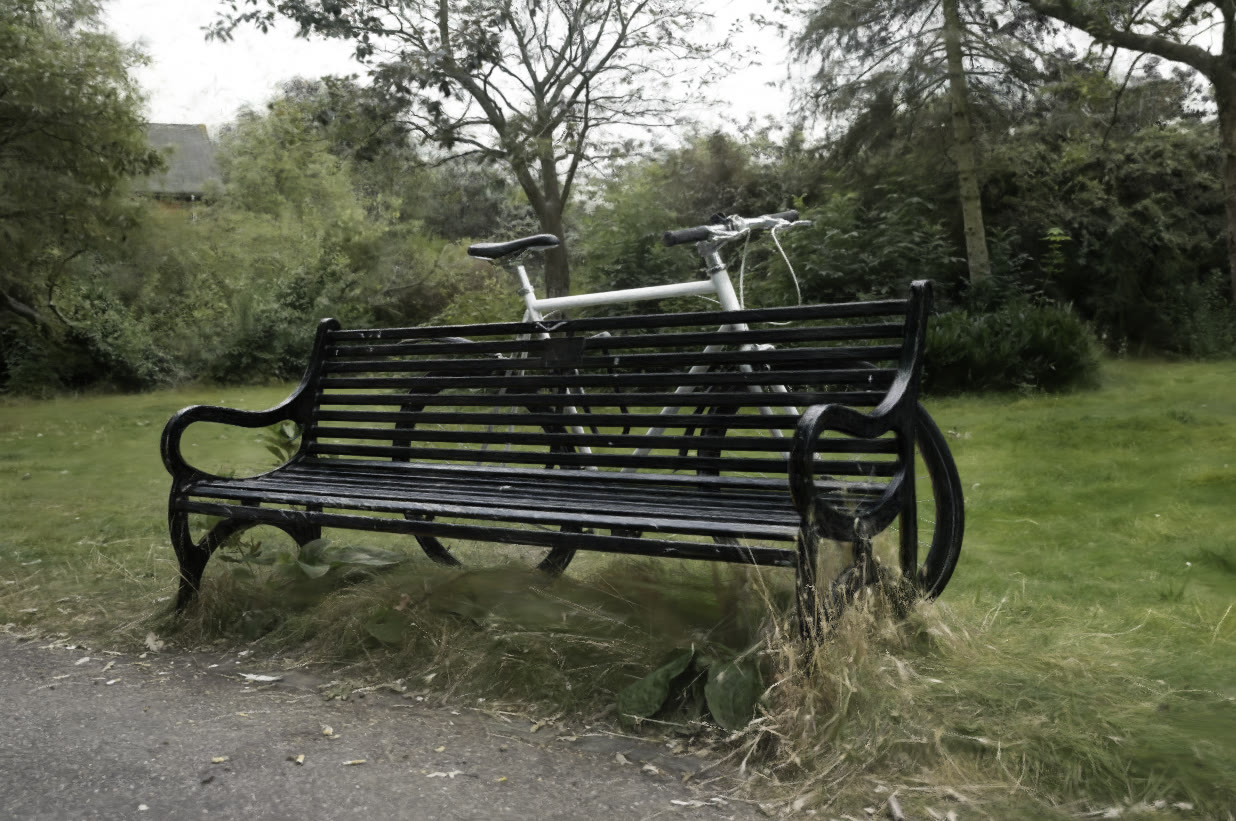} &
  \includegraphics[width=3cm]{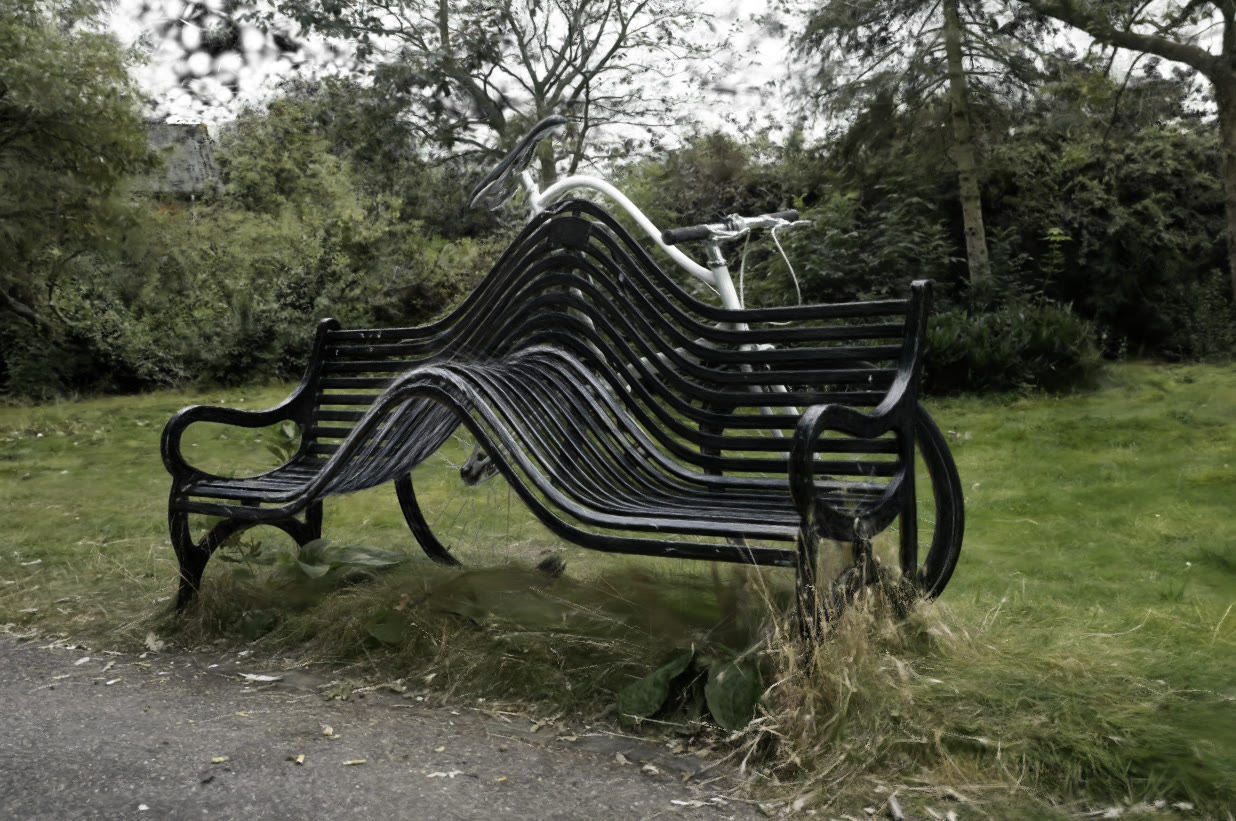} \\
  
  \raisebox{0.6cm}{\rotatebox{90}{\tiny Garden}} & 
  \includegraphics[width=3cm]{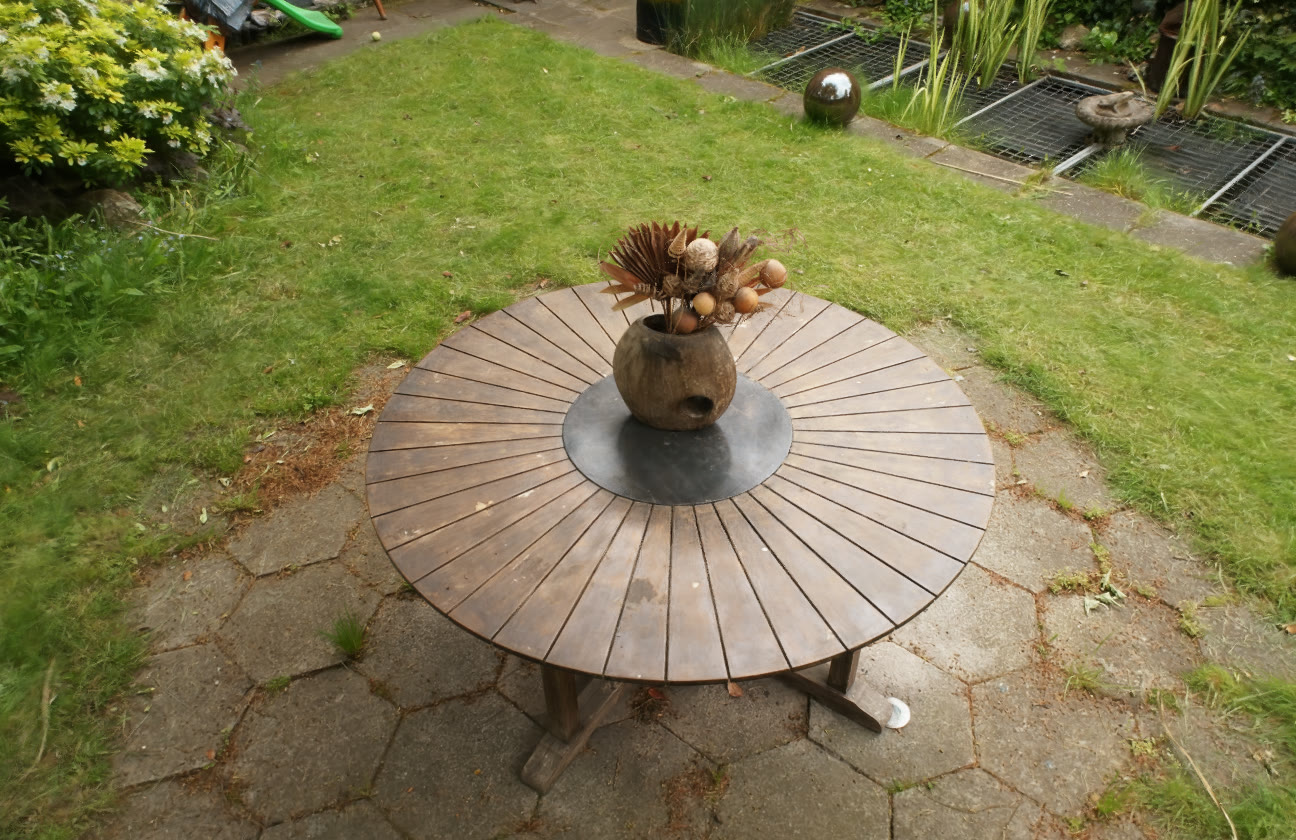} &
  \includegraphics[width=3cm]{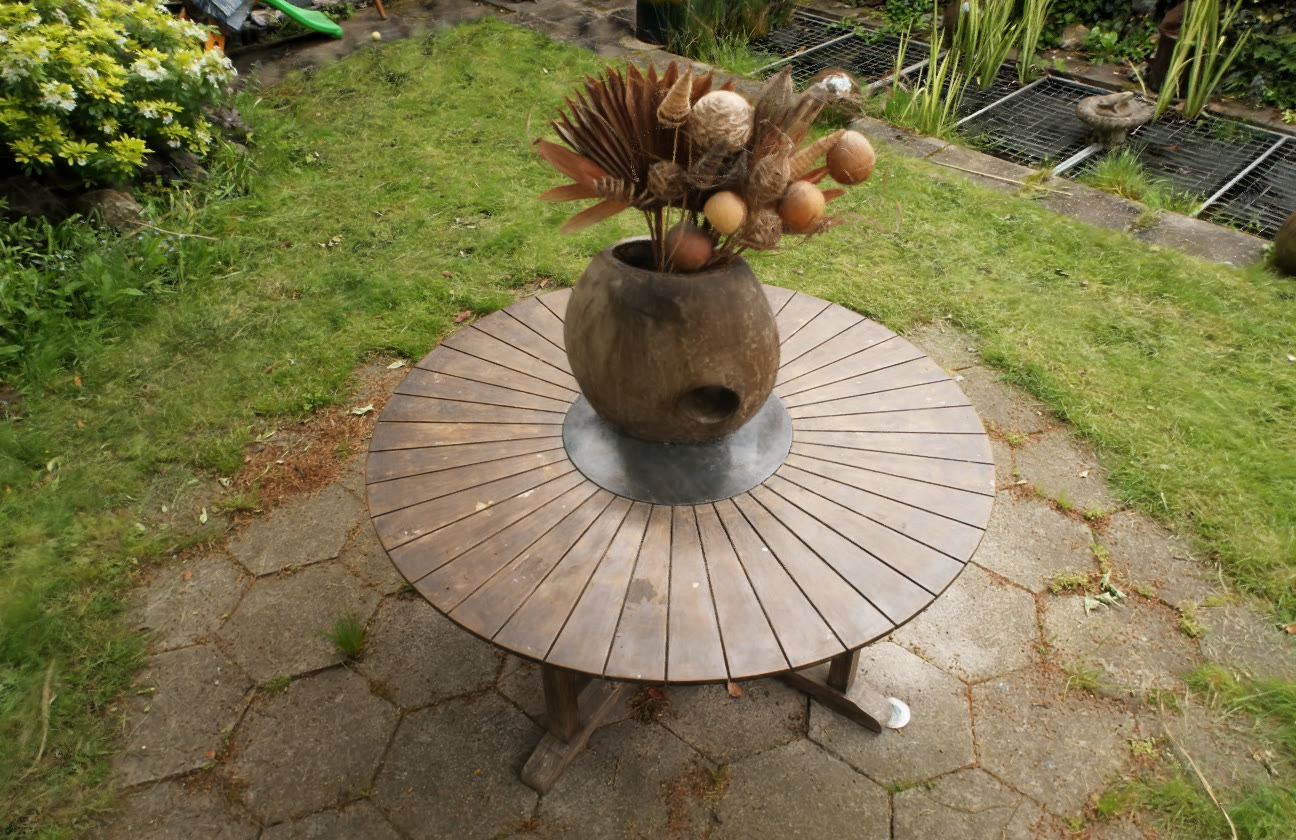} \\
 \end{tabular}
 }
 }
 \caption{Editing on large-scale scenes. Non-rigid warping (top) and object scaling (bottom) are demonstrated on \textit{Bicycle} and \textit{Garden} scenes. It is observed that high fidelity and geometric consistency are preserved in these challenging unbounded environments.}
 \label{fig:animation_real}
 \vspace{-0.9cm}
\end{wrapfigure}

These quantitative findings are corroborated by qualitative comparisons in Fig. \ref{fig:comp}, Fig. \ref{fig:comparisionmain} and Fig. \ref{fig:human}, where it is observed that \our{} preserves high-frequency details, such as thin structures and textures, that are often over-smoothed by competing editable approaches.

\textbf{Object-Centric Reconstruction}
Results averaged across the NeRF Synthetic dataset are summarized in Table \ref{tab:exptime}. While reconstruction fidelity remains comparable to the state-of-the-art editable baseline (EKS), a fundamental shift in efficiency is demonstrated. A drastic reduction in training time is achieved compared to other editable methods, and rendering speeds are accelerated by approximately $30\times$ relative to EKS. Notably, \our{} outperforms all listed methods in terms of FPS, effectively crossing the threshold for real-time interactive editing.

\textbf{Deformation Benchmark}
The geometric consistency of the representation under deformation is evaluated using the NeuralEditor benchmark. Detailed quantitative results are provided in the Appendix. Qualitatively, as shown in Fig. \ref{fig:edit_big_comparison}, it is observed that \our{} maintains high visual fidelity even under significant mesh-driven deformations. Artifacts such as texture degradation or geometric distortions, prevalent in baselines like NeuralEditor or GaMeS~\cite{waczynska2024games}, are effectively suppressed, confirming that the hybrid anchor-based representation supports robust topology-agnostic editing.

\textbf{Sampler Scalability Analysis}
To validate the architectural advantage of the Ray Intersection Selector (RIS), a focused analysis of sampling throughput was conducted against industry-standard strategies: the Proposal Sampler (Nerfacto) and the Volumetric Sampler (Instant-NGP). Detailed experimental configurations are provided in the Appendix.

\begin{wrapfigure}{r}{0.5\textwidth}
\vspace{-1.2cm}
 \centering
 \resizebox{\linewidth}{!}{
 \setlength{\tabcolsep}{2pt} 
 {\fontsize{6.8pt}{11pt}\selectfont
 \begin{tabular}{c c c c}
  & $t_1$ & $t_2$ & $t_3$ \\ 
  
  \raisebox{0.8cm}{\rotatebox{90}{\tiny \our{}}} & 
  \includegraphics[width=2cm, height=2cm]{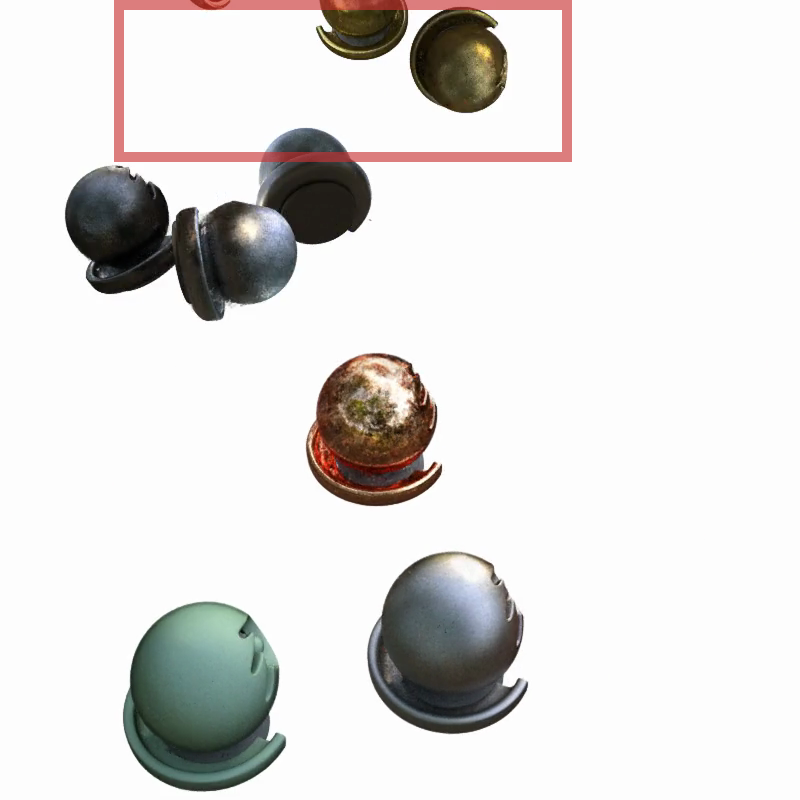} &
  \includegraphics[width=2cm, height=2cm]{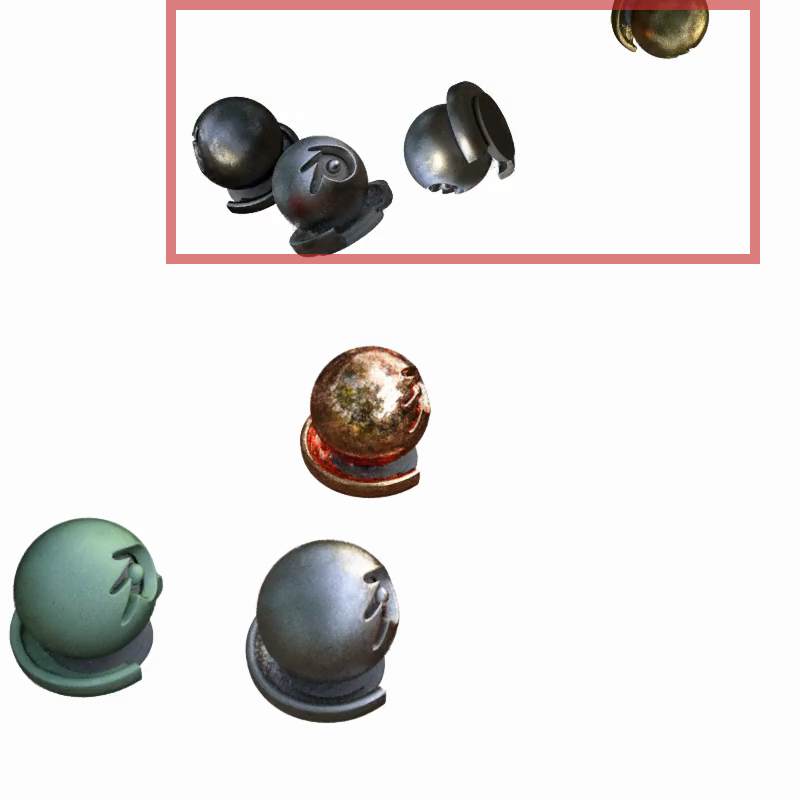} &
  \includegraphics[width=2cm, height=2cm]{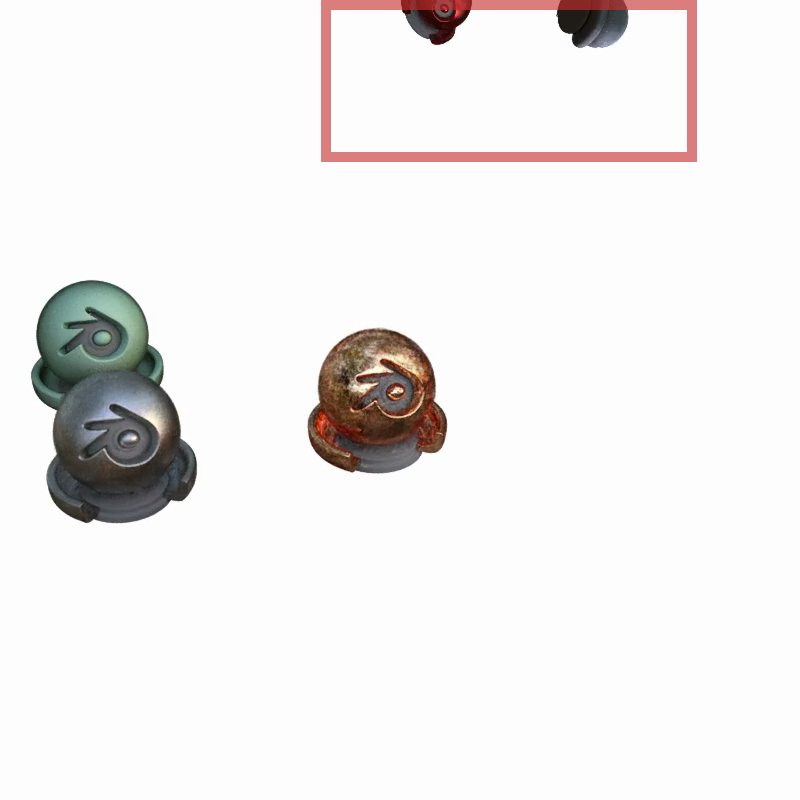} \\
  
  \raisebox{0.8cm}{\rotatebox{90}{\tiny EKS}} & 
  \includegraphics[width=2cm, height=2cm]{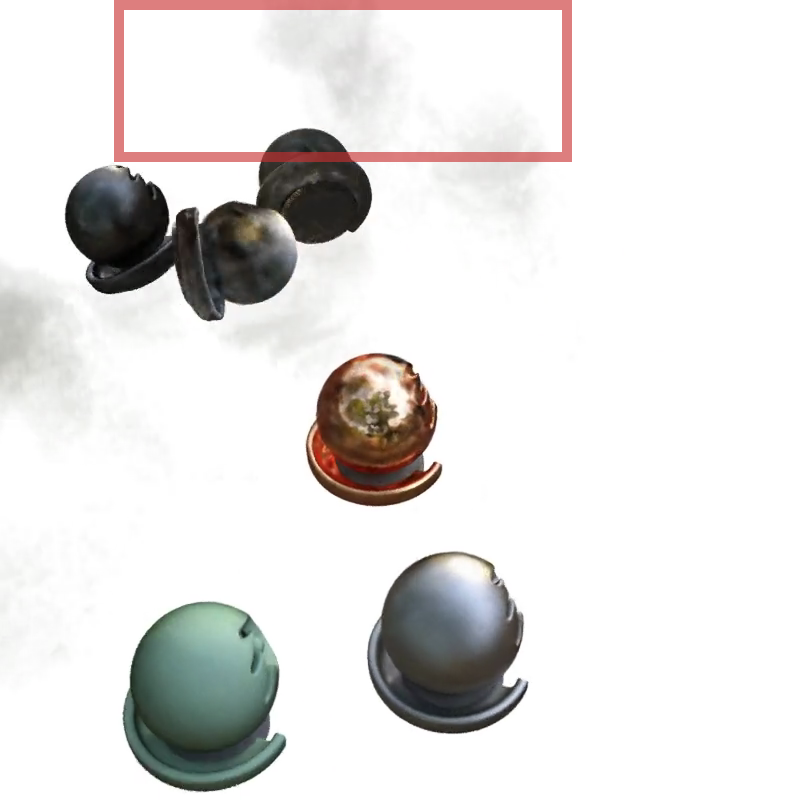} &
  \includegraphics[width=2cm, height=2cm]{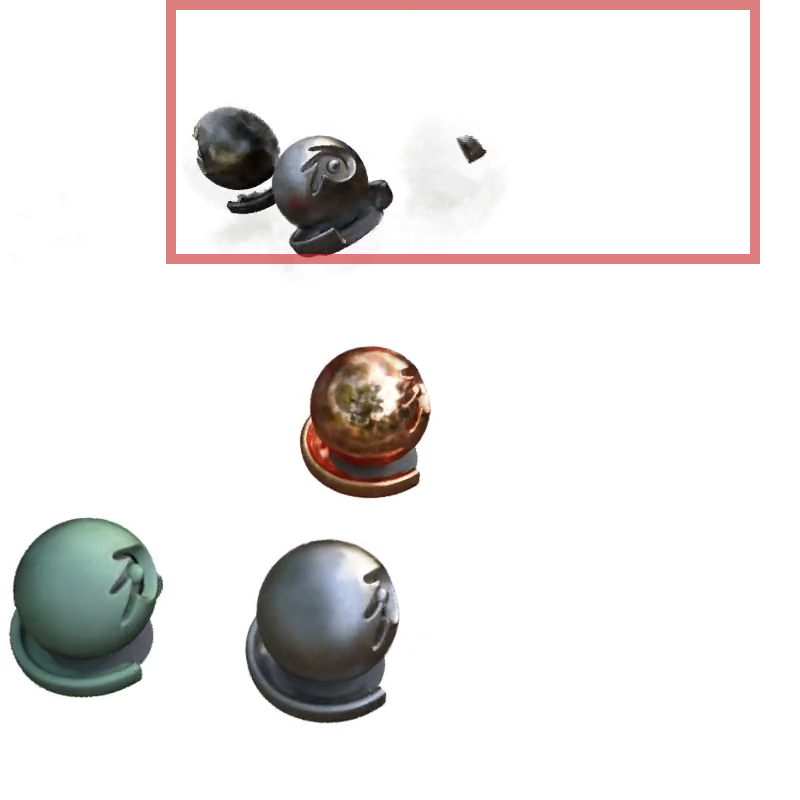} &
  \includegraphics[width=2cm, height=2cm]{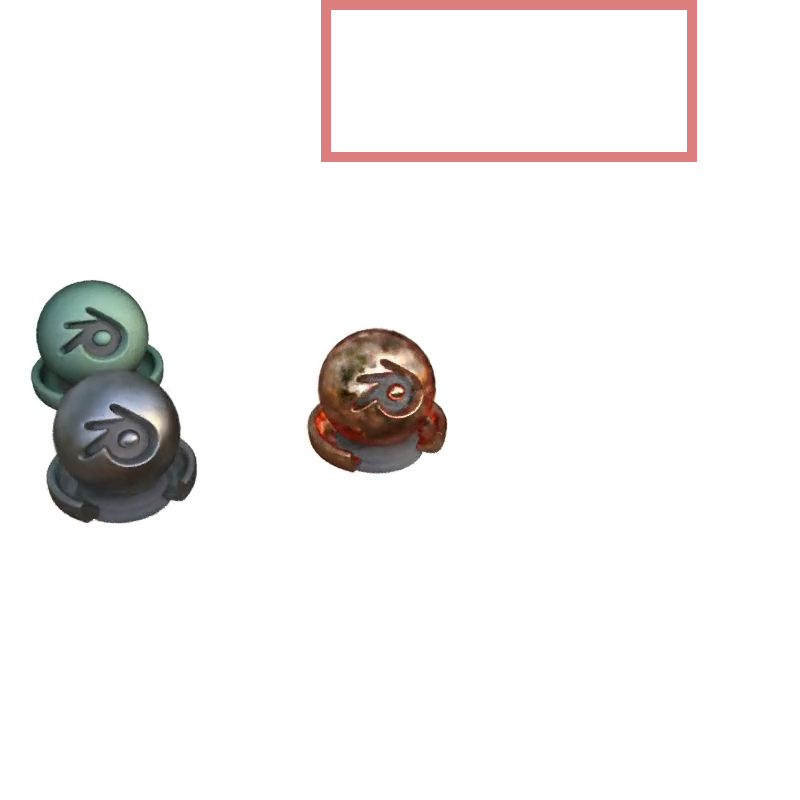} \\
 \end{tabular}
 }
 }
 \caption{
While standard NeRFs clip objects exiting the training volume, \our{} maintains full rendering integrity by dynamically sampling explicit geometry independent of static global bounds.
 }
 \label{fig:animation_bounded}
 \vspace{-0.7cm}
\end{wrapfigure}

The scalability results are visualized in Fig. \ref{fig:sampler}. Distinct behaviors are observed: the grid-based Volumetric Sampler exhibits the highest latency and fails to scale beyond a batch size of $2^{13}$ due to memory exhaustion. While the Proposal Sampler performs efficiently at lower workloads, a saturation point is reached around $2^{12}$ rays. In contrast, RIS demonstrates superior scalability, characterized by a downward linear trend in the log-log plot. This indicates that GPU utilization increases with larger workloads, confirming that the analytical intersection strategy effectively eliminates the computational redundancy of empty-space sampling.

\textbf{Scene Editing and Physics Simulation} 
The versatility of the representation is evaluated through manual and physics-driven editing tasks modeled within the Blender environment. To drive the animation, Neural Anchors are bound to low-resolution proxy geometries, allowing deformation fields to directly update the explicit scene structure. As visualized in Fig. \ref{fig:editables} and Fig, \ref{fig:animation_bounded}, scenarios ranging from non-rigid bending (\textit{Hotdog}) to complex soft-body dynamics (\textit{Cup and Rug}) are simulated. It is observed that \our{} maintains reconstruction fidelity comparable to, or exceeding, the EKS. A critical distinction is noted in the rendering cleanliness; while EKS exhibits persistent ghosting artifacts and trailing noise-attributed to the inertia of occupancy grid updates-\our{} produces sharp, artifact-free motion dure to the precision of the intersection-based sampler.

Furthermore, scalability is demonstrated on large-scene unbounded scenes (Fig.~\ref{fig:animation_real}) through direct manipulation of Neural Anchors, Operations such as warping (\textit{Bicycle}) or scaling (\textit{Garden}) are performed while preserving high visual fidelity and geometric consistency, validating the robustness of the hybrid representation in challenging real-world environments.

\textbf{Simulation Beyond Scene Bounds} 
To evaluate robustness under extreme transformations, a physics-based simulation was conducted on the \textit{Materials} scene, where Neural Anchors were rigidly bound to a dynamic mesh. As illustrated in Fig.~\ref{fig:animation_bounded}, a critical limitation is observed in EKS. Crucially, this constraint applies to most NeRF frameworks, where objects are abruptly clipped upon exiting the initial training volume because they rely on spatially restricted volumetric samplers. In contrast, full rendering integrity is maintained by \our{}. This capability is directly attributed to the synergy between Neural Anchors and the Ray Intersection Selector. By binding latent features to explicit primitives targeted via analytical intersections, the sampling domain is decoupled from static bounds and dynamically adapts to the moving geometry, thereby enabling true unbounded editing.

\begin{wrapfigure}{r}{0.5\textwidth}
\vspace{-1cm}
 \centering
\setlength{\tabcolsep}{1pt}
\begin{tabular}{ccccccccc}
 \tiny GT & \tiny\hspace{4mm}& \tiny16IPR &\tiny\hspace{4mm}& \tiny64IPR  &\tiny\hspace{4mm}&  \tiny{w/o hash}  &\tiny\hspace{4mm}&  \tiny full \\
 \includegraphics[width=0.09\textwidth]{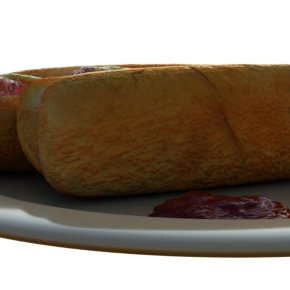} &&
  \includegraphics[width=0.09\textwidth]{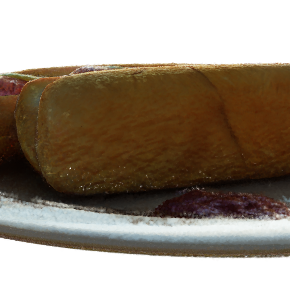} &&
   \includegraphics[width=0.09\textwidth]{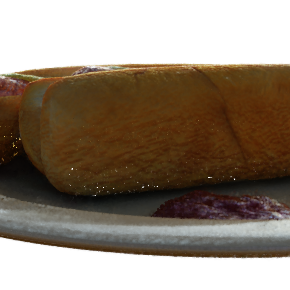} &&
    \includegraphics[width=0.09\textwidth]{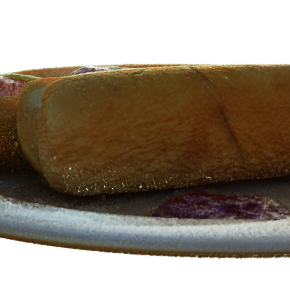} &&
     \includegraphics[width=0.09\textwidth]{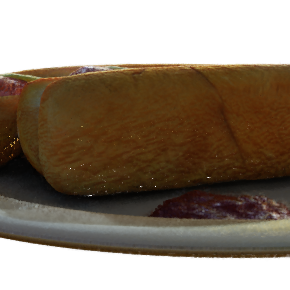} \\
      \includegraphics[width=0.09\textwidth]{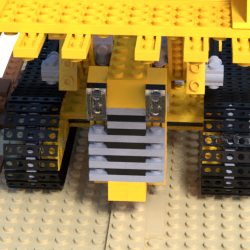} &&
  \includegraphics[width=0.09\textwidth]{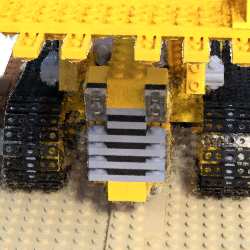} &&
   \includegraphics[width=0.09\textwidth]{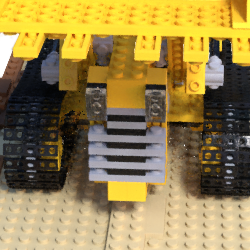} &&
    \includegraphics[width=0.09\textwidth]{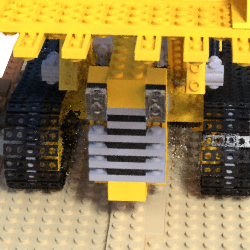} &&
     \includegraphics[width=0.09\textwidth]{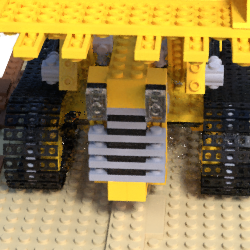} \\
\end{tabular}
\caption{Qualitative ablation on \textit{Hotdog} and \textit{Lego}. It is observed that restricted sampling (16 IPR) results in significant geometric noise. While increasing the budget (64  IPR) or removing hash initialization with full budget recovers structure, the full model is required to effectively resolve high-frequency details and specular highlights.}
\label{fig:ablationstudy}
\vspace{-0.7cm}
\end{wrapfigure}

\textbf{Ablation Study}
An ablation study is conducted on the NeRF Synthetic dataset to evaluate the impact of the sampling budget and the feature initialization strategy. The results are visualized in Figure \ref{fig:ablationstudy}. Detailed quantitative tables are provided in the Appendix.

\textbf{Sampling Budget} The influence of the Ray Intersection Selector (RIS) is analyzed by varying the maximum Intersections Per Ray (IPR). It is observed that a restricted budget of 16 IPR leads to significant geometric degradation and visual noise. While increasing the capacity to 64 IPS recovers structural coherence, the full budget is required to resolve high-frequency details and maximize reconstruction fidelity. This confirms that sufficient depth coverage is critical for the accurate integration of the neural field.

\textbf{Feature Initialization} The contribution of the Splash Grid strategy is assessed by substituting the multi-resolution hash grid with random feature initialization (``w/o hash``). Although the quantitative difference is marginal, qualitative inspection reveals that the inductive bias provided by the hash grid is beneficial for capturing fine textural details and accurate view-dependent effects, such as specular highlights, which appear sharper in the full model.
\vspace{-0.2cm}
 \section{Conclusion}
In this work, \our{} is introduced as a unified neural rendering framework that reconciles the speed of discrete splatting with the high-fidelity continuity of volumetric fields.
By coupling explicit Neural Anchors with an analytical Ray Intersection Selector (RIS), the computational redundancy of stochastic sampling is eliminated, enabling interactive inference rates. Crucially, discrete Gaussian intersections are transformed by the Ray-Coherent Aggregation (RCA) mechanism into a mathematically continuous volumetric function via sliding-window feature interpolation. This enables \our{} to resolve high-frequency details typical of implicit representations, even in challenging, unbounded environments. Finally, extensive experiments confirm that explicit geometric anchors, combined with RIS and Infinitesimal Surface Transformations, maintain view-dependent consistency under geometric changes. These capabilities establish \our{} as a robust and efficient solution for unbounded, interactive 3D scene manipulation.
\textbf{Limitation} The finite nature of explicit anchors limits the precision of far-field geometry reconstruction in unbounded outdoor environments.

\section*{Acknowledgements}
M. Zieliński and D. Belter were supported by the National Science Centre, Poland, under research project no UMO-2023/51/B/ST6/01646. The scientific work was carried out using the infrastructure of the Poznan Supercomputing and Networking Center. The work of P. Spurek was supported by the National Centre of Science (Poland) Grant No. 2023/50/E/ST6/00068.


%
%
\bibliographystyle{splncs04}

\clearpage
\appendix

\input{supplementary_body}

\end{document}

%% file: supplementary_body.tex
\definecolor{colorA}{RGB}{247, 166, 166} 
\definecolor{colorB}{RGB}{250, 195, 184} 
\definecolor{colorC}{RGB}{253, 225, 197} 
\definecolor{colorD}{RGB}{255, 239, 190} 
\definecolor{colorE}{RGB}{255, 253, 205} 

\newcommand{\ce}{\cellcolor{colorE}} 

\section{Implementation details}

In this section, an in-depth description of the ray geometry intersection algorithms and the detailed experimental setup utilized in the main paper is provided.
\\
\\
\textbf{Ray-Geometry Intersection Logic}

The samples are determined as the maximum response points of Gaussians intersected by each ray. For efficient intersection testing, the slightly modified 3DGRT~\cite{moenne20243d} OptiX API-based approach is followed. High-performance Ray Generation and Any-hit shaders are utilized, within which the samples are gathered and stored in the output buffer comprising the 2-tuples of the intersection $t_{\mathrm{sample}}$-parameter and the corresponding Gaussian index. The $t_{\mathrm{sample}}$-parameter is determined in the Any-hit shader via the following computationally-efficient formula:

\begin{equation}
t_{\mathrm{sample}} = - \frac{\left \langle \mathrm{O_{obj}}, \mathrm{v_{obj}} \right \rangle}{\left \langle \mathrm{v_{obj}}, \mathrm{v_{obj}} \right \rangle}
\label{eq:tsample_appendix}
\end{equation}
with $\mathrm{O_{obj}}$ and $\mathrm{v_{obj}}$ denoting the ray origin and direction in the object coordinate frame of the underlying Gaussian, respectively. To reduce the number of the computationally heavy \verb|optixTrace| calls, the majority of the samples aggregation logic is delegated to the Any-hit shader, where the intersection data is accumulated and sorted in the hit buffer. Once the 16-entry hit buffer capacity is reached, the samples are transferred from the hit buffer to the output buffer in the Ray generation shader.

The above-mentioned routine is performed iteratively until the total number of gathered samples reaches a predefined target sample quota, which serves as a hyperparameter of the model. Since the ray-geometry intersection tests constitute the bottleneck of the OptiX pipeline, to further benefit from the hardware-accelerated ray-triangle intersection tests with the intersection logic offloaded to dedicated RT cores, the proxy geometry approach is employed. Each Gaussian, whose size matches the confidence ellipsoid of the underlying $\chi^2$ distribution with 3 degrees of freedom corresponding to the quantile hyperparameter $\chi^2_3$, is conservatively bounded by a regular, affine-transformed icosahedron. 

To minimize the memory footprint, the OptiX geometry instancing mechanism is leveraged, whereby triangle vertices and indices are stored exclusively for a canonical regular icosahedron, whereas for each instance, a dedicated CUDA kernel populates the instance buffer entries by computing the affine transformation matrices that map the reference proxy to the resulting Gaussian primitive.
\\
\\
\textbf{Experiments setup}

The framework is implemented in PyTorch within the Nerfstudio~\cite{nerfstudio} ecosystem. All experiments were conducted on a single NVIDIA RTX 5090 GPU.

\textbf{Initialization.} Anchor initialization strategies are tailored to the specific scene topology. For the object-centric NeRF Synthetic dataset~\cite{mildenhall2020nerf}, point clouds generated by the NeuralEditor~\cite{chen2023neuraleditor} pipeline are employed to ensure consistent evaluation baselines. In contrast, for large-scale unbounded scenes such as Mip-NeRF 360~\cite{barron2022mipnerf360}, sparse geometry derived from a pre-trained 3D Gaussian Splatting`\cite{kerbl20233d} model is utilized, as it provides a robust starting distribution for complex background elements.

\textbf{Hyperparameters.} Parameter configurations are adapted to the specific characteristics of the dataset. For bounded, object-centric scenes (NeRF Synthetic), the \textbf{Ray Intersection Selector (RIS)} is configured with a hit buffer of 128 intersections per ray. The bounding proxy geometry for each anchor is defined by a squared Mahalanobis distance threshold of $\lambda=6.25$, ensuring coverage of approximately $90\%$ of the Gaussian probability mass. For unbounded, large-scale environments (Mip-NeRF 360), the intersection buffer is expanded to 256 entries, and the proxy extent is increased to $\lambda=11.3449$ ($99\%$ mass coverage) to robustly capture distant background geometry. 

\begin{table}[t]
\caption{Ablation study conducted on the static NeRF Synthetic dataset. The impact of the sampling budget is evaluated by varying the maximum number of intersections processed per ray (denoted as IPR), comparing restricted budgets of 16 and 64 against the full model's 128. Additionally, the role of feature initialization is assessed by substituting the Hash Grid strategy with random initialization ''w/o hash''). It is observed that increasing the intersection quota significantly improves reconstruction fidelity, validating the need of sufficient depth coverage, while the Hash Grid initialization further contributes to the optimal performance of the full model.}
\vspace{-4mm}
            {\small 
            \begin{center}
               \setlength{\tabcolsep}{3pt}
    \renewcommand{\arraystretch}{0.9}
    {\fontsize{6.8pt}{11pt}\selectfont
    \begin{tabular}{@{}lccccccccc}
  & Chair & Drums & Lego & Mic &
      Mater. & Ship &Hotdog& Ficus & Avg.\\
             \hline
            \multicolumn{9}{c}{ PSNR } \\ 
            \hline
            16 IPR & \cd 32.39 & \cd 24.81 & \cd 30.69 & \cd 32.87 & \cd 27.18 & \cd 28.35 & \cd 34.60 & \cd 28.05 & \cd 29.87 \\
            64 IPR & \cb 34.94 & \cb 25.61 & \cb 35.43 & \cc 35.16 & \cb 29.40 & \ca 30.80 & \cb 36.95 & \cc 31.45 & \cc 32.47 \\
            w/o hash & \cc 34.70 & \ca 25.91 & \cc 34.87 & \cb 35.79 & \cc 29.35 & \cc 30.58 & \cc 36.68 & \ca 32.47 & \cb 32.54 \\
            full & \ca 35.00 & \cc 25.51 & \ca 35.54 & \ca 35.87 & \ca 29.46 & \cb 30.69 & \ca 37.02 & \cb 32.12 & \ca 32.65 \\
            \hline
                \end{tabular}}
                \end{center}
                }
                \label{tab:ablation}
\vspace{-0.8cm}
\end{table}

Regarding the neural architecture, the \textbf{Ray-Coherent Aggregation (RCA)} window size is consistently set to $N=2$. Feature initialization is performed using a multi-resolution Hash Grid with 16 levels, spanning a resolution range from $N_{min}=16$ to $N_{max}=8192$ with a geometric growth factor. Each level stores 2 feature dimensions within a hash map of size $2^{21}$. The implicit geometry decoder is implemented as a shallow MLP with 2 hidden layers of 64 units and ReLU activations. It accepts a 32-dimensional aggregated feature input and produces a 16-dimensional output vector, where the first channel corresponds to the raw density $\sigma$, and the remaining 15 channels serve as the latent embedding for the subsequent color decoding head. This RGB network shares an identical architecture (2 layers, 64 hidden units) and predicts the final 3-channel color value based on the concatenation of the geometric latent embedding and the view direction, encoded via Spherical Harmonics (SH).
\vspace{-0.15cm}
\section{Additional Quantitative Results}
\vspace{-0.1cm}
\begin{table}[t]
    \centering
    {\small 
    \caption{Quantitative comparisons on the NeuralEditor benchmark. The evaluation is conducted on the NeRF Synthetic dataset, where ground-truth deformations are applied to the objects for rigorous comparison. It is observed that \our{} achieves reconstruction quality comparable to the state-of-the-art editable method EKS, even surpassing it on specific scenes.\label{tab:neuraleditor_baseline}}
    
    \setlength{\tabcolsep}{3pt}
    \renewcommand{\arraystretch}{1.0}
    {\fontsize{6.8pt}{11pt}\selectfont
    \begin{tabular}{@{}lccccccccc}
        & Chair & Drums & Lego & Mic & Mater. & Ship &Hotdog& Ficus & Avg.\\
        \hline
        \multicolumn{9}{c}{ PSNR } \\ 
        \hline
        Naive Plotting & \cd 24.58 & \ce 21.54 & \ce 25.38 & \cb 27.56 & \ce 21.59 & \cd 22.21 & \ce 26.72 & \ce 24.62 & \ce 24.27 \\
        Neuraleditor & \cc 25.29 & \cc 21.93 & \cc 27.14 & \cc 27.49 & \cb 23.04 & \cc 24.12 & \cd 27.14 & \cd 24.83 & \cc 25.12 \\
        GaMeS & \ce 24.51 & \cb 22.02 & \cd 26.65 & \ce 27.07 & \cd 21.73 & \ce 22.19 & \cc 27.26 & \cc 26.65 & \cd 24.76 \\
        EKS & \ca 26.03 & \ca 22.08 & \cb 28.04 & \ca 27.85 & \ca 23.14 & \ca 24.43 & \ca 28.23 & \ca 27.58 & \ca 25.92 \\
        \our{} (our) & \cb 25.97 & \cd 21.72 & \ca 28.16 & \cd 27.46 & \cc 22.89 & \cb 24.16 & \cb 27.93 & \cb 27.34 & \cb 25.70 \\
        \hline
        \multicolumn{9}{c}{ SSIM } \\ 
        \hline
        Naive Plotting & \ce 0.930 & \ce 0.892 & \ce 0.904 & \ce 0.956 & \cd 0.867 & \ce 0.807 & \ce 0.930 & \ce 0.925 & \ce 0.901 \\
        Neuraleditor & \cc 0.944 & \cd 0.900 & \cc 0.945 & \cd 0.958 & \cd 0.887 & \cc 0.832 & \cd 0.937 & \cd 0.927 & \cd 0.916 \\
        GaMeS & \cd 0.941 & \ca 0.914 & \cd 0.936 & \cc 0.960 & \cc 0.890 & \cd 0.811 & \cc 0.947 & \cb 0.947 & \cc 0.918 \\
        EKS & \cb 0.957 & \cc 0.910 & \cb 0.961 & \ca 0.964 & \ca 0.911 & \ca 0.855 & \ca 0.962 & \ca 0.951 & \ca 0.934 \\
        \our{} (our) & \ca 0.958 & \cb  0.911 & \ca 0.962 & \ca 0.964 & \cb 0.902 & \cb 0.847 & \cb 0.960 & \cc 0.932 & \cb 0.932 \\
        \hline
        \multicolumn{9}{c}{ LPIPS } \\ 
        \hline
        Naive Plotting & \ce 0.050 & \ce 0.107 & \ce 0.066 & \ce 0.053 & \ce 0.126 & \ce 0.187 & \ce 0.085 & \cd 0.072 & \ce 0.093 \\
        Neuraleditor & \cd 0.041 & \cd 0.100 & \cd 0.038 & \cd 0.050 & \cd 0.103 & \cc 0.158 & \cd 0.078 & \cc 0.069 & \cd 0.080 \\
        GaMeS & \cc 0.039 & \ca 0.067 & \cc 0.035 & \ca 0.032 & \cb 0.077 & \cd 0.177 & \cc 0.046 & \ca 0.036 & \cc 0.064 \\
        EKS & \cb 0.030 & \cb 0.071 & \cb 0.023 & \cc 0.036 & \ca 0.062 & \cb 0.143 & \ca 0.037 & \ca 0.036 & \ca 0.055 \\
        \our{} (our) & \ca 0.029 & \cc 0.079 & \ca 0.022 & \cb 0.035 & \cc 0.085 & \ca 0.121 & \cb 0.040 & \cb 0.056 & \cb 0.056 \\
        \hline
    \end{tabular}
    }
    } 
\end{table}

The results presented in the main paper are expanded upon this section, providing a detailed breakdown of the ablation study and the deformation benchmark metrics.
\\
\\
\textbf{Ablation Study}

In Tab.~\ref{tab:ablation}, the impact of the sampling budget and feature initialization on reconstruction quality is analyzed. The results demonstrate that the Ray Intersection Selector benefits significantly from a larger intersection budget. It is observed that increasing the maximum intersections per ray (IPR) from 16 to 128 yields a consistent improvement in PSNR across all scenes, confirming that capturing deep structural occlusions is critical for high-fidelity rendering. Furthermore, it is shown that removing Hash Grid initialization (``w/o hash``) results in a slight drop in average PSNR.
\\
\\
\textbf{NeuralEditor Benchmark}

Tab.~\ref{tab:neuraleditor_baseline} presents a scene-by-scene quantitative comparison against editable baselines on the NeRF Synthetic deformation task. State-of-the-art results are achieved on \textit{Lego} scene, and highly competitive performance is maintained relative to EKS~\cite{zielinski2025genie}. Notably, other methods like Naive Plotting~\cite{neuraleditor}, NeuralEditor or GaMeS~\cite{waczynska2024games} are outperformed.
\\
\\
\\
\\
\textbf{Additional Qualitative Results}

\begin{figure*}[t]
\centering
\setlength{\tabcolsep}{1pt}
{\fontsize{6.8pt}{11pt}\selectfont
\begin{tabular}{ccccccc}
 & GT & TS & 3DGS & INGP   & MipNeRF-360 & \our{} (our) \\
\rotatebox{90}{\footnotesize\hspace{12pt}{Room}} &\includegraphics[width=0.15\textwidth]{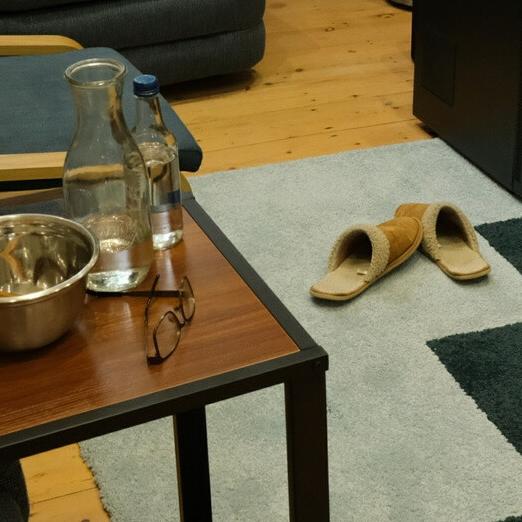}   & \includegraphics[width=0.15\textwidth]{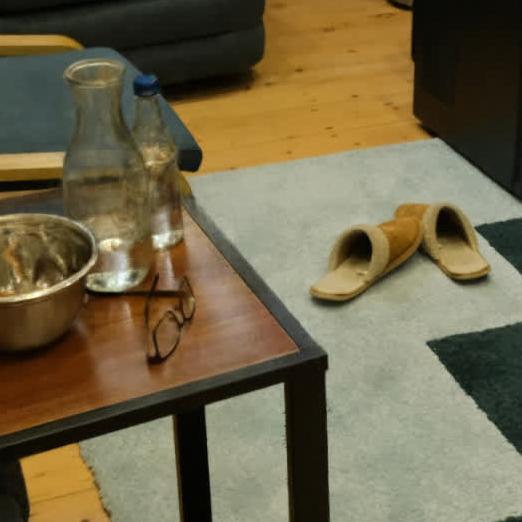}  & 
\includegraphics[width=0.15\textwidth]{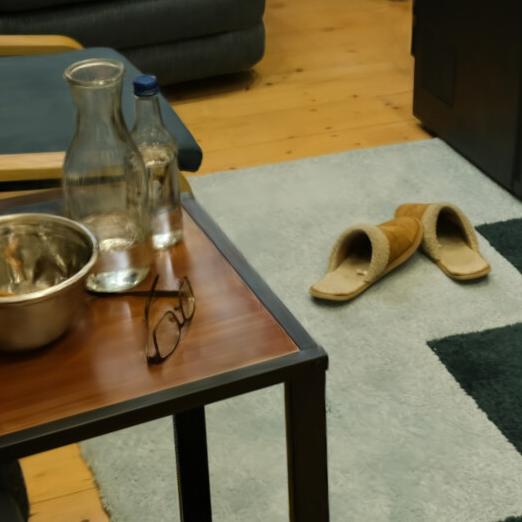}    &  \includegraphics[width=0.15\textwidth]{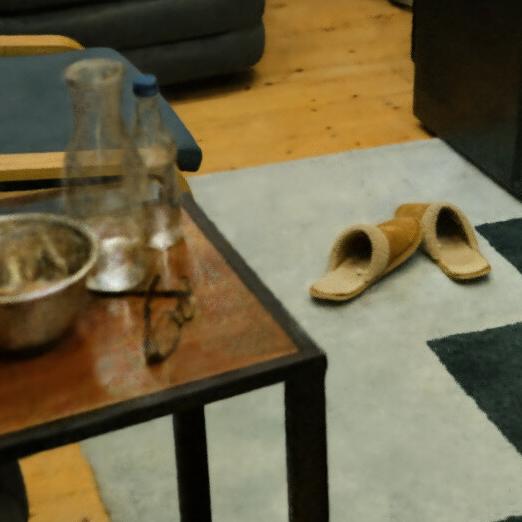}  &
\includegraphics[width=0.15\textwidth]{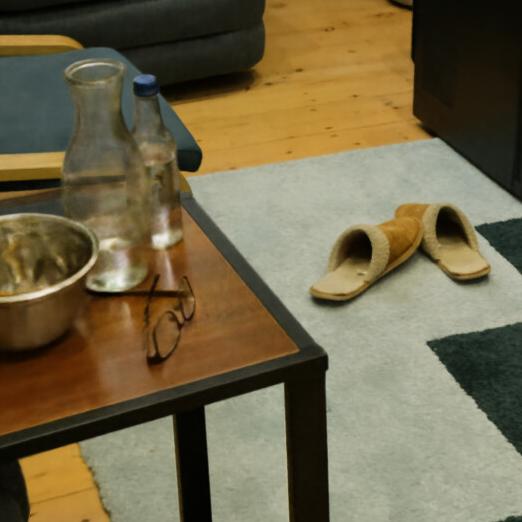}    & 
\includegraphics[width=0.15\textwidth]{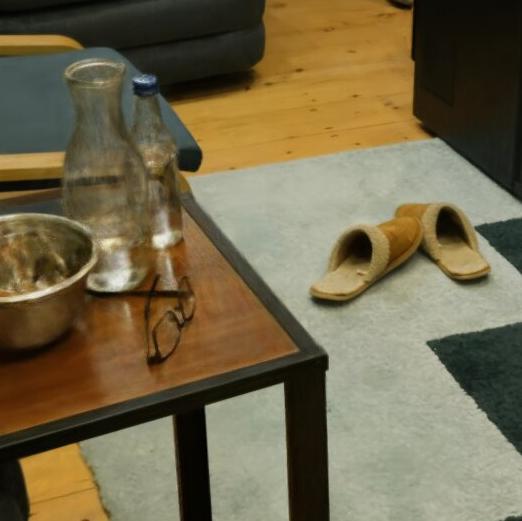}  \\
\rotatebox{90}{\footnotesize\hspace{6pt}{Playroom}} &\includegraphics[width=0.15\textwidth]{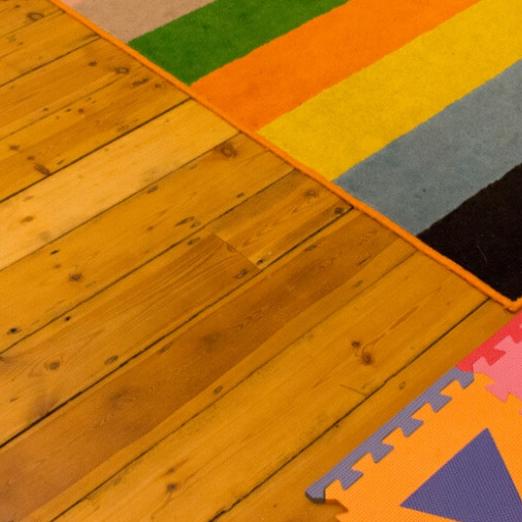}   & \includegraphics[width=0.15\textwidth]{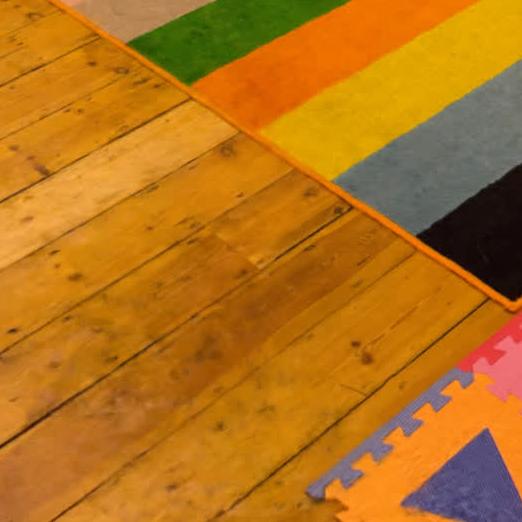}  & 
\includegraphics[width=0.15\textwidth]{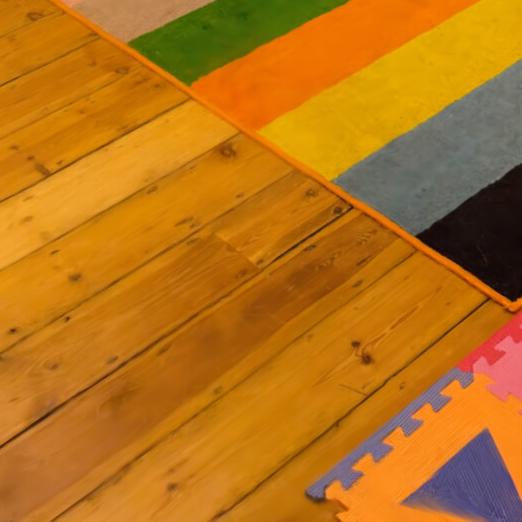}    &  \includegraphics[width=0.15\textwidth]{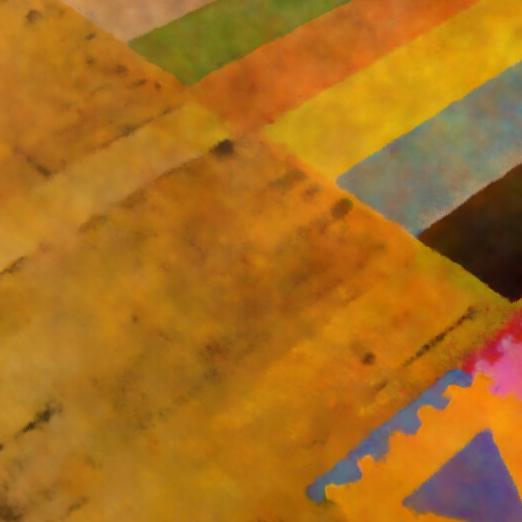}  &
\includegraphics[width=0.15\textwidth]{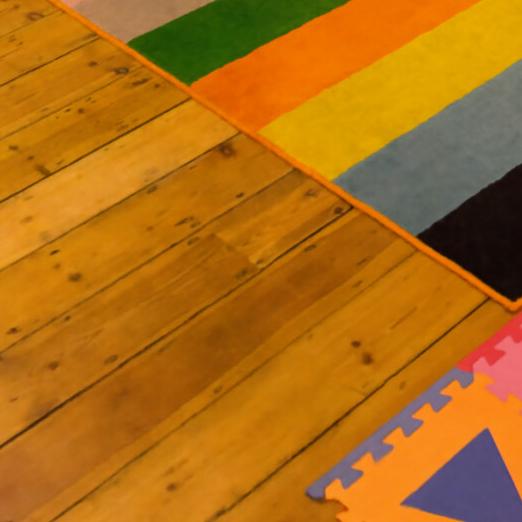}    & 
\includegraphics[width=0.15\textwidth]{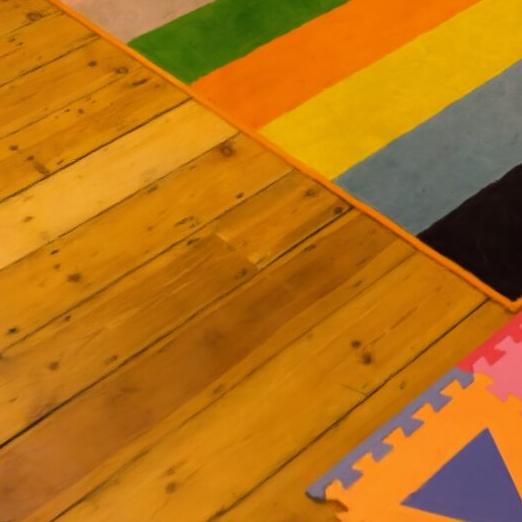}   \\ 
\end{tabular}}
\caption{Extended qualitative comparison (close-ups). This figure extends Fig. 5 from the main paper by providing additional crop views from scenes on the Mip-NeRF 360 and Deep Blending datasets. TS denotes Triangle Splatting. It is observed that \our{} preserves high-frequency structural details, achieving quality competitive with static state-of-the-art baselines like Mip-NeRF 360.}
\label{fig:comparison_crops}
\end{figure*}

\begin{figure*}[t]
\centering
\setlength{\tabcolsep}{1pt}
{\fontsize{6.8pt}{11pt}\selectfont
\begin{tabular}{ccccccc}
 & GT & TS & 3DGS & INGP   & MipNeRF-360 & \our{} (our) \\
\rotatebox{90}{\footnotesize\hspace{3pt}{Bonsai}} &
\includegraphics[width=0.15\textwidth]{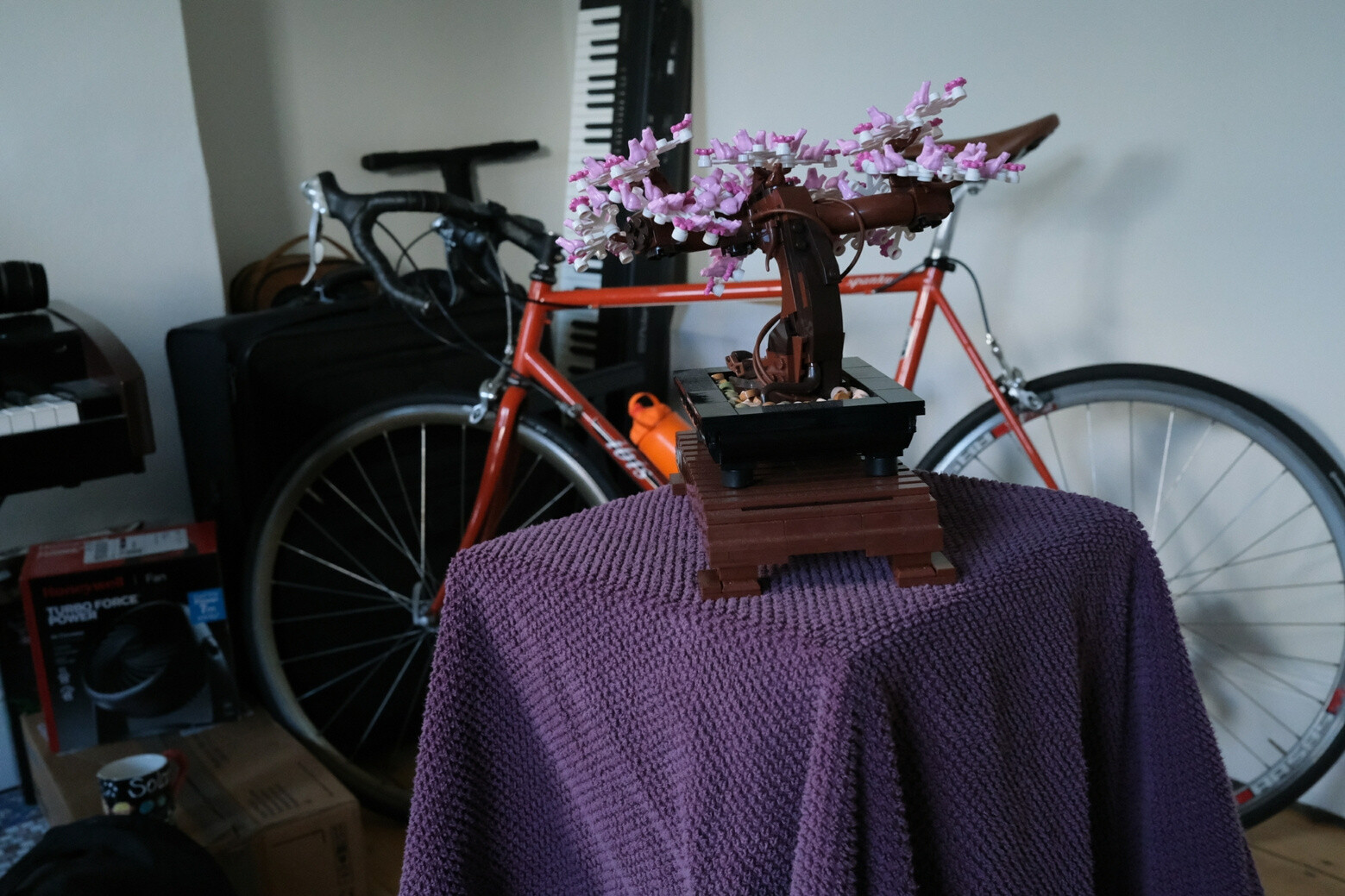}   & \includegraphics[width=0.15\textwidth]{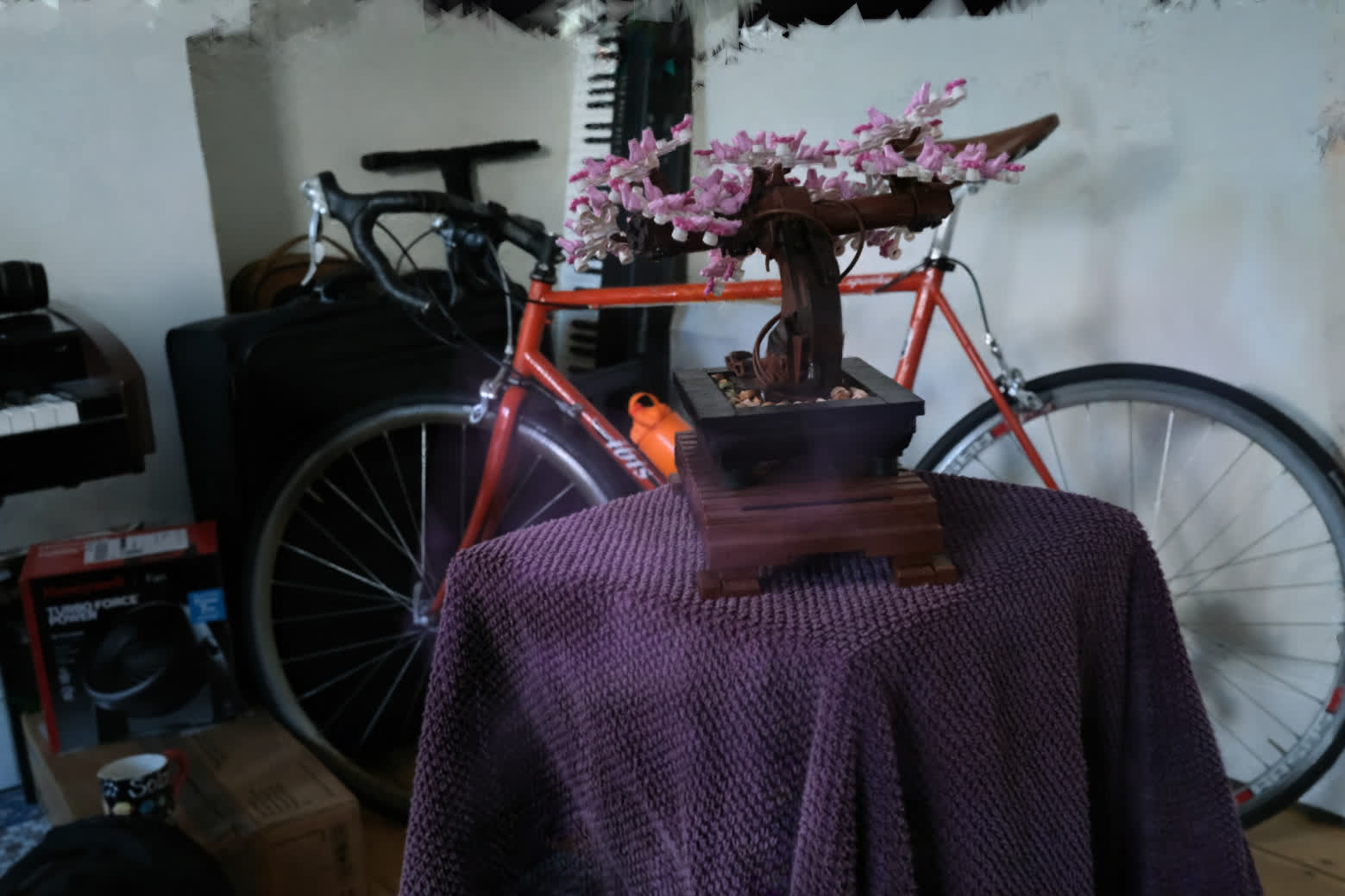}  & 
\includegraphics[width=0.15\textwidth]{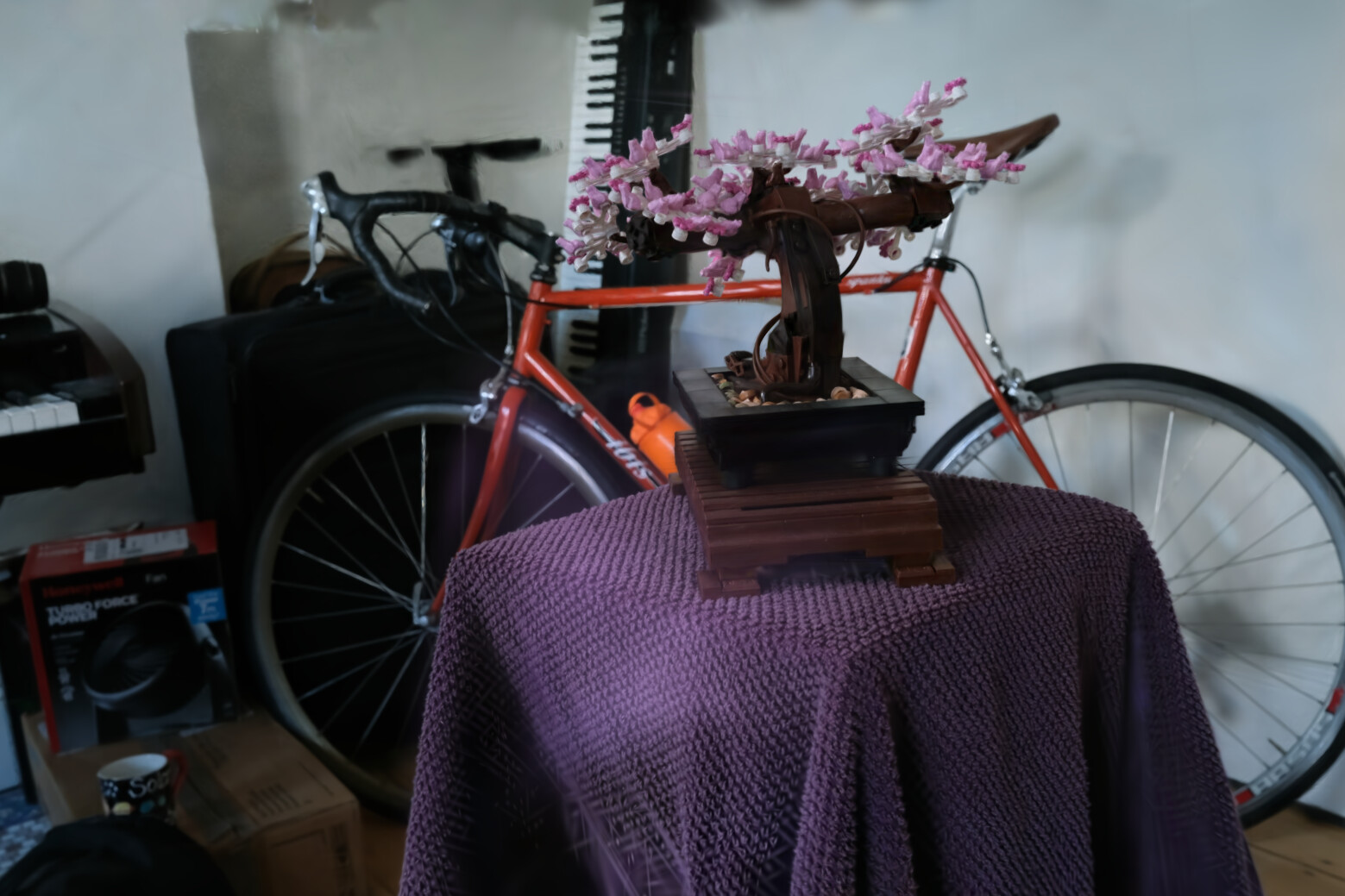}    &  \includegraphics[width=0.15\textwidth]{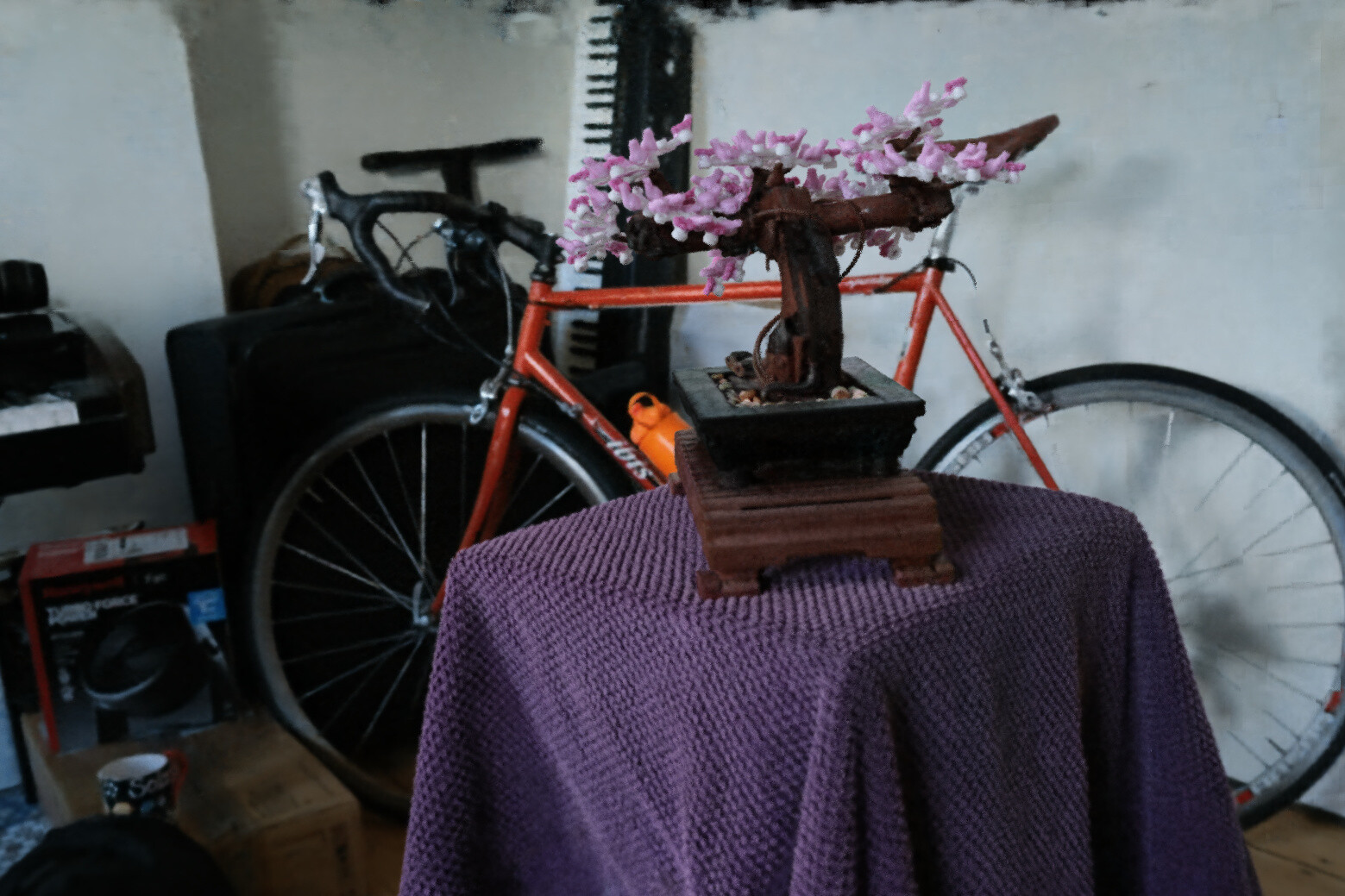}  &
\includegraphics[width=0.15\textwidth]{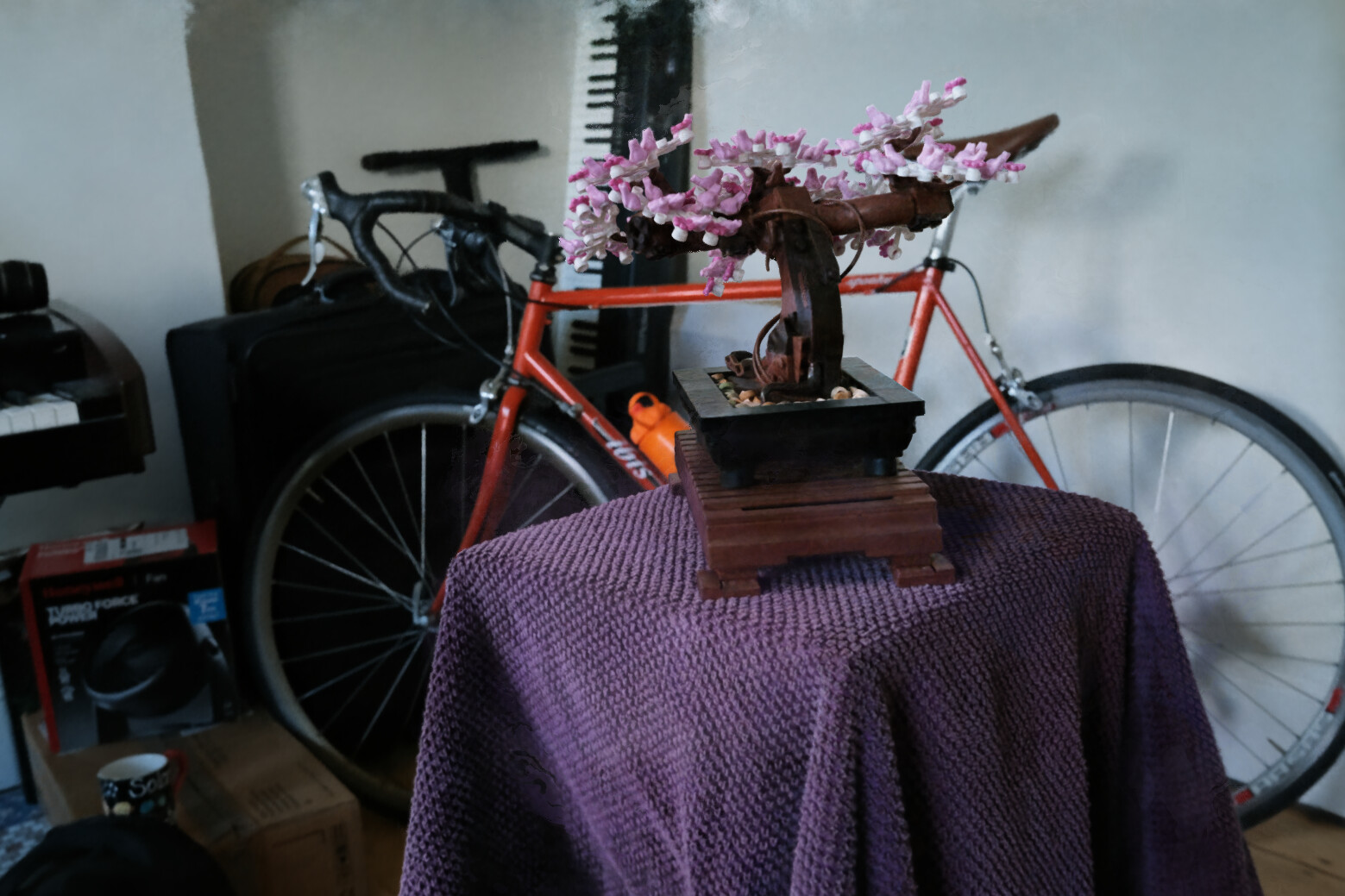}    & 
\includegraphics[width=0.15\textwidth]{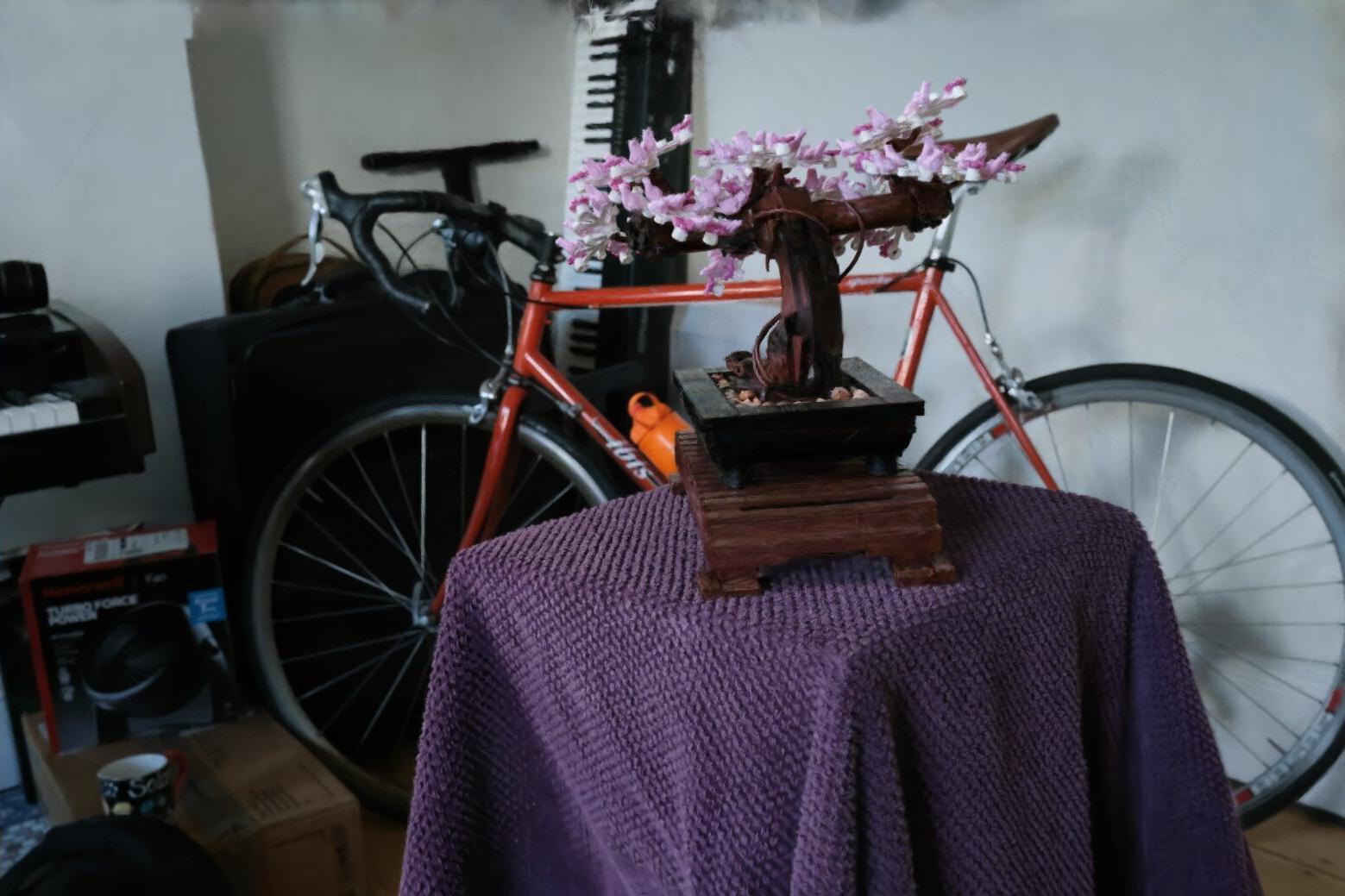}   \\
\rotatebox{90}{\footnotesize\hspace{3pt}{Bicycle}} &\includegraphics[width=0.15\textwidth]{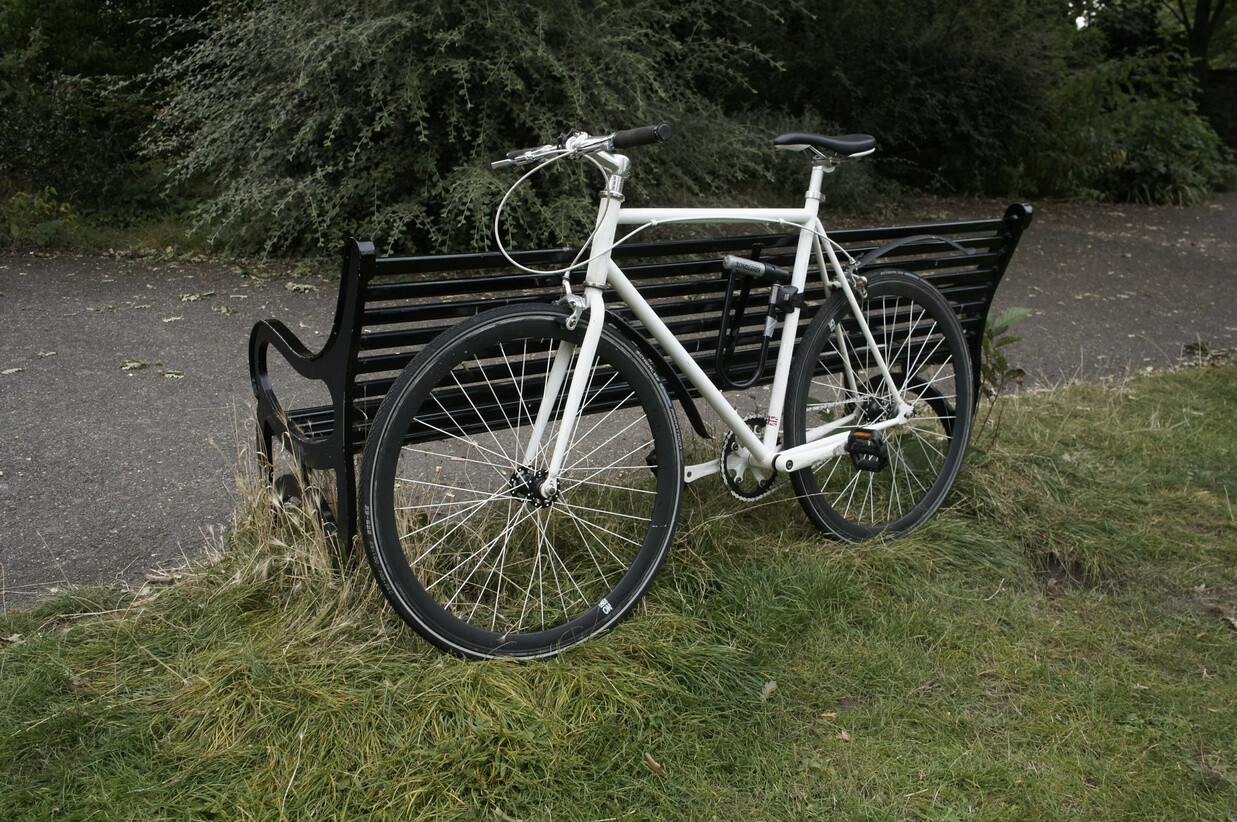}   & \includegraphics[width=0.15\textwidth]{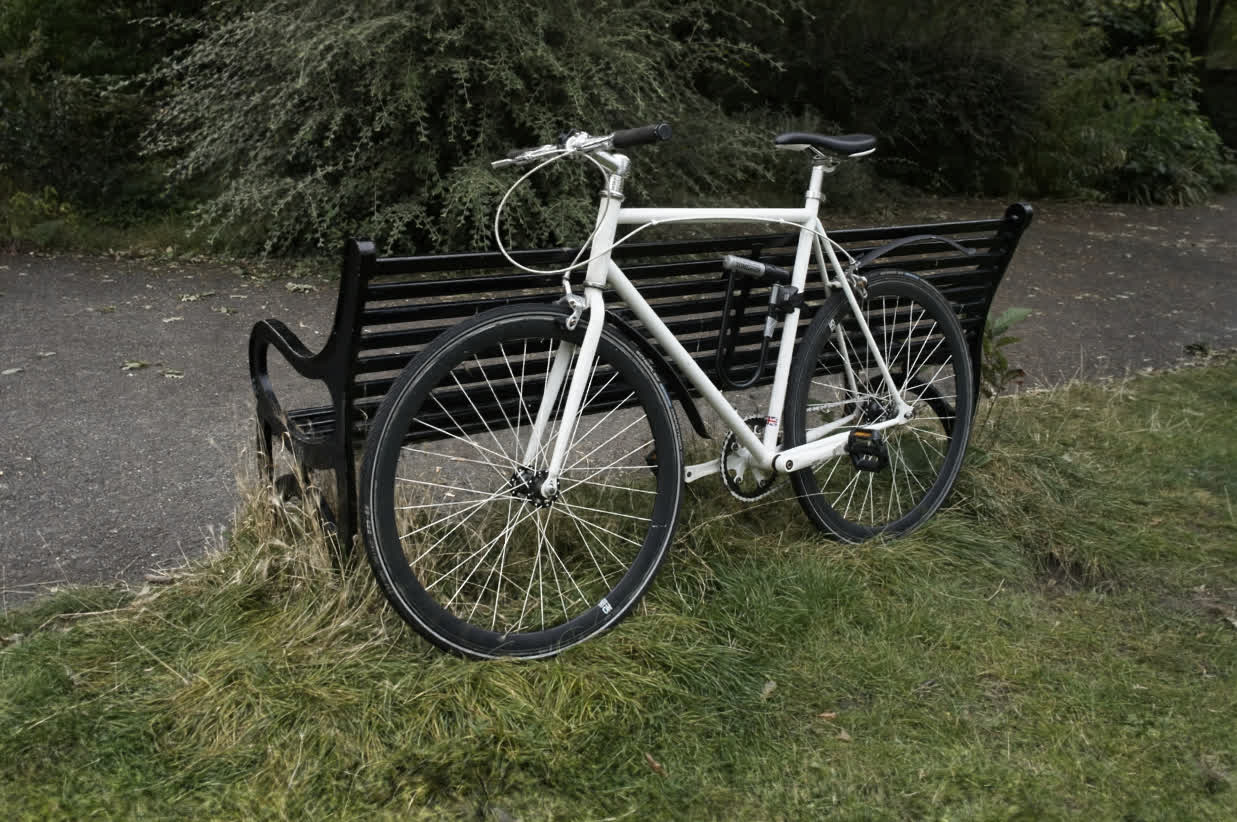}  & 
\includegraphics[width=0.15\textwidth]{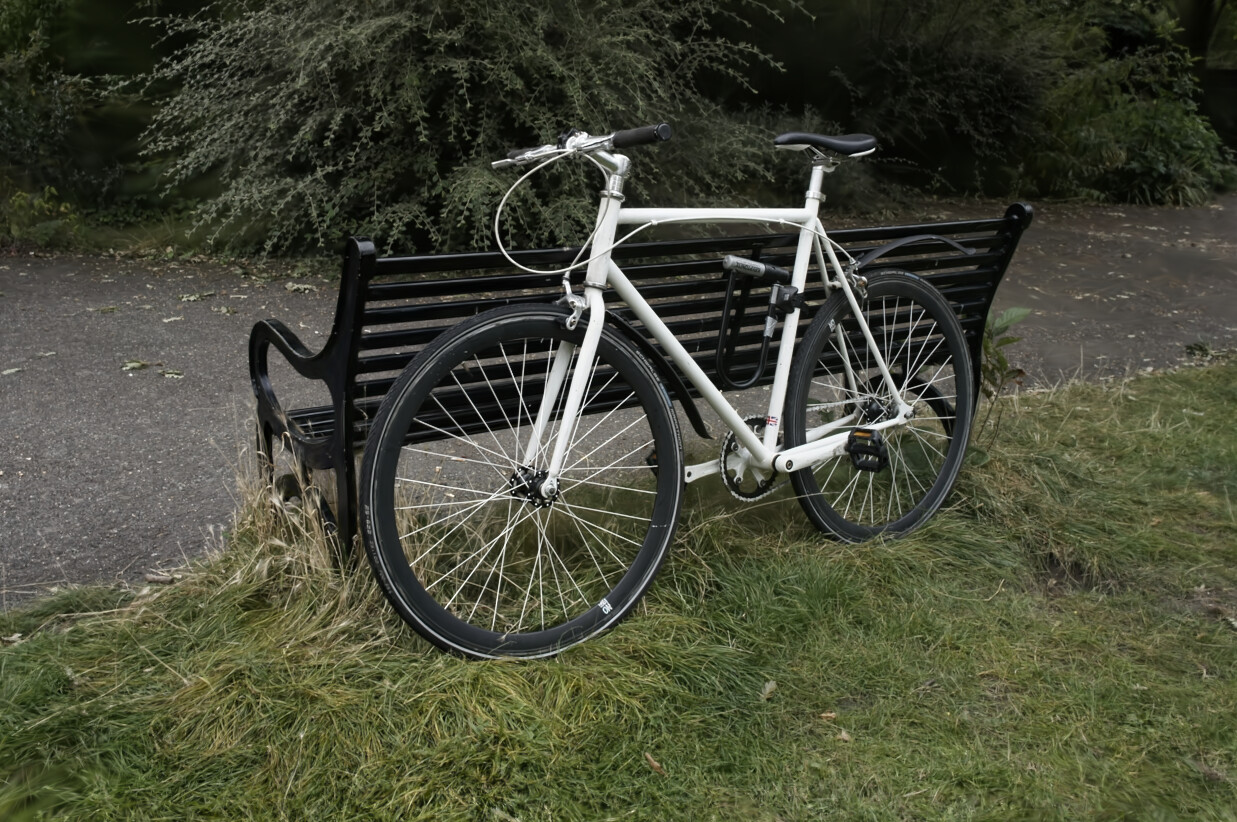}    &  \includegraphics[width=0.15\textwidth]{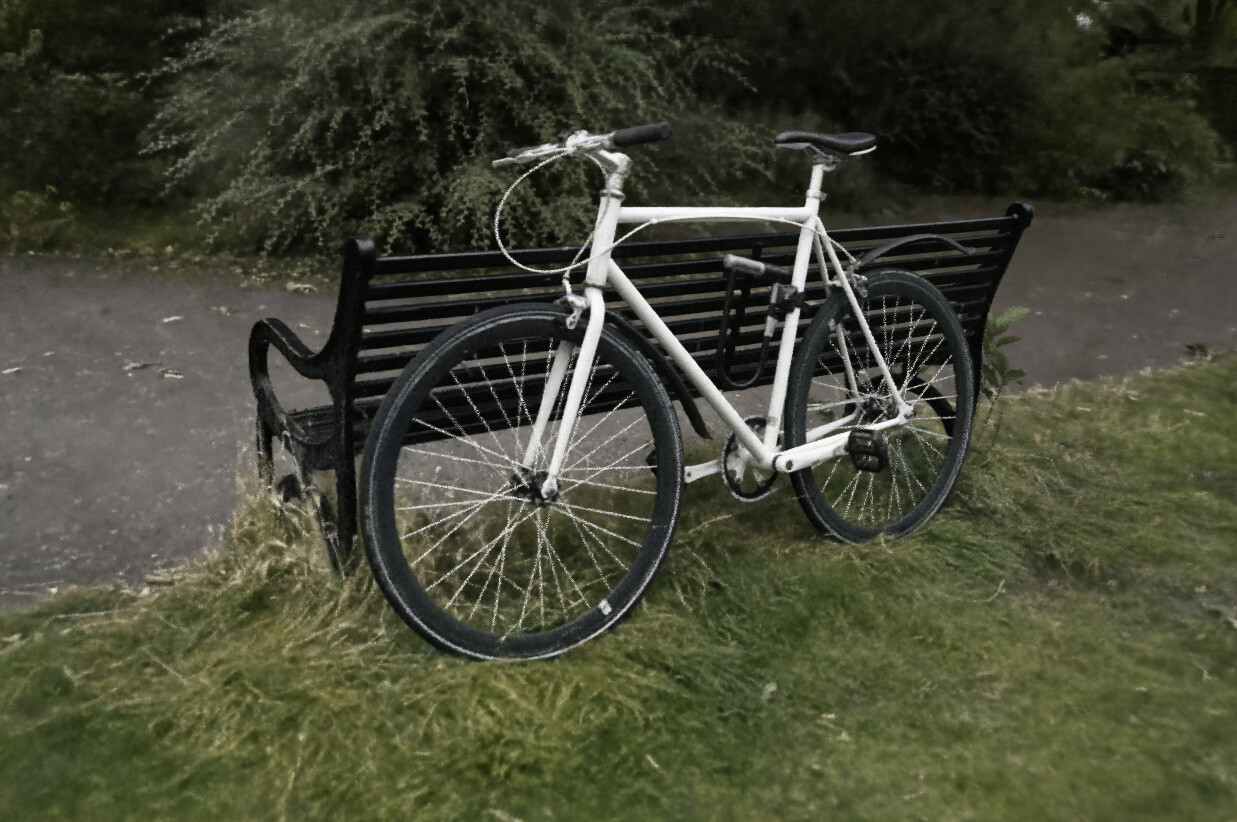}  &
\includegraphics[width=0.15\textwidth]{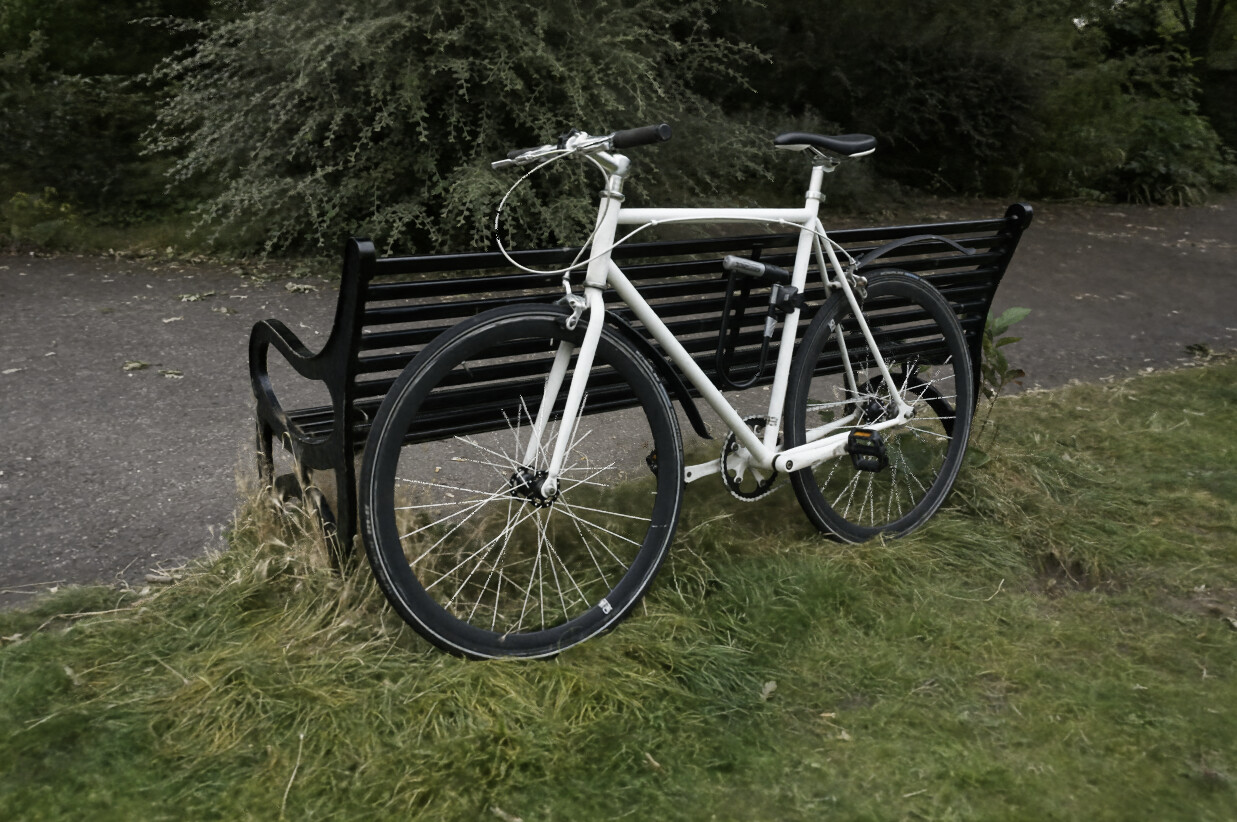}    & 
\includegraphics[width=0.15\textwidth]{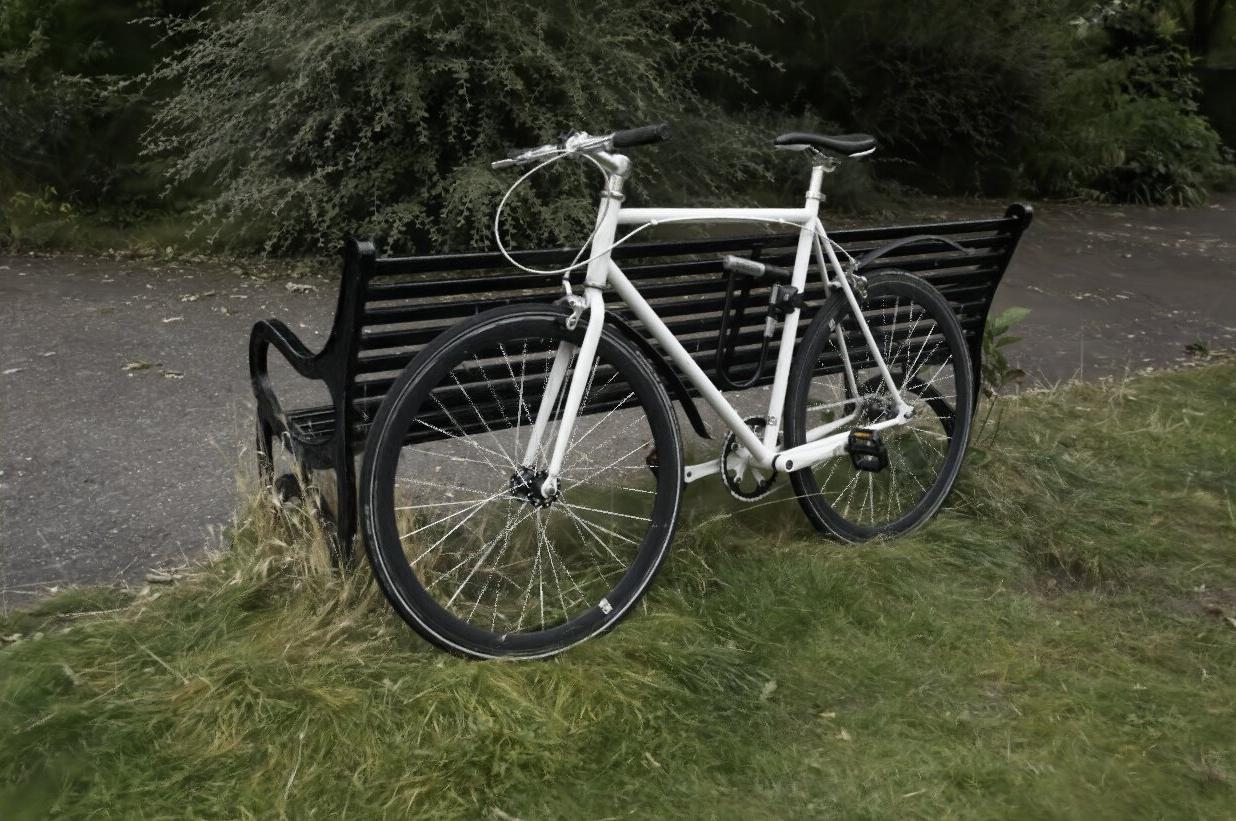}   \\
\rotatebox{90}{\footnotesize\hspace{4pt}{Playr.}} &\includegraphics[width=0.15\textwidth]{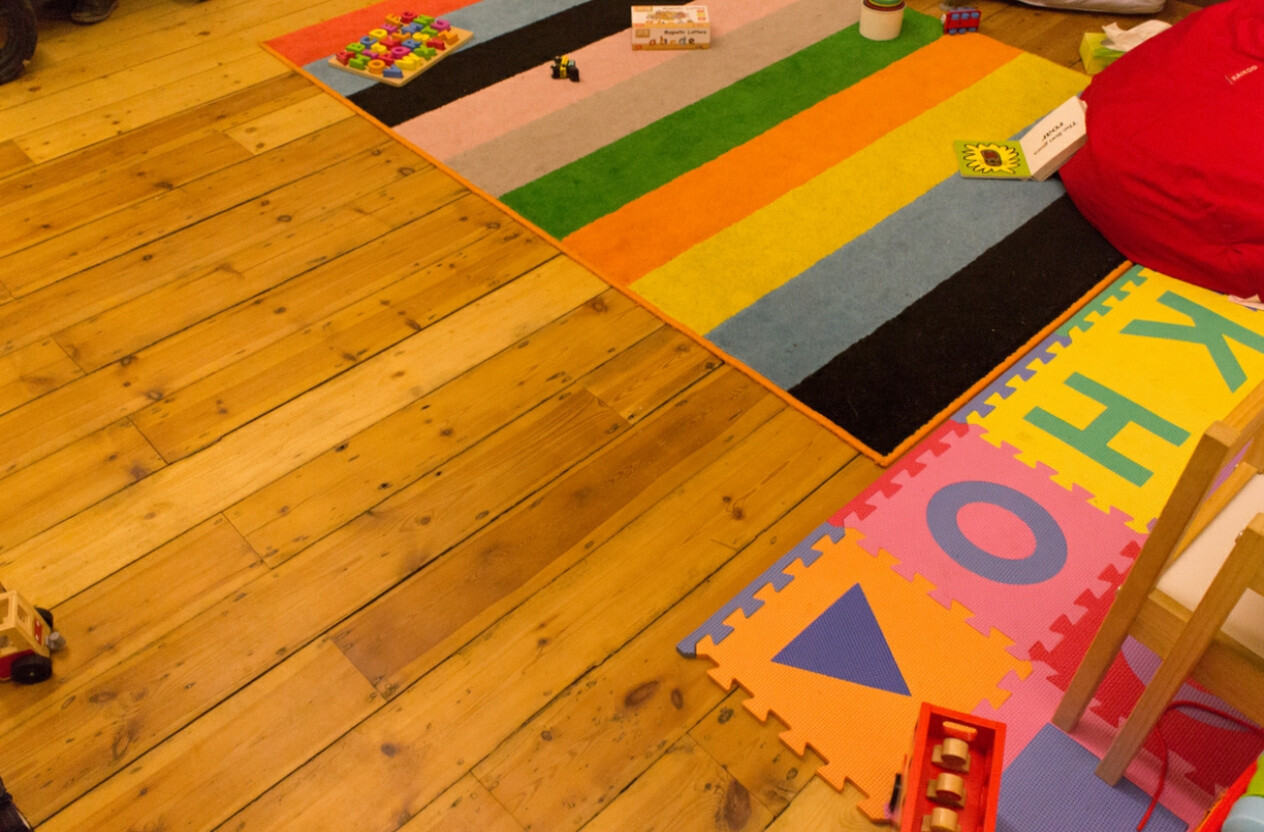}   & \includegraphics[width=0.15\textwidth]{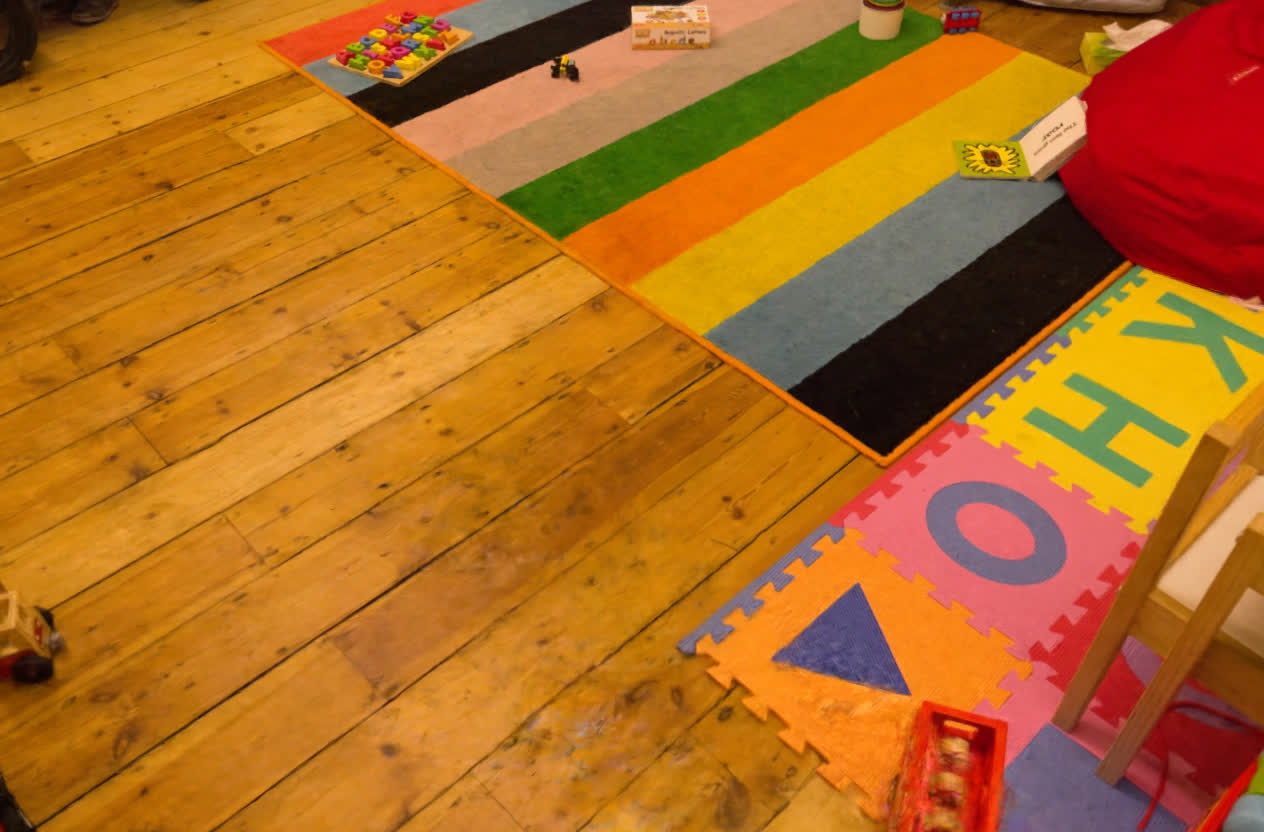}  & 
\includegraphics[width=0.15\textwidth]{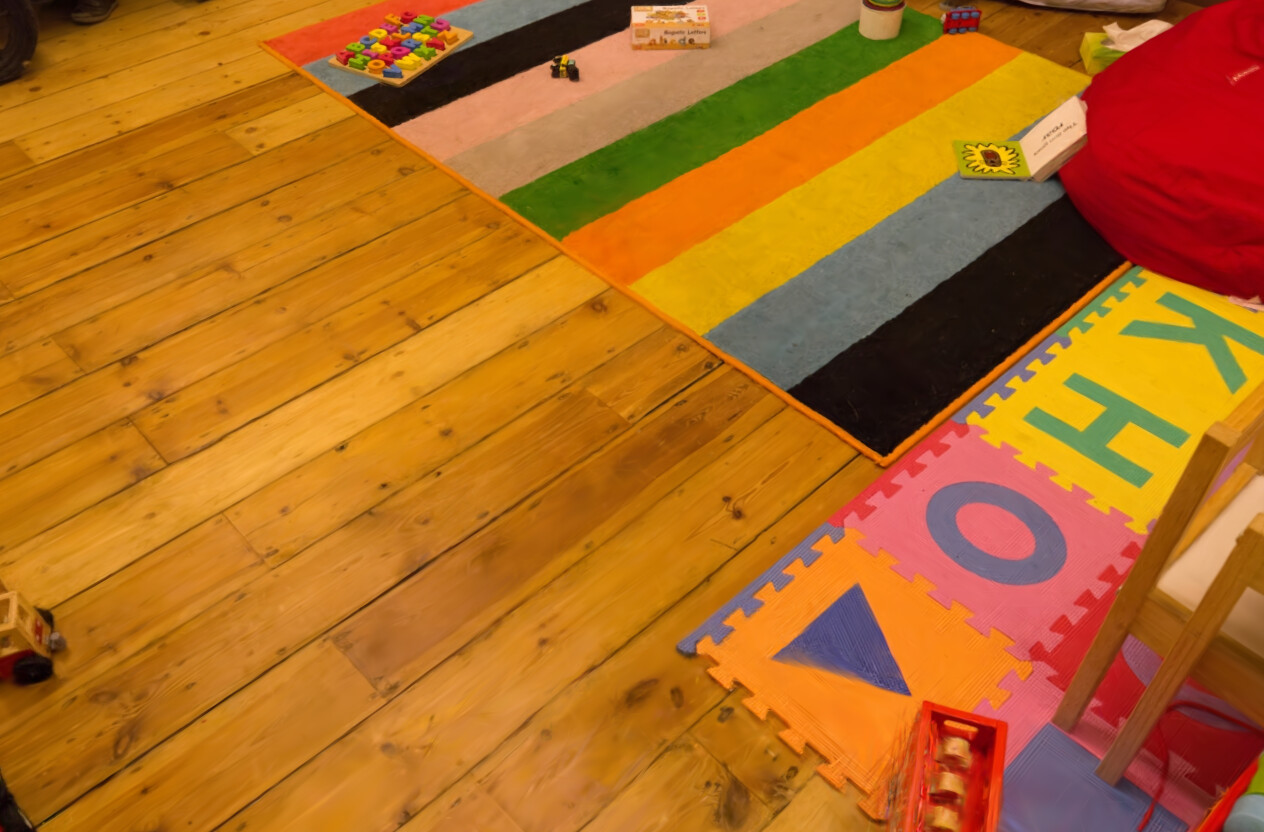}    &  \includegraphics[width=0.15\textwidth]{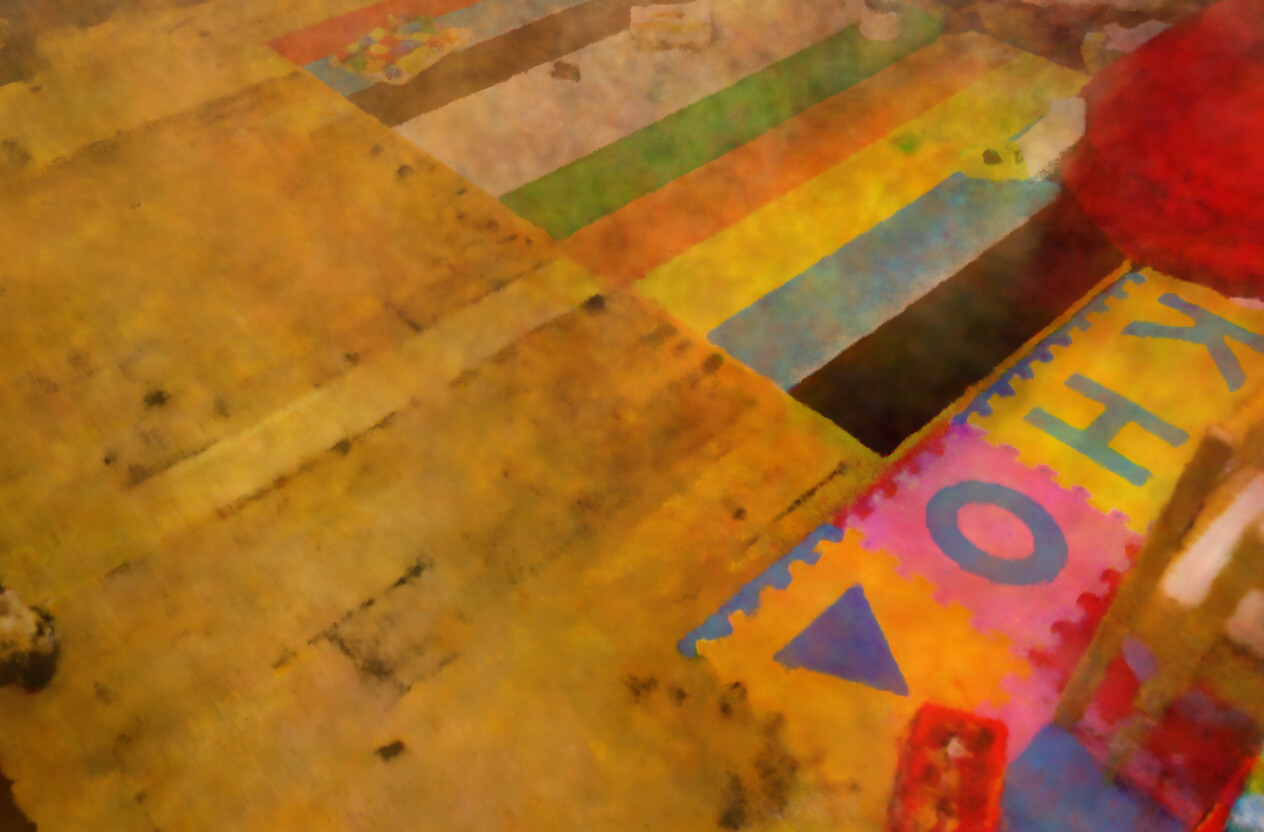}  &
\includegraphics[width=0.15\textwidth]{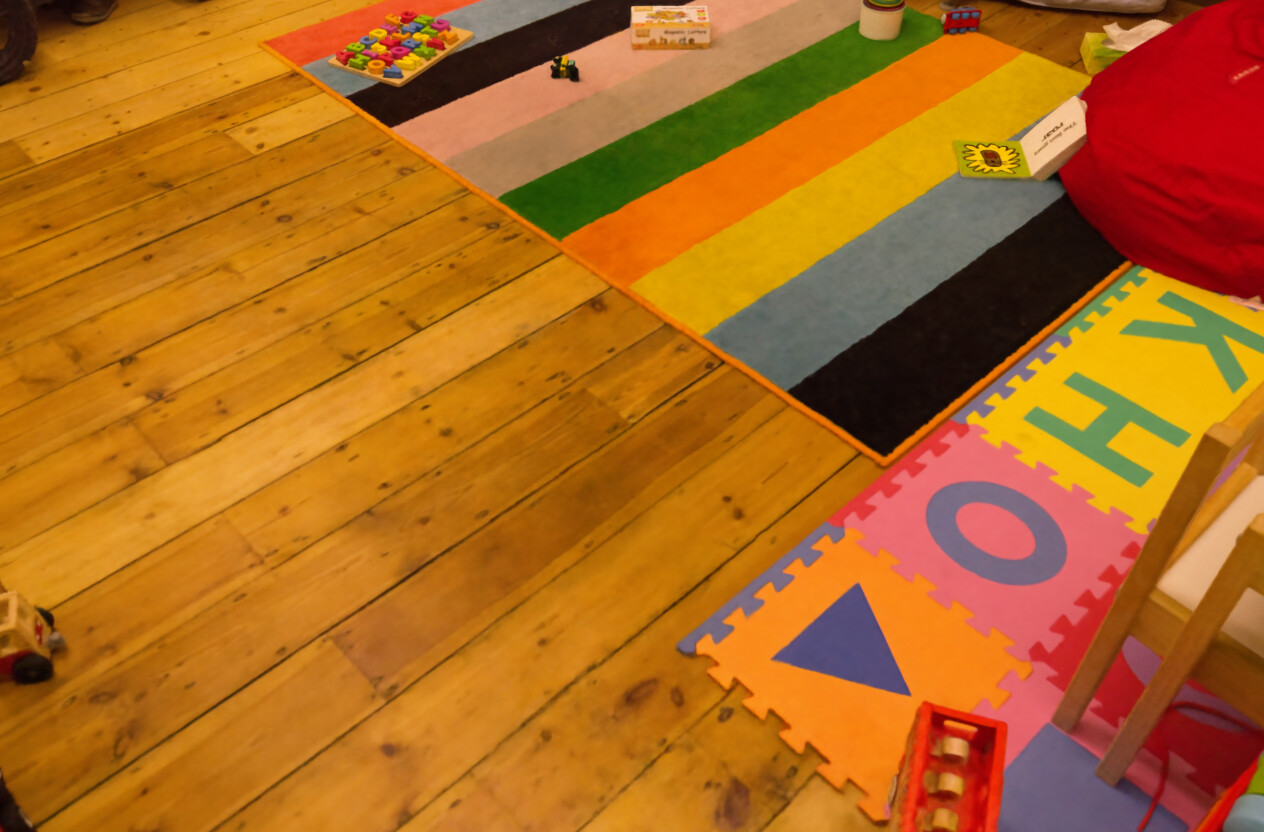}    & 
\includegraphics[width=0.15\textwidth]{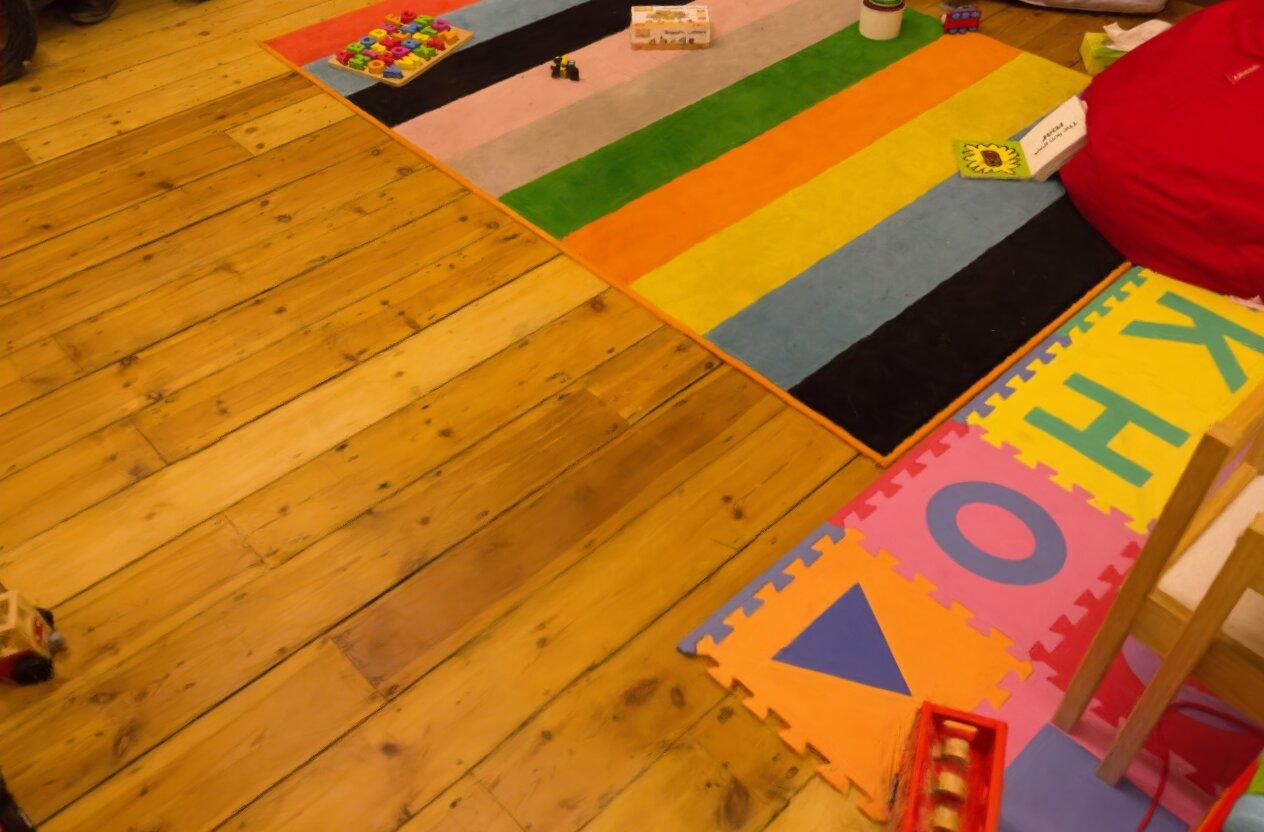}   \\
\rotatebox{90}{\footnotesize\hspace{5pt}{Room}} &\includegraphics[width=0.15\textwidth]{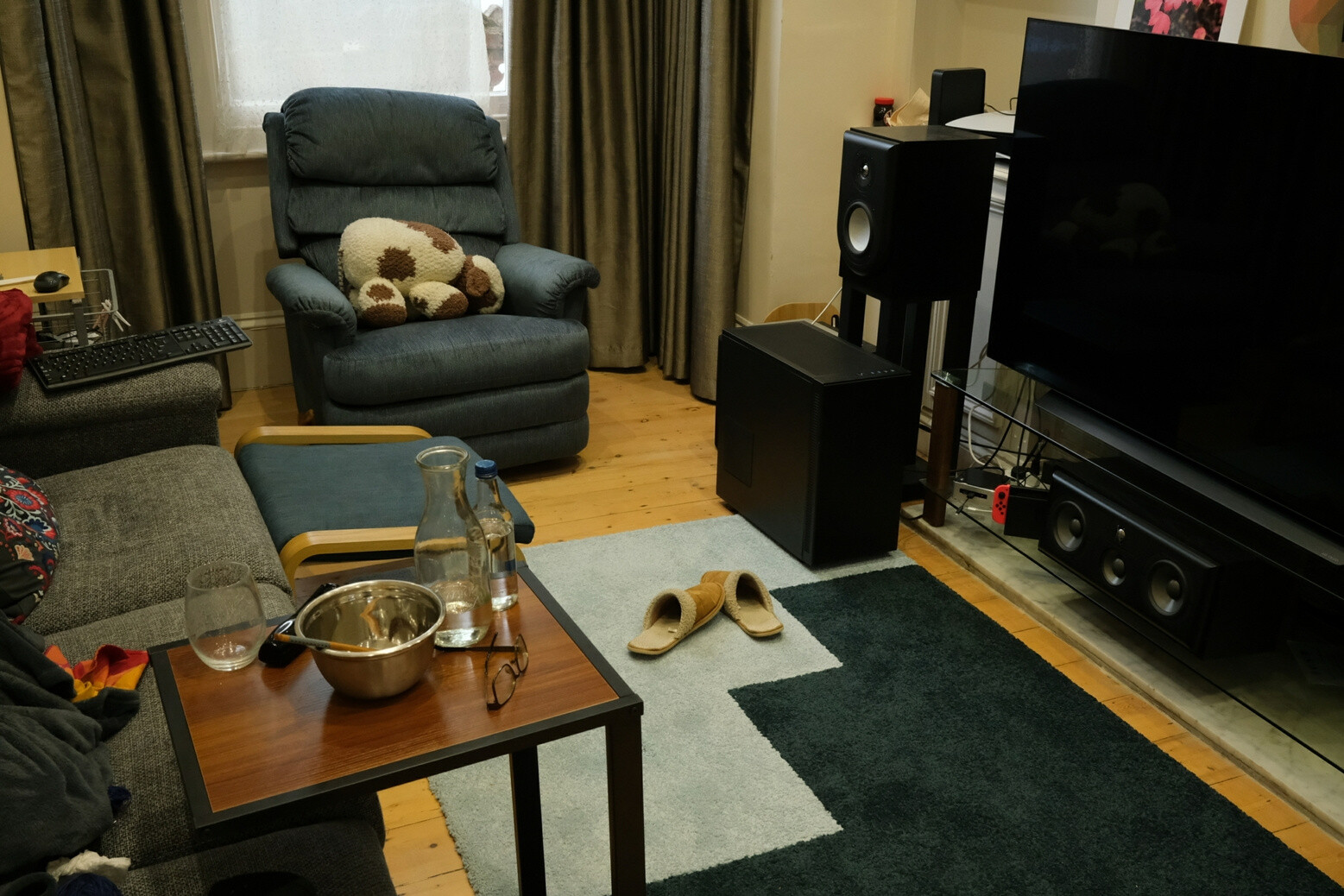}   & \includegraphics[width=0.15\textwidth]{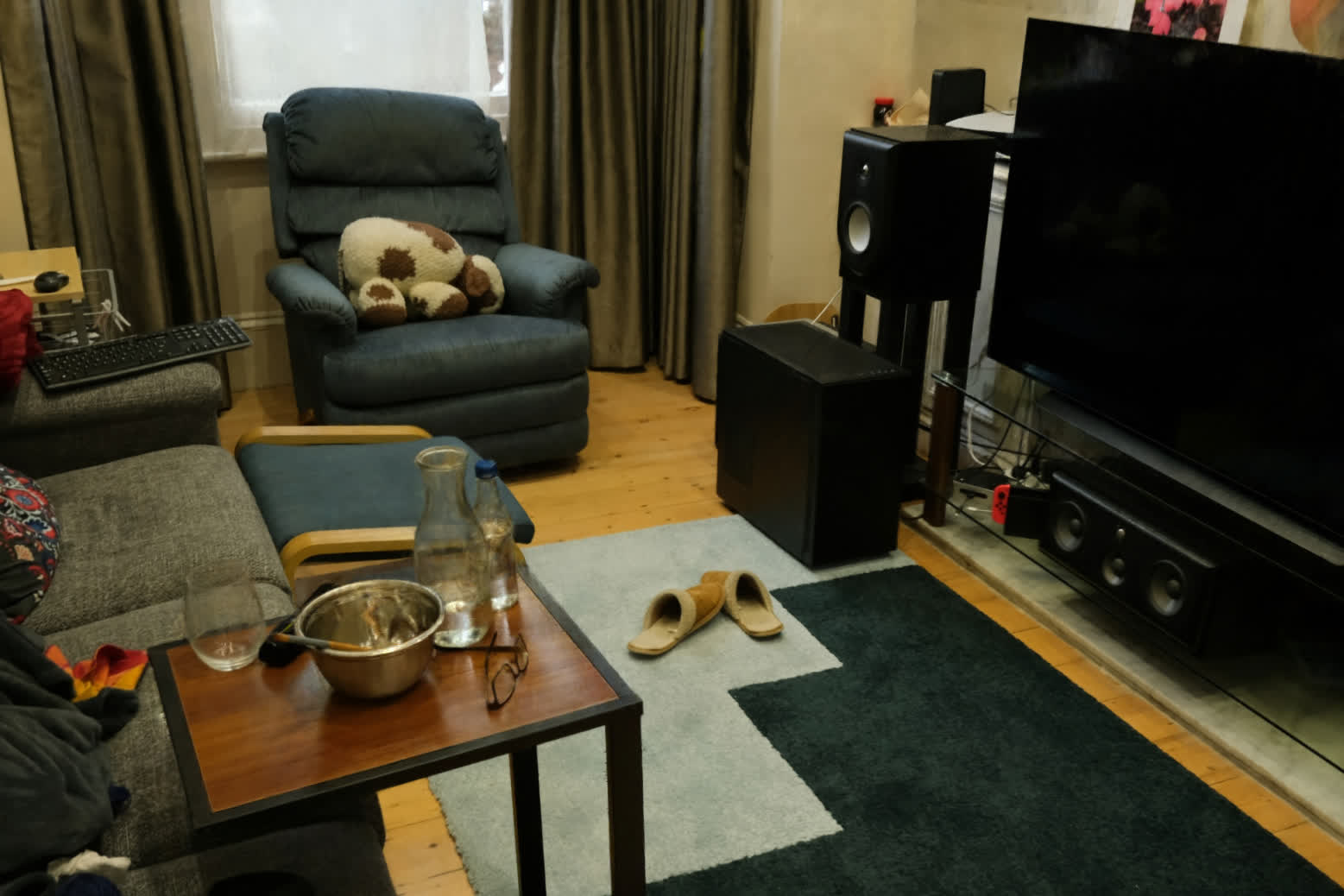}  & 
\includegraphics[width=0.15\textwidth]{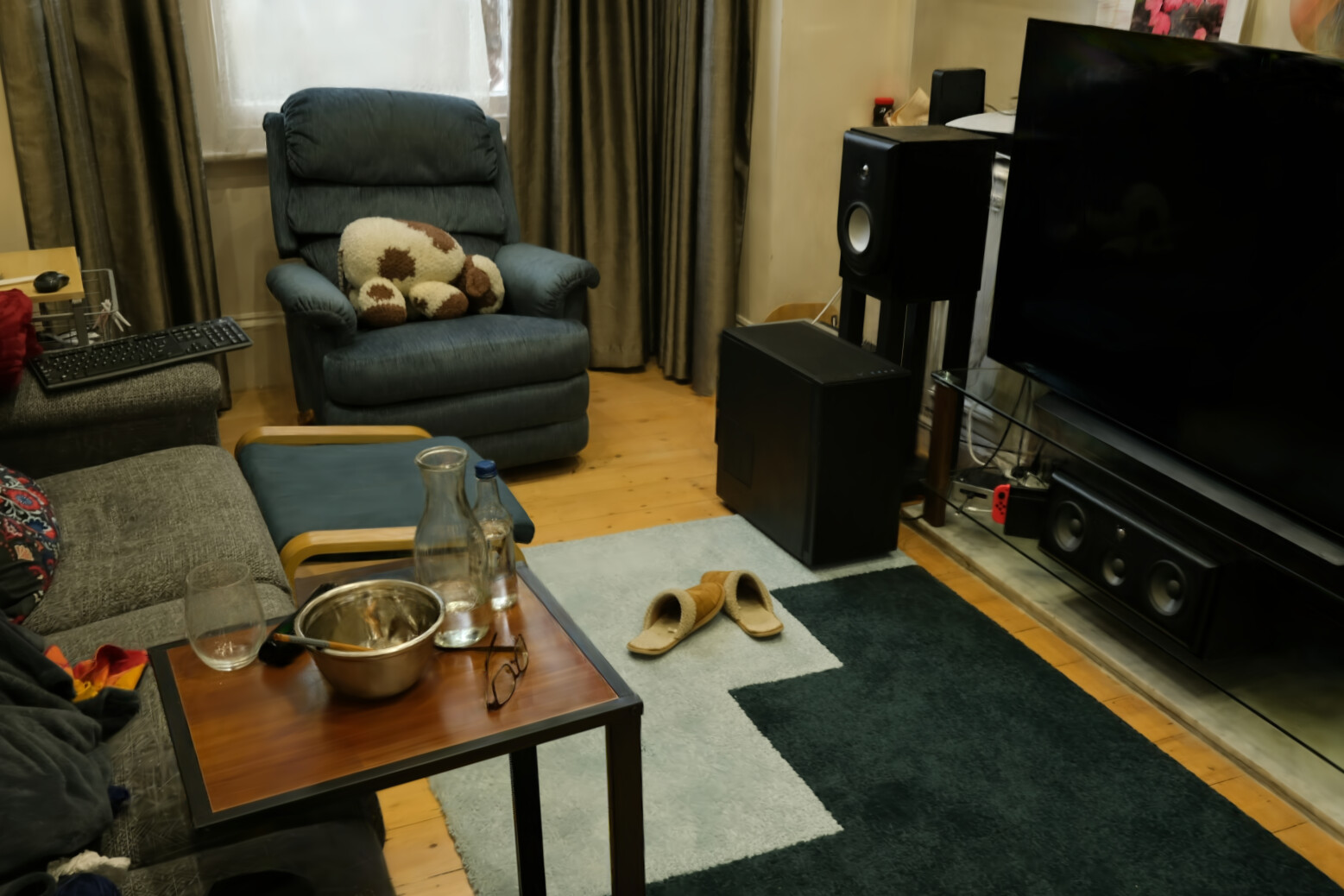}    &  \includegraphics[width=0.15\textwidth]{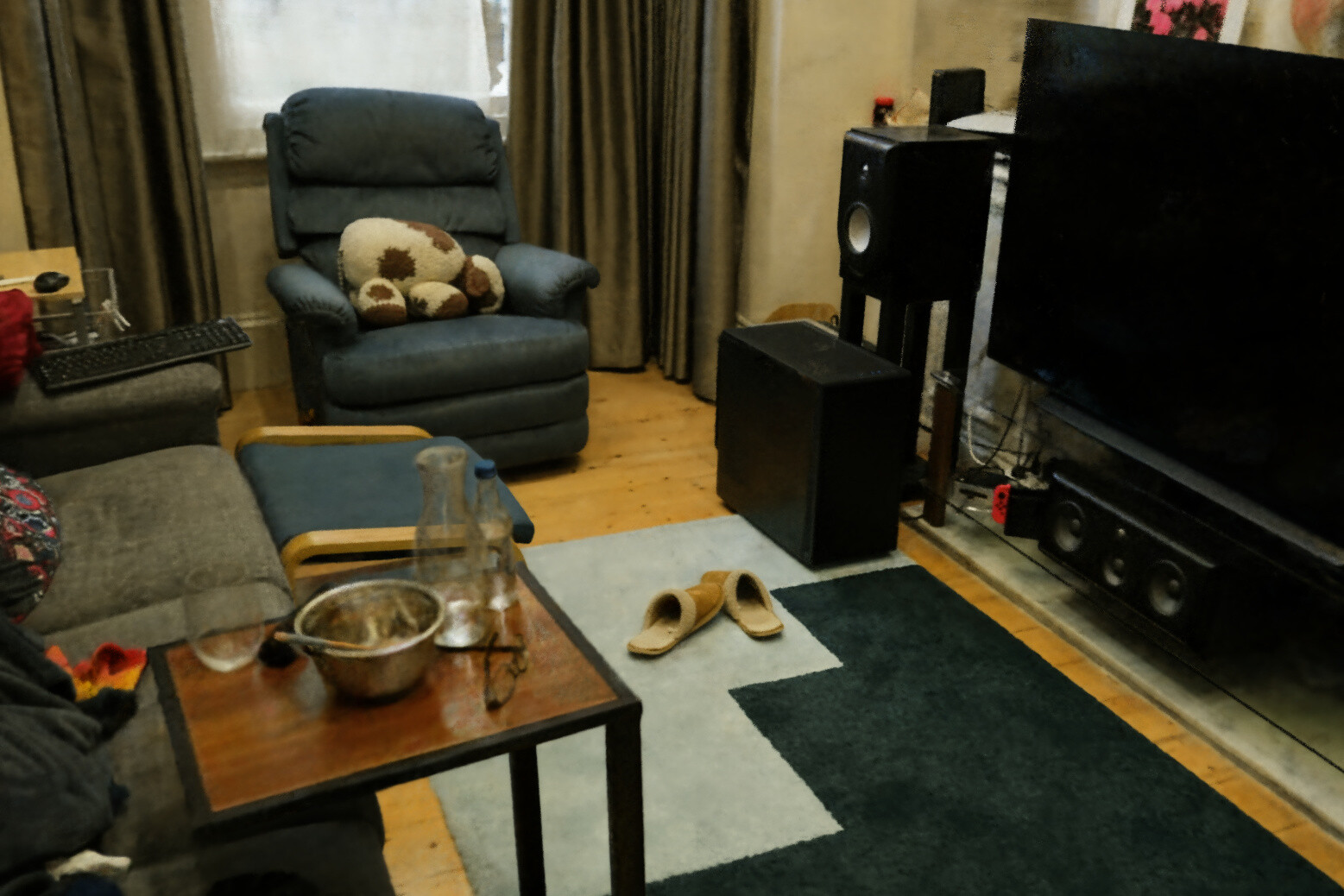}  &
\includegraphics[width=0.15\textwidth]{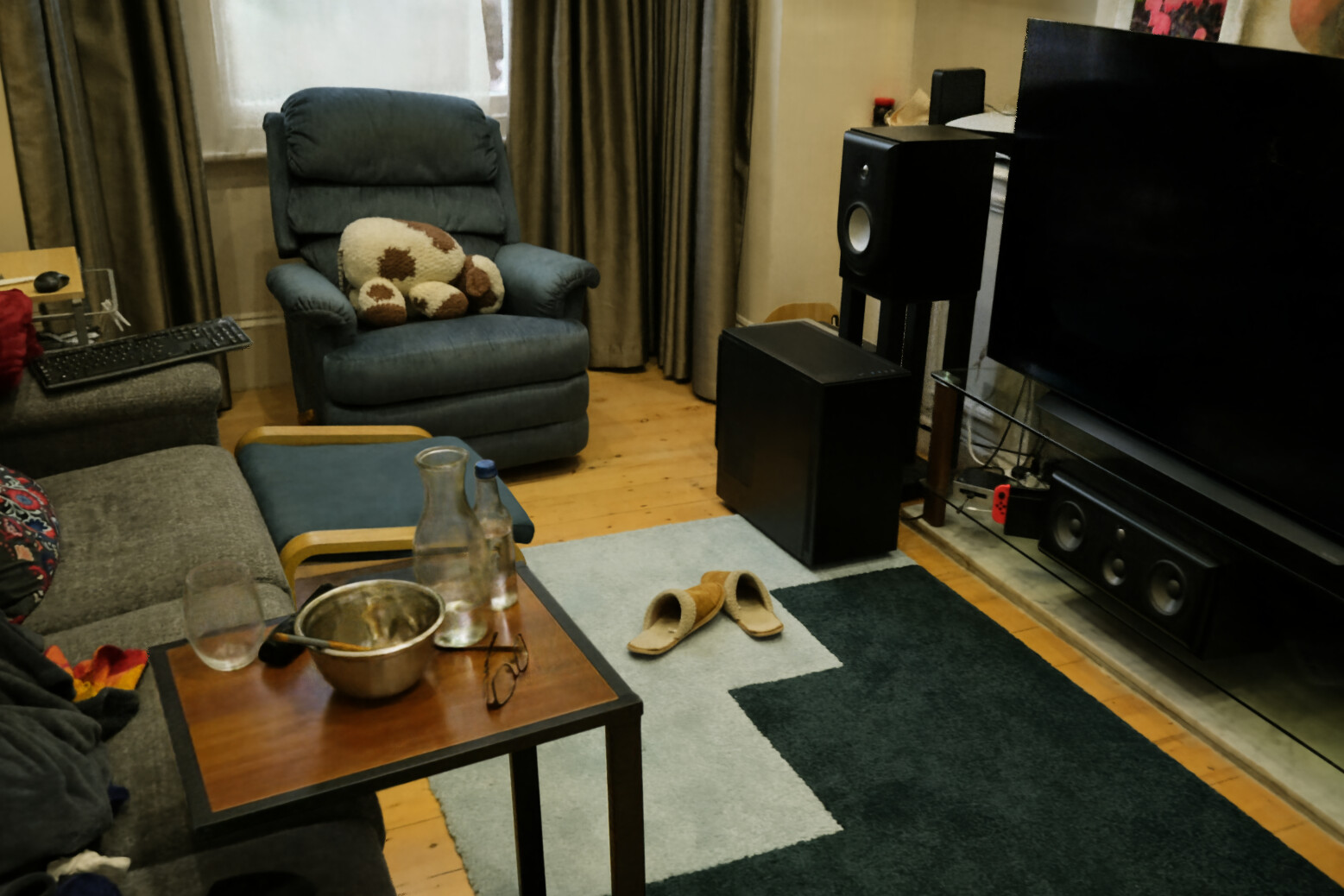}    & 
\includegraphics[width=0.15\textwidth]{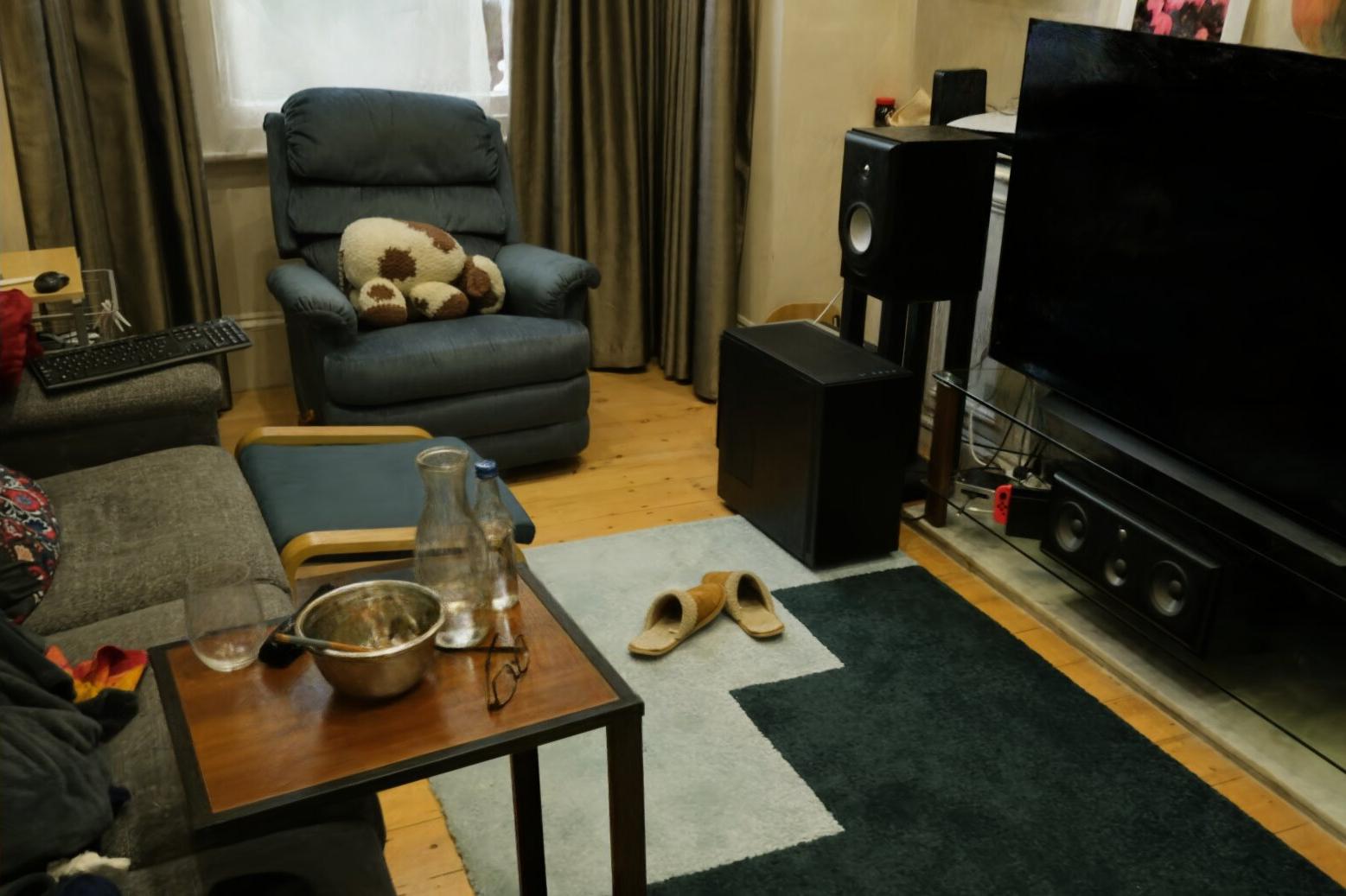}  \\
\rotatebox{90}{\footnotesize\hspace{3pt}{Truck}} &\includegraphics[width=0.15\textwidth]{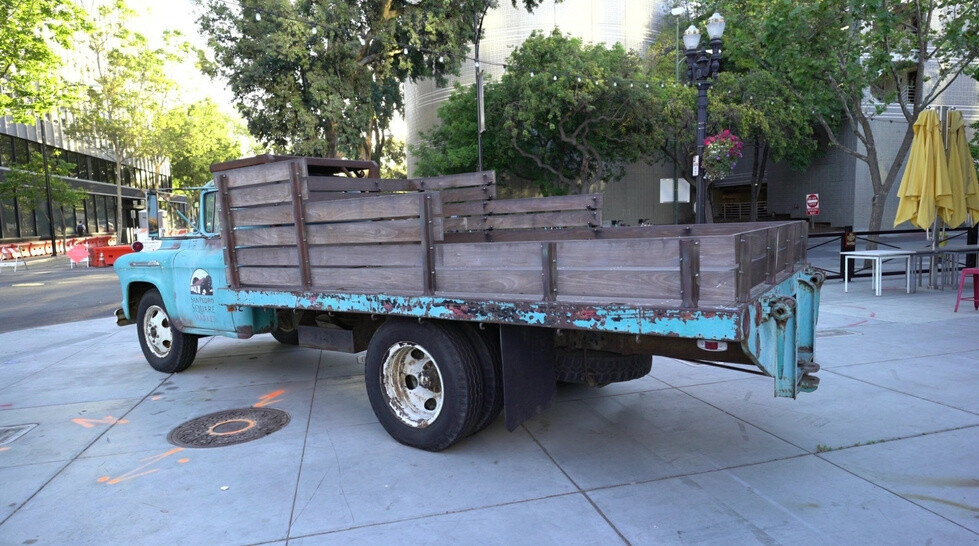}   & \includegraphics[width=0.15\textwidth]{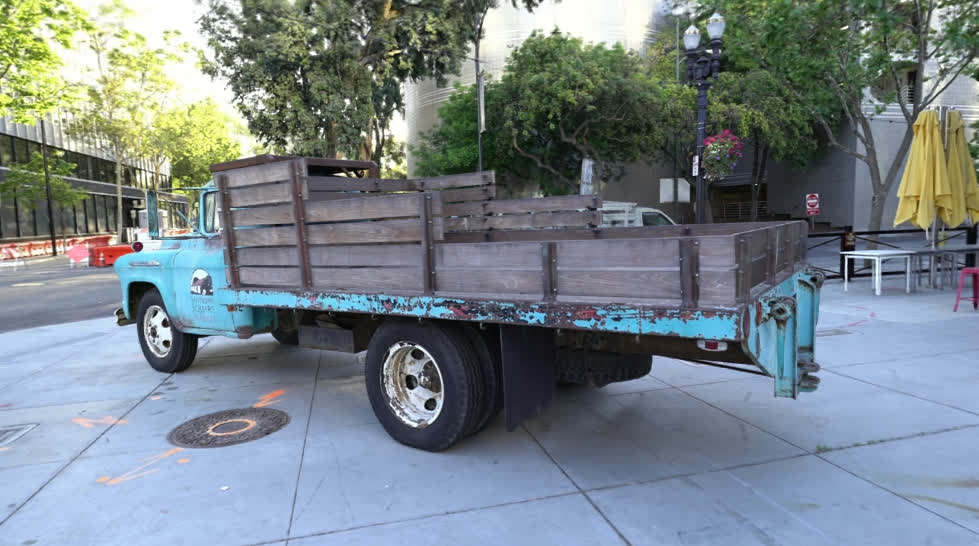}  & 
\includegraphics[width=0.15\textwidth]{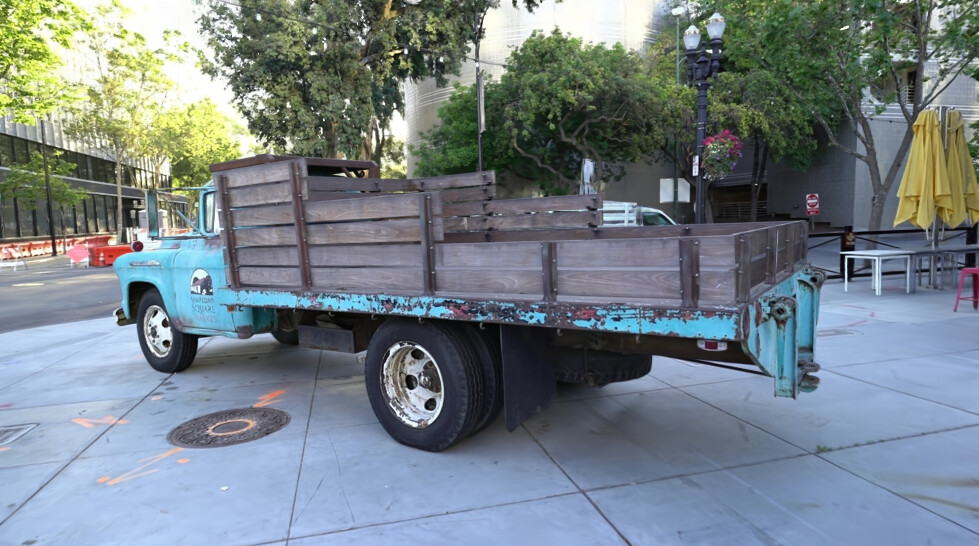}    &  \includegraphics[width=0.15\textwidth]{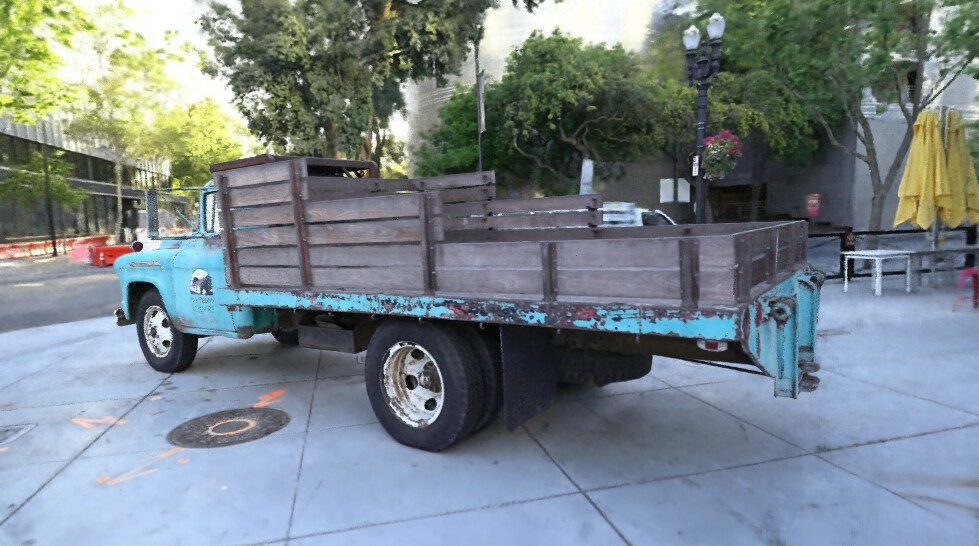}  &
\includegraphics[width=0.15\textwidth]{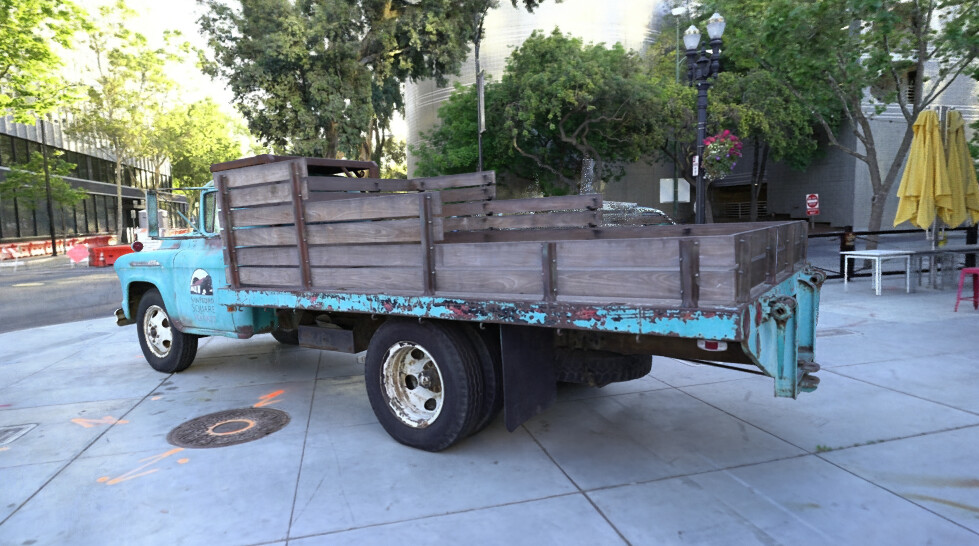}    & 
\includegraphics[width=0.15\textwidth]{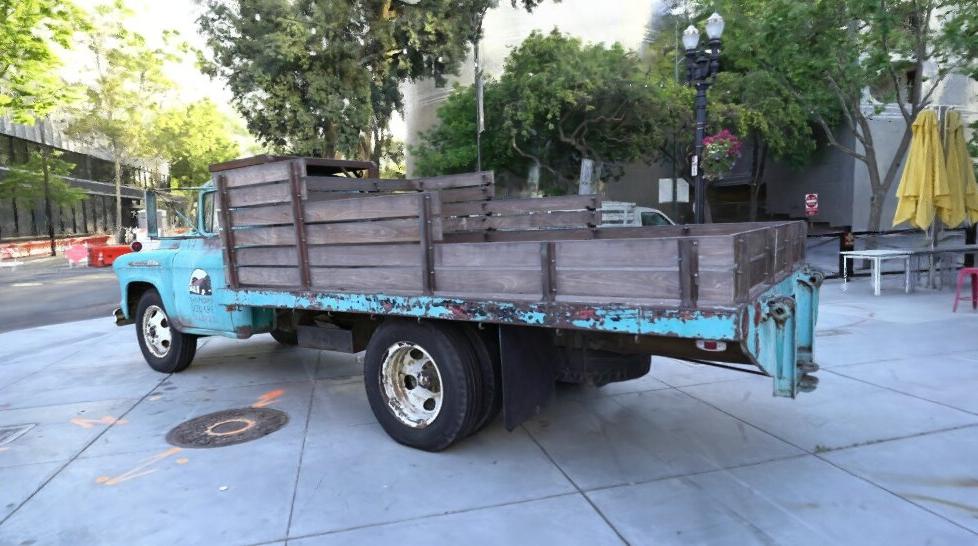}   \\
\rotatebox{90}{\footnotesize\hspace{3pt}{Train}} &\includegraphics[width=0.15\textwidth]{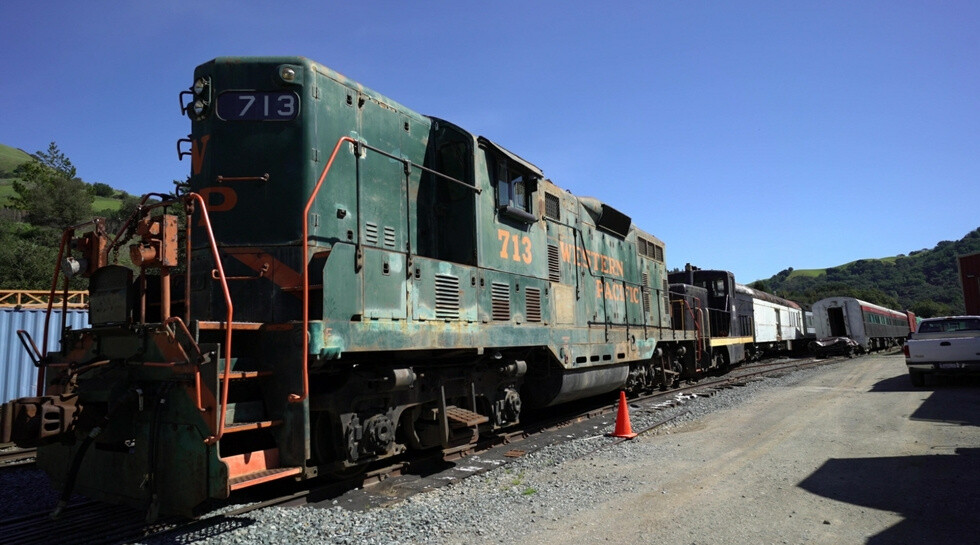}   & \includegraphics[width=0.15\textwidth]{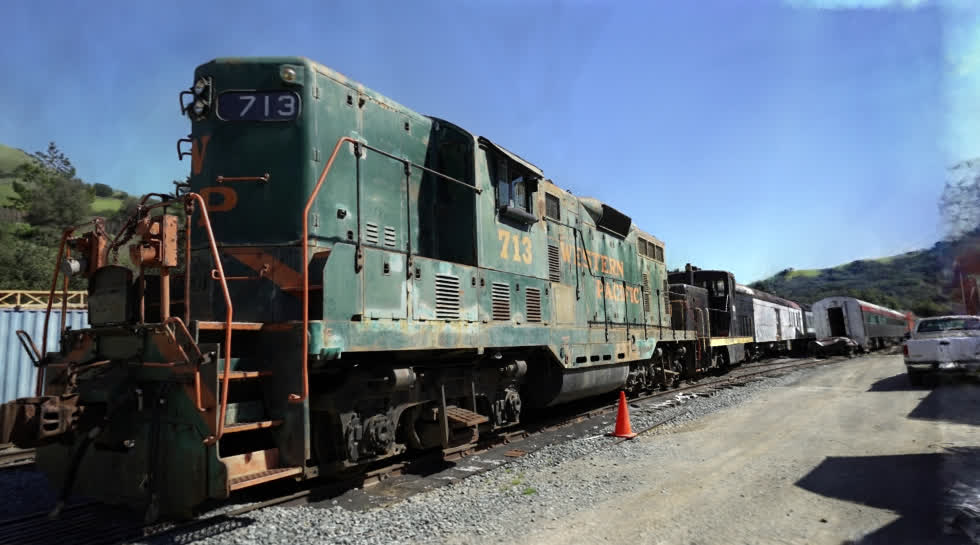}  & 
\includegraphics[width=0.15\textwidth]{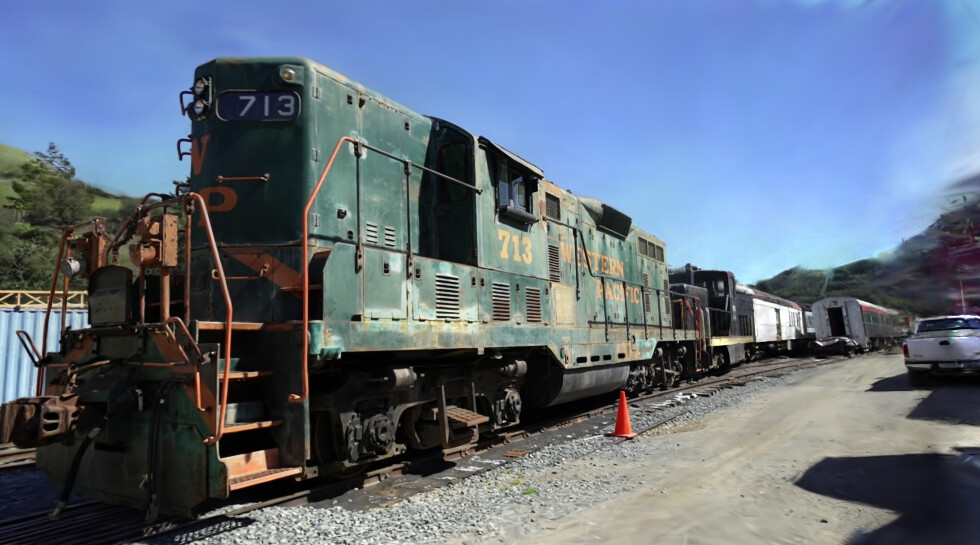}    &  \includegraphics[width=0.15\textwidth]{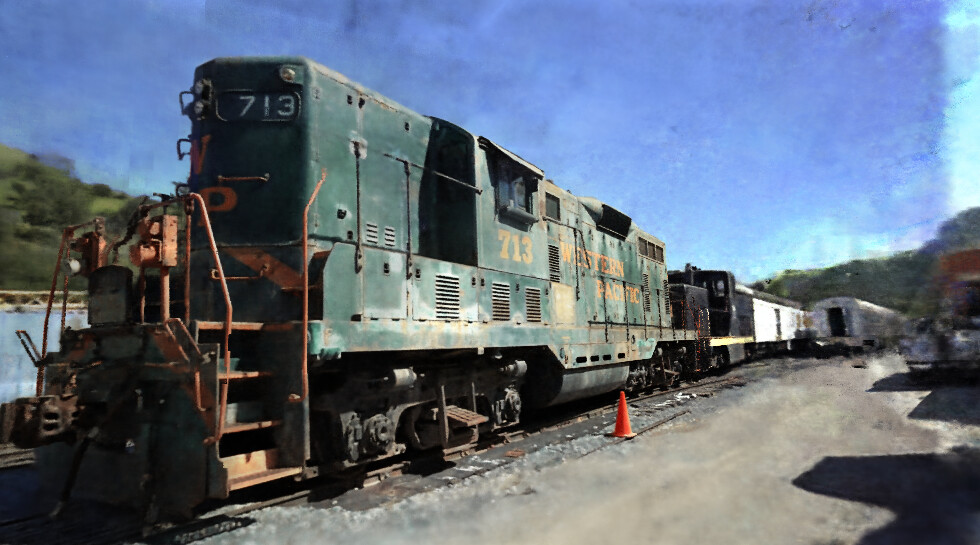}  &
\includegraphics[width=0.15\textwidth]{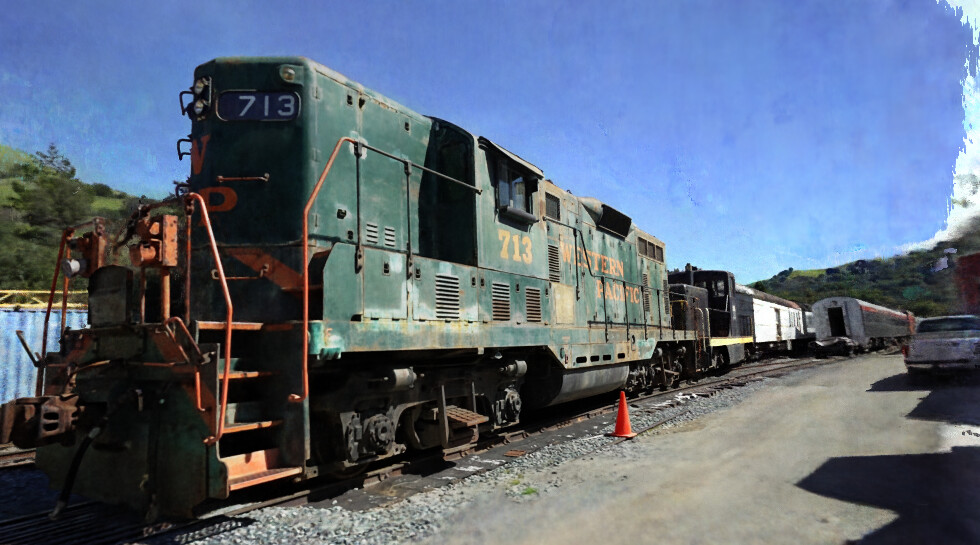}    & 
\includegraphics[width=0.15\textwidth]{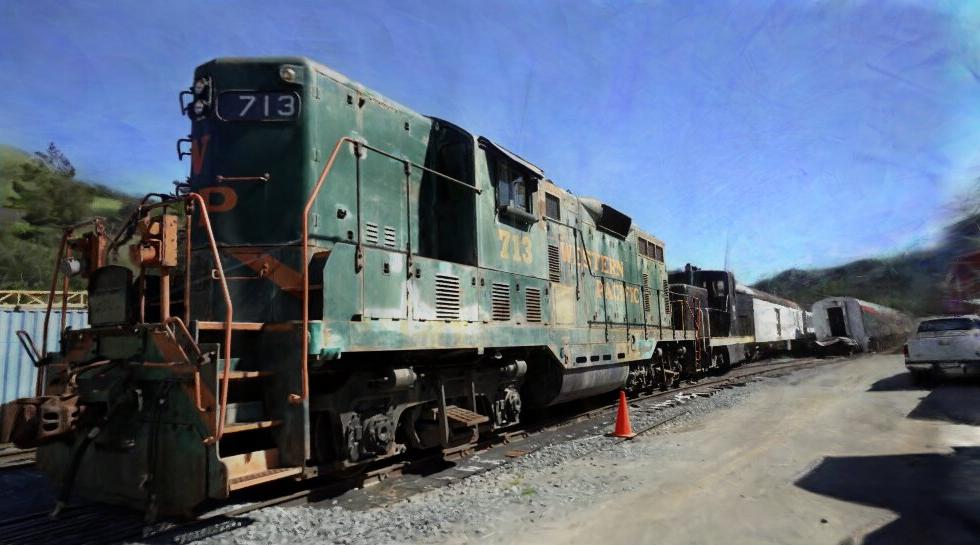}   
\end{tabular}}
\caption{Extended qualitative comparison (Full Views). Zoomed-out visualizations from Mip-NeRF 360, Deep Blending and Tanks\&Temples datasets are presented. These views demonstrate the global consistency of the reconstruction, confirming that \our{} maintains photometric accuracy across large spatial extends.}
\label{fig:comparison_full}
\end{figure*}

\begin{figure*}[t]
\centering
\setlength{\tabcolsep}{1pt}
{\fontsize{6.8pt}{11pt}\selectfont
\begin{tabular}{cccc}
 & $t_1$ & $t_2$ & $t_3$ \\
\rotatebox{90}{\footnotesize\hspace{15pt}{Bicycle}} &
\includegraphics[width=0.30\textwidth]{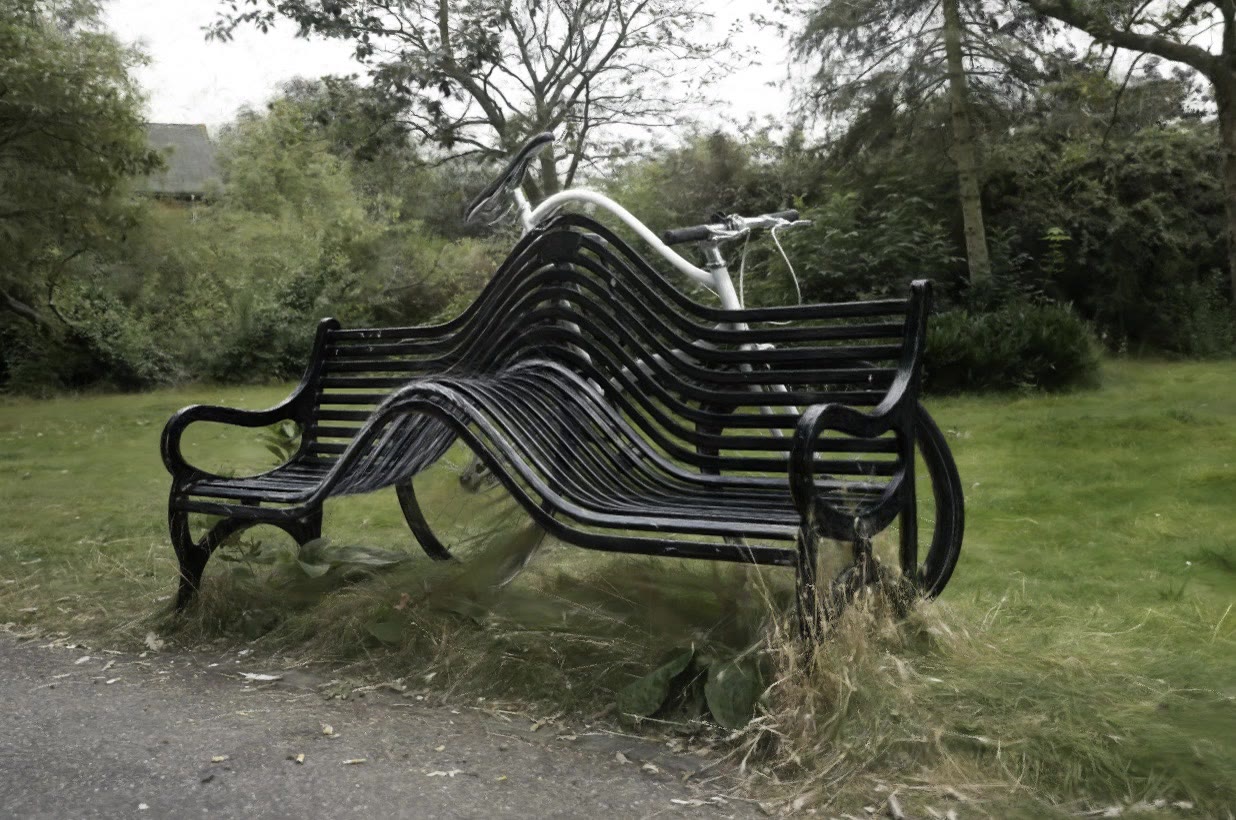}   & \includegraphics[width=0.30\textwidth]{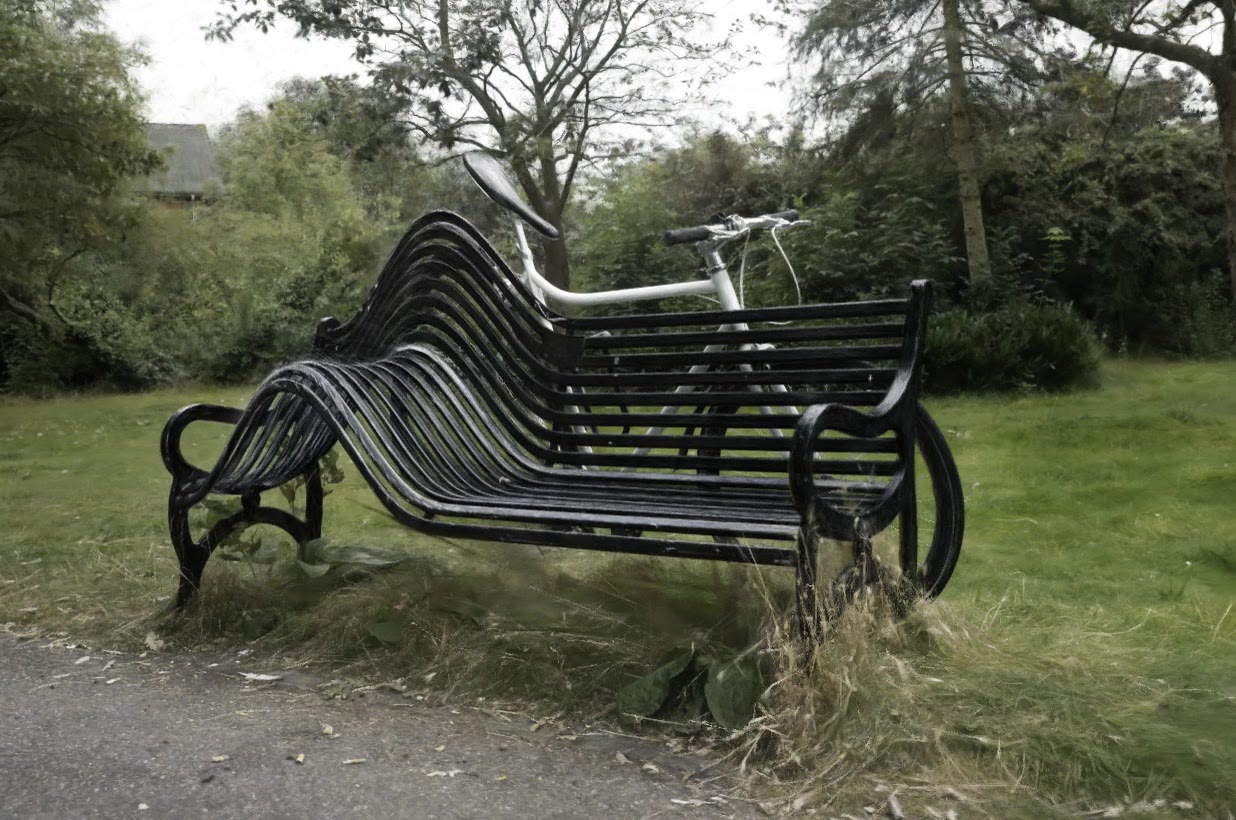}  & 
\includegraphics[width=0.30\textwidth]{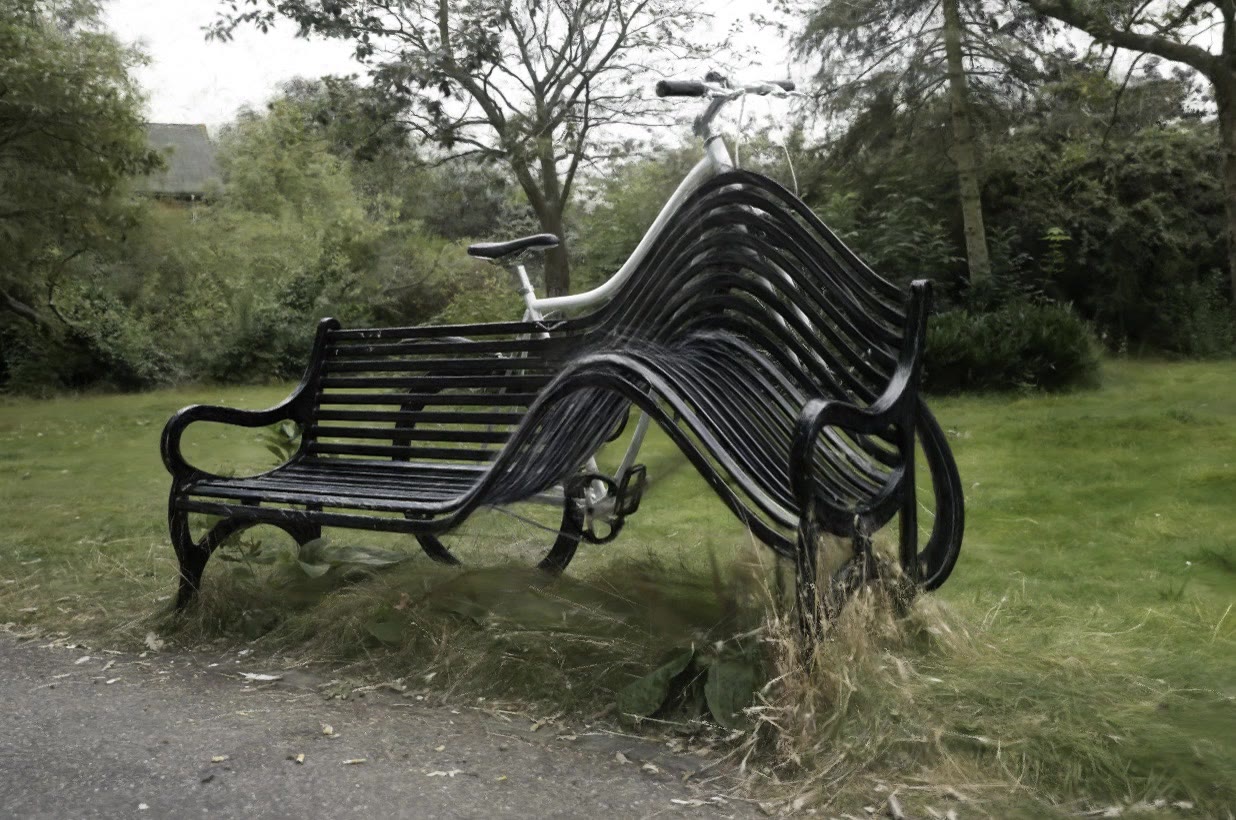}\\
\rotatebox{90}{\footnotesize\hspace{15pt}{Counter}} &\includegraphics[width=0.30\textwidth]{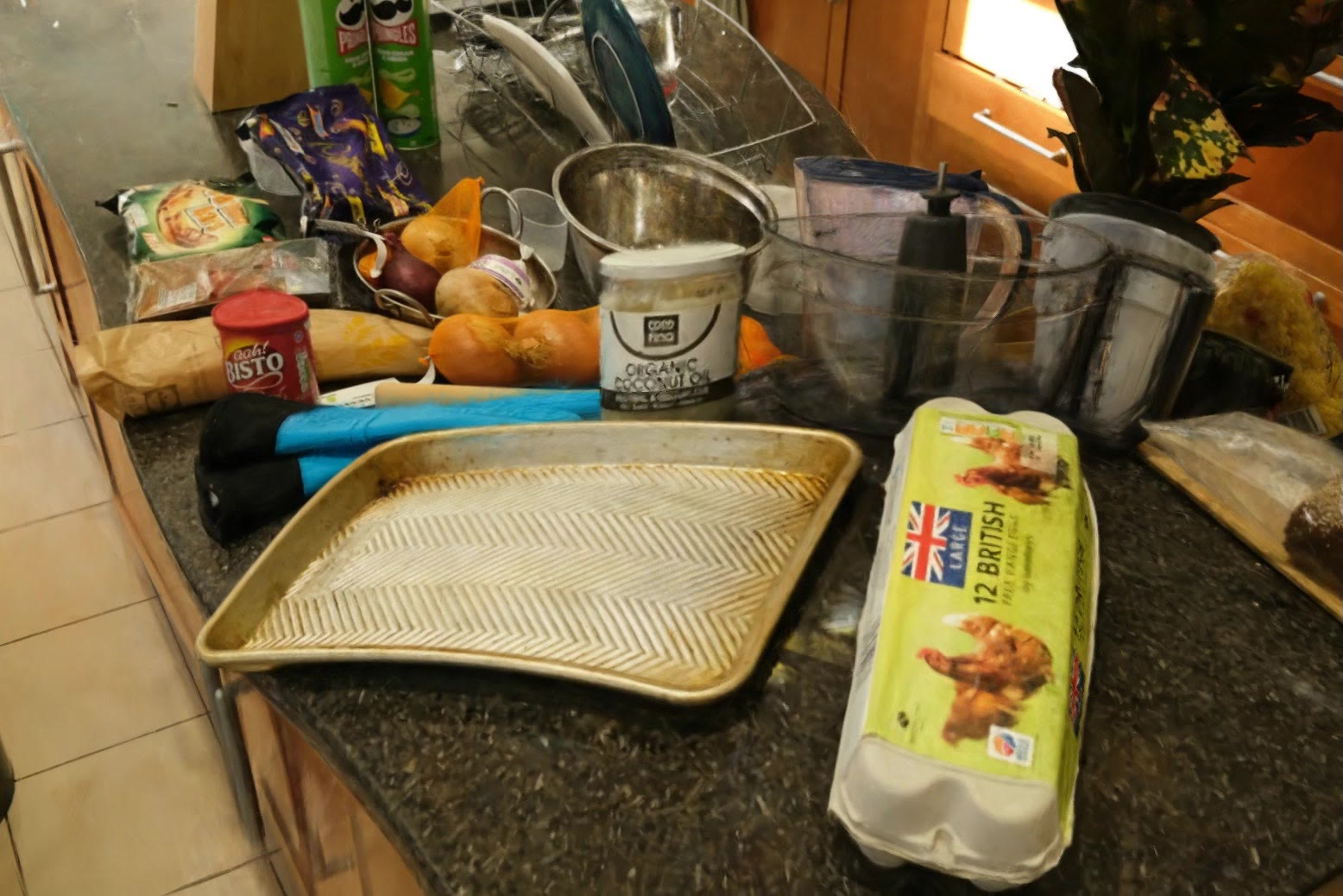}   & \includegraphics[width=0.30\textwidth]{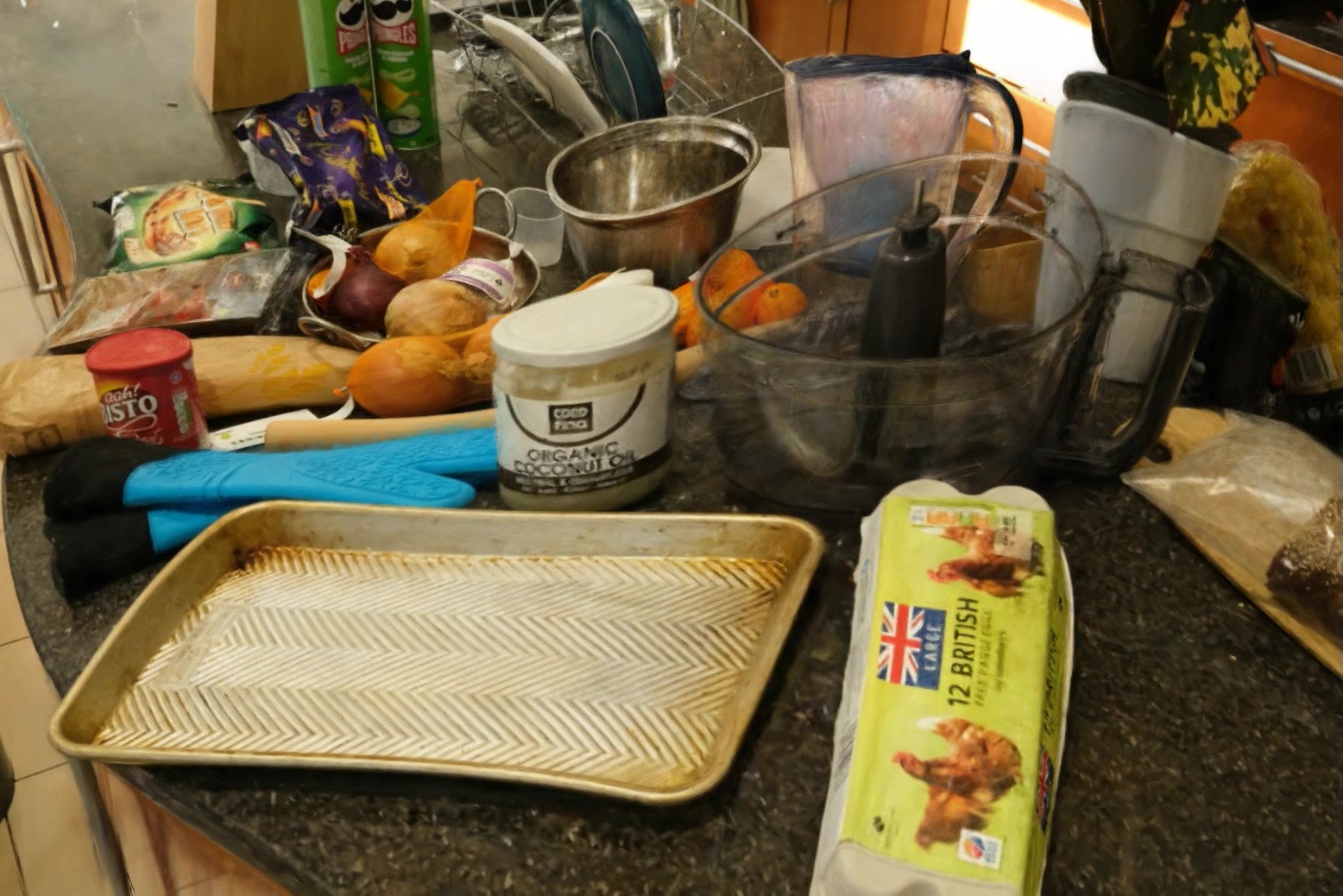}  & 
\includegraphics[width=0.30\textwidth]{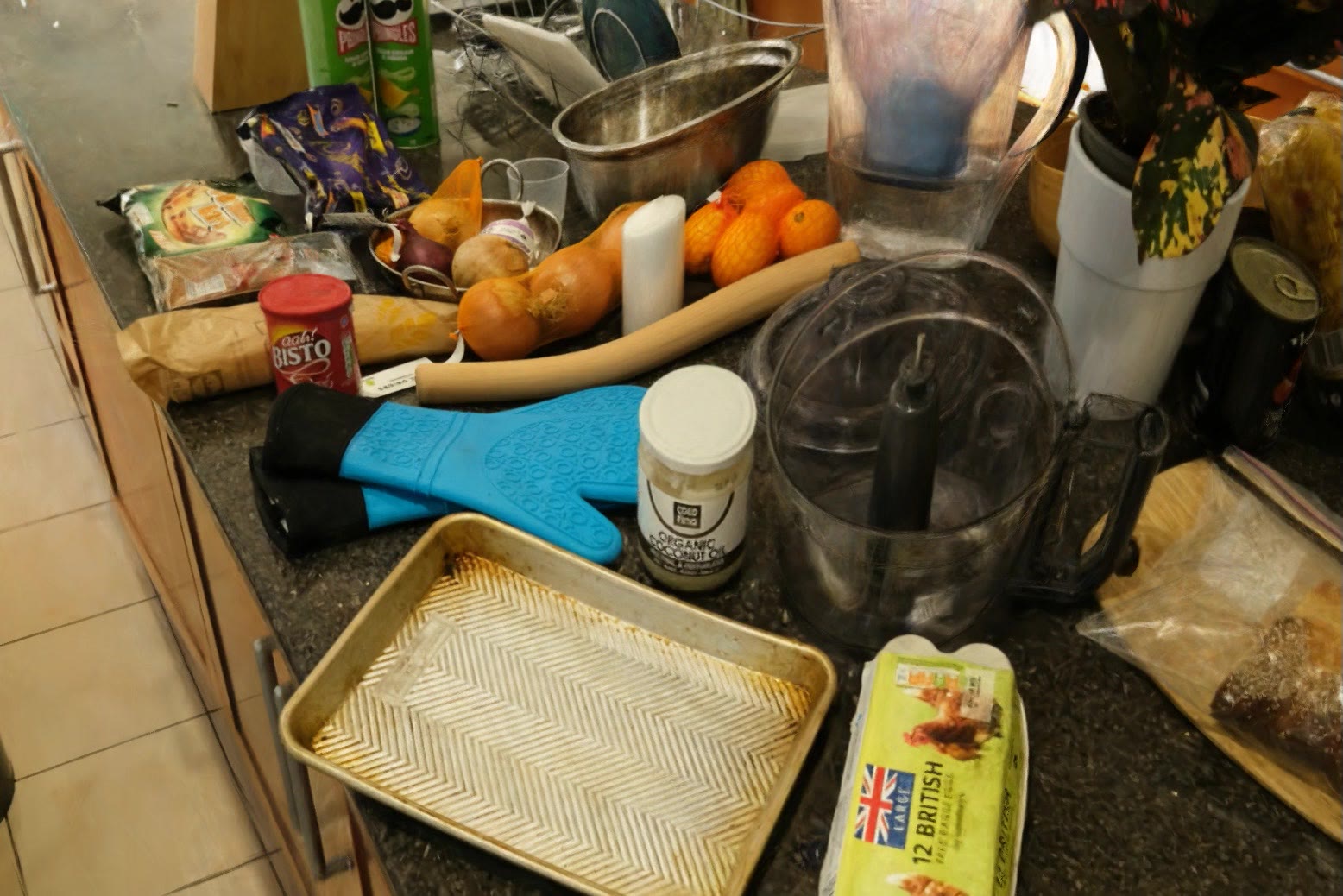} \\
\rotatebox{90}{\footnotesize\hspace{15pt}{Garden}} &\includegraphics[width=0.30\textwidth]{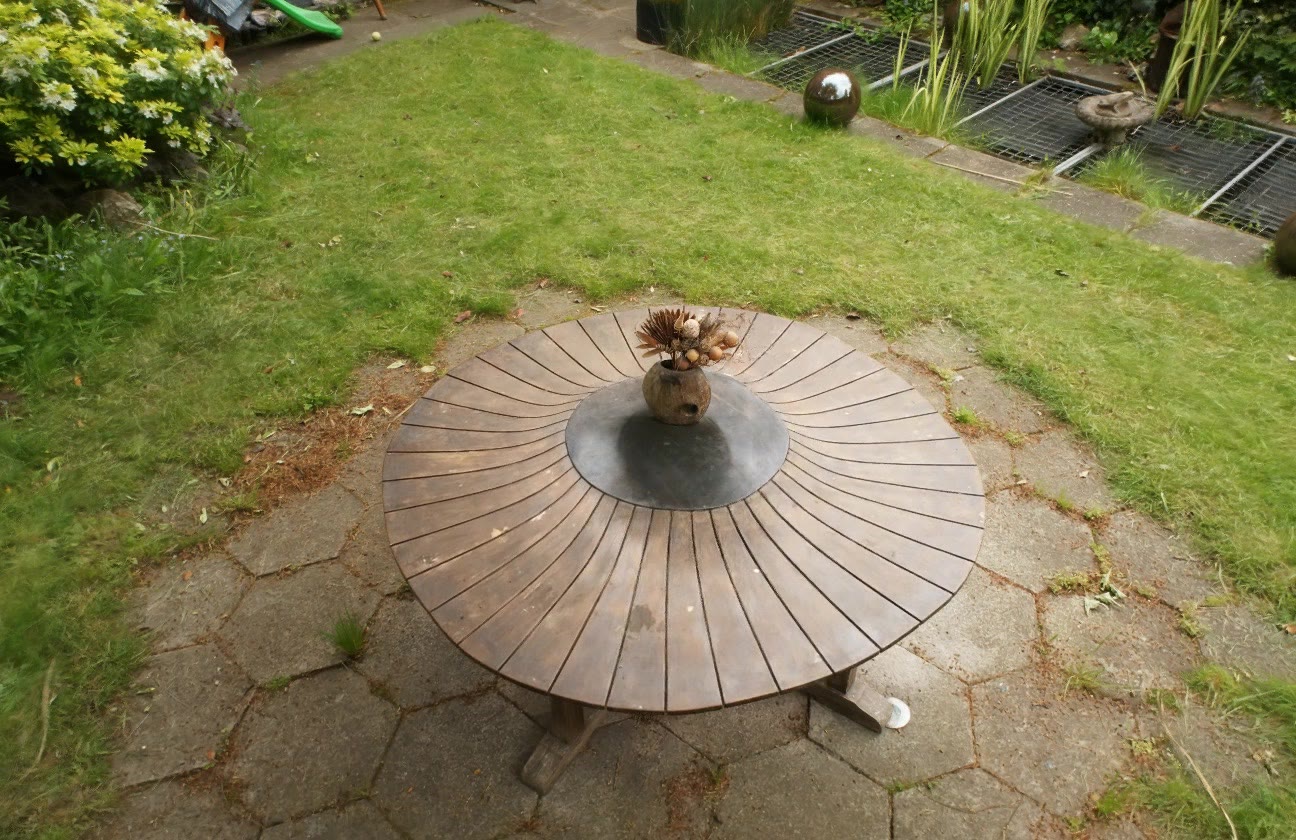}   & \includegraphics[width=0.30\textwidth]{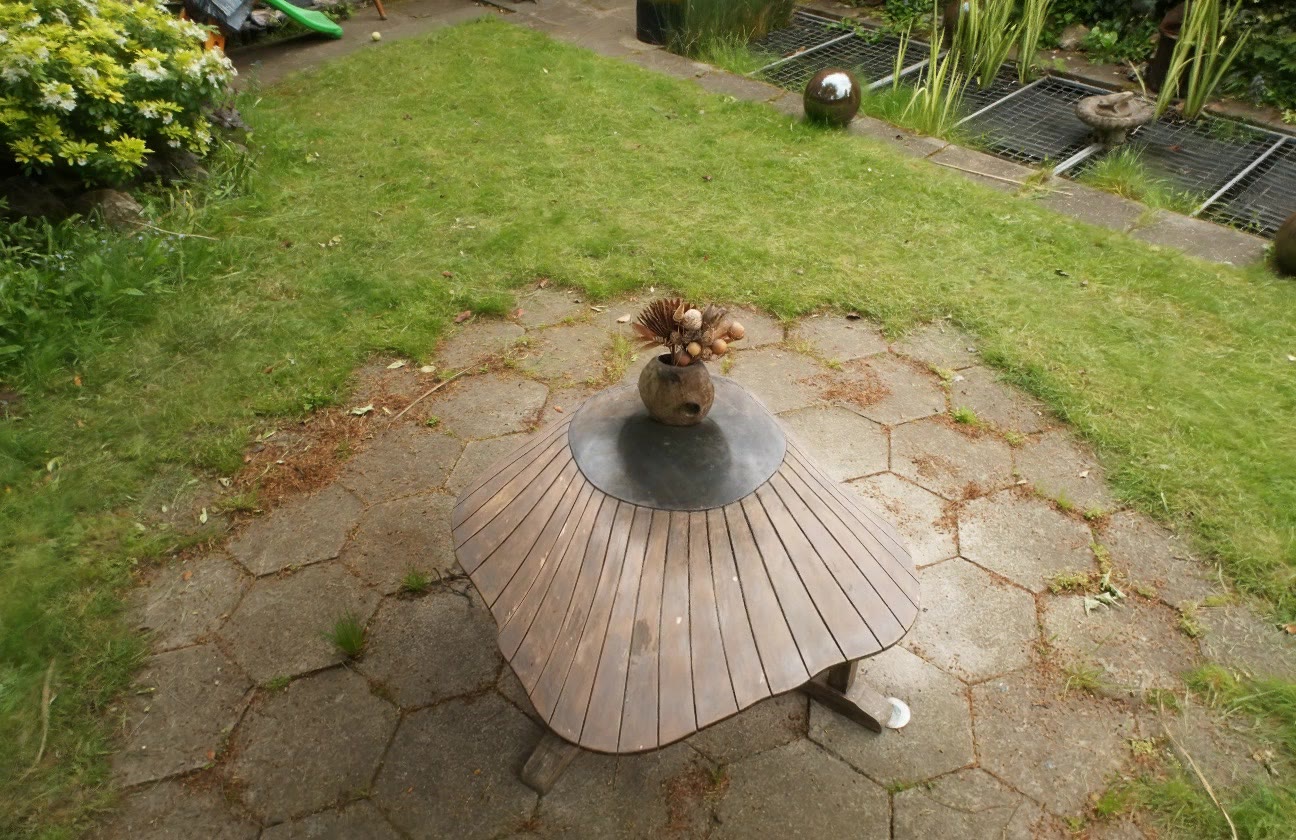}  & 
\includegraphics[width=0.30\textwidth]{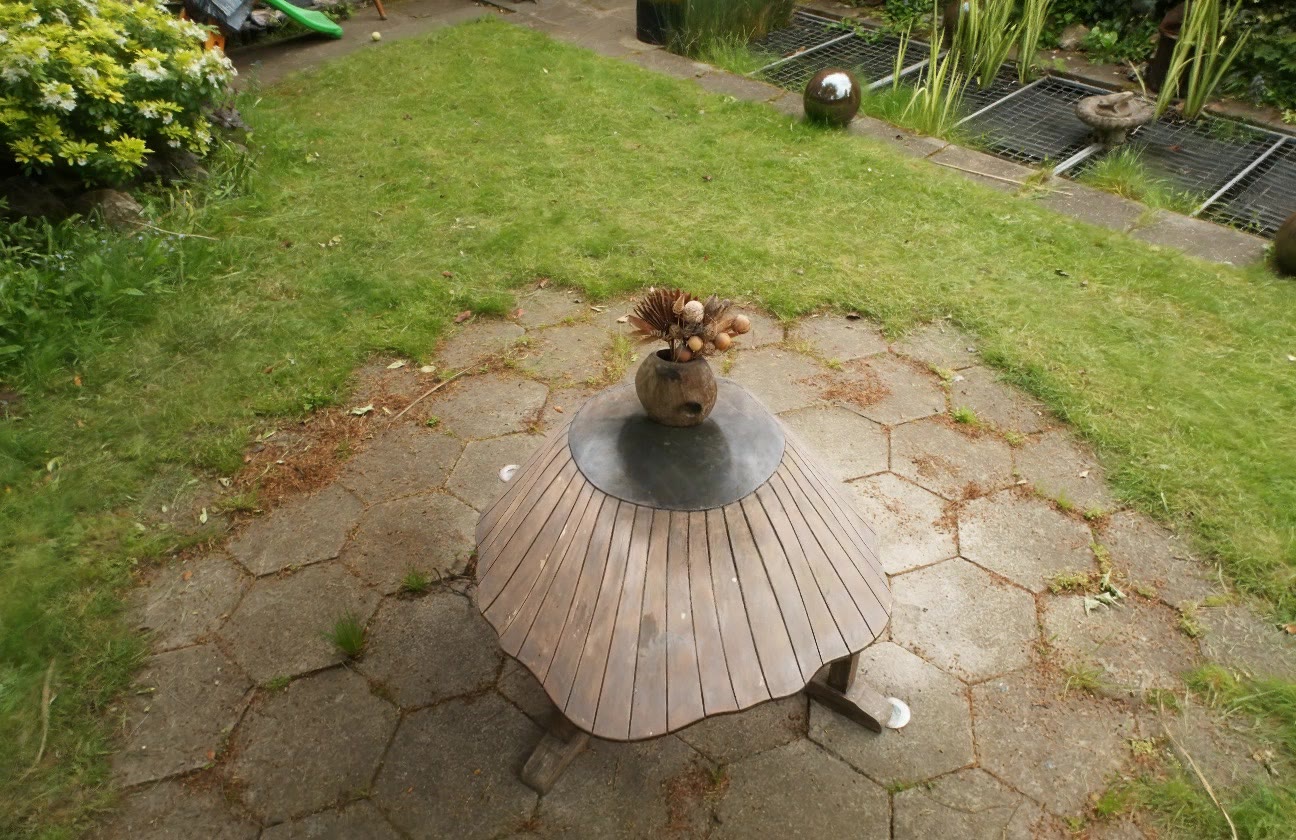} 
\end{tabular}}
\caption{Additional editing results on the Mip-NeRF 360 dataset. The capability of \our{} to perform complex, topology-agnostic manipulations in large-scale unbounded environments is demonstrated. Non-rigid deformations, such as the sinusoidal warping of the bench in the \textit{Bicycle} scene (top) and the rippling effect applied to the table in the \textit{Garden} scene (bottom), are executed while preserving high-frequency textural details. In all cases, geometric consistency and high fidelity are successfully maintained throughout the animation sequence.}
\label{fig:additional_anims}
\end{figure*}

\begin{figure*}[t]
\centering
\setlength{\tabcolsep}{1pt}

{\fontsize{6.8pt}{11pt}\selectfont
\begin{tabular}{c c c c c c}
 & $t_1$ & $t_2$ & $t_3$ & $t_4$   & $t_5$ \\
 
 \rotatebox[origin=c]{90}{\footnotesize Ficus} &
 \includegraphics[width=0.175\textwidth, valign=m]{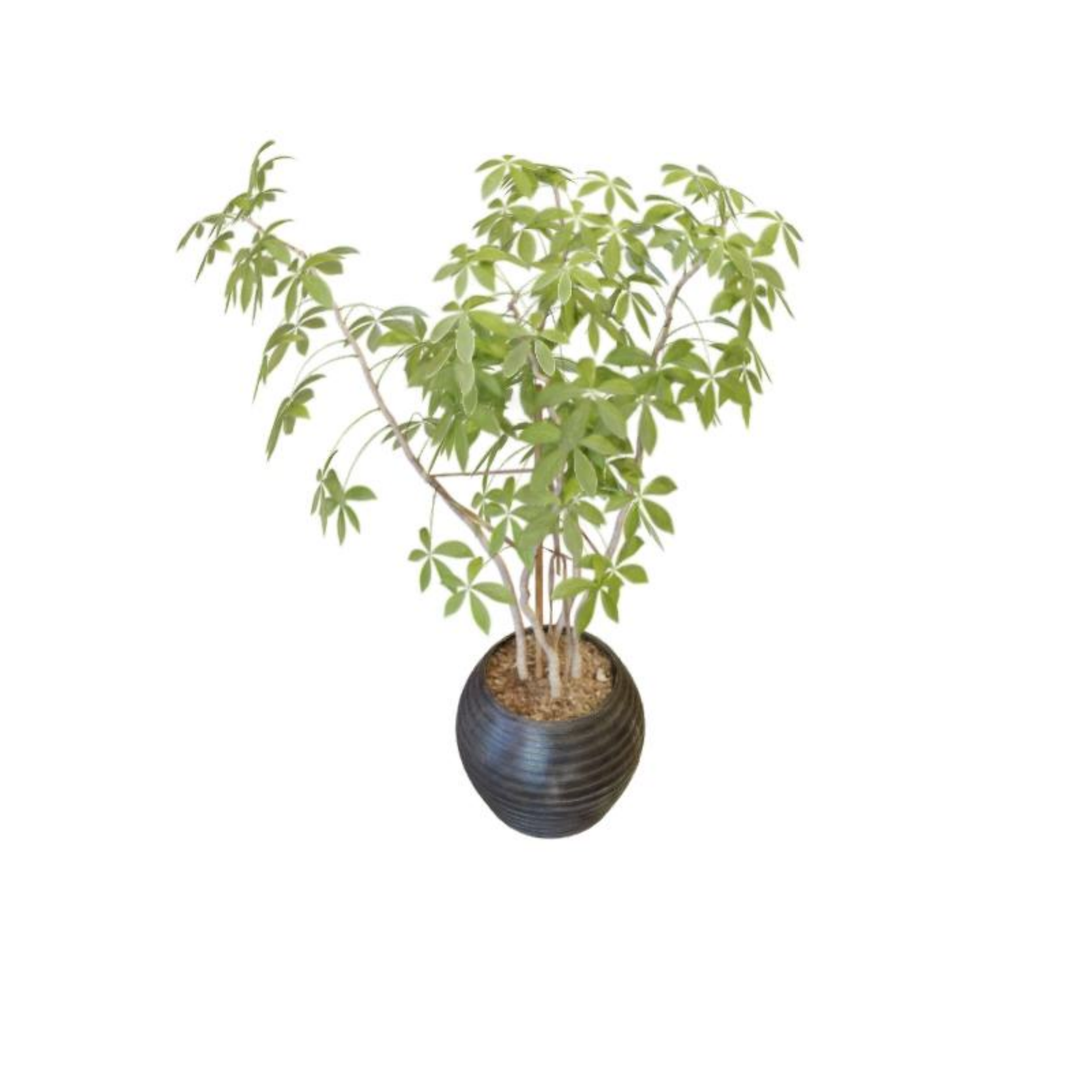} & 
 \includegraphics[width=0.175\textwidth, valign=m]{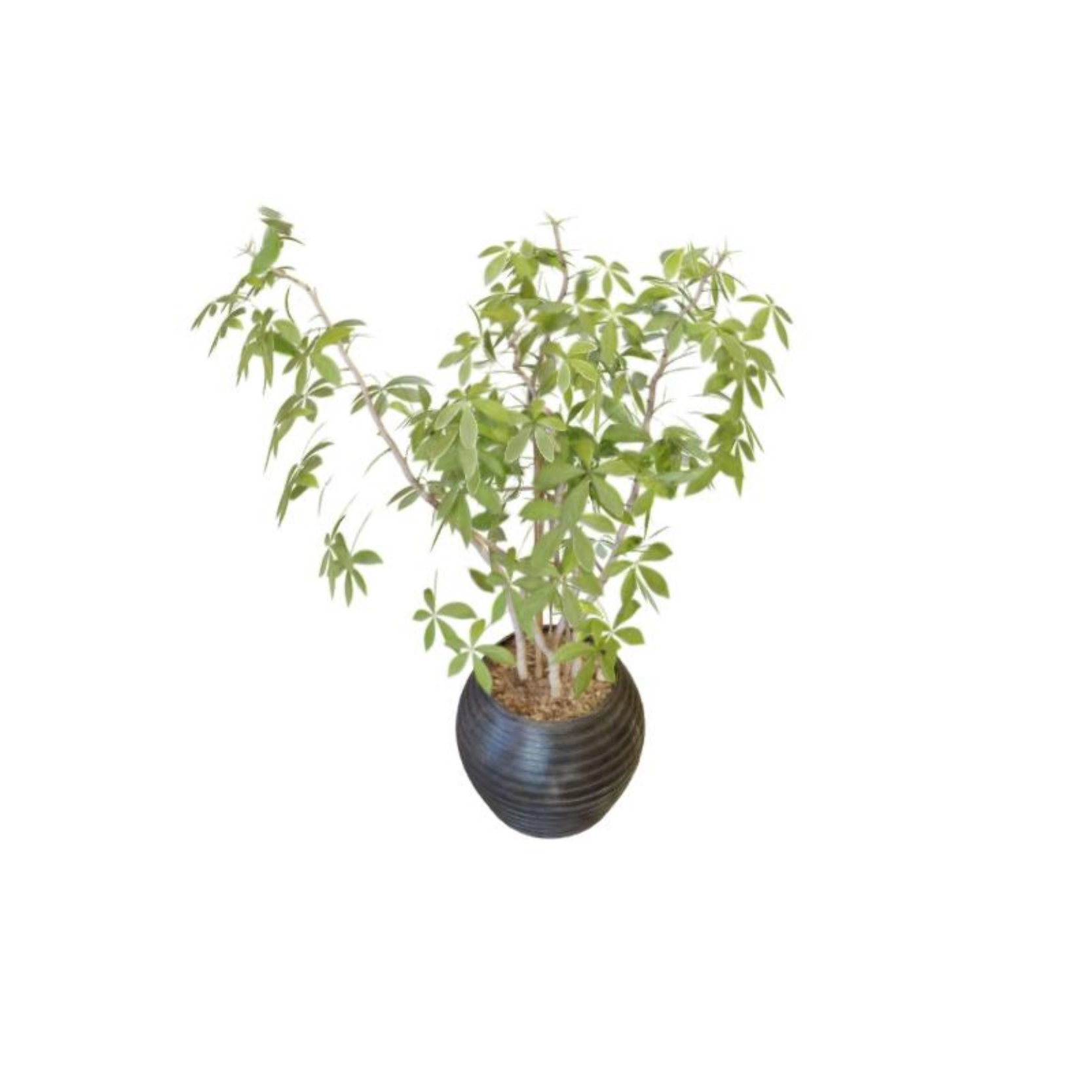} & 
 \includegraphics[width=0.175\textwidth, valign=m]{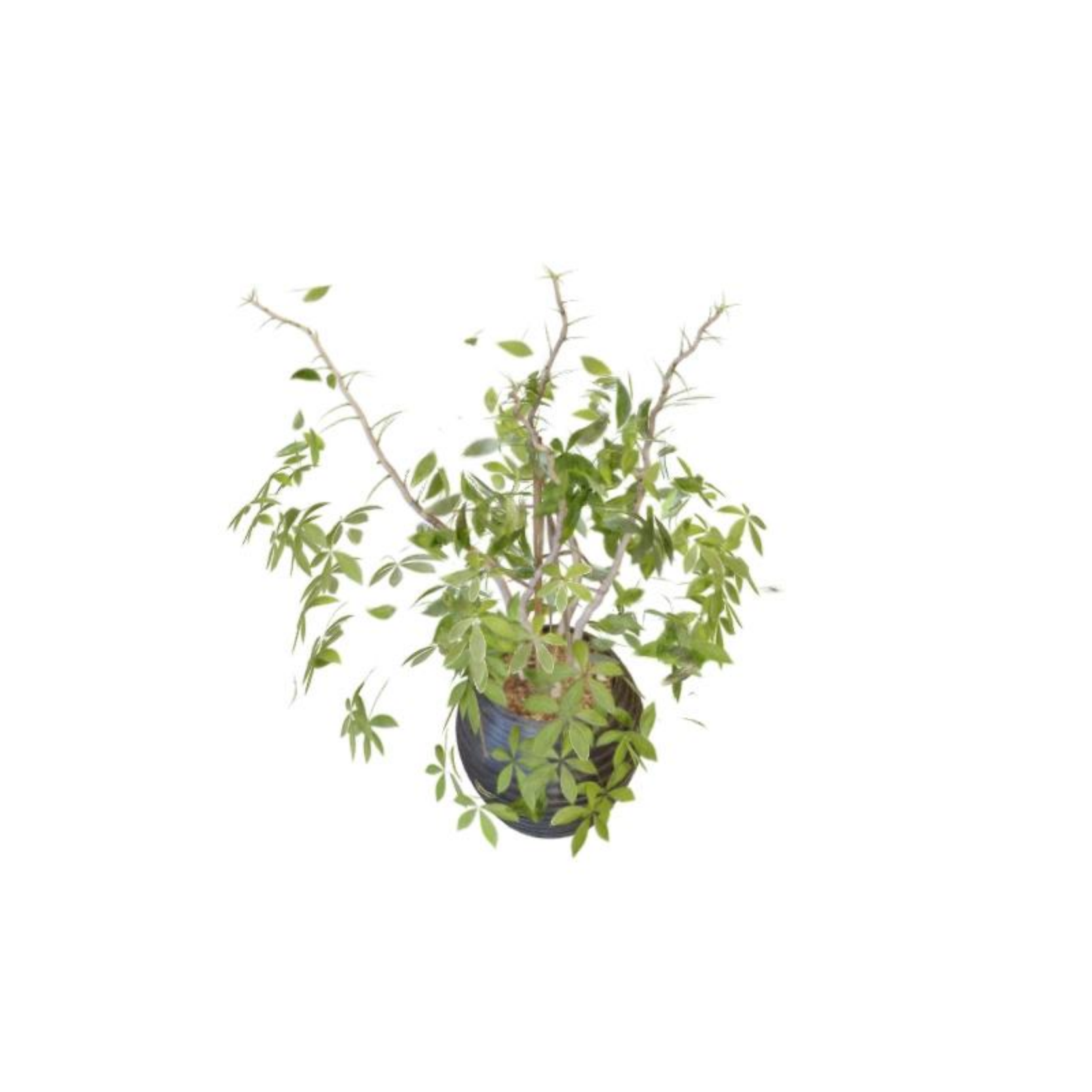} &  
 \includegraphics[width=0.175\textwidth, valign=m]{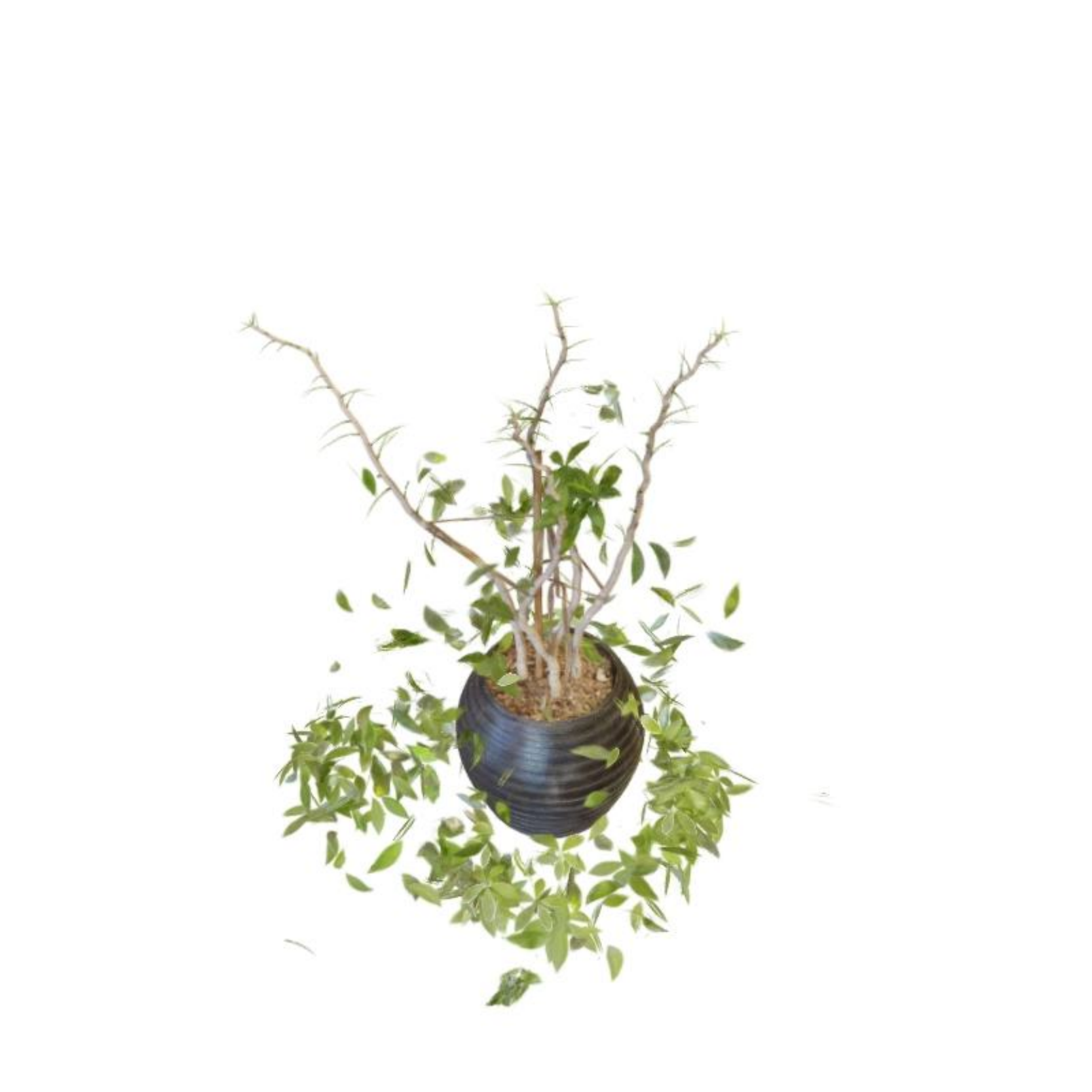} &
 \includegraphics[width=0.175\textwidth, valign=m]{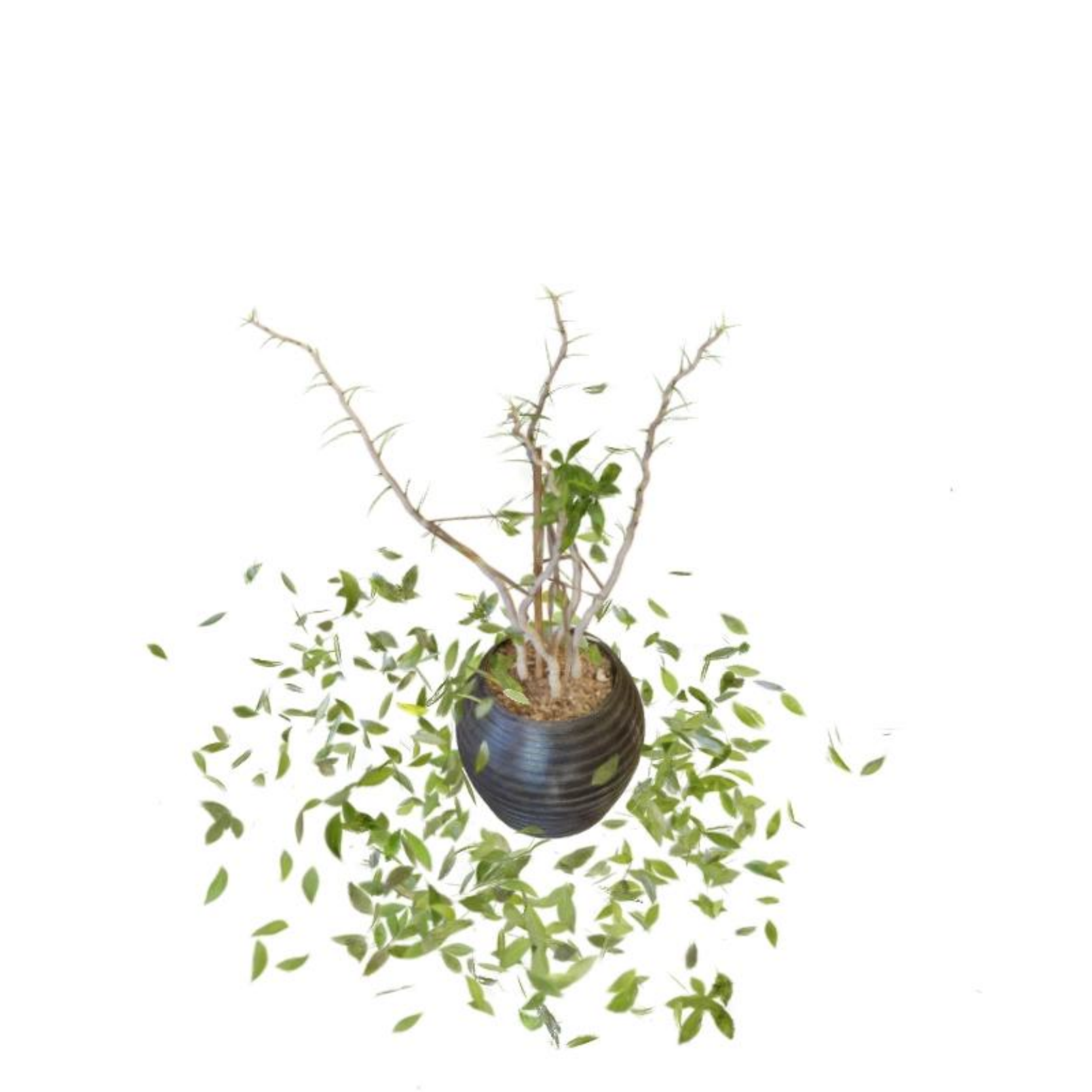} \\
 
 \rotatebox[origin=c]{90}{\footnotesize Lego} &
 \includegraphics[width=0.175\textwidth, valign=m]{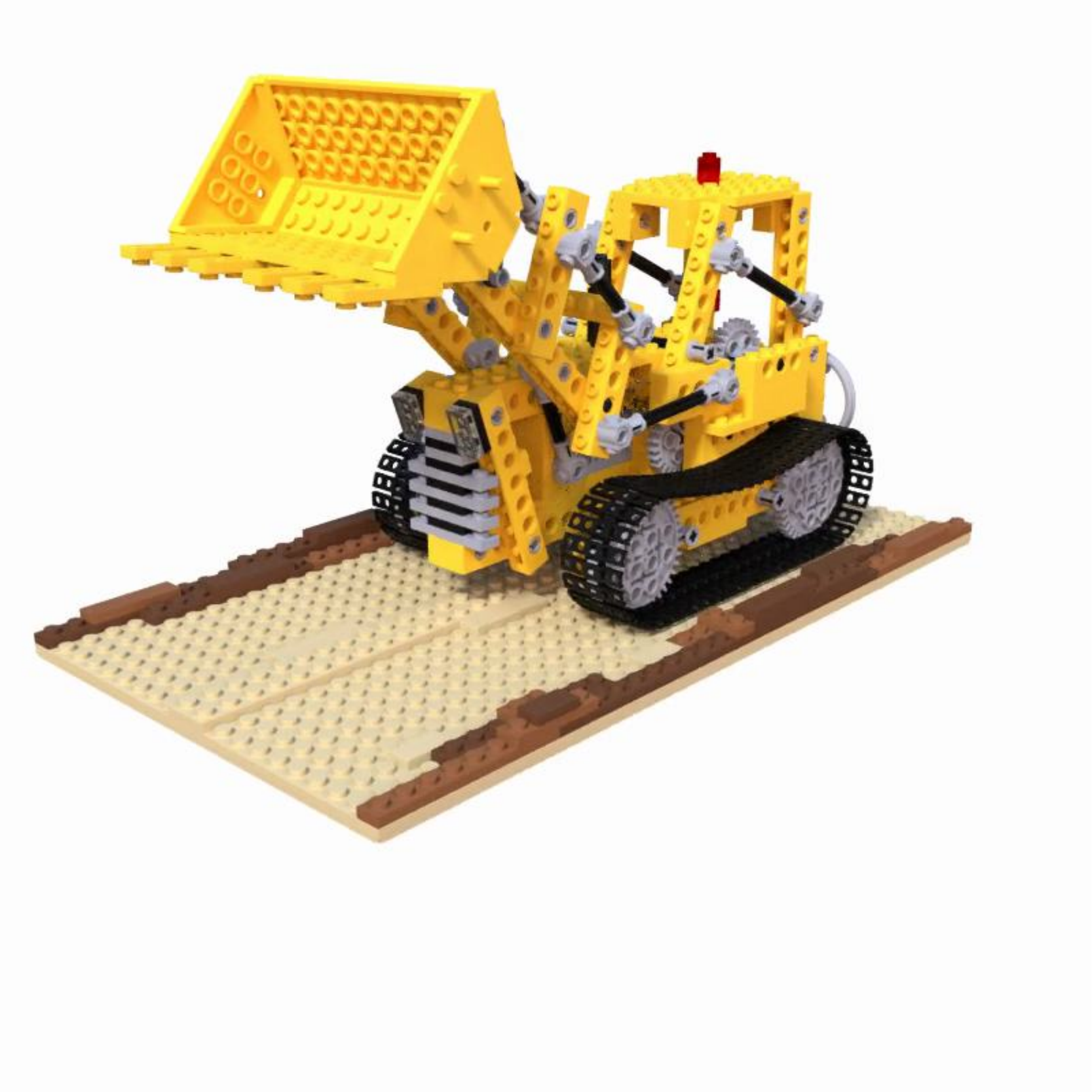} & 
 \includegraphics[width=0.175\textwidth, valign=m]{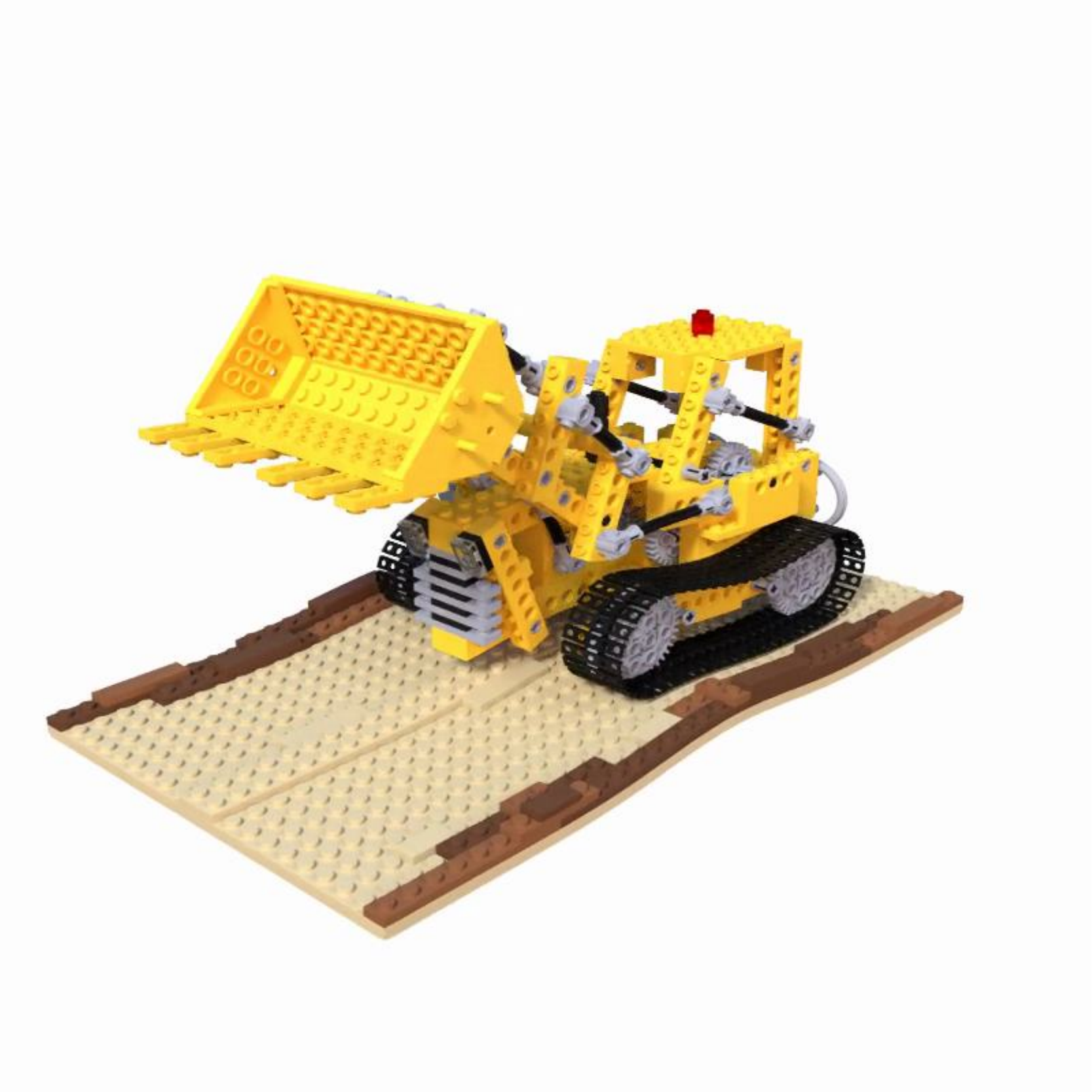} & 
 \includegraphics[width=0.175\textwidth, valign=m]{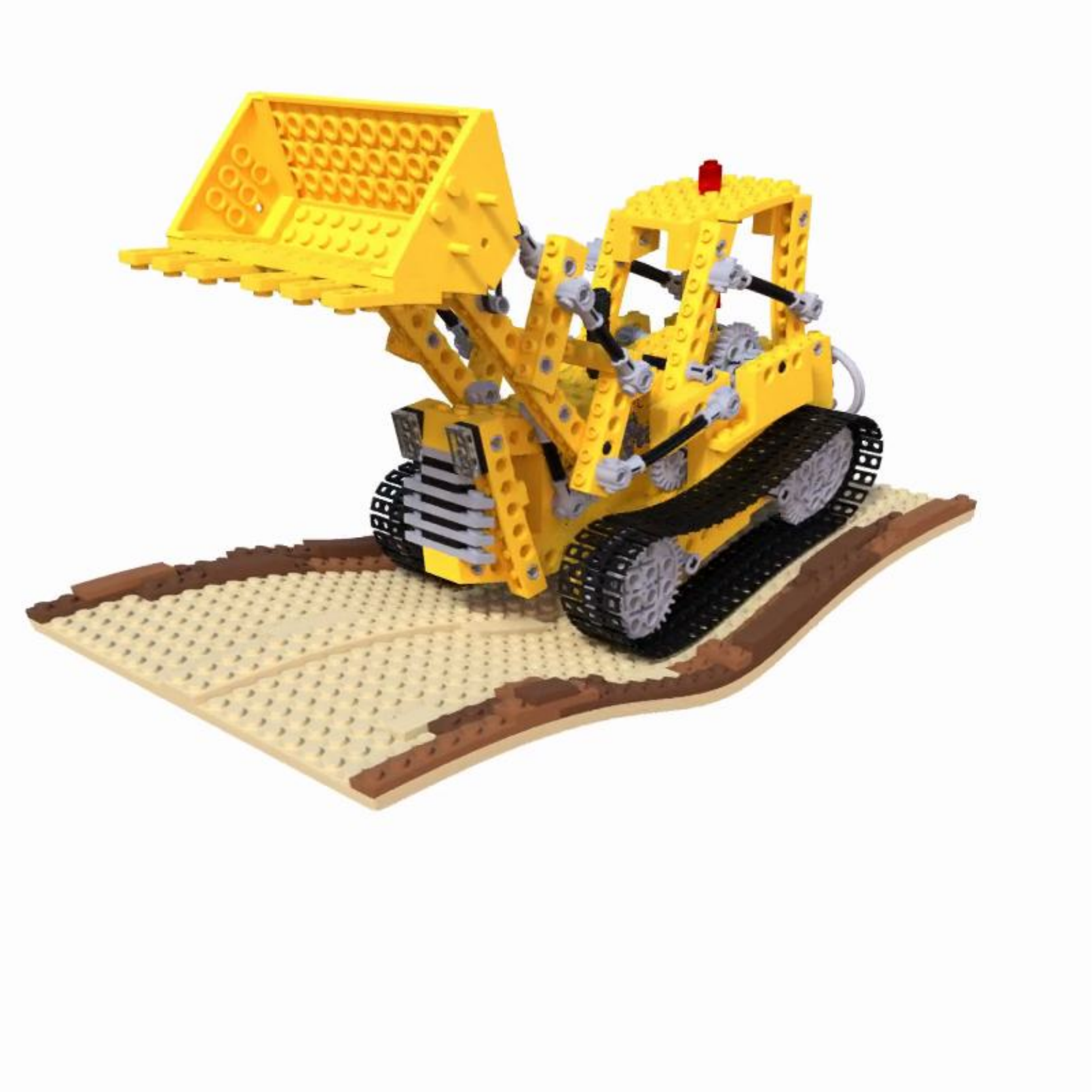} &  
 \includegraphics[width=0.175\textwidth, valign=m]{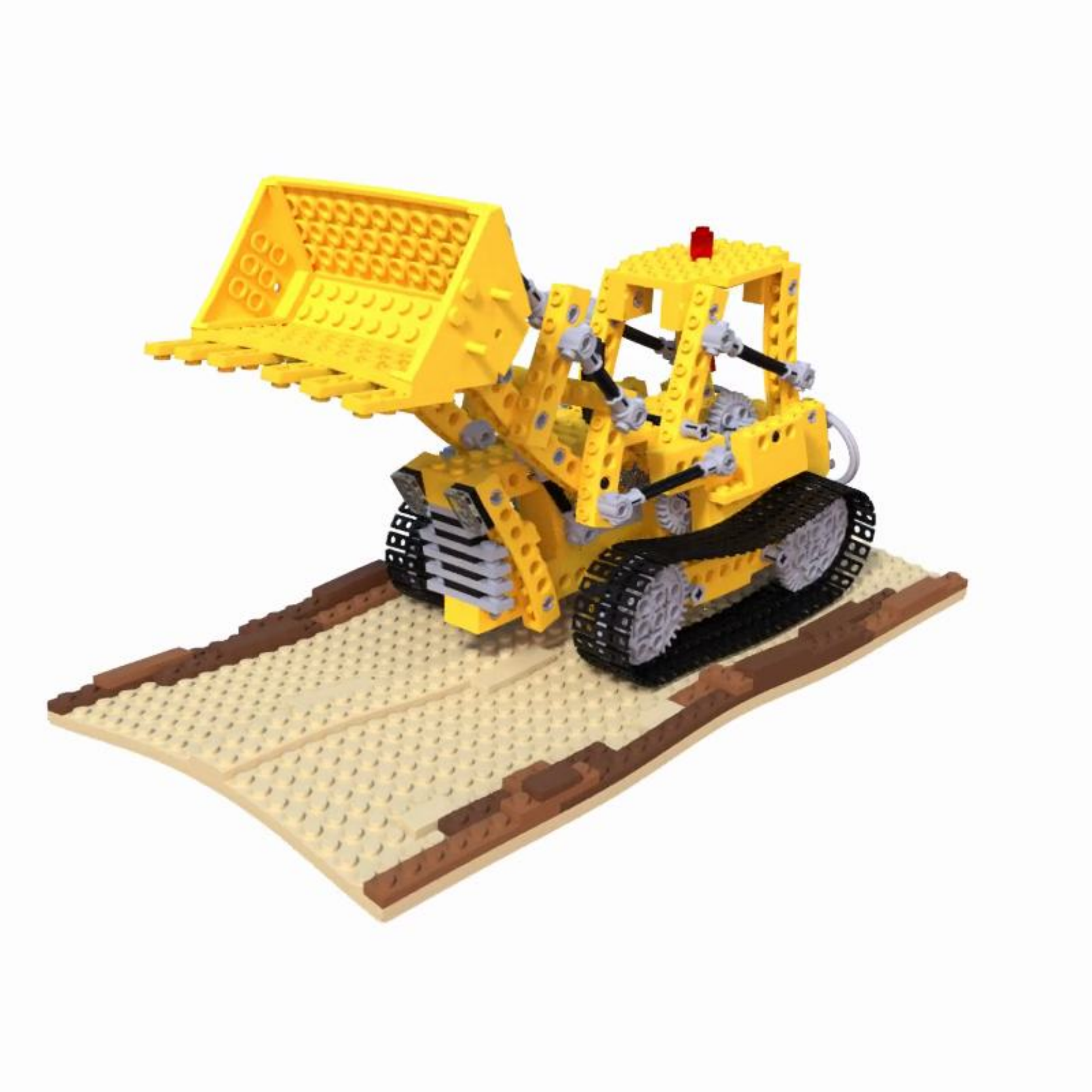} &
 \includegraphics[width=0.175\textwidth, valign=m]{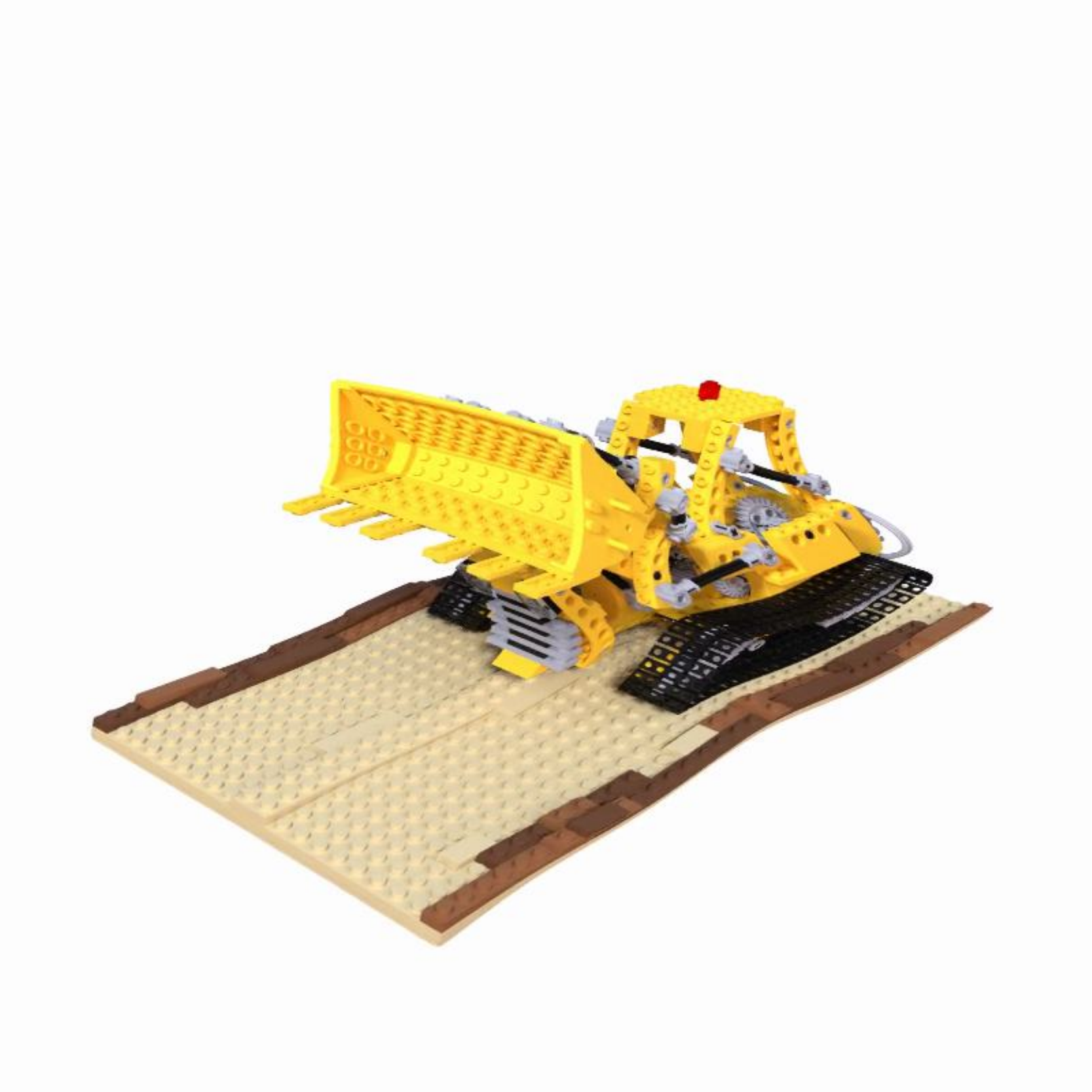} \\ 
\end{tabular}}

\caption{Physics-based simulations on \our{} representations. The explicit point clouds optimized by our method were exported to Blender to drive dynamic deformations. The \textit{Ficus} sequence (top) demonstrates a gravity simulation where leaves detach and fall realistically. The \textit{Lego} sequence (bottom) showcases soft-body dynamics; the truck is dropped onto a platform, exhibiting elastic deformation and bouncing behavior upon collision. These examples highlight the capability of \our{} to support complex physical interactions while maintaining rendering fidelity.}
\label{fig:anim_appendix}
\end{figure*}

\begin{figure}[t]
\centering
\setlength{\tabcolsep}{1pt}
\renewcommand{\arraystretch}{0.5} 

\newcommand{\imgw}{0.15\textwidth}

{\fontsize{6.8pt}{11pt}\selectfont
\begin{tabular}{c c c c c c c}
 & \tiny \textbf{GT} & \tiny \textbf{Naive} & \tiny \textbf{NE} & \tiny \textbf{GaMeS} &\tiny \textbf{EKS} & \tiny \textbf{\our{}} \\
 
\rotatebox[origin=c]{90}{Mat.} &
\includegraphics[width=\imgw, valign=m]{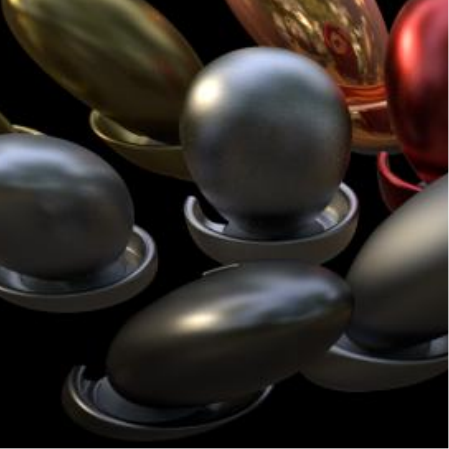} &
\includegraphics[width=\imgw, valign=m]{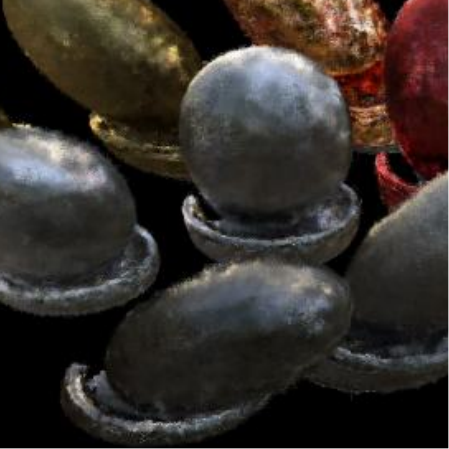} &
\includegraphics[width=\imgw, valign=m]{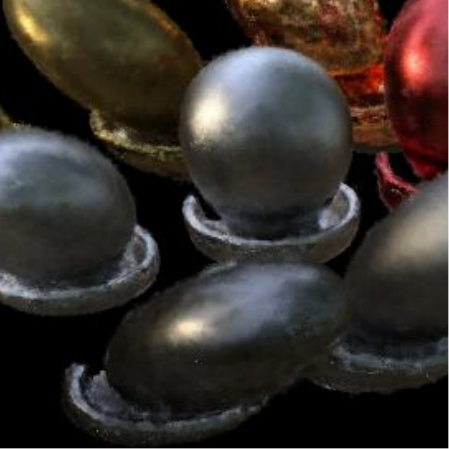} &
\includegraphics[width=\imgw, valign=m]{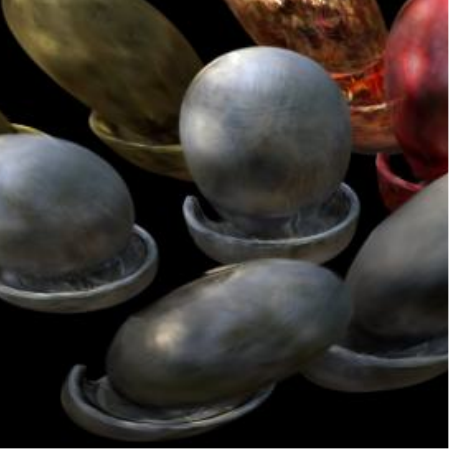} &
\includegraphics[width=\imgw, valign=m]{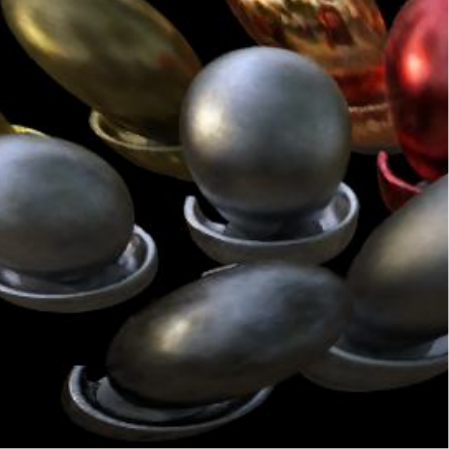} &
\includegraphics[width=\imgw, valign=m]{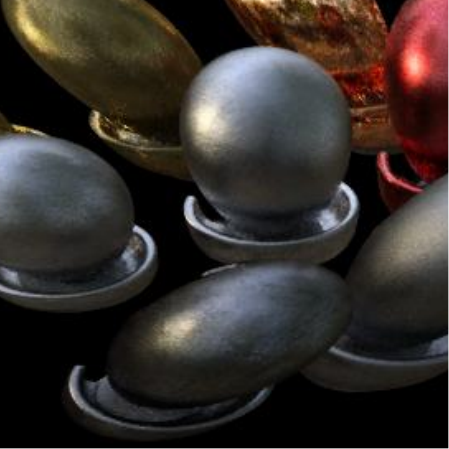} \\

\rotatebox[origin=c]{90}{Lego} &
\includegraphics[width=\imgw, valign=m]{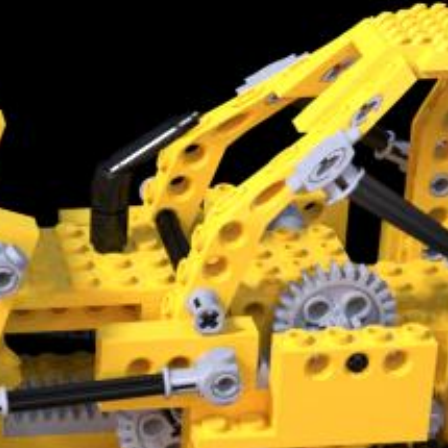} &
\includegraphics[width=\imgw, valign=m]{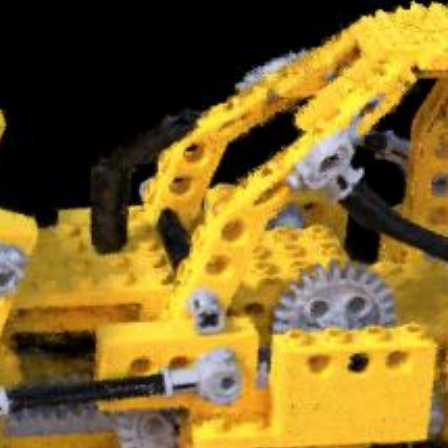} &
\includegraphics[width=\imgw, valign=m]{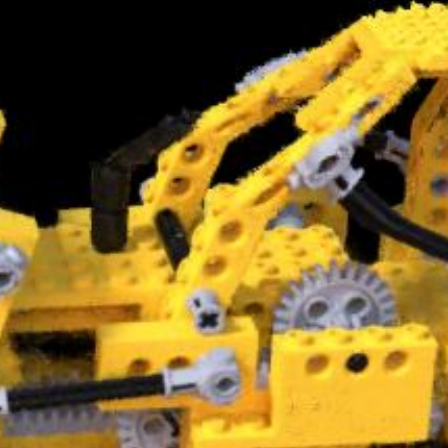} &
\includegraphics[width=\imgw, valign=m]{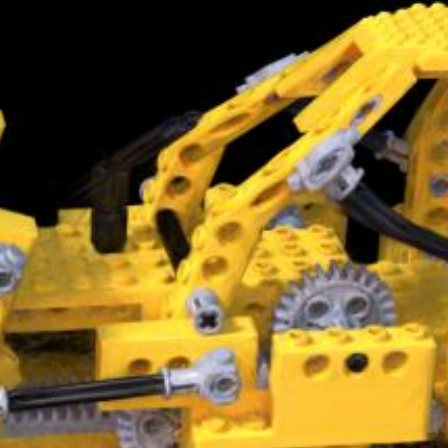} &
\includegraphics[width=\imgw, valign=m]{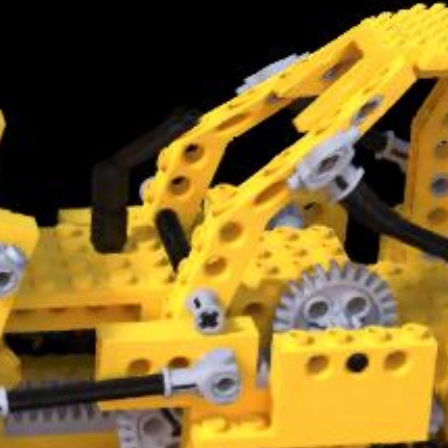} &
\includegraphics[width=\imgw, valign=m]{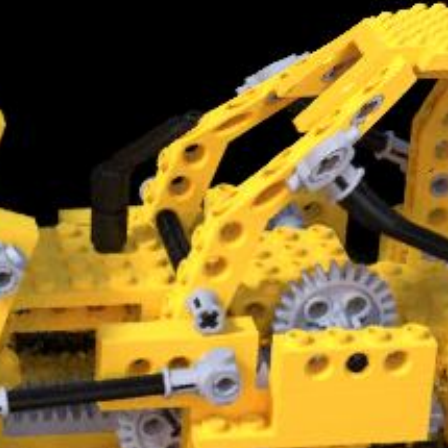} \\

\rotatebox[origin=c]{90}{Drums} &
\includegraphics[width=\imgw, valign=m]{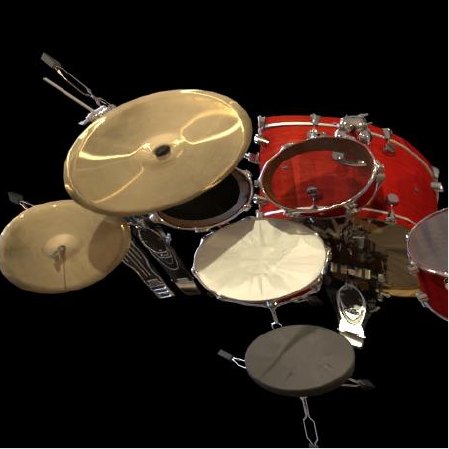} &
\includegraphics[width=\imgw, valign=m]{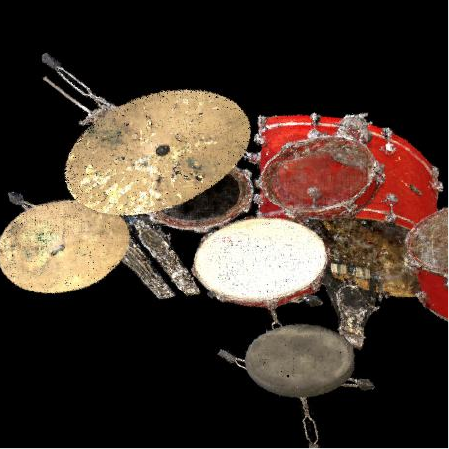} &
\includegraphics[width=\imgw, valign=m]{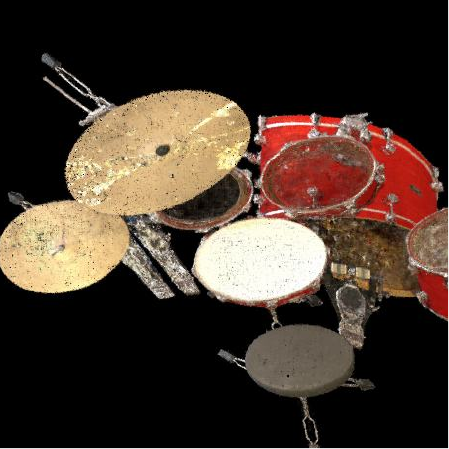} &
\includegraphics[width=\imgw, valign=m]{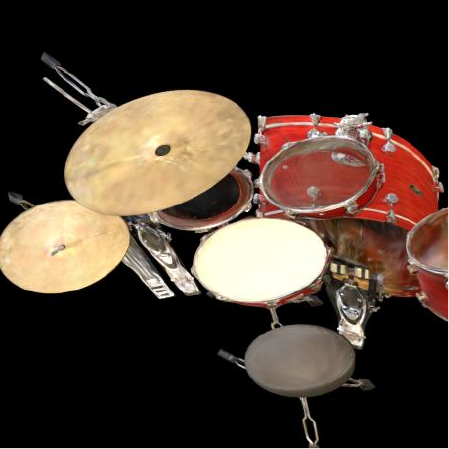} &
\includegraphics[width=\imgw, valign=m]{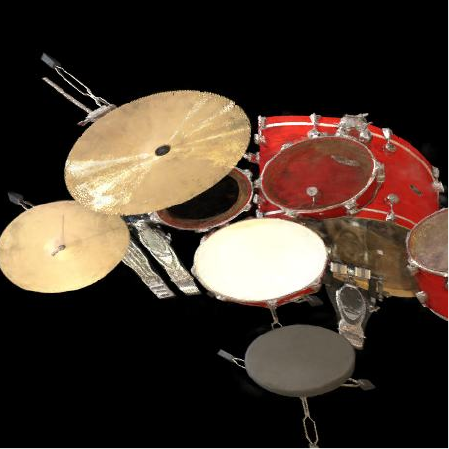} &
\includegraphics[width=\imgw, valign=m]{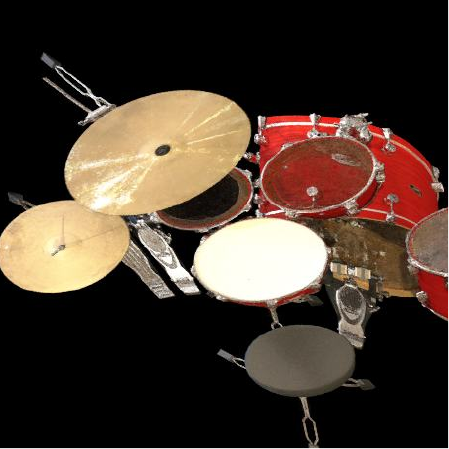} \\

\rotatebox[origin=c]{90}{Mic} &
\includegraphics[width=\imgw, valign=m]{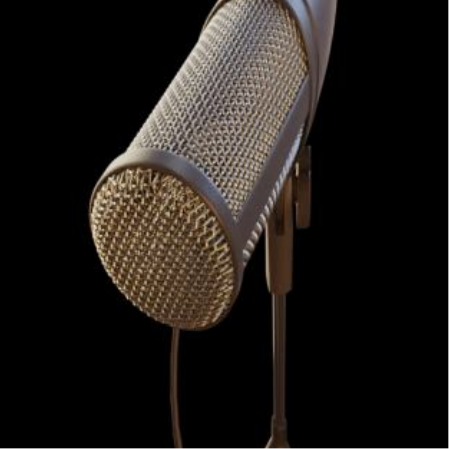} &
\includegraphics[width=\imgw, valign=m]{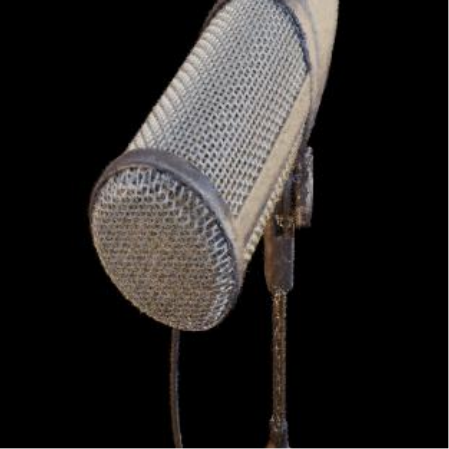} &
\includegraphics[width=\imgw, valign=m]{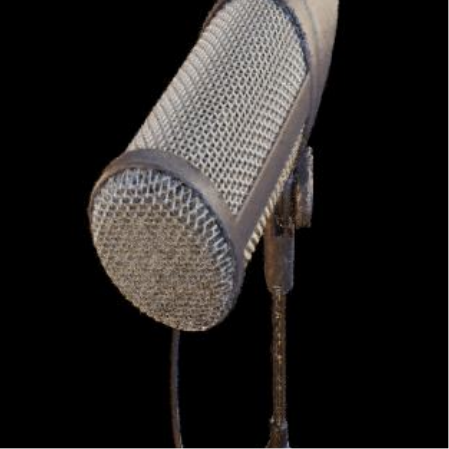} &
\includegraphics[width=\imgw, valign=m]{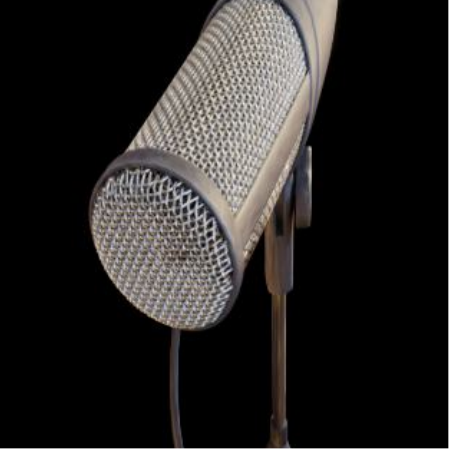} &
\includegraphics[width=\imgw, valign=m]{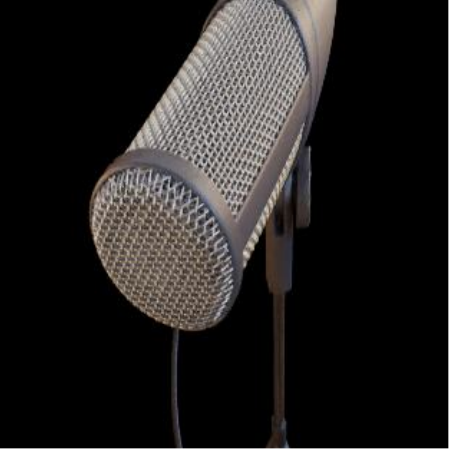} &
\includegraphics[width=\imgw, valign=m]{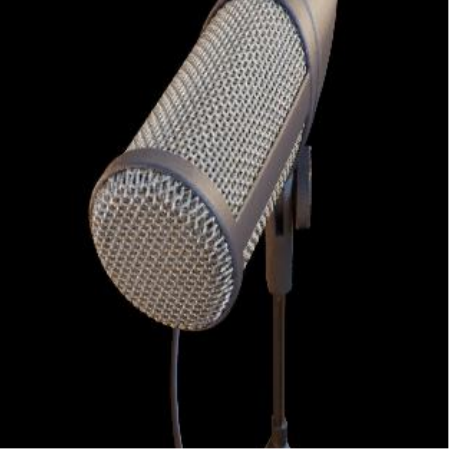}  \\

\rotatebox[origin=c]{90}{Chair} &
\includegraphics[width=\imgw, valign=m]{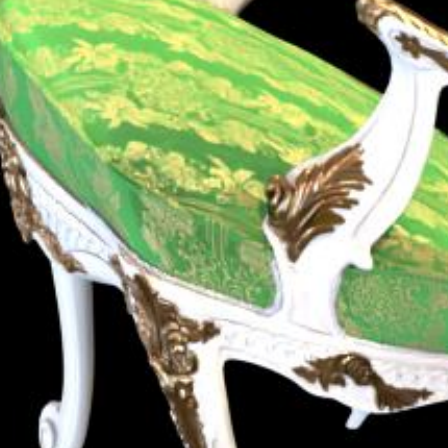} &
\includegraphics[width=\imgw, valign=m]{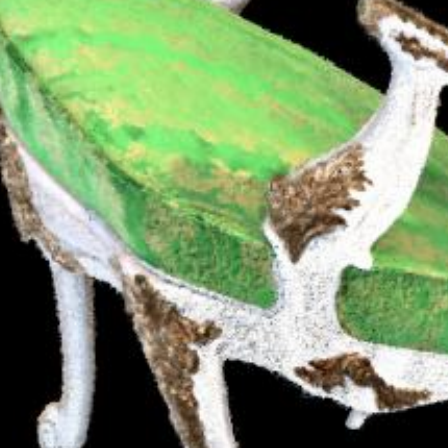} &
\includegraphics[width=\imgw, valign=m]{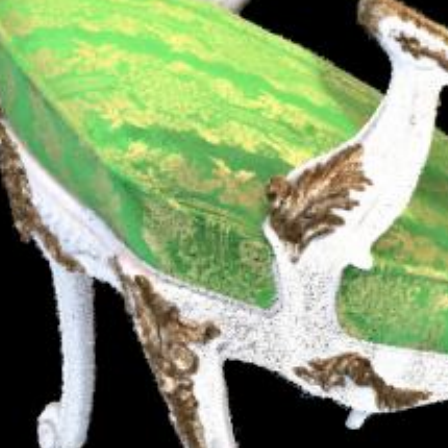} &
\includegraphics[width=\imgw, valign=m]{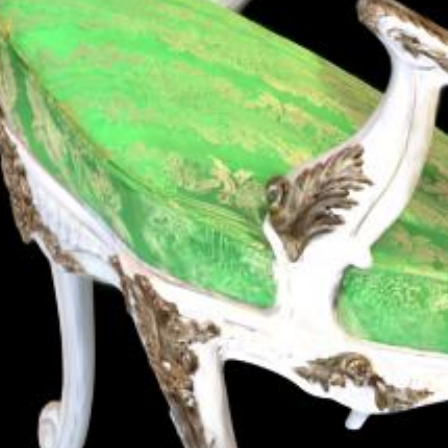} &
\includegraphics[width=\imgw, valign=m]{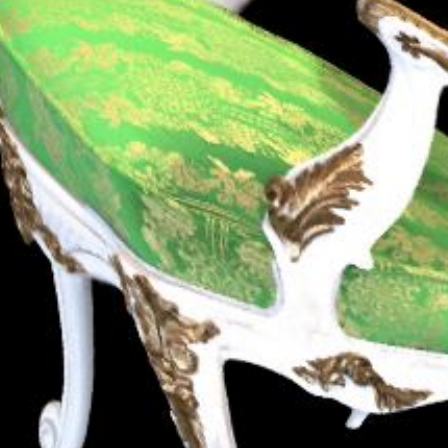} &
\includegraphics[width=\imgw, valign=m]{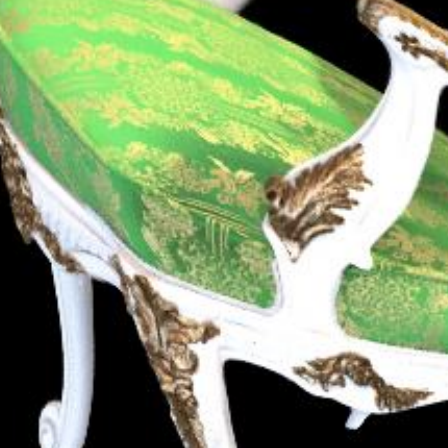}  \\

\rotatebox[origin=c]{90}{Ship} &
\includegraphics[width=\imgw, valign=m]{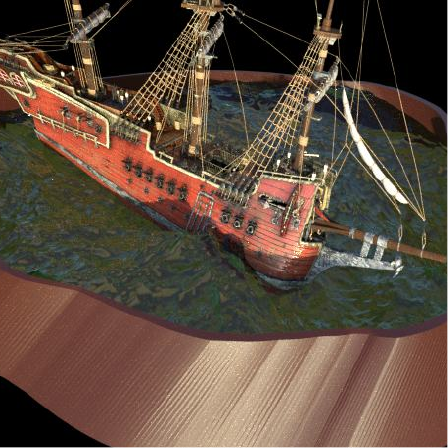} &
\includegraphics[width=\imgw, valign=m]{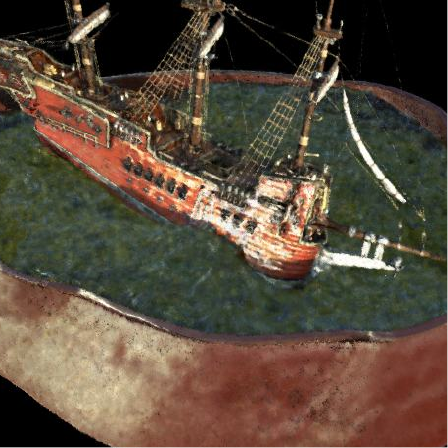} &
\includegraphics[width=\imgw, valign=m]{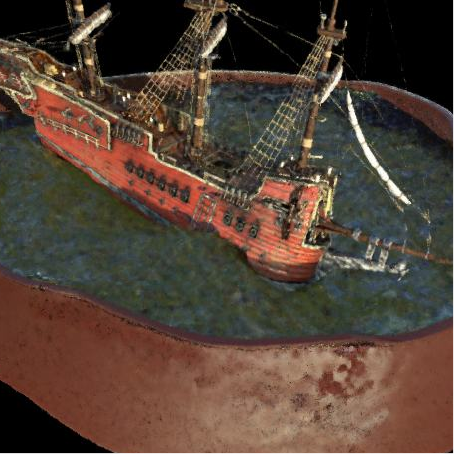} &
\includegraphics[width=\imgw, valign=m]{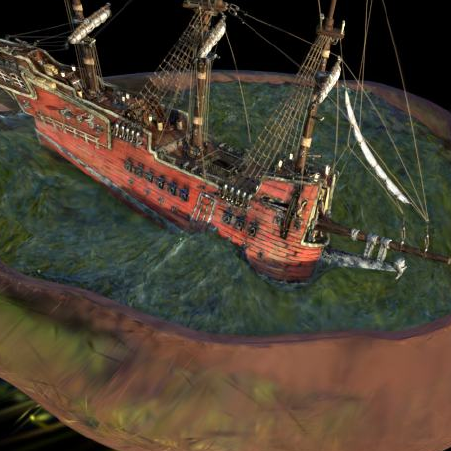} &
\includegraphics[width=\imgw, valign=m]{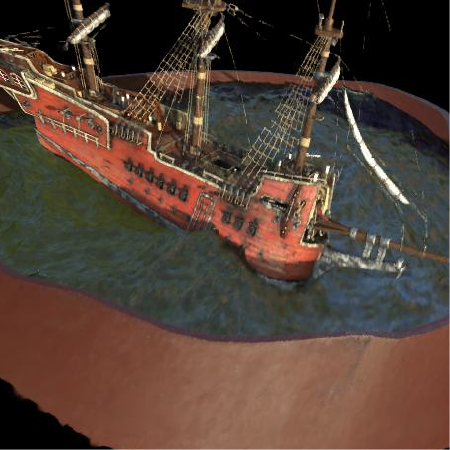} &
\includegraphics[width=\imgw, valign=m]{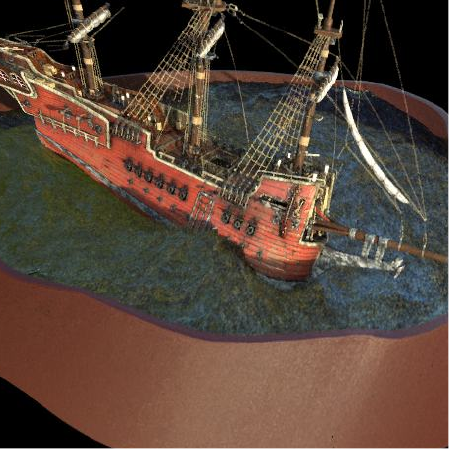} \\

\rotatebox[origin=c]{90}{Hotdog} &
\includegraphics[width=\imgw, valign=m]{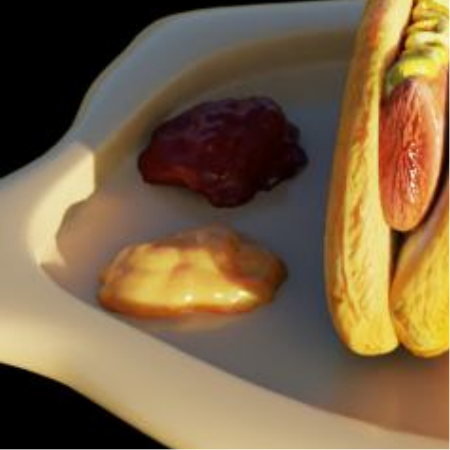} &
\includegraphics[width=\imgw, valign=m]{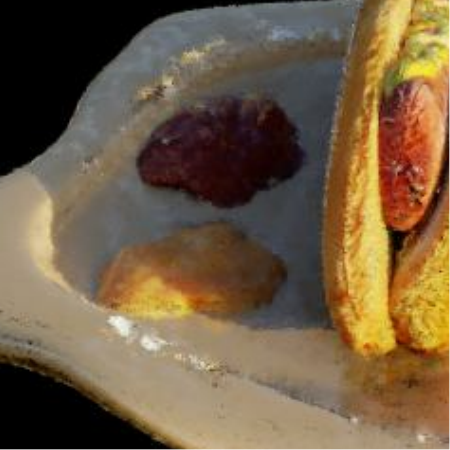} &
\includegraphics[width=\imgw, valign=m]{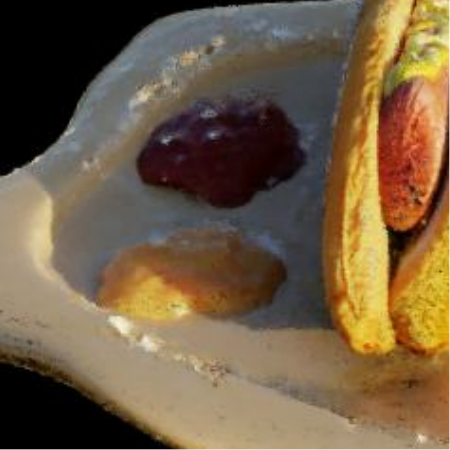} &
\includegraphics[width=\imgw, valign=m]{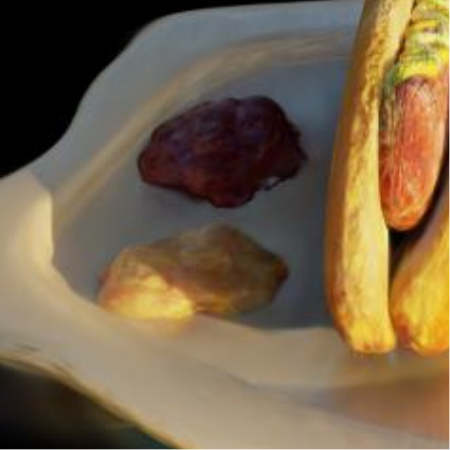} &
\includegraphics[width=\imgw, valign=m]{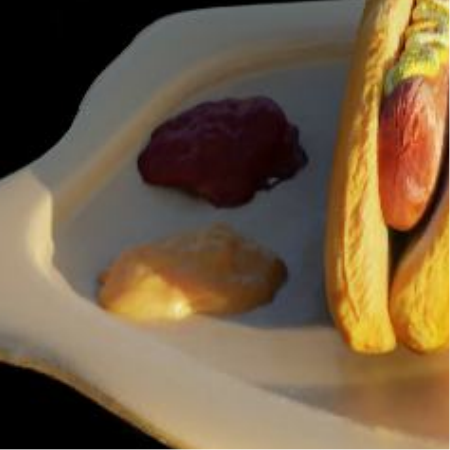} &
\includegraphics[width=\imgw, valign=m]{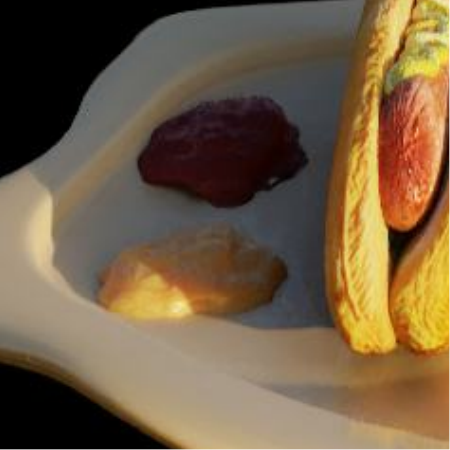}\\

\rotatebox[origin=c]{90}{Ficus} &
\includegraphics[width=\imgw, valign=m]{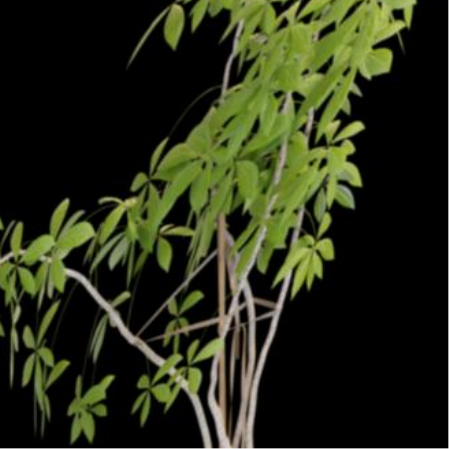} &
\includegraphics[width=\imgw, valign=m]{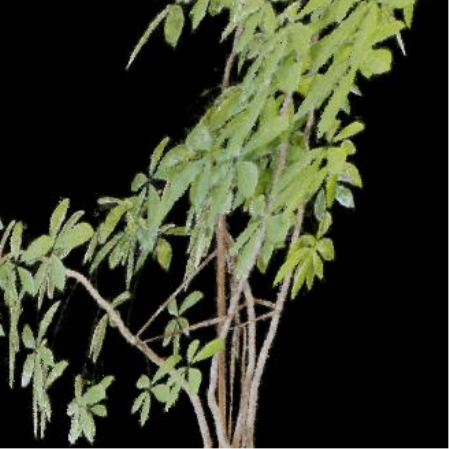} &
\includegraphics[width=\imgw, valign=m]{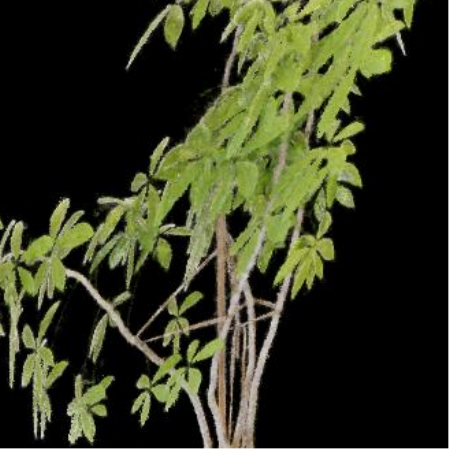} &
\includegraphics[width=\imgw, valign=m]{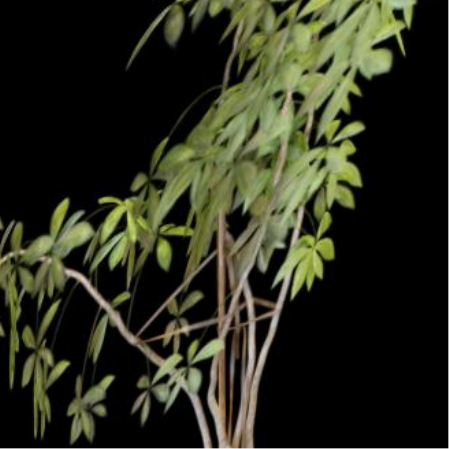} &
\includegraphics[width=\imgw, valign=m]{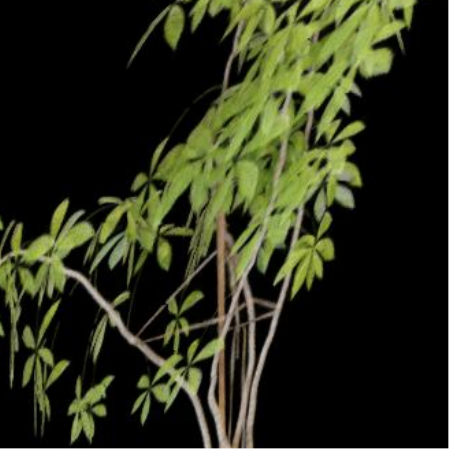} &
\includegraphics[width=\imgw, valign=m]{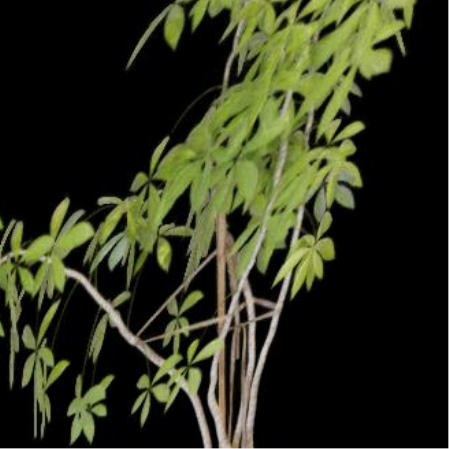}\\
\end{tabular}}
\caption{\textbf{Qualitative results on the NeuralEditor benchmark.} Comprehensive comparison across all scenes from the deformation dataset. NE denotes NeuralEditor. It is observed that \our{} maintains high visual fidelity and geometric consistency, avoiding the blurring artifacts seen in Naive and NeuralEditor baselines, while achieving quality comparable to the state-of-the-art EKS method.}
\label{fig:benchmark_neuraleditor}
\end{figure}

Further visual evidence supporting the robustness of the framework is presented in this section. Fig.~\ref{fig:comparison_crops} provides an extended qualitative comparison on the Mip-NeRF 360 and Deep Blending~\cite{deepblending} datasets. These close-up views highlight the ability of the model to reconstruct fine details. Fig.~\ref{fig:comparison_full} complements the close-up analysis by showing the full, zoomed-out views of these scenes, demonstrating the global geometric consistency of the reconstruction. Extending this analysis to dynamic scenarios, Fig.~\ref{fig:additional_anims} illustrates complex, topology-agnostic manipulations applied directly to the unbounded environments of the Mip-NeRF 360 dataset, preserving high-frequency details and realistic lighting during both rigid translations and non-rigid deformations.
Fig.~\ref{fig:anim_appendix} showcases the physics-based editing capabilities on the NeRF Synthetic dataset, demonstrating robust behavior under gravity and soft-body collisions.
Finally, Fig.~\ref{fig:benchmark_neuraleditor} provides a comprehensive comparison across all scenes from the deformation benchmark, illustrating the visual fidelity of \our{} compared to methods like NeuralEditor or GaMeS.

